\documentclass[10pt,twocolumn,letterpaper]{article}

\usepackage[pagenumbers]{wacv} 

\usepackage{graphicx}
\usepackage{amsmath}
\usepackage{amssymb}
\usepackage{booktabs}

\usepackage[pagebackref,breaklinks,colorlinks]{hyperref}

\usepackage[capitalize]{cleveref}
\crefname{section}{Sec.}{Secs.}
\Crefname{section}{Section}{Sections}
\Crefname{table}{Table}{Tables}
\crefname{table}{Tab.}{Tabs.}

\usepackage{graphicx}
\usepackage{amsmath}
\usepackage{amssymb}
\usepackage{colortbl}
\usepackage[usenames, dvipsnames]{xcolor}
\usepackage{pgf}

\usepackage{caption}
\usepackage{times}
\usepackage{epsfig}
\usepackage{bm}
\usepackage{pifont}%

\newcommand{\dec}[1]{\ensuremath{_{\text{\textcolor{blue}{(-#1)}}}}}

\definecolor{Gray}{gray}{0.90}
\definecolor{white}{rgb}{1.0, 1.0, 1.0}
\definecolor{LightCyan}{RGB}{240, 224, 238}

\usepackage{sidecap}

\usepackage{float}

\newcommand{\heatmapcolor}[1]{%
\pgfmathsetmacro{\opacity}{#1} 
\pgfmathsetmacro{\bluecomponent}{100 - \opacity} 
\pgfmathsetmacro{\greencomponent}{2*\opacity} 
\edef\x{\noexpand\cellcolor{blue!\bluecomponent!red!\greencomponent!white!10}}\x #1 
}

\newcommand{\heatmapcolors}[1]{%
\pgfmathsetmacro{\opacity}{min(100,#1)} 
\pgfmathsetmacro{\bluecomponent}{100 - \opacity} 
\pgfmathsetmacro{\greencomponent}{min(100,2*\opacity)} 
\edef\x{\noexpand\cellcolor{blue!\bluecomponent!red!\greencomponent!white!10}}\x #1 
}

\definecolor{citecolor}{RGB}{30,130,255}
\hypersetup{colorlinks=true, linkcolor=red, citecolor=blue, urlcolor=blue}
\usepackage{multibib}
\newcites{supp}{Supplementary References}

\def\wacvPaperID{2469} 
\def\confName{WACV}
\def\confYear{2025}

\begin{document}

\title{Towards Evaluating the Robustness of Visual State Space Models}



\author{
Hashmat Shadab Malik$^1$ \and 
Fahad Shamshad$^1$ \and 
Muzammal Naseer$^2$ \and 
Karthik Nandakumar$^1$ \and 
Fahad Shahbaz Khan$^{1,3}$ \and 
Salman Khan$^{1,4}$
\\[2ex]
$^1$Mohamed Bin Zayed University of AI, UAE\\
$^2$Center of Secure Cyber-Physical Security Systems, Khalifa University, UAE\\
$^3$Link{\"o}ping University, Sweden\\
$^4$Australian National University, Australia\\[2ex]
\tt\small
\{hashmat.malik, fahad.shamshad, karthik.nandakumar, fahad.khan, salman.khan\}@mbzuai.ac.ae\\
\tt\small muzammal.naseer@ku.ac.ae
}


\maketitle

\begin{abstract}
  Vision State Space Models (VSSMs), a novel architecture that combines the strengths of recurrent neural networks and latent variable models, have demonstrated remarkable performance in visual perception tasks by efficiently capturing long-range dependencies and modeling complex visual dynamics. However, their robustness under natural and adversarial perturbations remains a critical concern. In this work, we present a comprehensive evaluation of VSSMs' robustness under various perturbation scenarios, including occlusions, image structure, common corruptions, and adversarial attacks, and compare their performance to well-established architectures such as transformers and Convolutional Neural Networks.  Furthermore, we investigate the resilience of VSSMs to object-background compositional changes on sophisticated benchmarks designed to test model performance in complex visual scenes. We also assess their robustness on object detection and segmentation tasks using corrupted datasets that mimic real-world scenarios. To gain a deeper understanding of VSSMs' adversarial robustness, we conduct a frequency-based analysis of adversarial attacks, evaluating their performance against low-frequency and high-frequency perturbations. Our findings highlight the strengths and limitations of VSSMs in handling complex visual corruptions, offering valuable insights for future research.  Our code and models will be available at \href{https://github.com/HashmatShadab/MambaRobustness}{\color{Magenta}{{https://github.com/HashmatShadab/MambaRobustness}}}.
\end{abstract}

\section{Introduction}

Deep learning models such as Convolutional Neural Networks (CNNs)~\cite{krizhevsky2012imagenet} and Vision Transformers~\cite{dosovitskiy2020image}  have achieved remarkable success across various visual perception tasks, including image classification, object detection, and semantic segmentation. However, their robustness across different distribution shifts of the data remains a significant concern for their deployment in security-critical applications. Several works~\cite{hendrycks2019benchmarking,zhou2022understanding,naseer2021intriguing,bai2021transformers}  have extensively evaluated the robustness of CNNs and Transformers against common corruptions, domain shifts, information drop, and adversarial attacks, highlighting that a model's design impacts its ability to handle adversarial and natural corruptions, with robustness varying across different architectures. This observation motivates us to investigate the robustness of the recently proposed Vision State-Space Models (VSSMs)~\cite{zhu2024vision,liu2024vmamba,gu2023mamba}, a novel architecture designed to efficiently capture long-range dependencies in visual data.

CNNs are particularly adept at extracting hierarchical image features due to their shared weights across features, which help in capturing local-level information. In contrast, transformer-based models employ an attention mechanism that captures global information, effectively increasing the model's receptive field \cite{dosovitskiy2020image}. This allows transformers to excel at modeling long-range dependencies. However, a significant drawback of transformers is their quadratic computational scaling with input size, which makes them computationally expensive for downstream tasks \cite{luo2016understanding}. Recently, state space sequence models (SSMs) have been adapted from the natural language domain to vision tasks. Unlike transformers, vision-based SSMs offer the capability to handle long-range dependencies while maintaining a linear computational cost, providing a more efficient alternative for vision applications \cite{gu2021efficiently, gu2023mamba, liu2024vmamba, zhu2024vision, li2024videomamba, chen2024video}.

VSSMs, such as the VMamba~\cite{liu2024vmamba} and the hybrid Mamba-Transformer variant MambaVision~\cite{hatamizadeh2024mambavision}, have gained attention in the vision domain due to their impressive performance. These models offer a unique approach to managing spatial dependencies, which is critical for handling dynamic visual environments. Their ability to selectively adjust interactions between states promises enhanced adaptability, a trait that could be pivotal in improving resilience against perturbations. Given their potential in safety-critical applications such as autonomous vehicles, robotics, and healthcare, it is crucial to thoroughly assess the robustness of these models.

In this paper, we present a comprehensive analysis of the performance of VSSMs, Vision Transformers, and CNNs in handling various nuisances for classification, detection, and segmentation tasks, aiming to provide valuable insights into their robustness and suitability for real-world applications. Our evaluation is divided into three main parts, each addressing a crucial aspect of model robustness.

\begin{itemize}
    \item[\ding{202}] \textbf{\textcolor{cyan}{Occlusions and Information Loss:}}  We rigorously assess the robustness of pure and hybrid VSSMs against information loss along scanning directions and severe occlusions affecting foreground objects, non-salient background regions, and random patch drops at multiple levels. This analysis is crucial for understanding how well VSSMs can handle partial information drop and maintain performance despite occlusions. Additionally, we explore the sensitivity of VSSMs to the overall image structure and global composition through patch shuffling experiments, providing insights into their ability to capture global context.
    \vspace{-0em}
    
    \textit{{\color{Magenta}\textbf{Findings:}} Our experiments reveal that ConvNext~\cite{liu2022convnet} and VSSM models are superior in handling of sequential information loss along the scanning direction compared to ViT and Swin models. In scenarios involving random, salient, and non-salient patch drops, VSSMs exhibit the highest overall robustness, although Swin models perform better under extreme information loss. Additionally, VSSM models show greater resilience to spatial structure disturbances caused by patch shuffling compared to Swin models.}

    \item[\ding{203}] \textbf{\textcolor{cyan}{Common Corruptions:}} We evaluate the robustness of VSSM-based classification models against common corruptions that mimic real-world scenarios. This includes both \textit{\textbf{global corruptions}} such as noise, blur, weather, and digital-based corruptions at multiple intensity levels, and \textit{\textbf{fine-grained corruptions}} like object attribute editing and background manipulations. Furthermore, we extend the evaluation to VSSM-based detection and segmentation models to demonstrate their robustness in dense prediction tasks. By testing the models under these diverse and challenging conditions, we aim to provide a comprehensive understanding of their resilience in real-world applications.
    \vspace{-0em}
    
    \textit{\textbf{{\color{Magenta}Findings:}} For global corruptions, VSSM models experience the least average performance drop compared to Swin and ConvNext models. When subjected to fine-grained corruptions, the VSSM family outperforms all transformer-based variants and maintains performance that is either better than or comparable to the advanced ConvNext models. In dense prediction tasks such as detection and segmentation, VSSM-based models generally demonstrate greater resilience and outperform other models.}

    \item[\ding{204}] \textbf{\textcolor{cyan}{Adversarial Attacks:}} We analyze the robustness of VSSMs against adversarial attacks in both white-box and black-box settings. In addition to the standard adversarial evaluation, we conduct a frequency analysis to demonstrate the resilience of VSSM models against low-frequency and high-frequency adversarial perturbations. This analysis provides insights into VSSMs ability to withstand adversarial perturbations at different frequency levels.

    \textit{\textbf{{\color{Magenta}Findings:}} For adversarial attacks, smaller VSSM models exhibit higher robustness against white-box attacks than Swin Transformers, though this does not scale to larger VSSMs. VSSMs maintain over 90$\%$ robustness against low-frequency perturbations, even at high perturbation strengths, but degrade quickly under high-frequency attacks. Across standard attacks, VSSMs outperform ConvNext, ViT, and Swin models under smaller perturbation budgets. In adversarial fine-tuning, VSSMs excel in both clean and robust accuracy on high-resolution images, but ViT models outperform them on low-resolution datasets like CIFAR.}    
\end{itemize}

Our findings reveal that VSSM models have both strengths and limitations in handling various nuisances and adversarial attacks. They also indicate that a variety of metrics is essential to fully evaluate the diverse capabilities of different architectures. While VSSMs often demonstrate superior robustness, ConvNext and ViT architectures occasionally outperform them. These insights can inform model selection for specific applications, considering robustness requirements in real-world scenarios.

\section{Related Work}
\noindent

\noindent \textbf{Robustness of Deep Learning Models:} Robustness refers to a conventionally trained model's ability to maintain satisfactory performance under natural and adversarial distribution shifts~\cite{geirhos2018imagenet,drenkow2021systematic}.  In practice, deep learning-based models often encounter various types of corruptions, such as noise, blur, compression artifacts, and adversarial perturbations, which can significantly degrade their performance. To ensure the reliability and robustness of these models, it is essential to systematically evaluate their performance under such challenging conditions. Recent studies have investigated the robustness of deep learning-based models across a wide range of areas, including image classification~\cite{hendrycks2019benchmarking, hendrycks2021natural}, semantic segmentation~\cite{kamann2020benchmarking}, object detection~\cite{michaelis2019benchmarking}, video classification~\cite{yi2021benchmarking}, point cloud processing \cite{ji2023benchmarking}, and transformer-based architectures~\cite{paul2022vision,Kumar2024MultimodalGV,pinto2022impartial}. However, there is a lack of similar investigations for vision state space models (VSSMs), despite their growing popularity and potential applications~\cite{liu2024vision,yue2024medmamba,xu2024survey,chen2024rsmamba}. In this work, we aim to bridge this gap by examining how the performance of VSSMs is affected by adversarial and common corruptions.
Considering the increasing adoption of VSSMs, our findings can provide valuable insights for researchers and practitioners working on developing robust and reliable vision systems.

\noindent \textbf{State Space Models:} State space models (SSMs)~\cite{mehta2022long,wang2023selective} have emerged as a promising method for modeling sequential data in deep learning. These models map a 1-dimensional sequence $x(t) \in \mathbb{R}^{L}$ to $y(t) \in \mathbb{R}^{L}$ via an implicit latent state $h(t) \in \mathbb{R}^{N}$ as:

\begin{align}
\label{eq:ode}
h'(t) = \bm{A}h(t) + \bm{B}x(t), \quad\quad y(t) = \bm{C}h(t),
\end{align}

where $\bm{A}\in\mathbb{R}^{N\times N}$, $\bm{B}\in\mathbb{R}^{N\times 1}$, and $\bm{C}\in\mathbb{R}^{N\times 1}$ are continuous parameters governing the dynamics and output mapping.
To enhance computational efficiency, the continuous SSM is discretized using a zero-order hold assumption, leading to a discretized form:
\begin{align} \label{eq:discretization}
h_t = \bm{\overline{A}}h_{t-1} + \bm{\overline{B}}x_t, \quad\quad y_t = \bm{C}h_t,
\end{align}
where $\bm{\overline{A}}$ and $\bm{\overline{B}}$ are the discrete counterparts of $\bm{A}$ and $\bm{B}$, obtained via a specified sampling timescale $\bm{\Delta} \in \mathbb{R}{>0}$.
The iterative process in Eq. \ref{eq:discretization} can be further expedited through parallel computation using a global convolution operation:
\begin{small}
\begin{align}
\bm{y} &= \bm{x} \circledast \bm{\overline{K}}, \quad\text{with}\quad \bm{\overline{K}} = (\bm{C}\bm{\overline{B}},\bm{C}\overline{\bm{A}\bm{B}}, ..., \bm{C}\bm{\overline{A}}^{L-1}\bm{\overline{B}}),
\label{eq:Cumulative}
\end{align}
\end{small}
where $\bm{\overline{K}} \in \mathbb{R}^L$ is the kernel used and $\circledast$ denotes convolution operator.
Recent advancements in SSMs, like Mamba models~\cite{gu2023mamba}, introduce dynamic, input-dependent parameterization for managing sequential state interactions. This inspired vision models such as VMamba~\cite{liu2024vmamba} and MambaVision~\cite{hatamizadeh2024mambavision}, which combine Mamba with ViT-like hierarchies. While these models have been studied for tasks like detection and segmentation, their robustness against natural and adversarial corruptions remains underexplored. We aim to evaluate their resilience to these perturbations, which is crucial for understanding their potential in real-world applications and identifying areas for improvement.

\section{Robustness of Vision State Space Models}

We have broadly categorized the experiments into natural and adversarial corruption categories to evaluate the robustness of CNNs, transformers, and VSSMs across tiny, small, and base-model families. For natural corruptions, we conduct experiments on classification, detection, and segmentation tasks. For the classification task, we employ the recent ConvNext model~\cite{liu2022convnet} as a representative of the CNN family, while selecting the ViT~\cite{dosovitskiy2020image} and Swin~\cite{liu2021swin} family models for transformer architectures. For VSSMs, we report results on pure VMamba \emph{v2} pretrained models~\cite{liu2024vmamba} and hybrid MambaVision models~\cite{hatamizadeh2024mambavision}.
For detection and segmentation, we report results using ImageNet-pretrained backbones of the specified models. These models are fine-tuned with the MMDetection~\cite{chen2019mmdetection} and MMSegmentation~\cite{contributors2020mmsegmentation} frameworks. For detection, we utilize Mask-RCNN~\cite{he2017mask}, and for segmentation, we use UperNet~\cite{xiao2018unified}  as the network architecture. To evaluate the robustness of VSSMs against adversarial attacks, we consider imagenet trained classification models. Furthermore, we use imagenet pretrained models for adversarial fine-tuning on two downstream datasets; CIFAR-10~\cite{article} and Imagenette~\cite{imagenette}. 
Evaluations are done on 224×224 images for classification (except 32×32 for CIFAR-10), 800×1216 for detection, and 512×512 for segmentation.


\subsection{Robustness against Natural Corruptions} 
\noindent We categorize natural corruptions into \textbf{information drop} and \textbf{ImageNet-based corruption benchmarks}. Information drop experiments assess models' robustness against various patch perturbations, such as occlusions, random patch drops, and patch shuffling, assessing their ability to handle partial information loss and local distortions. All information drop experiments are conducted on 5000 images from the ImageNet validation set, following~\cite{naseer2021improving}. ImageNet-based corruptions mimic real-world issues such as noise, blur, weather, and digital corruptions at various intensities. We also evaluate \textbf{VSSM-based detectors and segmentation models} under these natural corruptions.

\begin{figure*}[t]
\small \centering
\begin{minipage}{0.69\textwidth}
    \begin{minipage}{0.49\textwidth}
        \centering
        \includegraphics[height=3.5cm, keepaspectratio]{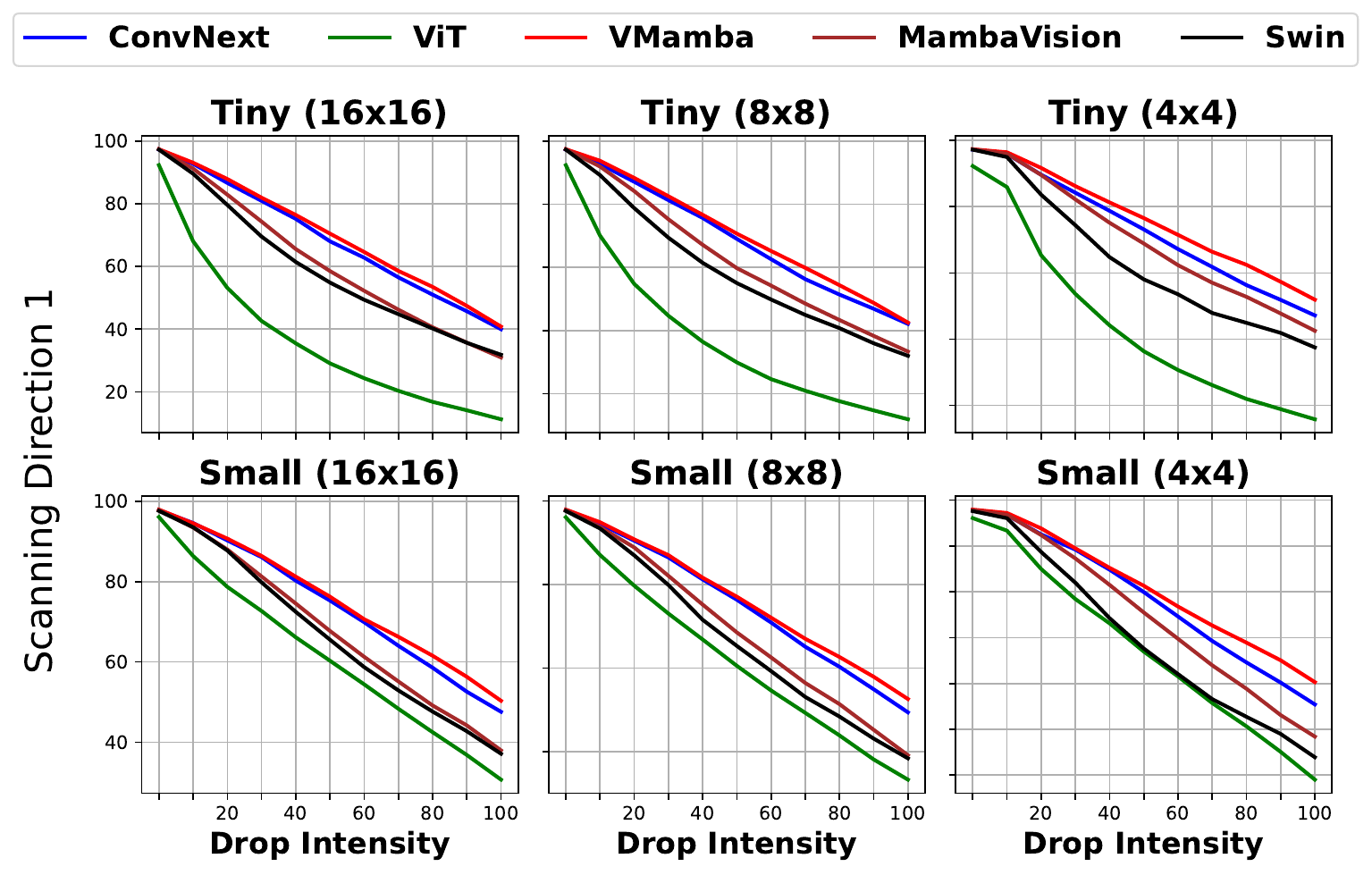}
    \end{minipage}
    \begin{minipage}{0.49\textwidth}
        \centering
        \includegraphics[height=3.5cm, keepaspectratio]{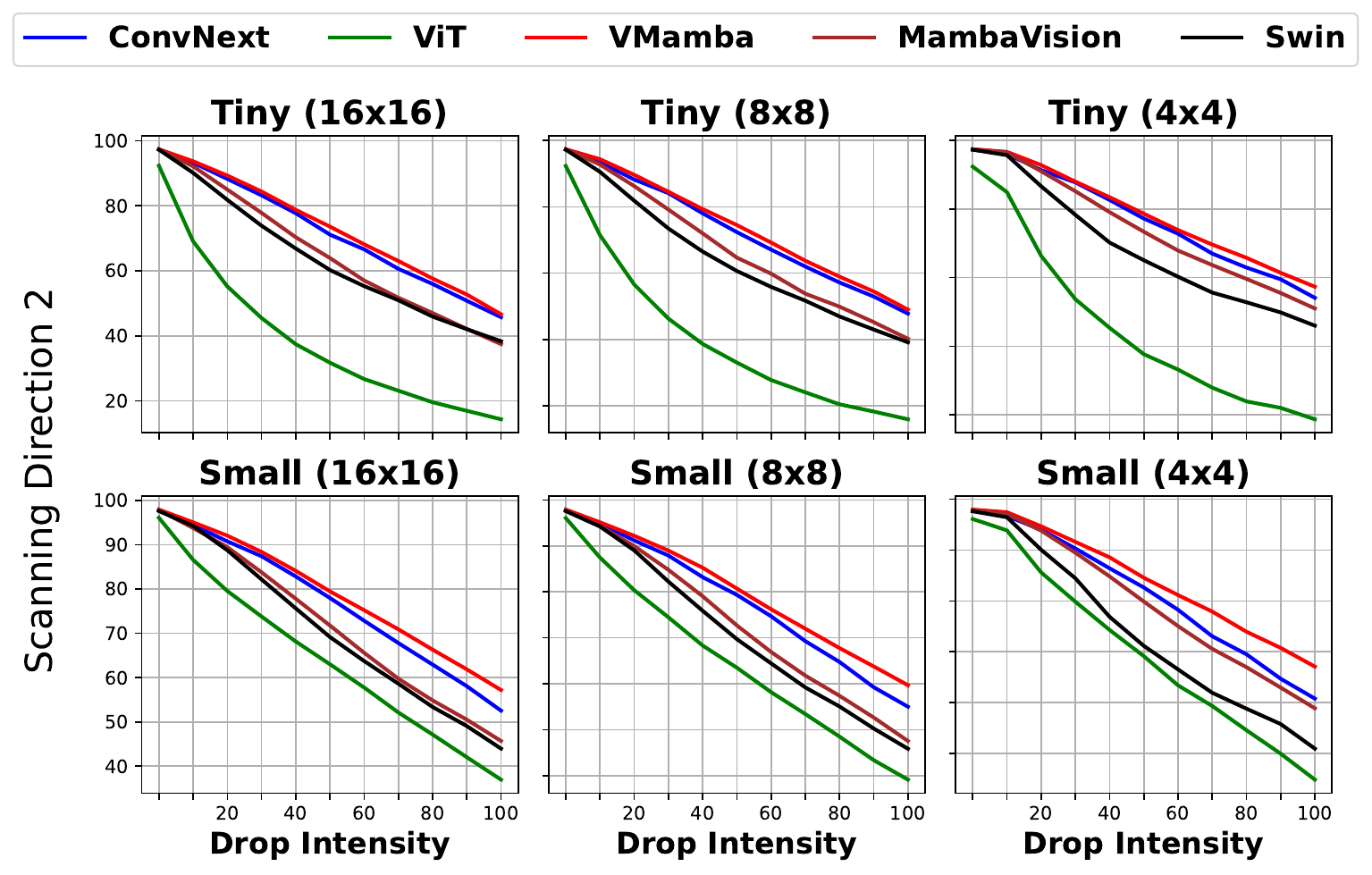}
    \end{minipage}
    \begin{minipage}{0.49\textwidth}
        \centering
        \includegraphics[height=3.2cm, keepaspectratio]{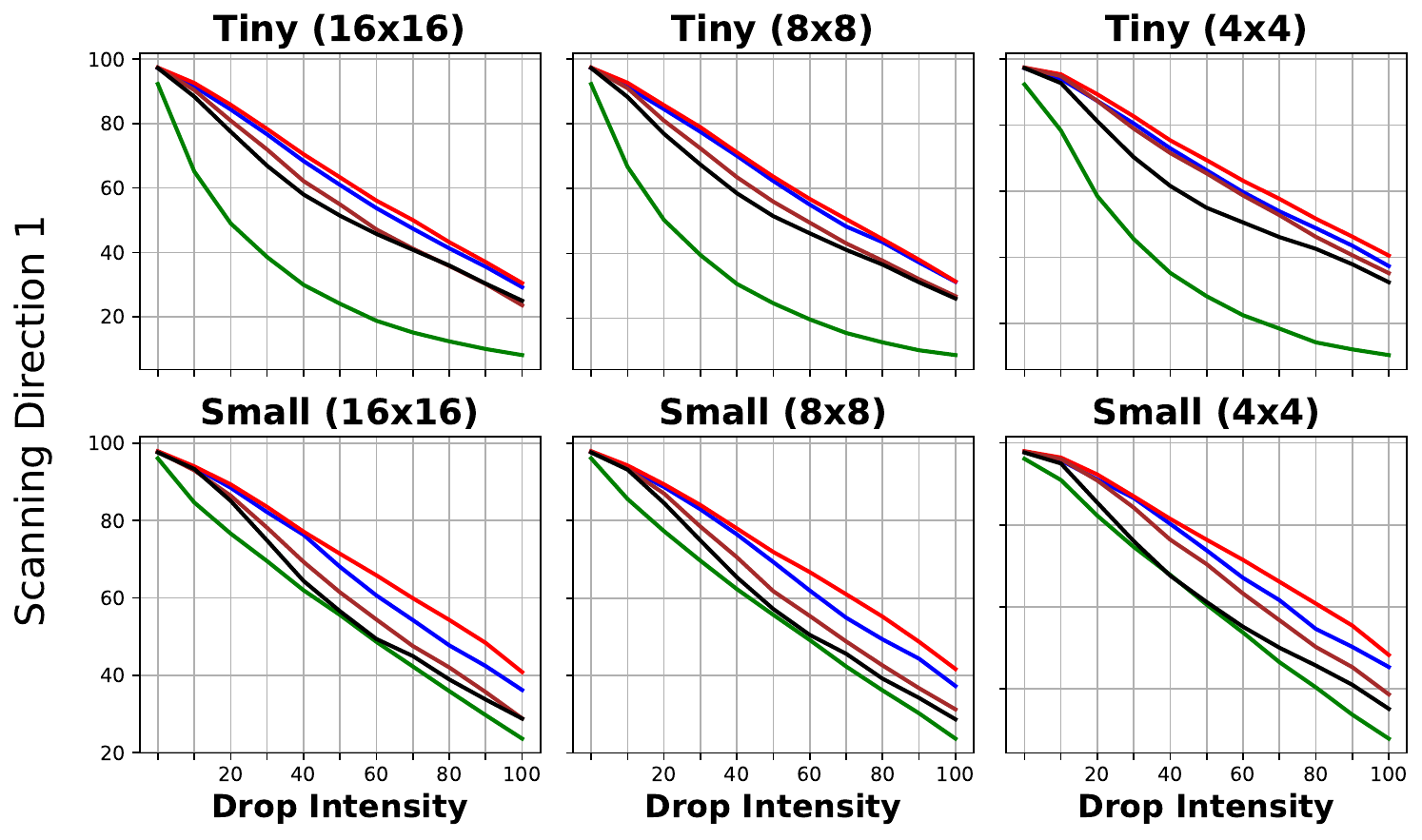}
    \end{minipage}
    \hspace{0.4em}
    \begin{minipage}{0.49\textwidth}
        \centering
        \includegraphics[height=3.2cm, keepaspectratio]{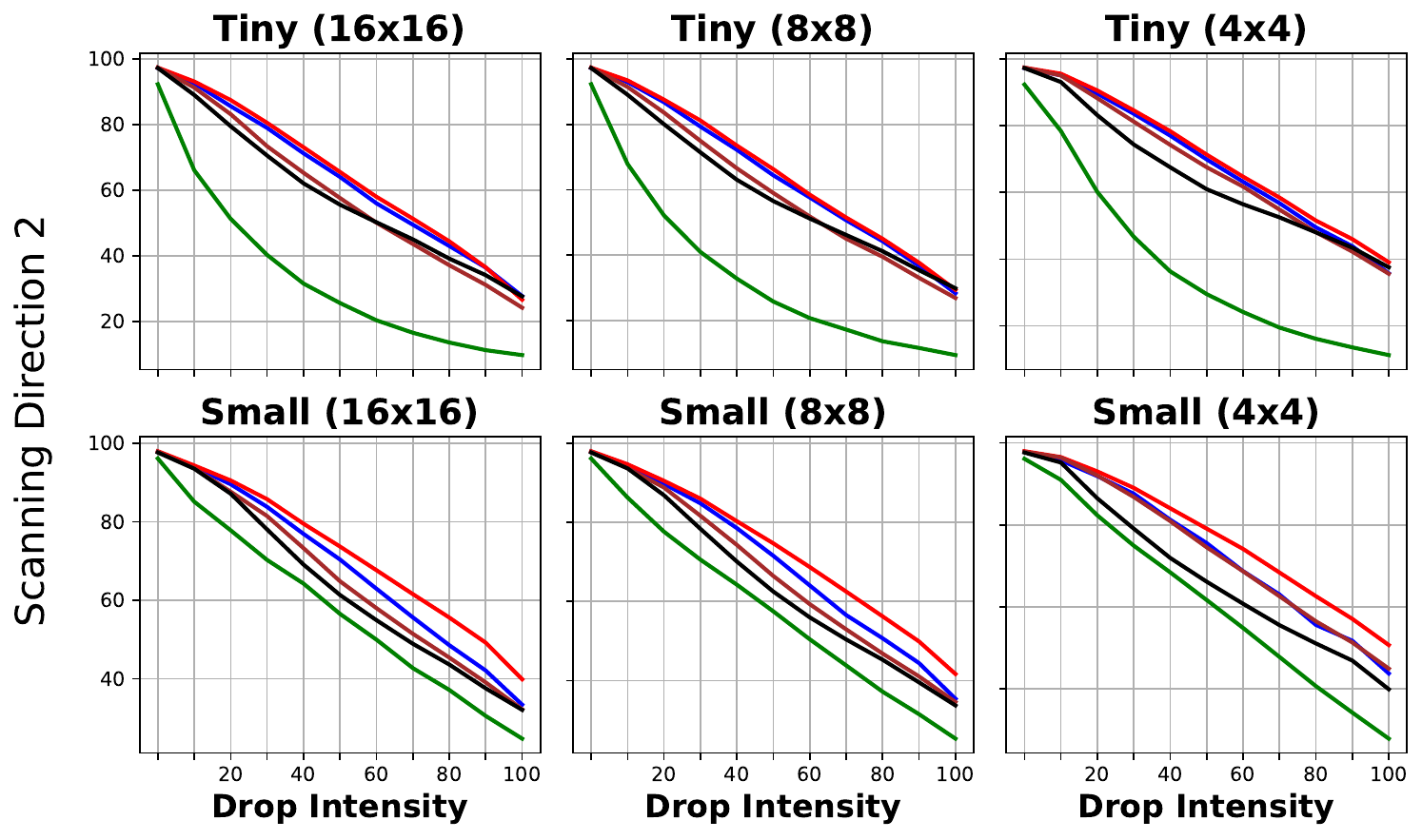}
    \end{minipage}
\end{minipage}
\hfill
\begin{minipage}{0.29\textwidth}
    \caption{\small Information drop of Tiny and Small family of models along the scanning direction: the image is split into a sequence of fixed-size non-overlapping patches of size 16x16, 8x8, and 4x4. The first row shows the results of linearly increasing the number of pixels dropped from each patch to the maximum threshold (\emph{Drop Intensity}) along different scanning directions. The bottom row presents results of linearly increasing the number of pixels dropped from each patch to the maximum threshold (\emph{Drop Intensity}) until the center of the scanning direction. More detailed analysis is provided in Section \ref{app:scandrop} of the Appendix.}
    \label{fig:scan_exp}

\end{minipage}
\end{figure*}

\subsubsection{\textbf{Robustness against Information Drop}} 
\underline{\textbf{Information Drop along the Scanning Axis:}} 
VSSM models scan image patches sequentially along four paths (top-bottom, bottom-top, left-right, right-left) to capture spatial information. To study the effectiveness of this 2D-Selective Scan operation, we investigate the models' response to information drop along these scanning directions. We consider two settings: (1) linearly increasing information drop along the scanning direction, with maximum drop in last-scanned patches, and (2) linearly increasing drop from start to the center, then linearly decreasing until the end.

We split the image into $n\times n$ non-overlapping patches and perform the information drop experiments, with the maximum drop in a patch (\emph{Drop Intensity}) varying from 10$\%$ to 100$\%$. In Fig. \ref{fig:scan_exp}, we report results on information drop along top-to-bottom \emph{(Direction 1)} and left-to-right \emph{(Direction 2)} scanning directions. For both settings (1) and (2) we observe that VMamba and ConvNext models show high robustness to sequential information drop across various thresholds. Overall, \textit{`T'} and \textit{`S'} versions of VMmamba  model demonstrate superior performance compared to their counterparts across different patch sizes. Pure transformer-based ViT models show poor performance in this experimental setup. We also observe that as we reduce the patch size for splitting the image, leading to a gradual loss of information in the scanning direction, the performance of all the models improves. This implies that handling more abrupt information loss in fewer and larger patches is challenging for these models. In Section \ref{app:scandrop} of Appendix, we expand the analysis across base models and all the scanning directions with varying number of patch sizes. \textit{\textbf{Overall, we observe that both VMamba and ConvNext are more adept at handling sequential drop of information along different scanning directions, compared to hybrid MambaVision, ViT and Swin Models.}}

\noindent \underline{\textbf{Random Patch Drop:}}
We assess the robustness of VSSMs in occluded scenarios by randomly dropping patches from images. We split the image into $n\times n$ patches and randomly select the patches whose values will be set to zero. As shown in Tab.\ref{tab:random_drop_main} (\textit{top}), when image is split into $16 \times 16$ patches, VSSMs consistently outperform MambaVision, ResNet, ConvNeXt, and ViT models in maintaining accuracy with increasing numbers of dropped patches. However, under conditions of extreme spatial information loss, Swin models demonstrate superior performance, whereas the hybrid architecture-based MambaVision performs worst. \textit{\textbf{This trend persists when the image is split into $8 \times 8$ patches, as illustrated in Tab.\ref{tab:random_drop_main} (\textit{bottom}), highlighting the robustness of VSSMs and the exceptional resilience of Swin models in extreme patch drop scenarios.}} Results on different patch sizes showing similar trend are reported in Appendix \ref{app:randomdrop}.

\begin{table*}[t]
\centering
\caption{\small Top-1 classification accuracy of various architectures under random patch drop occlusions.}
\label{tab:random_drop_main}
\setlength{\tabcolsep}{2.0pt}
\resizebox{\linewidth}{!}{
\begin{tabular}{|c c c c c c c c c c c c c c c c|}
\hline
\rowcolor{LightCyan}
\rotatebox{0}{\scriptsize \textbf{ResNet-50}} & \rotatebox{0}{\scriptsize \textbf{ConvNext-T}} & \rotatebox{0}{\scriptsize \textbf{ConvNext-S}} & \rotatebox{0}{\scriptsize \textbf{ConvNext-B}} & \rotatebox{0}{\scriptsize \textbf{ViT-T}} & \rotatebox{0}{\scriptsize \textbf{ViT-S}} & \rotatebox{0}{\scriptsize \textbf{ViT-B}} & \rotatebox{0}{\scriptsize \textbf{VMamba-T}} & \rotatebox{0}{\scriptsize \textbf{VMamba-S}} & \rotatebox{0}{\scriptsize \textbf{VMamba-B}} & \rotatebox{0}{\scriptsize \textbf{MambaVision-T}}& \rotatebox{0}{\scriptsize \textbf{MambaVision-S}}& \rotatebox{0}{\scriptsize \textbf{MambaVision-B}} & \rotatebox{0}{\scriptsize \textbf{Swin-T}} & \rotatebox{0}{\scriptsize \textbf{Swin-S}} & \rotatebox{0}{\scriptsize \textbf{Swin-B}} \\
\hline

\rowcolor{gray!5}
\multicolumn{16}{|c|}{Patch Size $16\times16$ \emph{(Percentage of patch drop increasing from top to bottom (10\% to 90\%))}} \\
\hline

\heatmapcolor{96.70} & \heatmapcolor{97.24}  & \heatmapcolor{97.78}  & \heatmapcolor{97.84}  & \heatmapcolor{92.30}  & \heatmapcolor{96.08}  & \heatmapcolor{97.54}  & \heatmapcolor{97.38}  & \heatmapcolor{97.94}  & \heatmapcolor{97.96}  & \heatmapcolor{97.36} & \heatmapcolor{97.82} & \heatmapcolor{97.60} & \heatmapcolor{97.24}  & \heatmapcolor{97.60} & \heatmapcolor{97.60}  \\
\hline

\heatmapcolor{75.27}  & \heatmapcolor{96.49}  & \heatmapcolor{97.19}  & \heatmapcolor{97.37}  & \heatmapcolor{90.83}  & \heatmapcolor{95.39}  & \heatmapcolor{96.85}  & \heatmapcolor{96.49}  & \heatmapcolor{96.61}  & \heatmapcolor{97.25}  & \heatmapcolor{96.24} & \heatmapcolor{96.80} & \heatmapcolor{97.09} & \heatmapcolor{96.76}  & \heatmapcolor{97.38}  & \heatmapcolor{97.32}   \\
\hline

\heatmapcolor{39.93}  & \heatmapcolor{94.63}  & \heatmapcolor{95.27}  & \heatmapcolor{96.48}  & \heatmapcolor{88.09}  & \heatmapcolor{94.29}  & \heatmapcolor{96.27}  & \heatmapcolor{95.16}  & \heatmapcolor{92.97}  & \heatmapcolor{96.25}  & \heatmapcolor{92.91} & \heatmapcolor{94.67} & \heatmapcolor{95.74} & \heatmapcolor{96.11}  & \heatmapcolor{96.84}  & \heatmapcolor{96.79}   \\
\hline

\heatmapcolor{17.91}  & \heatmapcolor{89.99}  & \heatmapcolor{91.12}  & \heatmapcolor{95.29}  & \heatmapcolor{85.26}  & \heatmapcolor{92.35}  & \heatmapcolor{95.08}  & \heatmapcolor{93.45}  & \heatmapcolor{89.74}  & \heatmapcolor{95.21}   & \heatmapcolor{87.14} & \heatmapcolor{91.19} & \heatmapcolor{93.25}& \heatmapcolor{94.88}  & \heatmapcolor{95.88}  & \heatmapcolor{96.17}  \\
\hline

\heatmapcolor{6.73}  & \heatmapcolor{81.43} & \heatmapcolor{84.63} & \heatmapcolor{93.03}  & \heatmapcolor{80.08}  & \heatmapcolor{90.15}  & \heatmapcolor{92.78}  & \heatmapcolor{90.52}  & \heatmapcolor{84.82}  & \heatmapcolor{93.46} & \heatmapcolor{77.02} & \heatmapcolor{84.56} & \heatmapcolor{88.05} & \heatmapcolor{93.38}  & \heatmapcolor{94.35} & \heatmapcolor{95.03}    \\
\hline
\heatmapcolor{2.43}  & \heatmapcolor{70.07} & \heatmapcolor{74.44} & \heatmapcolor{88.76}  & \heatmapcolor{72.49}  & \heatmapcolor{85.10}  & \heatmapcolor{89.21}  & \heatmapcolor{86.52}  & \heatmapcolor{78.41}  & \heatmapcolor{90.89}  & \heatmapcolor{61.79} & \heatmapcolor{76.07} & \heatmapcolor{78.92}& \heatmapcolor{91.05}  & \heatmapcolor{92.21} & \heatmapcolor{93.25}    \\
\hline
\heatmapcolor{1.05}  & \heatmapcolor{57.59} & \heatmapcolor{60.15} & \heatmapcolor{82.35}  & \heatmapcolor{61.34}  & \heatmapcolor{76.56}  & \heatmapcolor{82.08}  & \heatmapcolor{80.52}  & \heatmapcolor{68.40}  & \heatmapcolor{87.03}  & \heatmapcolor{42.16} & \heatmapcolor{62.44} & \heatmapcolor{63.89}& \heatmapcolor{87.96}  & \heatmapcolor{87.84} & \heatmapcolor{90.43}    \\
\hline
\heatmapcolor{0.56}  & \heatmapcolor{44.67} & \heatmapcolor{44.71} & \heatmapcolor{71.25}  & \heatmapcolor{45.63}  & \heatmapcolor{62.63}  & \heatmapcolor{70.25}  & \heatmapcolor{70.39}  & \heatmapcolor{52.72}  & \heatmapcolor{79.96}  & \heatmapcolor{20.73} & \heatmapcolor{42.16} & \heatmapcolor{43.33}& \heatmapcolor{80.65}  & \heatmapcolor{79.71} & \heatmapcolor{84.70}    \\
\hline
\heatmapcolor{0.45}  & \heatmapcolor{31.29} & \heatmapcolor{28.82} & \heatmapcolor{57.71}  & \heatmapcolor{25.86}  & \heatmapcolor{41.73}  & \heatmapcolor{50.08}  & \heatmapcolor{56.23}  & \heatmapcolor{34.51}  & \heatmapcolor{67.56}  & \heatmapcolor{5.62} & \heatmapcolor{21.07} & \heatmapcolor{22.21}& \heatmapcolor{70.37}  & \heatmapcolor{66.60} & \heatmapcolor{74.38}    \\
\hline
\heatmapcolor{0.43}  & \heatmapcolor{16.73} & \heatmapcolor{14.98} & \heatmapcolor{33.67}  & \heatmapcolor{7.85}  & \heatmapcolor{15.86}  & \heatmapcolor{19.68}  & \heatmapcolor{34.83}  & \heatmapcolor{16.82}  & \heatmapcolor{41.55} & \heatmapcolor{1.95} & \heatmapcolor{7.08} & \heatmapcolor{11.08} & \heatmapcolor{47.16}  & \heatmapcolor{47.85} & \heatmapcolor{53.54}    \\
\hline

\rowcolor{gray!5}
\multicolumn{16}{|c|}{Patch Size $8\times8$ \emph{(Percentage of patch drop increasing from top to bottom (10\% to 90\%))}} \\
\hline
\heatmapcolor{96.70} & \heatmapcolor{97.24}  & \heatmapcolor{97.78}  & \heatmapcolor{97.84}  & \heatmapcolor{92.32}  & \heatmapcolor{96.08}  & \heatmapcolor{97.54}  & \heatmapcolor{97.38}  & \heatmapcolor{97.94}  & \heatmapcolor{97.96}  & \heatmapcolor{97.36} & \heatmapcolor{97.82} & \heatmapcolor{97.60} & \heatmapcolor{97.24}  & \heatmapcolor{97.60} & \heatmapcolor{97.60}  \\
\hline

\heatmapcolor{44.91}  & \heatmapcolor{86.69}  & \heatmapcolor{91.39}  & \heatmapcolor{95.63}  & \heatmapcolor{70.18}  & \heatmapcolor{88.90}  & \heatmapcolor{94.23}  & \heatmapcolor{87.86}  & \heatmapcolor{85.93}  & \heatmapcolor{90.20}  & \heatmapcolor{89.57} & \heatmapcolor{91.25} & \heatmapcolor{92.32} & \heatmapcolor{96.44}  & \heatmapcolor{96.77}  & \heatmapcolor{96.76}   \\
\hline

\heatmapcolor{12.44}  & \heatmapcolor{68.38}  & \heatmapcolor{81.13}  & \heatmapcolor{90.83}  & \heatmapcolor{42.19}  & \heatmapcolor{79.72}  & \heatmapcolor{88.37}  & \heatmapcolor{79.91}  & \heatmapcolor{78.23}  & \heatmapcolor{84.43}  & \heatmapcolor{73.34} & \heatmapcolor{79.03} & \heatmapcolor{84.32} & \heatmapcolor{95.04}  & \heatmapcolor{94.87}  & \heatmapcolor{95.84}   \\
\hline

\heatmapcolor{3.79}  & \heatmapcolor{55.12}  & \heatmapcolor{68.35}  & \heatmapcolor{84.08}  & \heatmapcolor{16.91}  & \heatmapcolor{65.39}  & \heatmapcolor{76.63}  & \heatmapcolor{70.40}  & \heatmapcolor{70.95}  & \heatmapcolor{78.47}  & \heatmapcolor{51.09} & \heatmapcolor{60.21} & \heatmapcolor{71.70} & \heatmapcolor{92.87}  & \heatmapcolor{92.08}  & \heatmapcolor{94.49}  \\
\hline

\heatmapcolor{1.35}  & \heatmapcolor{39.05} & \heatmapcolor{54.58} & \heatmapcolor{73.51}  & \heatmapcolor{4.59}  & \heatmapcolor{ 46.17}  & \heatmapcolor{60.12}  & \heatmapcolor{57.34}  & \heatmapcolor{59.09}  & \heatmapcolor{70.04} & \heatmapcolor{29.64} & \heatmapcolor{39.81} & \heatmapcolor{55.01} & \heatmapcolor{90.13}  & \heatmapcolor{88.21} & \heatmapcolor{92.40}    \\
\hline
\heatmapcolor{0.56}  & \heatmapcolor{23.89} & \heatmapcolor{37.94} & \heatmapcolor{58.05}  & \heatmapcolor{1.25}  & \heatmapcolor{25.91}  & \heatmapcolor{39.97}  & \heatmapcolor{43.09}  & \heatmapcolor{44.08}  & \heatmapcolor{58.29}  & \heatmapcolor{13.73} & \heatmapcolor{23.86} & \heatmapcolor{34.67}& \heatmapcolor{85.40}  & \heatmapcolor{81.87} & \heatmapcolor{88.36}    \\
\hline
\heatmapcolor{0.33}  & \heatmapcolor{13.33} & \heatmapcolor{21.97} & \heatmapcolor{40.37}  & \heatmapcolor{0.45}  & \heatmapcolor{11.03}  & \heatmapcolor{21.56}  & \heatmapcolor{28.25}  & \heatmapcolor{27.85}  & \heatmapcolor{43.00}  & \heatmapcolor{4.90} & \heatmapcolor{11.86} & \heatmapcolor{16.70}& \heatmapcolor{78.76}  & \heatmapcolor{71.63} & \heatmapcolor{81.85}    \\
\hline
\heatmapcolor{0.21}  & \heatmapcolor{5.95} & \heatmapcolor{9.85} & \heatmapcolor{21.51}  & \heatmapcolor{0.21}  & \heatmapcolor{3.78}  & \heatmapcolor{8.85}  & \heatmapcolor{14.69}  & \heatmapcolor{12.45}  & \heatmapcolor{25.75}  & \heatmapcolor{1.51} & \heatmapcolor{4.59} & \heatmapcolor{5.90} & \heatmapcolor{68.40}  & \heatmapcolor{53.81} & \heatmapcolor{70.03}    \\
\hline
\heatmapcolor{0.24}  & \heatmapcolor{2.08} & \heatmapcolor{2.31} & \heatmapcolor{7.85}  & \heatmapcolor{0.14}  & \heatmapcolor{1.02}  & \heatmapcolor{ 2.49}  & \heatmapcolor{5.11}  & \heatmapcolor{2.45}  & \heatmapcolor{10.07} & \heatmapcolor{0.50} & \heatmapcolor{1.06} & \heatmapcolor{1.16} & \heatmapcolor{52.41}  & \heatmapcolor{29.58} & \heatmapcolor{49.35}    \\
\hline
\heatmapcolor{0.25}  & \heatmapcolor{0.46} & \heatmapcolor{0.56} & \heatmapcolor{1.33}  & \heatmapcolor{0.16}  & \heatmapcolor{0.39}  & \heatmapcolor{0.59}  & \heatmapcolor{0.75}  & \heatmapcolor{0.43}  & \heatmapcolor{1.65} & \heatmapcolor{0.22} & \heatmapcolor{0.31} & \heatmapcolor{0.21} & \heatmapcolor{28.46}  & \heatmapcolor{8.32} & \heatmapcolor{23.67}    \\
\hline

\end{tabular}
}
\end{table*}

\begin{table*}[t]
\centering
\caption{\small  Top-1 classification accuracy reported under salient patch drop  occlusion and patch shuffling.
}
\label{tab:random_dino_drop_and_shuffle_main}
\setlength{\tabcolsep}{2.0pt}
\resizebox{\linewidth}{!}{
\begin{tabular}{|c c c c c c c c c c c c c c c c|}
\hline
\rowcolor{LightCyan}
\rotatebox{0}{\scriptsize \textbf{ResNet-50}} & \rotatebox{0}{\scriptsize \textbf{ConvNext-T}} & \rotatebox{0}{\scriptsize \textbf{ConvNext-S}} & \rotatebox{0}{\scriptsize \textbf{ConvNext-B}} & \rotatebox{0}{\scriptsize \textbf{ViT-T}} & \rotatebox{0}{\scriptsize \textbf{ViT-S}} & \rotatebox{0}{\scriptsize \textbf{ViT-B}} & \rotatebox{0}{\scriptsize \textbf{VMamba-T}} & \rotatebox{0}{\scriptsize \textbf{VMamba-S}} & \rotatebox{0}{\scriptsize \textbf{VMamba-B}} & \rotatebox{0}{\scriptsize \textbf{MambaVision-T}}& \rotatebox{0}{\scriptsize \textbf{MambaVision-S}}& \rotatebox{0}{\scriptsize \textbf{MambaVision-B}} & \rotatebox{0}{\scriptsize \textbf{Swin-T}} & \rotatebox{0}{\scriptsize \textbf{Swin-S}} & \rotatebox{0}{\scriptsize \textbf{Swin-B}} \\
\hline
\rowcolor{gray!5}
\multicolumn{16}{|c|}{Salient Patch Drop \emph{(Percentage of patch drop from top to bottom(10\% to 100\%))}} \\
\hline

\heatmapcolor{92.70} & \heatmapcolor{96.88}  & \heatmapcolor{97.38}  & \heatmapcolor{97.50}  & \heatmapcolor{90.46}  & \heatmapcolor{95.36}  & \heatmapcolor{94.90}  & \heatmapcolor{97.04}  & \heatmapcolor{97.66}  & \heatmapcolor{97.40} & \heatmapcolor{96.88}  & \heatmapcolor{97.24} & \heatmapcolor{97.20} & \heatmapcolor{96.94}  & \heatmapcolor{97.44} & \heatmapcolor{97.26}  \\
\hline

\heatmapcolor{84.86}  & \heatmapcolor{95.98}  & \heatmapcolor{96.98}  & \heatmapcolor{96.86}  & \heatmapcolor{88.62}  & \heatmapcolor{94.56}  & \heatmapcolor{93.92}  & \heatmapcolor{96.58}  & \heatmapcolor{96.98}  & \heatmapcolor{97.14} & \heatmapcolor{95.78}  & \heatmapcolor{96.30} & \heatmapcolor{96.62} & \heatmapcolor{96.14}  & \heatmapcolor{97.02}  & \heatmapcolor{96.90}   \\
\hline

\heatmapcolor{73.64}  & \heatmapcolor{94.60}  & \heatmapcolor{95.86}  & \heatmapcolor{96.52}  & \heatmapcolor{85.40}  & \heatmapcolor{92.70}  & \heatmapcolor{92.52}  & \heatmapcolor{95.34}  & \heatmapcolor{95.82}  & \heatmapcolor{96.74} & \heatmapcolor{93.56}  & \heatmapcolor{94.82} & \heatmapcolor{95.34} & \heatmapcolor{95.02}  & \heatmapcolor{96.22}  & \heatmapcolor{96.24}   \\
\hline

\heatmapcolor{60.16}  & \heatmapcolor{92.52}  & \heatmapcolor{94.04}  & \heatmapcolor{94.92}  & \heatmapcolor{80.94}  & \heatmapcolor{89.82}  & \heatmapcolor{89.72}  & \heatmapcolor{93.88}  & \heatmapcolor{93.14}  & \heatmapcolor{95.04} & \heatmapcolor{90.08}  & \heatmapcolor{91.88} & \heatmapcolor{93.62} & \heatmapcolor{93.28}  & \heatmapcolor{94.92}  & \heatmapcolor{94.76}  \\
\hline

\heatmapcolor{44.78}  & \heatmapcolor{88.22} & \heatmapcolor{90.06} & \heatmapcolor{92.60}  & \heatmapcolor{75.12}  & \heatmapcolor{85.26}  & \heatmapcolor{85.34}  & \heatmapcolor{91.04}  & \heatmapcolor{89.64}  & \heatmapcolor{92.74} & \heatmapcolor{85.00}  & \heatmapcolor{87.90} & \heatmapcolor{89.84}& \heatmapcolor{90.24}  & \heatmapcolor{92.74} & \heatmapcolor{92.98}    \\
\hline
\heatmapcolor{29.16}  & \heatmapcolor{81.40} & \heatmapcolor{84.02} & \heatmapcolor{88.64}  & \heatmapcolor{65.42}  & \heatmapcolor{78.06}  & \heatmapcolor{78.36}  & \heatmapcolor{86.34}  & \heatmapcolor{83.66}  & \heatmapcolor{89.30}& \heatmapcolor{76.24}  & \heatmapcolor{80.32} & \heatmapcolor{83.98} & \heatmapcolor{86.76}  & \heatmapcolor{88.88} & \heatmapcolor{89.78}    \\
\hline
\heatmapcolor{16.12}  & \heatmapcolor{72.32} & \heatmapcolor{74.64} & \heatmapcolor{82.36}  & \heatmapcolor{51.60}  & \heatmapcolor{67.60}  & \heatmapcolor{68.34}  & \heatmapcolor{79.24}  & \heatmapcolor{74.26}  & \heatmapcolor{83.44} & \heatmapcolor{63.44}  & \heatmapcolor{70.16} & \heatmapcolor{74.80}& \heatmapcolor{80.32}  & \heatmapcolor{82.08} & \heatmapcolor{83.54}    \\
\hline
\heatmapcolor{7.38}  & \heatmapcolor{56.96} & \heatmapcolor{58.28} & \heatmapcolor{69.44}  & \heatmapcolor{35.10}  & \heatmapcolor{50.18}  & \heatmapcolor{50.80}  & \heatmapcolor{66.42}  & \heatmapcolor{59.26}  & \heatmapcolor{72.88} & \heatmapcolor{45.52}  & \heatmapcolor{52.86} & \heatmapcolor{59.08}& \heatmapcolor{68.84}  & \heatmapcolor{71.18} & \heatmapcolor{73.30}    \\
\hline
\heatmapcolor{2.14}  & \heatmapcolor{33.80} & \heatmapcolor{34.96} & \heatmapcolor{46.52}  & \heatmapcolor{14.96}  & \heatmapcolor{24.26}  & \heatmapcolor{24.72}  & \heatmapcolor{43.90}  & \heatmapcolor{35.38}  & \heatmapcolor{50.94}& \heatmapcolor{22.26}  & \heatmapcolor{29.24} & \heatmapcolor{35.12} & \heatmapcolor{46.42}  & \heatmapcolor{49.58} & \heatmapcolor{51.80}    \\
\hline
\heatmapcolor{0.10}  & \heatmapcolor{0.10} & \heatmapcolor{0.10} & \heatmapcolor{0.10}  & \heatmapcolor{0.10}  & \heatmapcolor{0.10}  & \heatmapcolor{0.10}  & \heatmapcolor{0.10}  & \heatmapcolor{0.10}  & \heatmapcolor{0.10} & \heatmapcolor{0.10}  & \heatmapcolor{0.10} & \heatmapcolor{0.10} & \heatmapcolor{0.10}  & \heatmapcolor{0.10} & \heatmapcolor{0.10}    \\
\hline
\rowcolor{gray!5}
\multicolumn{16}{|c|}{Patch Shuffling \emph{(From top to bottom, the image is split into $4$, $8$, $16$, $32$, $64$, and $256$ patches)}} \\
\hline

\heatmapcolor{90.79} & \heatmapcolor{95.62}  & \heatmapcolor{96.57}  & \heatmapcolor{96.69}  & \heatmapcolor{85.71}  & \heatmapcolor{92.88}  & \heatmapcolor{95.18}  & \heatmapcolor{95.75}  & \heatmapcolor{96.73}  & \heatmapcolor{96.88}  & \heatmapcolor{95.71}  & \heatmapcolor{96.33}  & \heatmapcolor{96.75} & \heatmapcolor{95.41}  & \heatmapcolor{96.05} & \heatmapcolor{96.31}  \\
\hline

\heatmapcolor{82.01}  & \heatmapcolor{93.61}  & \heatmapcolor{94.81}  & \heatmapcolor{95.43}  & \heatmapcolor{78.67}  & \heatmapcolor{88.81}  & \heatmapcolor{92.55}  & \heatmapcolor{94.66}  & \heatmapcolor{95.75}  & \heatmapcolor{96.12}  & \heatmapcolor{93.94}  & \heatmapcolor{95.15}  & \heatmapcolor{95.69} & \heatmapcolor{93.99}  & \heatmapcolor{94.55}  & \heatmapcolor{95.25}   \\
\hline

\heatmapcolor{67.09}  & \heatmapcolor{89.99}  & \heatmapcolor{91.66}  & \heatmapcolor{93.05}  & \heatmapcolor{67.95}  & \heatmapcolor{81.22}  & \heatmapcolor{87.88}  & \heatmapcolor{91.65}  & \heatmapcolor{93.26}  & \heatmapcolor{94.27} & \heatmapcolor{91.12}  & \heatmapcolor{92.40}  & \heatmapcolor{93.48}  & \heatmapcolor{90.69}  & \heatmapcolor{91.93}  & \heatmapcolor{92.77}   \\
\hline

\heatmapcolor{33.34}  & \heatmapcolor{80.29}  & \heatmapcolor{82.06}  & \heatmapcolor{85.31}  & \heatmapcolor{42.65}  & \heatmapcolor{63.97}  & \heatmapcolor{75.68}  & \heatmapcolor{85.34}  & \heatmapcolor{87.10}  & \heatmapcolor{90.08}  & \heatmapcolor{80.51}  & \heatmapcolor{84.34}  & \heatmapcolor{86.98} & \heatmapcolor{85.01}  & \heatmapcolor{86.33}  & \heatmapcolor{88.18}  \\
\hline

\heatmapcolor{10.93}  & \heatmapcolor{63.33} & \heatmapcolor{64.75} & \heatmapcolor{69.44}  & \heatmapcolor{19.05}  & \heatmapcolor{45.21}  & \heatmapcolor{55.52}  & \heatmapcolor{73.72}  & \heatmapcolor{73.68}  & \heatmapcolor{80.79}  & \heatmapcolor{63.59}  & \heatmapcolor{68.87}  & \heatmapcolor{73.87}& \heatmapcolor{76.60}  & \heatmapcolor{75.46} & \heatmapcolor{80.10}    \\
\hline
\heatmapcolor{1.07}  & \heatmapcolor{7.57} & \heatmapcolor{5.45} & \heatmapcolor{8.02}  & \heatmapcolor{1.00}  & \heatmapcolor{3.14}  & \heatmapcolor{5.00}  & \heatmapcolor{15.45}  & \heatmapcolor{13.83}  & \heatmapcolor{22.50}  & \heatmapcolor{3.50}  & \heatmapcolor{5.09}  & \heatmapcolor{4.62} & \heatmapcolor{25.79}  & \heatmapcolor{12.83} & \heatmapcolor{21.08}    \\
\hline

\end{tabular}
}
\end{table*}

\noindent \underline{\textbf{Salient and Non-Salient Patch Drop:}}
We evaluate the robustness of VSSMs against salient (foreground) and non-salient (background) patch drop. Using a self-supervised ViT model, DINO~\cite{caron2021emerging}, we effectively segment salient objects by exploiting the spatial positions of information flowing into the final feature vector within the last attention block. This allows us to control the amount of salient information captured within the selected patches by thresholding. Similarly, we also select the least salient regions of the image and drop the patches containing the lowest foreground information~\cite{naseer2021intriguing}. Similar to \cite{naseer2021intriguing}, the patch size is fixed to $16 \times 16$ for this experiment. Tab.~\ref{tab:random_dino_drop_and_shuffle_main} \emph{(top)} shows that \textit{\textbf{VSSM models, including VMamba and MambaVision, demonstrate notable robustness when foreground content is removed, outperforming both convolutional (ResNet and ConvNeXt) and the ViT transformer family. Their performance is on par with the Swin family until a 50$\%$ salient patch drop, beyond which Swin transformers exhibit better robustness}}, maintaining higher accuracy compared to VSSMs. The trend for non-salient patch drops is similar and is shown in the Appendix~\ref{app:dinodrop}.

\noindent \underline{\textbf{Patch Shuffling:}}
VSSMs process images as a sequence of patches, and the order of these patches represents the overall image structure and global composition. To evaluate the robustness of VSSMs to patch permutations, we define a shuffling operation on the input image patches, which destroys the image structure by changing the order of the patch sequence. Based on the dimensions of the patch size, we split the image into either $4,8,16,32,64$ and $256$ patches, which is then followed by random shuffling of the patches to evaluate the performance of models on spatial rearrangement of information. In Tab. ~\ref{tab:random_dino_drop_and_shuffle_main} \emph{(bottom)}, we observe that VMamba family overall performs better than other models when the spatial structure of the input image is disturbed. \textit{\textbf{VMamba models generally demonstrate greater resilience to spatial structure disturbances than Swin models.}}



\subsubsection{Robustness against ImageNet Corruptions}

To evaluate the robustness of VSSMs in real-world scenarios, we performed experiments on several corrupted versions of the ImageNet dataset, which can be grouped into two categories based on the type of changes they introduce to the images. The \textbf{first category} includes datasets that make overall global compositional changes, such as \textit{\textbf{ImageNet-C}}~\cite{hendrycks2019benchmarking}, which evaluates robustness against 19 common distortions across five categories (noise, blur, weather, and digital-based corruptions) with varying severity levels. Additionally, datasets like \textit{\textbf{ImageNetV2}}~\cite{recht2019imagenet}, \textit{\textbf{ImageNet-A}}~\cite{hendrycks2021natural}, \textit{\textbf{ImageNet-R}}~\cite{hendrycks2021many}, and \textit{\textbf{ImageNet-S}}~\cite{gao2022large} also fall into this category, as they introduce domain shifts that affect the global composition of the images. For details of these datasets, see Appendix~\ref{app:imagent_domain}. 
The \textbf{second category} consists of datasets that provide more control in editing the images and focus on fine-grained details. \textit{\textbf{ImageNet-E}}~\cite{li2023imagenet} assesses classifiers' robustness to changes in background, object size, position, and direction, while \textit{\textbf{ImageNet-B}}~\cite{malik2024objectcompose} introduces diverse object-to-background changes using text-to-image, image-to-text, and image-to-segment models, preserving original object semantics while varying backgrounds to include both natural and adversarial changes. These datasets use diffusion models to generate various object-to-background compositional changes to test the resilience of the models.

\begin{table}[t]
 \centering
    \caption{\small Top-1 classification accuracy for the domain generalization setting across various architectures and datasets. Models trained on ImageNet are evaluated on datasets with domain shifts.} 
        \label{tab:robustness}
    \small
 \setlength{\tabcolsep}{12.0pt}
    \resizebox{\linewidth}{!}{
    \begin{tabular}{|l|c|cl|}
    \toprule
    \rowcolor{LightCyan}
    Model $\downarrow$ & ImageNet   & OOD Average($\uparrow$) & mCE($\downarrow$)\\
    \midrule
     \cellcolor{gray!15}ConvNext-T  &  \cellcolor{gray!5}81.87 & 67.55 &  36.90\dec{44.97}  \\
    \cellcolor{gray!15}ViT-T& \cellcolor{gray!5}75.35 & 153.7 &  26.88\dec{48.47}  \\
    \cellcolor{gray!15}Swin-T  & \cellcolor{gray!5}80.91 & 77.86 &  34.00\dec{46.91}  \\
    \cellcolor{gray!15}VMamba-T & \cellcolor{gray!5}82.28 & 65.05 &  37.32\dec{44.96}  \\
     \cellcolor{gray!15}MambaVision-T & \cellcolor{gray!5}82.10 & 66.58 &  37.66\dec{44.44}  \\
    \midrule
     \cellcolor{gray!15}ConvNext-S & \cellcolor{gray!5}82.82  & 59.52 &  39.81\dec{43.01} \\
    \cellcolor{gray!15}ViT-S & \cellcolor{gray!5}81.40 & 79.28 v&  36.74\dec{44.66}  \\
    \cellcolor{gray!15}Swin-S  & \cellcolor{gray!5}82.90 & 63.04 &  37.81\dec{45.09}  \\
    \cellcolor{gray!15}VMamba-S  & \cellcolor{gray!5}83.48 & 53.08 &  40.55\dec{42.93} \\
    \cellcolor{gray!15}MambaVision-S & \cellcolor{gray!5}83.22 &   50.50 &39.52\dec{43.70}  \\
    \midrule
    \cellcolor{gray!15}ConvNext-B & \cellcolor{gray!5}83.75 &   56.92 & 41.68\dec{42.07}  \\
    \cellcolor{gray!15}ViT-B & \cellcolor{gray!5}84.40 &   42.26 &45.49\dec{38.91}  \\
    \cellcolor{gray!15}Swin-B  & \cellcolor{gray!5}83.08 &   64.98 &38.97\dec{44.11}  \\
    \cellcolor{gray!15}VMamba-B  & \cellcolor{gray!5}83.76  & 53.76 &  41.52\dec{42.24} \\
            \cellcolor{gray!15}MambaVision-B & \cellcolor{gray!5}83.96  & 49.61 &  41.98\dec{41.98} \\
    \bottomrule
    \end{tabular}}
\end{table}

\begin{table*}[t]
\centering
\caption{\small Top-1 classification accuracy of various architectures on the ImageNet-E dataset~\cite{li2023imagenet} (\textit{left}) and ImageNet-B dataset~\cite{malik2024objectcompose} (\textit{right}). }

\label{tab:imagenet_e_b_main}
\setlength{\tabcolsep}{2.0pt}
\resizebox{\linewidth}{!}{
\begin{tabular}{|l| c c c c c c c c c | c c c c c|}
\hline

\rowcolor{LightCyan}
\multicolumn{1}{|l|}{\textbf{Dataset} $\rightarrow$} & \multicolumn{9}{|c|}{\textbf{ImageNet-E}} & \multicolumn{5}{|c|}{\textbf{ImageNet-B}} \\
\hline
\rowcolor{gray!5}
\rotatebox{0}{\textbf{Model} $\downarrow$} & \rotatebox{0}{\scriptsize $\lambda=-20$} & \rotatebox{0}{\scriptsize $\lambda=20$} & \rotatebox{0}{\scriptsize $\lambda=20$(adv)} & \rotatebox{0}{\scriptsize Random-BG} & \rotatebox{0}{\scriptsize $0.1$} & \rotatebox{0}{\scriptsize $0.08$} & \rotatebox{0}{\scriptsize $0.05$} & \rotatebox{0}{\scriptsize Random Pos.} & \rotatebox{0}{\scriptsize Original}  & \rotatebox{0}{\scriptsize Original}  & \rotatebox{0}{\scriptsize Caption}  & \rotatebox{0}{\scriptsize Class}  & \rotatebox{0}{\scriptsize Color} & \rotatebox{0}{\scriptsize Texture}  \\
\hline

ResNet-50 & \heatmapcolor{88.74} & \heatmapcolor{86.76}  & \heatmapcolor{73.02}  & \heatmapcolor{84.05}  & \heatmapcolor{89.19}  & \heatmapcolor{86.60}  & \heatmapcolor{77.34}  & \heatmapcolor{73.30}  & \heatmapcolor{94.55} & \heatmapcolor{98.60} & \heatmapcolor{94.00}  & \heatmapcolor{96.60}  & \heatmapcolor{88.20}  & \heatmapcolor{85.70}   \\
\hline

ConvNext-T & \heatmapcolor{90.95}  & \heatmapcolor{90.03}  & \heatmapcolor{76.88}  & \heatmapcolor{88.09}  & \heatmapcolor{93.01}  & \heatmapcolor{90.87}  & \heatmapcolor{83.09}  & \heatmapcolor{80.19}  & \heatmapcolor{96.09} & \heatmapcolor{98.20}  & \heatmapcolor{93.20}  & \heatmapcolor{95.10}  & \heatmapcolor{88.80}  & \heatmapcolor{87.40}    \\
\hline


ConvNext-S & \heatmapcolor{91.96}  & \heatmapcolor{90.76}  & \heatmapcolor{78.52}  & \heatmapcolor{88.99}  & \heatmapcolor{93.61}  & \heatmapcolor{91.66}  & \heatmapcolor{85.34}  & \heatmapcolor{82.19}  & \heatmapcolor{96.07}  & \heatmapcolor{98.80}  & \heatmapcolor{94.00}  & \heatmapcolor{96.70}  & \heatmapcolor{90.70}  & \heatmapcolor{89.60}     \\
\hline


ConvNext-B & \heatmapcolor{92.30}  & \heatmapcolor{91.52}  & \heatmapcolor{80.44}  & \heatmapcolor{90.00}  & \heatmapcolor{93.91}  & \heatmapcolor{93.01}  & \heatmapcolor{86.65}  & \heatmapcolor{83.75}  & \heatmapcolor{96.41}  & \heatmapcolor{99.20}  & \heatmapcolor{93.60}  & \heatmapcolor{96.40}  & \heatmapcolor{90.60}  & \heatmapcolor{91.40}    \\
\hline


ViT-T & \heatmapcolor{80.81}  & \heatmapcolor{77.07}  & \heatmapcolor{46.78}  & \heatmapcolor{69.07}  & \heatmapcolor{81.06}  & \heatmapcolor{76.55}  & \heatmapcolor{64.13}  & \heatmapcolor{57.86}  & \heatmapcolor{91.08} & \heatmapcolor{95.20}  & \heatmapcolor{85.50}  & \heatmapcolor{90.40}  & \heatmapcolor{67.30}  & \heatmapcolor{64.50}    \\
\hline

ViT-S & \heatmapcolor{86.77}  & \heatmapcolor{83.46}  & \heatmapcolor{63.19}  & \heatmapcolor{80.58}  & \heatmapcolor{87.98}  & \heatmapcolor{84.05}  & \heatmapcolor{74.29}  & \heatmapcolor{69.94}  & \heatmapcolor{94.74} & \heatmapcolor{97.70}  & \heatmapcolor{89.20}  & \heatmapcolor{94.30}  & \heatmapcolor{84.20}  & \heatmapcolor{80.60}     \\
\hline

ViT-B & \heatmapcolor{90.07}  & \heatmapcolor{87.48}  & \heatmapcolor{71.28}  & \heatmapcolor{84.88}  & \heatmapcolor{91.01}  & \heatmapcolor{88.64}  & \heatmapcolor{79.99}  & \heatmapcolor{76.42}  & \heatmapcolor{95.66} & \heatmapcolor{98.00}  & \heatmapcolor{90.40}  & \heatmapcolor{93.80}  & \heatmapcolor{86.20}  & \heatmapcolor{84.80}     \\
\hline

VMamba-T & \heatmapcolor{91.15}  & \heatmapcolor{89.87}  & \heatmapcolor{75.18}  & \heatmapcolor{87.41}  & \heatmapcolor{92.09}  & \heatmapcolor{91.06}  & \heatmapcolor{83.66}  & \heatmapcolor{79.71}  & \heatmapcolor{95.84}   & \heatmapcolor{98.50}  & \heatmapcolor{92.20}  & \heatmapcolor{96.30}  & \heatmapcolor{87.20}  & \heatmapcolor{86.80}   \\
\hline

VMamba-S & \heatmapcolor{92.03}  & \heatmapcolor{90.79}  & \heatmapcolor{76.15}  & \heatmapcolor{88.81}  & \heatmapcolor{93.22}  & \heatmapcolor{92.25}  & \heatmapcolor{85.57}  & \heatmapcolor{81.89}  & \heatmapcolor{96.37}  & \heatmapcolor{99.20}  & \heatmapcolor{94.10}  & \heatmapcolor{97.40}  & \heatmapcolor{90.90}  & \heatmapcolor{89.50}    \\
\hline

VMamba-B & \heatmapcolor{92.37}  & \heatmapcolor{91.27}  & \heatmapcolor{77.30}  & \heatmapcolor{89.11}  & \heatmapcolor{93.70}  & \heatmapcolor{92.64}  & \heatmapcolor{86.03}  & \heatmapcolor{83.62}  & \heatmapcolor{96.37}  & \heatmapcolor{99.10}  & \heatmapcolor{94.00}  & \heatmapcolor{96.50}  & \heatmapcolor{90.80}  & \heatmapcolor{89.80}     \\
\hline

MambaVision-T & \heatmapcolor{90.67}  & \heatmapcolor{88.83}  & \heatmapcolor{73.07}  & \heatmapcolor{86.40}  & \heatmapcolor{91.93}  & \heatmapcolor{90.07}  & \heatmapcolor{81.87}  & \heatmapcolor{78.42}  & \heatmapcolor{95.73}  & \heatmapcolor{98.60}  & \heatmapcolor{93.70}  & \heatmapcolor{96.60}  & \heatmapcolor{89.10}  & \heatmapcolor{87.70}     \\
\hline

MambaVision-S & \heatmapcolor{91.22}  & \heatmapcolor{90.19}  & \heatmapcolor{75.18}  & \heatmapcolor{88.19}  & \heatmapcolor{92.78}  & \heatmapcolor{91.19}  & \heatmapcolor{84.01}  & \heatmapcolor{81.18}  & \heatmapcolor{96.03}  & \heatmapcolor{99.40}  & \heatmapcolor{94.40}  & \heatmapcolor{97.80}  & \heatmapcolor{91.40}  & \heatmapcolor{90.10}     \\
\hline

MambaVision-B & \heatmapcolor{91.77}  & \heatmapcolor{90.65}  & \heatmapcolor{78.42}  & \heatmapcolor{89.18}  & \heatmapcolor{93.70}  & \heatmapcolor{92.81}  & \heatmapcolor{86.35}  & \heatmapcolor{83.78}  & \heatmapcolor{96.30}  & \heatmapcolor{99.10}  & \heatmapcolor{94.60}  & \heatmapcolor{97.20}  & \heatmapcolor{91.40}  & \heatmapcolor{90.60}     \\
\hline

Swin-T & \heatmapcolor{90.05}  & \heatmapcolor{88.83}  & \heatmapcolor{71.51}  & \heatmapcolor{86.19}  & \heatmapcolor{91.08}  & \heatmapcolor{88.94}  & \heatmapcolor{79.39}  & \heatmapcolor{76.49}  & \heatmapcolor{95.27}  & \heatmapcolor{97.90}  & \heatmapcolor{91.70}  & \heatmapcolor{95.30}  & \heatmapcolor{85.50}  & \heatmapcolor{84.00}     \\
\hline
Swin-S & \heatmapcolor{90.67}  & \heatmapcolor{88.86}  & \heatmapcolor{73.35}  & \heatmapcolor{87.25}  & \heatmapcolor{91.91}  & \heatmapcolor{89.68}  & \heatmapcolor{81.55}  & \heatmapcolor{78.81}  & \heatmapcolor{96.25}   & \heatmapcolor{98.30}  & \heatmapcolor{91.80}  & \heatmapcolor{95.50}  & \heatmapcolor{86.10}  & \heatmapcolor{85.40}   \\
\hline

Swin-B & \heatmapcolor{91.08}  & \heatmapcolor{89.96}  & \heatmapcolor{75.09}  & \heatmapcolor{87.87}  & \heatmapcolor{92.62}  & \heatmapcolor{91.22}  & \heatmapcolor{83.43}  & \heatmapcolor{80.65}  & \heatmapcolor{95.95} & \heatmapcolor{98.60}  & \heatmapcolor{92.30}  & \heatmapcolor{95.60}  & \heatmapcolor{89.20}  & \heatmapcolor{86.70}      \\
\hline


\end{tabular}
}
\end{table*}

\noindent \underline{\textbf{Robustness against Global Corruptions}:} The performance of various models on ImageNet-C is reported in terms of the mean corruption error (mCE) in Tab.~\ref{tab:robustness}. The mCE (\textit{lower is better}) represents the average error of the model across various common corruptions at multiple intensity levels, normalized by the ResNet-50's standard accuracy. VMamba-T has the lowest mCE among all the \textit{`T'} versions, followed by MambaVision-S, which performs significantly better than its \textit{`S'} counterparts.  However, among the \textit{`B'} family, ViT-B achieves the lowest mCE score, , possibly due to ImageNet21k pretraining and ImageNet 1k finetuning.
The table also shows top-1 accuracy on domain-shifted datasets (ImageNetV2, ImageNet-A, ImageNet-R, ImageNet-S). \textit{`T'} and \textit{`S'} versions of VMamba and MambaVision demonstrate the least average performance drop compared to Swin and ConvNext counterparts. For \textit{`B'} models, ViT performs best, likely due to ImageNet21k pretraining. For more 
results, see Appendix~\ref{app:imagent_domain}. Furthermore, in Appendix \ref{app:calibration}, we report Expected Calibration Error(ECE) of models across the mentioned datasets to quantify the reliability of model's predicted confidence levels, accompanied by reliability diagrams for visualization.



\noindent \underline{\textbf{Robustness against Fine-grained Corruptions}:} 
Tab.~\ref{tab:imagenet_e_b_main} (left) shows model robustness on ImageNet-E against variations in background complexity, object size, and positioning. Lower $\lambda$ values indicate low background texture complexity, higher values indicate high complexity, with $\lambda=20$ (adv) representing adversarially optimized high texture complexity. Objects are resized to (0.1, 0.08, 0.05) of original size and randomly placed.
VSSM models (VMamba and MambaVision) demonstrate higher resilience to background changes compared to Swin and ViT transformer families, and perform comparably to ConvNeXt models. As background complexity increases or object size decreases, all models' performance declines, but the drop is less significant for VSSM and ConvNeXt models, showing their robustness to object size variations.
Similar trends are observed for ImageNet-B (Tab.~\ref{tab:imagenet_e_b_main}, right), where backgrounds are modified using a diffusion model with textual guidance (class/caption information) or color/texture prompts.
\textit{\textbf{The VSSM family, including VMamba and MambaVision, demonstrates superior performance compared to all Transformer-based variants and maintains performance better or comparable to the advanced ConvNext models. Pure VMamba and hybrid MambaVision exhibit comparable robustness against ImageNet corruption when no information is dropped.}} For further analysis across more models, see Appendix \ref{app:corruptions}.

\subsubsection{Robustness on Object Detection} 
We evaluate VSSM, transformer, and CNN robustness for object detection using COCO-O~\cite{debenedetti2023light}, COCO-DC~\cite{malik2024objectcompose}, and COCO-C datasets. COCO-O includes 6,782 images with 26,624 bounding boxes across six domains (sketch, cartoon, painting, weather, handmake, tattoo). COCO-DC contains 1,127 COCO 2017 validation images with diffusion model-induced background changes. COCO-C applies ImageNet-C style corruptions to the COCO-2017 evaluation set at various intensities. Performance is assessed using Average Precision (AP) and AP by object sizes (APs, APm, APl) on five models: ConvNext-T, Swin-T, Swin-S, VMamba-T, and VMamba-S.

\begin{figure*}[t]
\includegraphics[width=\textwidth]{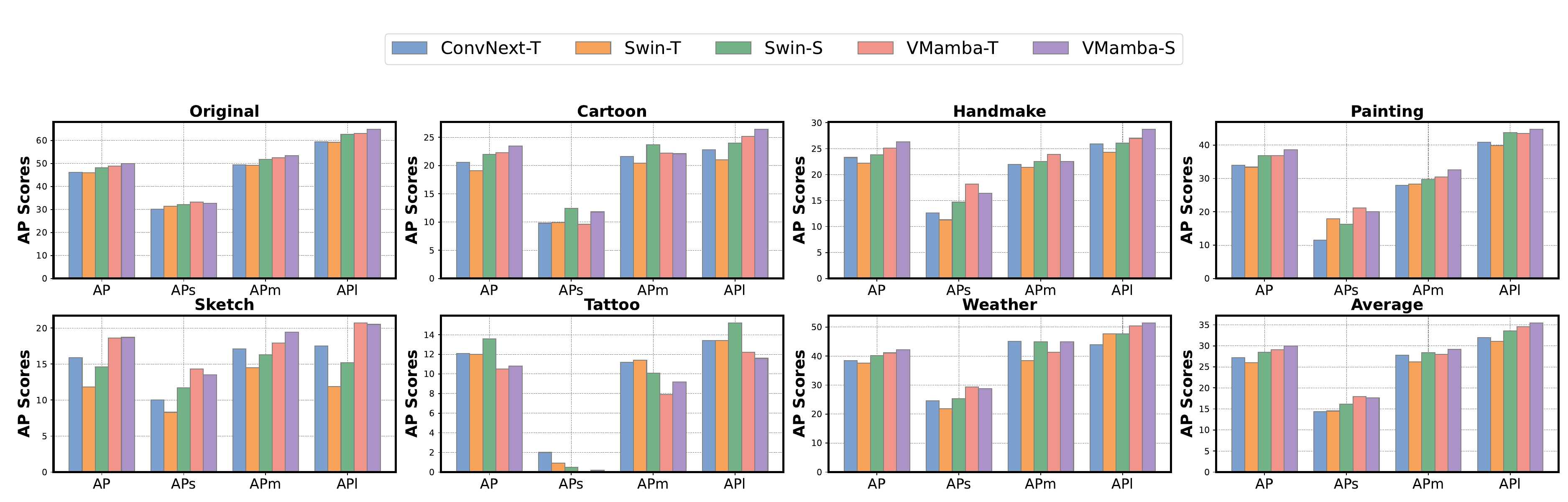}
\caption{\small Average Precision (AP) scores for different architectures on the COCO-O dataset~\cite{debenedetti2023light}, detailing results for small (APs), medium (APm), and large objects (APl).}
\label{fig:coco_o}
\vspace{-1.5em}
\end{figure*}

\begin{figure*}[t]
\centering
\vspace{-1.4em}
\includegraphics[height=3.3cm,width=\textwidth]{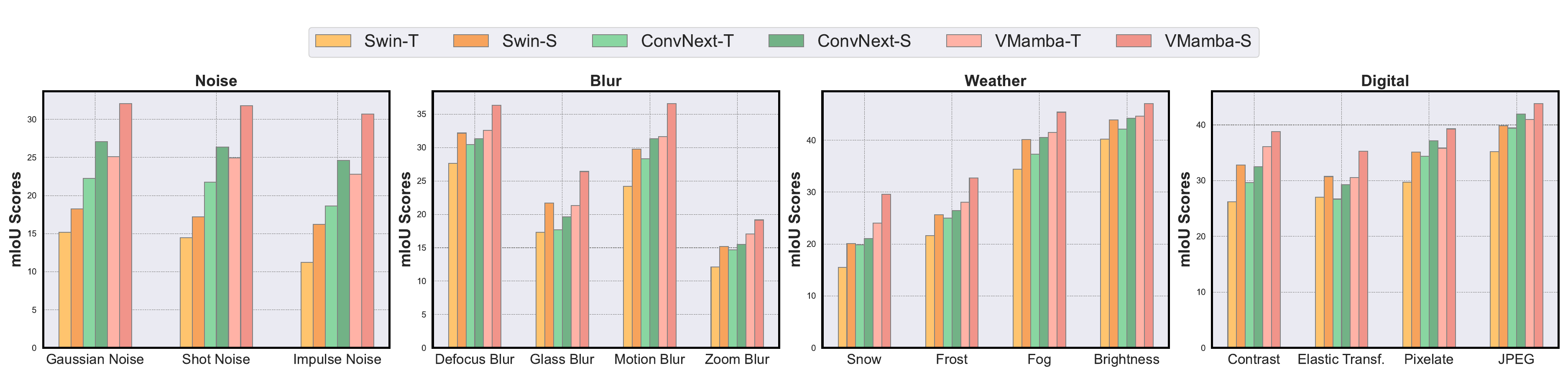}
\vspace{-.4em}
\includegraphics[height=2.6cm,width=\textwidth, trim=0cm 0cm 0cm 3cm, clip]{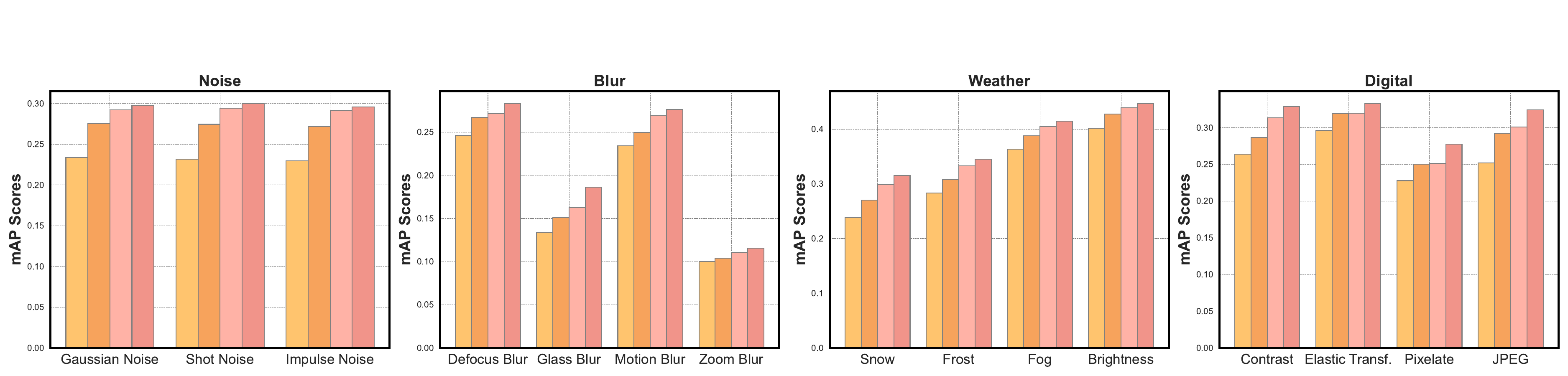}
\caption{\small Performance comparison of different architectures on the AED20k-C  and COCO-C datasets for segmentation (top) and detection (bottom) tasks. The top figure shows the Mean Intersection over Union (mIoU) score on the AED20k-C dataset, while the Mean Average Precision (mAP) score on the COCO-C dataset.}
\label{fig:ade_c_coco_c}
\end{figure*}

Fig.~\ref{fig:coco_o} shows VMamba-S and VMamba-T consistently outperforming other architectures in most COCO-O scenarios, leading in original validation and out-of-distribution domains (cartoon, painting, sketch, weather). However, all models struggle with the tattoo domain. On average, VMamba-S and VMamba-T achieve AP scores of 42.2$\%$ and 41.1$\%$, surpassing Swin and ConvNext models.
Fig.~\ref{fig:ade_c_coco_c} demonstrates VMamba models' superior robustness across various common corruptions in object detection, with even VMamba-T outperforming larger Swin-S in most scenarios. All models show performance drops with  \textit{`Glass Blur'} and \textit{`Zoom Blur'} corruptions. For further details and results on COCO-DC, see Appendix \ref{app:detection_corruptions}.

\subsubsection{Robustness on Semantic Segmentation}
We assess segmentation model robustness using 2,000 images from the ADE20K~\cite{zhou2017scene} validation set, corrupted with ImageNet-C~\cite{hendrycks2019benchmarking} at various intensities. Performance is evaluated using mean Intersection over Union (mIoU).


Fig.~\ref{fig:ade_c_coco_c} shows VMamba-T and VMamba-S consistently outperforming Swin counterparts across various ImageNet-C corruptions in segmentation, mirroring trends seen in object detection. 
The high performance of VMamba models on the original dataset also transfers effectively to the corrupted version of the dataset. Notably, VMamba-T surpasses the larger Swin-S in most corruption scenarios. Additional results are provided in Appendix \ref{app:detection_corruptions}.

\subsection{Robustness against Adversarial Attacks}

\begin{table*}[t]
\centering
\caption{\small Robust accuracy of models under white-box and black-box settings for FGSM attack. Adversarial examples  crafted on surrogate models are used to evaluate robustness of target models.}
\label{tab:adv_transf_main_only_fgsm}
\setlength{\tabcolsep}{2.0pt}
\resizebox{\linewidth}{!}{
\begin{tabular}{|l| c c c c c c c c c c c c c c c c|}
\hline
\rowcolor{LightCyan}
\multicolumn{1}{|l|}{\scriptsize Target $\rightarrow$} & \rotatebox{0}{\scriptsize \textbf{VMamba-T}} & \rotatebox{0}{\scriptsize \textbf{VMamba-S}} & \rotatebox{0}{\scriptsize \textbf{VMamba-B}} & \rotatebox{0}{\scriptsize \textbf{MambaVision-T}} & \rotatebox{0}{\scriptsize \textbf{MambaVision-S}} & \rotatebox{0}{\scriptsize \textbf{MambaVision-B}} & \rotatebox{0}{\scriptsize \textbf{ResNet-50}} & \rotatebox{0}{\scriptsize \textbf{ConvNext-T}} & \rotatebox{0}{\scriptsize \textbf{ConvNext-S}} & \rotatebox{0}{\scriptsize \textbf{ConvNext-B}} & \rotatebox{0}{\scriptsize \textbf{ViT-T}} & \rotatebox{0}{\scriptsize \textbf{ViT-S}} & \rotatebox{0}{\scriptsize \textbf{ViT-B}} & \rotatebox{0}{\scriptsize \textbf{Swin-T}} & \rotatebox{0}{\scriptsize \textbf{Swin-S}} & \rotatebox{0}{\scriptsize \textbf{Swin-B}} \\
\hline
\rowcolor{gray!5}
\multicolumn{1}{|l|}{\scriptsize Surrogate $\downarrow$} & \multicolumn{16}{c|}{Fast Gradient Sign Method (FGSM) at $\epsilon=\frac{8}{255}$ } \\
\hline
{\scriptsize VMamba-T} &   \heatmapcolor{42.90}  & \heatmapcolor{66.34}  & \heatmapcolor{65.10}  & \heatmapcolor{74.44}  & \heatmapcolor{73.00}  & \heatmapcolor{73.16}  & \heatmapcolor{80.20}  & \heatmapcolor{72.66}  & \heatmapcolor{73.80}  & \heatmapcolor{72.84}  & \heatmapcolor{79.46}  & \heatmapcolor{83.64}  & \heatmapcolor{86.94}  & \heatmapcolor{72.22} & \heatmapcolor{76.38} & \heatmapcolor{74.60}    \\
\hline
{\scriptsize VMamba-S} &  \heatmapcolor{62.24}  & \heatmapcolor{48.42}  & \heatmapcolor{63.00}  & \heatmapcolor{71.84}  & \heatmapcolor{71.92}  & \heatmapcolor{71.34}  & \heatmapcolor{79.70}  & \heatmapcolor{71.40}  & \heatmapcolor{71.04}  & \heatmapcolor{70.18}  & \heatmapcolor{78.42}  & \heatmapcolor{81.56}  & \heatmapcolor{84.58}  & \heatmapcolor{70.96}  & \heatmapcolor{72.92}  & \heatmapcolor{71.14}   \\
\hline
{\scriptsize VMamba-B} &   \heatmapcolor{65.52}  & \heatmapcolor{66.96}  & \heatmapcolor{51.24}  & \heatmapcolor{75.22}  & \heatmapcolor{73.82}  & \heatmapcolor{73.08}  & \heatmapcolor{81.22}  & \heatmapcolor{73.54}  & \heatmapcolor{73.20}  & \heatmapcolor{72.32}  & \heatmapcolor{79.12}  & \heatmapcolor{83.62}  & \heatmapcolor{86.24}  & \heatmapcolor{73.24}  & \heatmapcolor{76.30}  & \heatmapcolor{73.88} \\
\hline

{\scriptsize MambaVision-T} &   \heatmapcolor{77.74}  & \heatmapcolor{79.04}  & \heatmapcolor{79.78}   & \heatmapcolor{46.18}  & \heatmapcolor{73.30}  & \heatmapcolor{75.64} & \heatmapcolor{82.06}  & \heatmapcolor{79.50}  & \heatmapcolor{80.88}  & \heatmapcolor{81.82}  & \heatmapcolor{78.30}  & \heatmapcolor{84.20}  & \heatmapcolor{87.86}  & \heatmapcolor{77.78}  & \heatmapcolor{81.24}  & \heatmapcolor{80.80} \\
\hline

{\scriptsize MambaVision-S} &   \heatmapcolor{77.86}  & \heatmapcolor{79.56}  & \heatmapcolor{79.24}  & \heatmapcolor{74.96}  & \heatmapcolor{53.42}  & \heatmapcolor{73.84}  & \heatmapcolor{83.60}  & \heatmapcolor{80.40}  & \heatmapcolor{81.14}  & \heatmapcolor{81.26}  & \heatmapcolor{79.90}  & \heatmapcolor{85.64}  & \heatmapcolor{88.52}  & \heatmapcolor{78.50}  & \heatmapcolor{82.40}  & \heatmapcolor{81.88} \\
\hline

{\scriptsize MambaVision-B} &   \heatmapcolor{75.90}  & \heatmapcolor{77.84}  & \heatmapcolor{77.00}  & \heatmapcolor{75.46}  & \heatmapcolor{71.96}  & \heatmapcolor{52.68}  & \heatmapcolor{83.46}  & \heatmapcolor{80.24}  & \heatmapcolor{80.52}  & \heatmapcolor{79.92}  & \heatmapcolor{79.82}  & \heatmapcolor{85.18}  & \heatmapcolor{88.12}  & \heatmapcolor{78.46}  & \heatmapcolor{81.60}  & \heatmapcolor{80.30} \\
\hline

{\scriptsize ResNet-50} &   \heatmapcolor{81.38}  & \heatmapcolor{83.24}  & \heatmapcolor{83.84}  & \heatmapcolor{81.06}  & \heatmapcolor{82.54}  & \heatmapcolor{84.88}  & \heatmapcolor{30.46}  & \heatmapcolor{80.30}  & \heatmapcolor{82.20}  & \heatmapcolor{83.38}  & \heatmapcolor{75.94}  & \heatmapcolor{84.74}  & \heatmapcolor{89.12}  & \heatmapcolor{80.64}  & \heatmapcolor{85.00} & \heatmapcolor{85.42}  \\
\hline


{\scriptsize ConvNext-T} &  \heatmapcolor{69.00} & \heatmapcolor{71.46} & \heatmapcolor{71.18}  & \heatmapcolor{73.50}  & \heatmapcolor{73.92}  & \heatmapcolor{74.40}  & \heatmapcolor{77.96}  & \heatmapcolor{36.36}  & \heatmapcolor{61.96}  & \heatmapcolor{63.76}  & \heatmapcolor{76.92}  & \heatmapcolor{82.74} & \heatmapcolor{85.58}  & \heatmapcolor{67.06} & \heatmapcolor{71.88}   & \heatmapcolor{70.78}  \\
\hline


{\scriptsize ConvNext-S} &  \heatmapcolor{69.54} & \heatmapcolor{70.62} & \heatmapcolor{70.48}  & \heatmapcolor{74.68}  & \heatmapcolor{74.38}  & \heatmapcolor{75.24}  & \heatmapcolor{79.48}  & \heatmapcolor{63.48}  & \heatmapcolor{49.10}  & \heatmapcolor{63.62}  & \heatmapcolor{78.48}  & \heatmapcolor{82.94} & \heatmapcolor{85.02}  & \heatmapcolor{69.24} & \heatmapcolor{71.54}  & \heatmapcolor{69.74}  \\
\hline


{\scriptsize ConvNext-B} &  \heatmapcolor{70.54} & \heatmapcolor{71.78} & \heatmapcolor{69.78}  & \heatmapcolor{76.78}  & \heatmapcolor{75.02}  & \heatmapcolor{74.82}  & \heatmapcolor{81.72}  & \heatmapcolor{67.34}  & \heatmapcolor{66.24}  & \heatmapcolor{51.32}  & \heatmapcolor{80.26}  & \heatmapcolor{83.98} & \heatmapcolor{86.22}  & \heatmapcolor{69.86} & \heatmapcolor{73.44}  & \heatmapcolor{70.84}   \\
\hline

{\scriptsize ViT-T} &   \heatmapcolor{85.92} & \heatmapcolor{88.30} & \heatmapcolor{88.88}  & \heatmapcolor{82.16}  & \heatmapcolor{85.06}  & \heatmapcolor{87.24}  & \heatmapcolor{82.68}  & \heatmapcolor{85.08}  & \heatmapcolor{86.56}  & \heatmapcolor{88.06}  & \heatmapcolor{2.28}  & \heatmapcolor{50.04} & \heatmapcolor{69.90}  & \heatmapcolor{75.66} & \heatmapcolor{79.72}  & \heatmapcolor{82.32}  \\
\hline
{\scriptsize ViT-S} &    \heatmapcolor{81.74} & \heatmapcolor{83.76} & \heatmapcolor{84.46}  & \heatmapcolor{78.94}  & \heatmapcolor{80.78}  & \heatmapcolor{82.68}  & \heatmapcolor{81.70}  & \heatmapcolor{82.34}  & \heatmapcolor{82.78}  & \heatmapcolor{84.38}  & \heatmapcolor{45.40}  & \heatmapcolor{11.02} & \heatmapcolor{54.82}  & \heatmapcolor{72.50} & \heatmapcolor{75.98} & \heatmapcolor{77.88}  \\
\hline
{\scriptsize ViT-B} &    \heatmapcolor{82.46} & \heatmapcolor{84.02} & \heatmapcolor{84.48}  & \heatmapcolor{80.40}  & \heatmapcolor{81.82}  & \heatmapcolor{82.46}  & \heatmapcolor{81.90}  & \heatmapcolor{82.04}  & \heatmapcolor{82.24}  & \heatmapcolor{83.62}  & \heatmapcolor{58.50}  & \heatmapcolor{53.84} & \heatmapcolor{24.24}  & \heatmapcolor{75.82} & \heatmapcolor{77.62}  & \heatmapcolor{78.54} \\
\hline
{\scriptsize Swin-T} &    \heatmapcolor{72.40} & \heatmapcolor{76.06} & \heatmapcolor{75.58}  & \heatmapcolor{76.80}  & \heatmapcolor{77.16}  & \heatmapcolor{78.06}  & \heatmapcolor{82.82}  & \heatmapcolor{71.46}  & \heatmapcolor{72.12}  & \heatmapcolor{71.96}  & \heatmapcolor{76.48}  & \heatmapcolor{80.66} & \heatmapcolor{85.46}  & \heatmapcolor{28.86} & \heatmapcolor{56.22}  & \heatmapcolor{55.12}  \\
\hline
{\scriptsize Swin-S} &   \heatmapcolor{78.24} & \heatmapcolor{79.10} & \heatmapcolor{79.30}  & \heatmapcolor{80.12}  & \heatmapcolor{81.90}  & \heatmapcolor{81.40}  & \heatmapcolor{85.46}  & \heatmapcolor{77.96}  & \heatmapcolor{77.10}  & \heatmapcolor{77.02}  & \heatmapcolor{79.38}  & \heatmapcolor{83.04} & \heatmapcolor{86.58}  & \heatmapcolor{61.94} & \heatmapcolor{48.00}  & \heatmapcolor{63.90} \\
\hline
{\scriptsize Swin-B} &   \heatmapcolor{78.88} & \heatmapcolor{79.40} & \heatmapcolor{79.04}  & \heatmapcolor{81.48}  & \heatmapcolor{82.16}  & \heatmapcolor{81.04}  & \heatmapcolor{86.60}  & \heatmapcolor{78.16}  & \heatmapcolor{78.46}  & \heatmapcolor{77.32}  & \heatmapcolor{81.48}  & \heatmapcolor{84.74} & \heatmapcolor{87.64}  & \heatmapcolor{66.82} & \heatmapcolor{68.28}  & \heatmapcolor{54.76} \\
\hline

\end{tabular}
}
\vspace{-0em}

\end{table*}




In this section, we evaluate the robustness of VSSMs against spatial and frequency-based adversarial attacks. We also compare the performance of adversarially fine-tuned VSSM models on CIFAR-10~\cite{article} and Imagenette~\cite{imagenette} downstream dataset with ViT and Swin models.

\noindent \underline{\textbf{Adversarial Attacks in Spatial Domain:}}
We conduct experiments in both white-box and black-box settings using the Fast Gradient Sign Method (FGSM)~\cite{goodfellow2014explaining} and Projected Gradient Descent (PGD)~\cite{madry2017towards} with an $l_\infty$-norm and $\epsilon = 8/255$. FGSM is a single-step process, while PGD operates as a multistep method, iterating for 20 steps with a step size of $2/255$. Tab.~\ref{tab:adv_transf_main_only_fgsm} (\textit{top}) displays the robust accuracy scores under white-box (diagonal entries) and black-box (off-diagonal entries) adversarial attacks for FGSM. \textbf{{\textit{`T'} and \textit{`S'} versions of VMamba and MambaVision models exhibit higher white-box attack robustness compared to their Swin Transformer counterparts, but this pattern does not extend to the larger \textit{`B'} models}}. This indicates that VSSM's robustness advantage over Swin Transformers may not consistently scale with increased model size. For black-box settings, attacks transfer rate is high within the same architecture family  than across different architectures. As expected, under stronger iteraive attack (PGD), all models' performance almost drops to zero in white-box settings. For the black-box transferability,  we observe similar trends to FGSM attack (see Appendix \ref{app:adv_attack}).

\noindent \underline{\textbf{Adversarial Attacks in the Frequency Domain:}} 
We evaluate VMamba and MambaVision's robustness against frequency-specific PGD attacks and report the results in Appendix \ref{app:adv_attack}. VMamba and MambaVision maintain above 90$\%$ robustness for low-frequency perturbations up to $\epsilon=16$, comparable to ConvNext and Swin. For high-frequency attacks, all models' robustness decreases rapidly, with ViT models showing highest resilience. In standard full-frequency attacks, VSSM models display higher robustness than ConvNext, ViT, and Swin. 

\begin{figure}[h]
\small \centering
        \includegraphics[height=5cm,width=\linewidth,  keepaspectratio=true ]{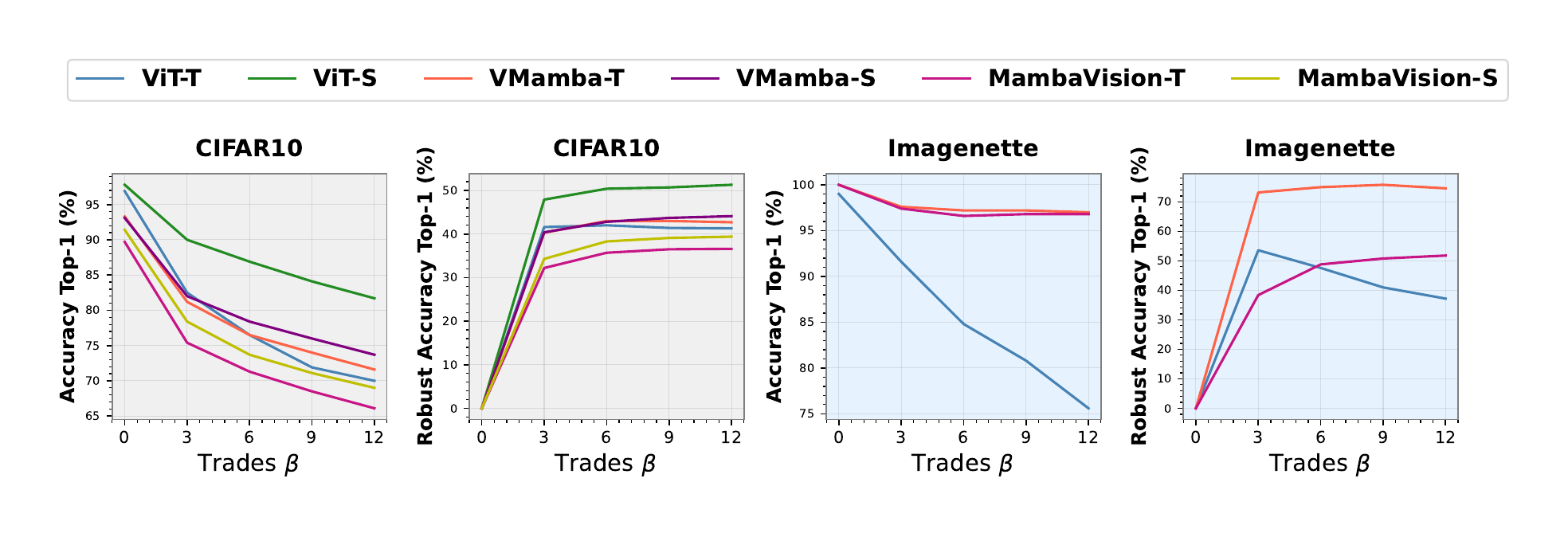}
    \caption{\small Clean and Robust accuracy  of models evaluated on CIFAR-10 \emph{(left)} and Imagenette \emph{(right)}.}
\label{fig: advft}
\end{figure}

\noindent \underline{\textbf{Adversarial Fine-tuning on Downstream Datasets:}} We adversarially fine-tune ImageNet pretrained VSSM, ViT, and Swin on downstream datasets using the TRADES~\cite{zhang2019theoretically} objective with varying robustness strength $\beta$ and an $l_{\infty}$ perturbation budget $\epsilon=\frac{8}{255}$. In Fig.~\ref{fig: advft}, we plot the clean and robust accuracy under PGD-100 at $\epsilon=\frac{8}{255}$. On Imagennette~\cite{imagenette}, VSSM-T shows strong performance in both clean and robust accuracy across different $\beta$ levels, followed closely by Swin-T, while ViT-T exhibits the lowest robustness. However, on the low-resolution CIFAR-10 dataset with significantly lower number of patches, ViT models perform better than Mamba-based VSSM models. 

\section{Conclusion}

In conclusion, we present a comprehensive evaluation of the robustness of Vision State-Space Models (VSSMs) under diverse natural and adversarial perturbations, highlighting both their strengths and weaknesses compared to transformers and CNNs. Through rigorous experiments, we demonstrated the capabilities and limitations of VSSMs classifiers in handling occlusions, common corruptions, and adversarial attacks, as well as their resilience to object-background compositional changes in complex visual scenes. Additionally, we show that VSSM-based models are generally more robust to real-world corruptions in the dense prediction tasks, including detection and segmentation. 
As an early work in this area, our findings significantly enhance the understanding of VSSMs' robustness and set the stage for future research aimed at improving the reliability and effectiveness of visual perception systems in real-world scenarios.



{\small
\bibliographystyle{ieee_fullname}
\bibliography{review}
}

\clearpage
\appendix

\noindent\begin{LARGE} \textbf{Appendix} \vspace{4mm} \end{LARGE}


We provide additional insights into our study through various sections in the appendix, which include results for both versions of the VMamba models \cite{liu2024vmamba}: \emph{v2} (reported in the main paper) and \emph{v0}. Results that do not specify the VMamba version should be assumed to refer to \emph{v2} by default.

 In Section \ref{non-adversarial}, we provide detailed results on \emph{Information Drop} experiments. Section \ref{app:scandrop} explores information drop along the scanning direction. Sections \ref{app:randomdrop} and \ref{app:dinodrop}   offer additional detailed results on Salient and Non-Salient Patch Drop and Random Patch Drop, respectively. Section \ref{app:shuffle} presents further results on patch shuffling.
 Section \ref{app:image_corruptions} provides results on \emph{Image Corruptions}. In Section \ref{app:imagent_domain}, robustness against global corruptions such as common corruption, and out-of-distribution datasets is evaluated. Additionally, in Section \ref{app:calibration}, we provide model calibration results on in-distribition and out-of-distribution datasets.
 In Section \ref{app:corruptions}, we report results on fine-grained corruptions across different models. In Section \ref{app:detection_corruptions}, we report image corruption results for the task of object detection and semantic segmentation. Finally, we expand our analysis of \emph{Adversarial Attacks} across models in Section \ref{app:adv_attack}. We present results for white-box attacks, transfer-based black-box attacks, and frequency-based attacks.

\section{Robustness against Information Drop}\label{non-adversarial}

\subsection{Information Drop along the Scanning Axis}
\label{app:scandrop}

\begin{figure*}[t]
\centering
\includegraphics[width=\textwidth]{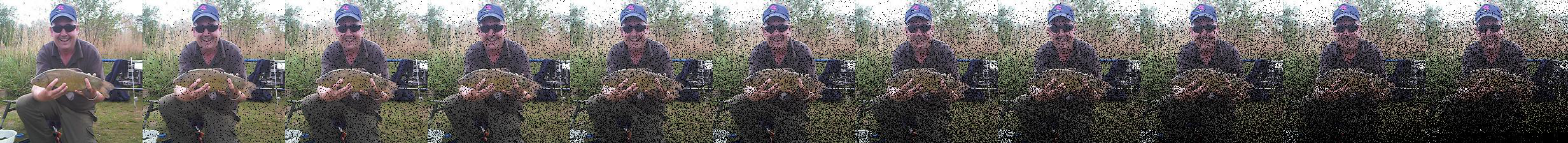}
\includegraphics[width=\textwidth]{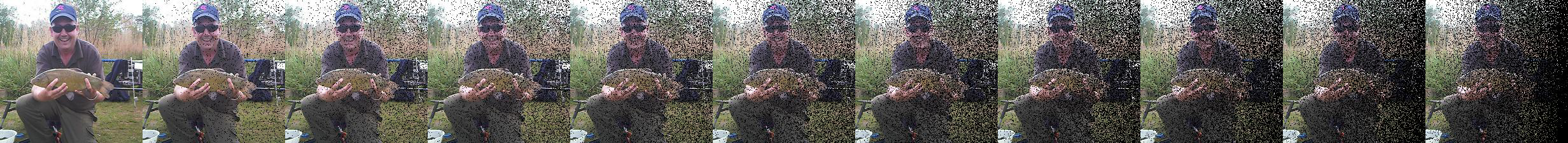}
\includegraphics[width=\textwidth]{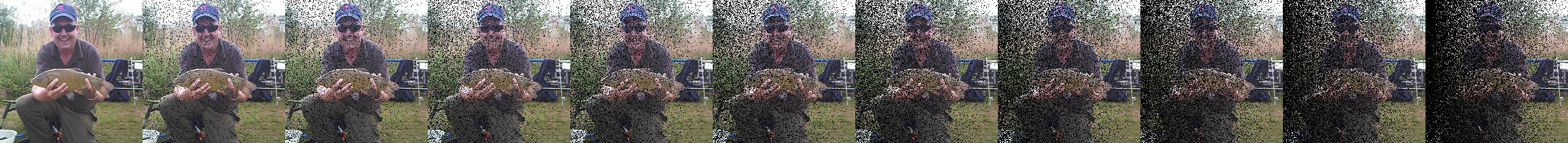}
\includegraphics[width=\textwidth]{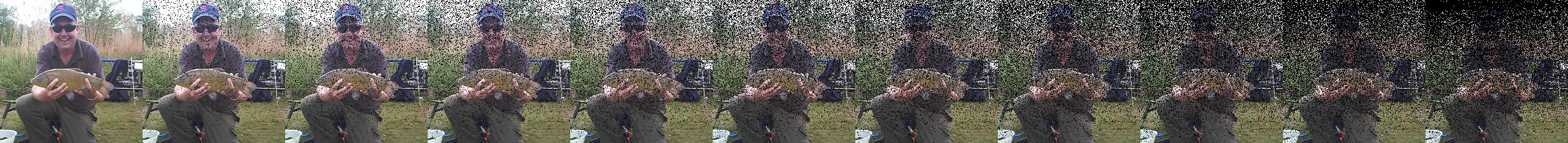}
\includegraphics[width=\textwidth]{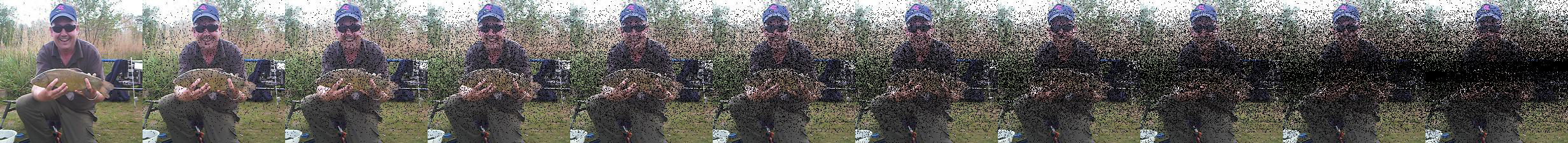}
\includegraphics[width=\textwidth]{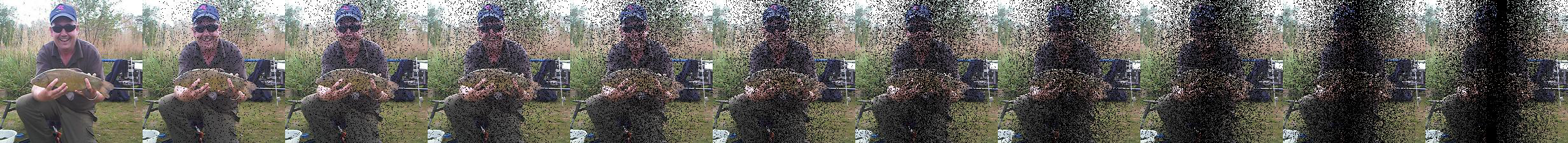}
\includegraphics[width=\textwidth]{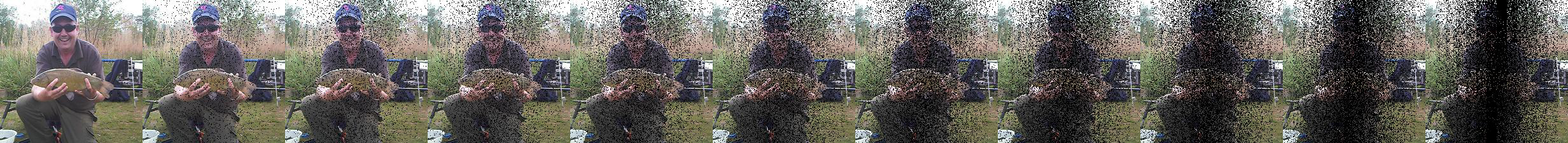}
\includegraphics[width=\textwidth]{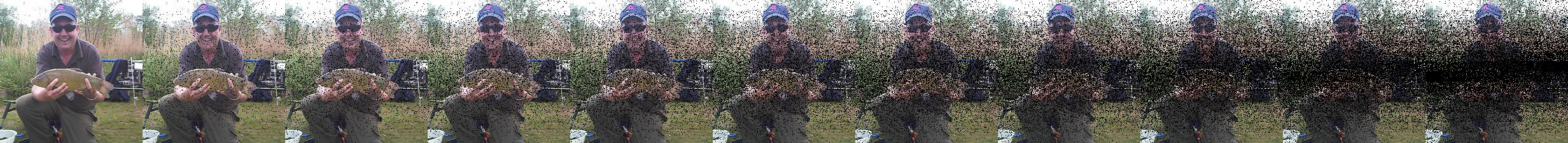}
\includegraphics[width=\textwidth]{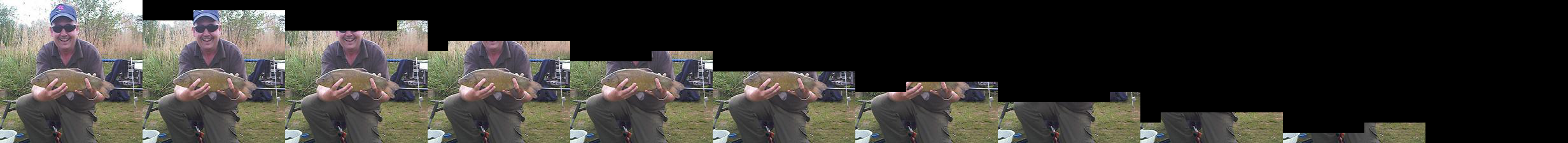}
\includegraphics[width=\textwidth]{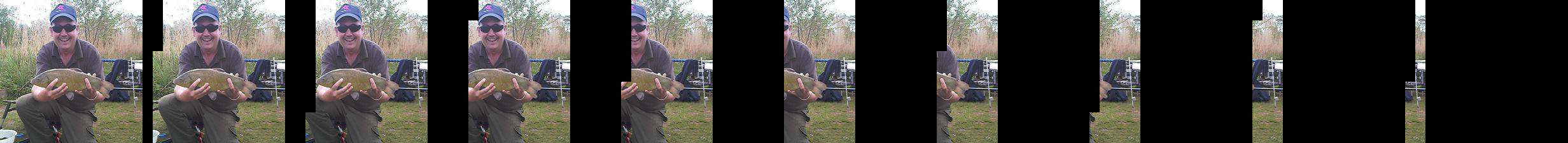}
\includegraphics[width=\textwidth]{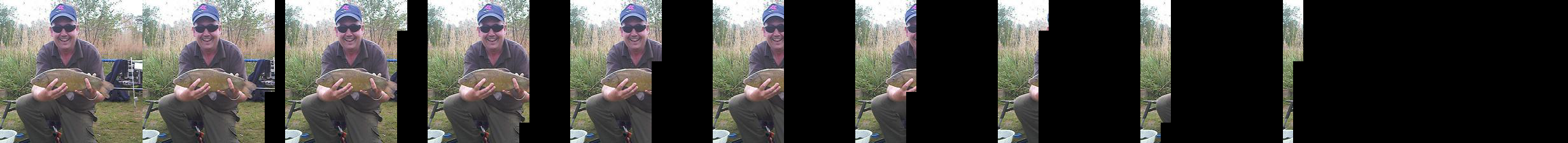}
\includegraphics[width=\textwidth]{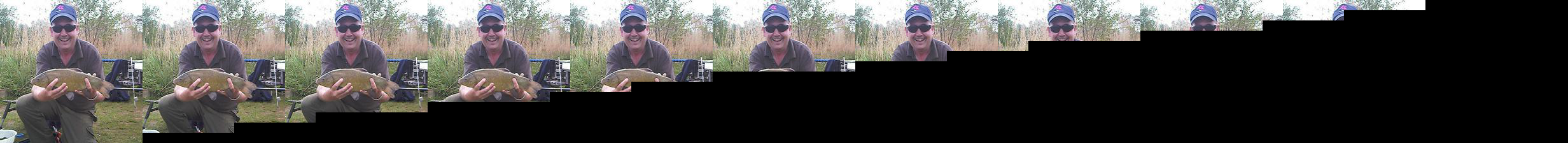}

\caption{\small Information drop along scanning direction: The top four rows represent linearly increasing information drop along the four scanning directions at patch size $14 \times 14$. The center four rows represent linearly increasing information drop till the center  along the four scanning directions at patch size $14 \times 14$. The bottom four rows represent  sequentially dropping patches along the four scanning directions at patch size $14 \times 14$.}
\label{fig:scanning_example}
\end{figure*}

\begin{figure*}[!t]
\small \centering
\begin{minipage}{0.49\textwidth}
    \begin{minipage}{\textwidth}
        \centering
        \includegraphics[height=4.5cm, keepaspectratio]{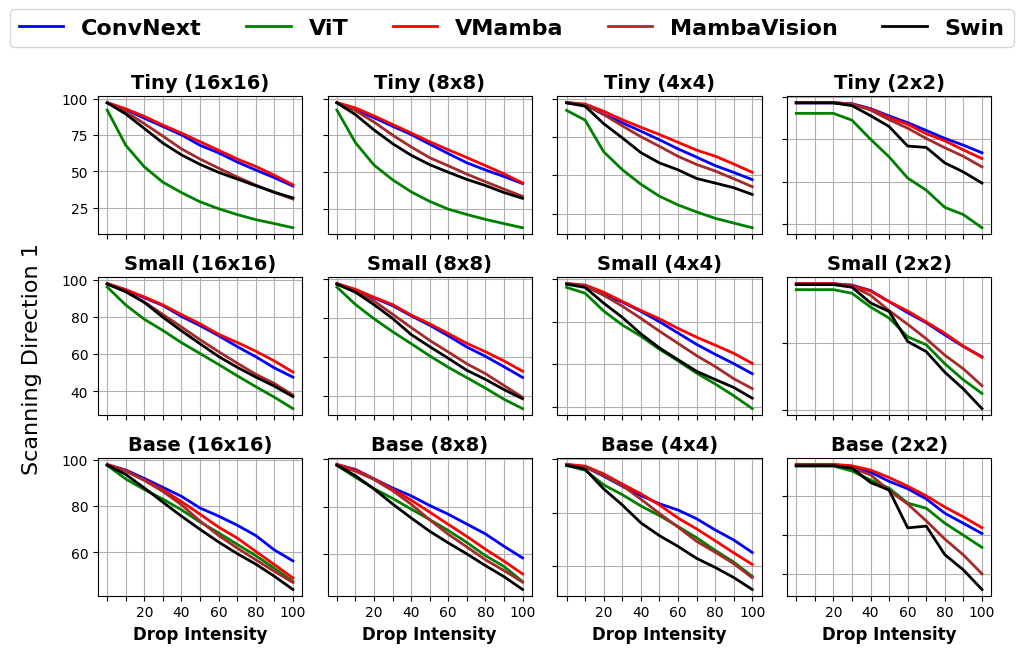}
    \end{minipage}
\end{minipage}
\begin{minipage}{0.49\textwidth}
 \begin{minipage}{\textwidth}
        \centering
        \includegraphics[height=4.5cm, keepaspectratio]{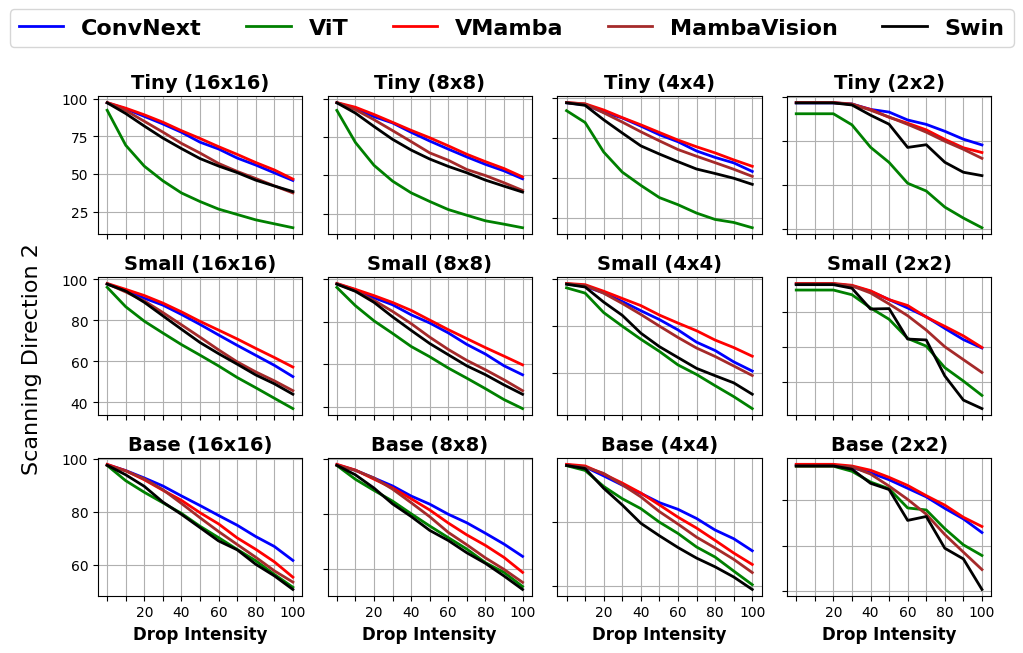}
    \end{minipage}

\end{minipage}
\begin{minipage}{0.49\textwidth}
    \begin{minipage}{\textwidth}
        \centering
        \includegraphics[height=4.5cm, keepaspectratio]{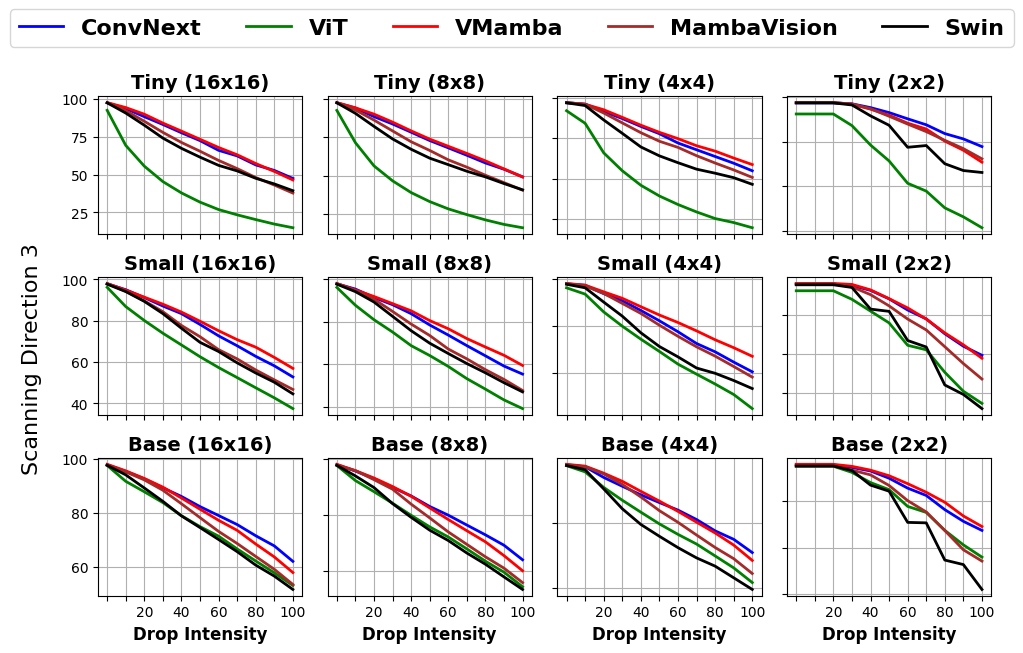}
    \end{minipage}
\end{minipage}
\begin{minipage}{0.49\textwidth}
 \begin{minipage}{\textwidth}
        \centering
        \includegraphics[height=4.5cm, keepaspectratio]{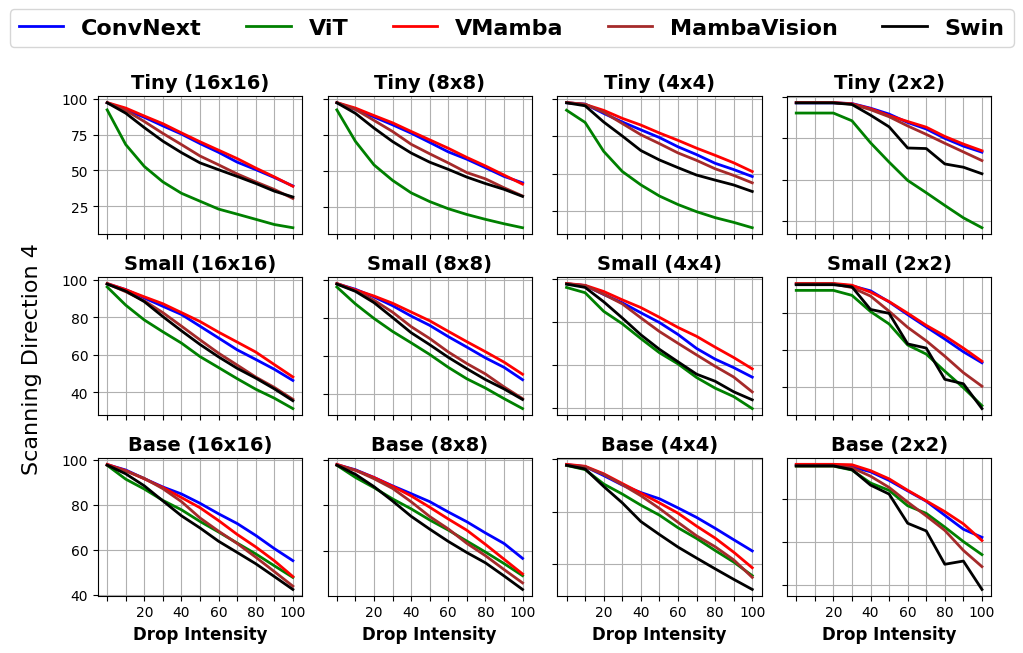}
    \end{minipage}

\end{minipage}
    \caption{\small Information drop of Tiny and Small family of models along the scanning direction: the image is split into a sequence of fixed-size non-overlapping patches of size 16x16, 8x8. 4x4, and 2x2. We report results of linearly increasing the number of pixels dropped from each patch to the maximum threshold (\emph{Drop Intensity}) along the scanning direction. Top row shows results for top-to-bottom \emph{(Direction 1)} and left-to-right direction\emph{(Direction 2)}. Bottom row shows results for right-to-left \emph{(Direction 3)} and bottom-to-top direction\emph{(Direction 4)}.}
\label{fig:scan_exp_linear_increase_all}

\end{figure*}

\begin{figure*}[!t]
\small \centering
\begin{minipage}{0.49\textwidth}
    \begin{minipage}{\textwidth}
        \centering
        \includegraphics[height=4.5cm, keepaspectratio]{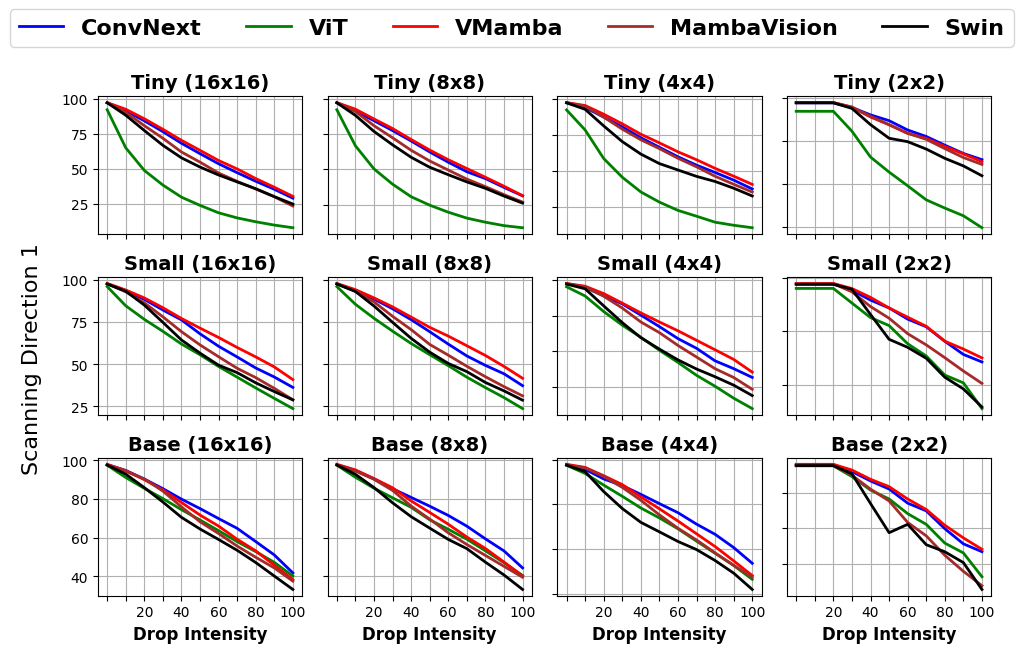}
    \end{minipage}
\end{minipage}
\begin{minipage}{0.49\textwidth}
 \begin{minipage}{\textwidth}
        \centering
        \includegraphics[height=4.5cm, keepaspectratio]{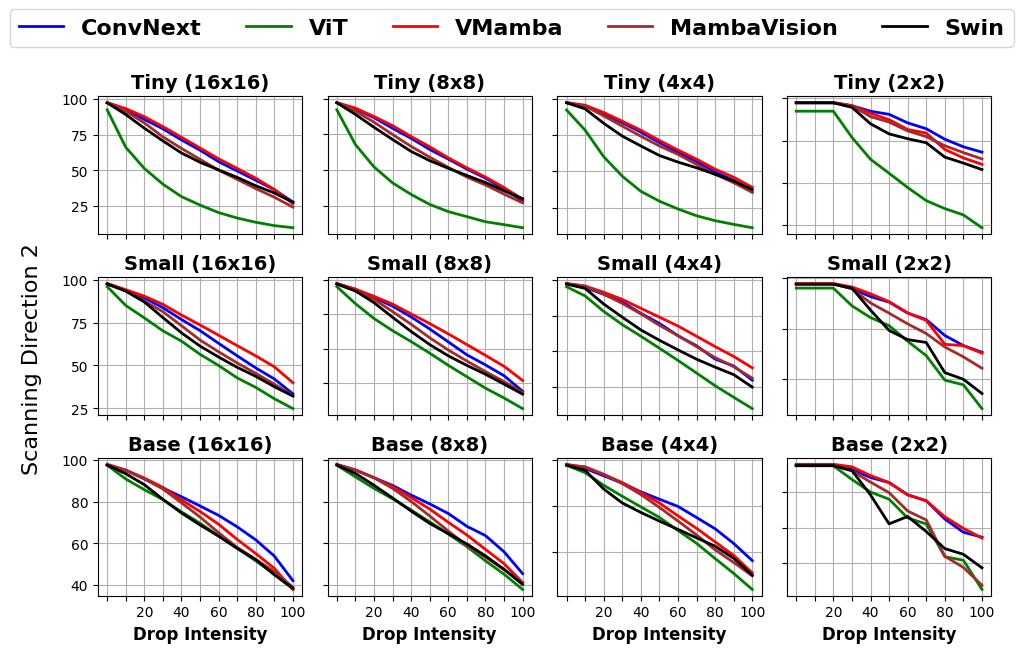}
    \end{minipage}

\end{minipage}
\begin{minipage}{0.49\textwidth}
    \begin{minipage}{\textwidth}
        \centering
        \includegraphics[height=4.5cm, keepaspectratio]{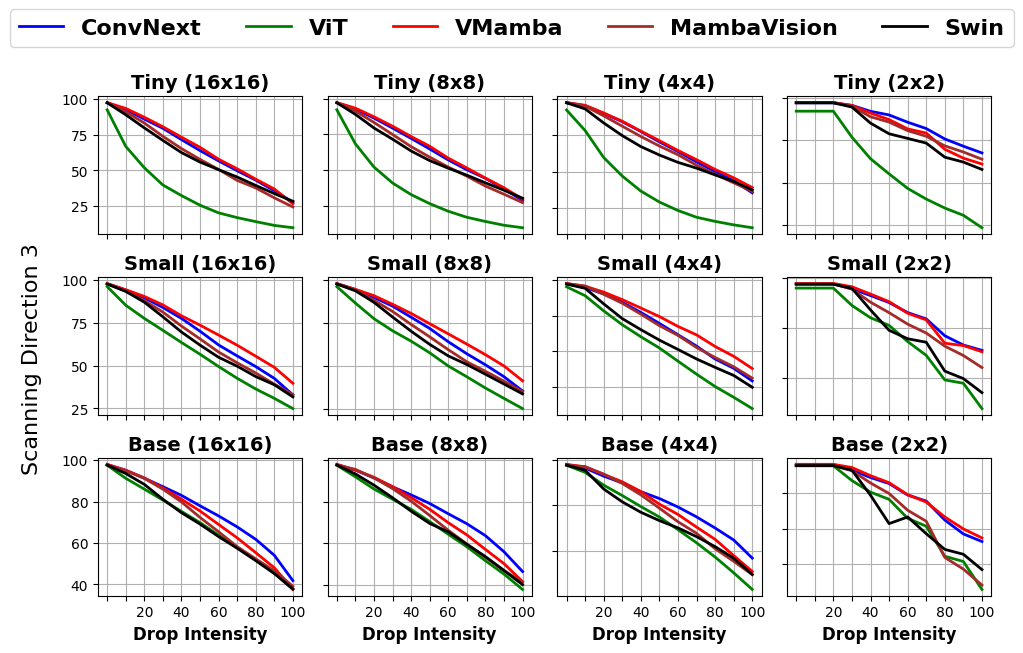}
    \end{minipage}
\end{minipage}
\begin{minipage}{0.49\textwidth}
 \begin{minipage}{\textwidth}
        \centering
        \includegraphics[height=4.5cm, keepaspectratio]{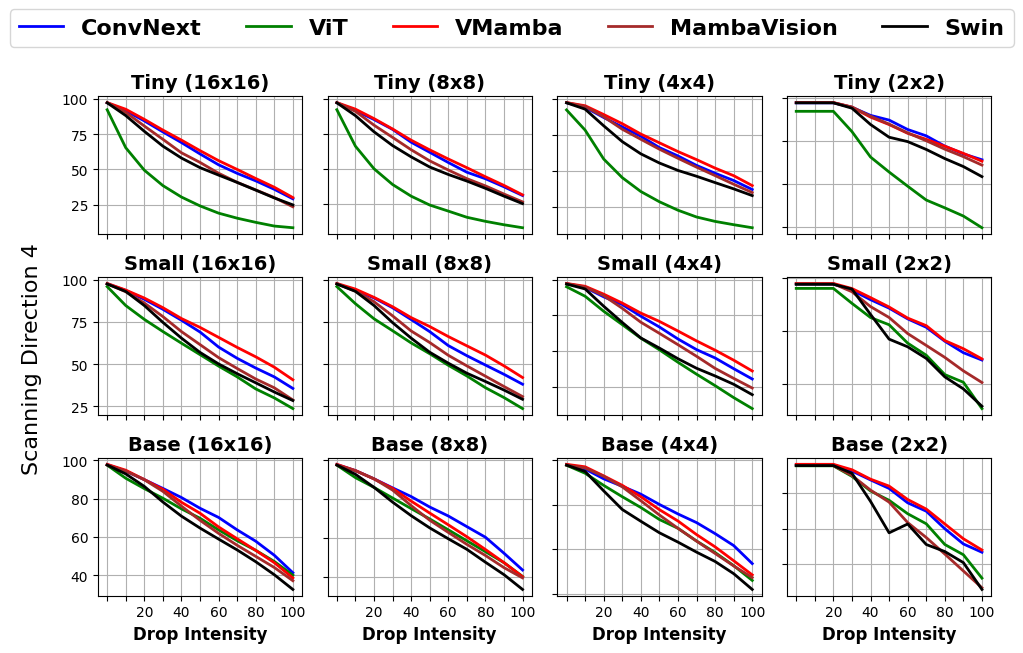}
    \end{minipage}

\end{minipage}
    \caption{\small Information drop of Tiny and Small family of models along the scanning direction: the image is split into a sequence of fixed-size non-overlapping patches of size 16x16, 8x8. 4x4, and 2x2. We report results of linearly increasing the number of pixels dropped from each patch to the maximum threshold (\emph{Drop Intensity}) at the center of the scanning direction and then again linearly decreased till the end. The top row shows results for top-to-bottom \emph{(Direction 1)} and left-to-right direction\emph{(Direction 2)}.  The bottom row shows results for right-to-left \emph{(Direction 3)} and bottom-to-top direction\emph{(Direction 4)}.}
\label{fig:scan_exp_max_center_all}

\end{figure*}

\begin{figure*}[!t]
\small \centering
\begin{minipage}{0.49\textwidth}
    \begin{minipage}{\textwidth}
        \centering
        \includegraphics[height=4.5cm, keepaspectratio]{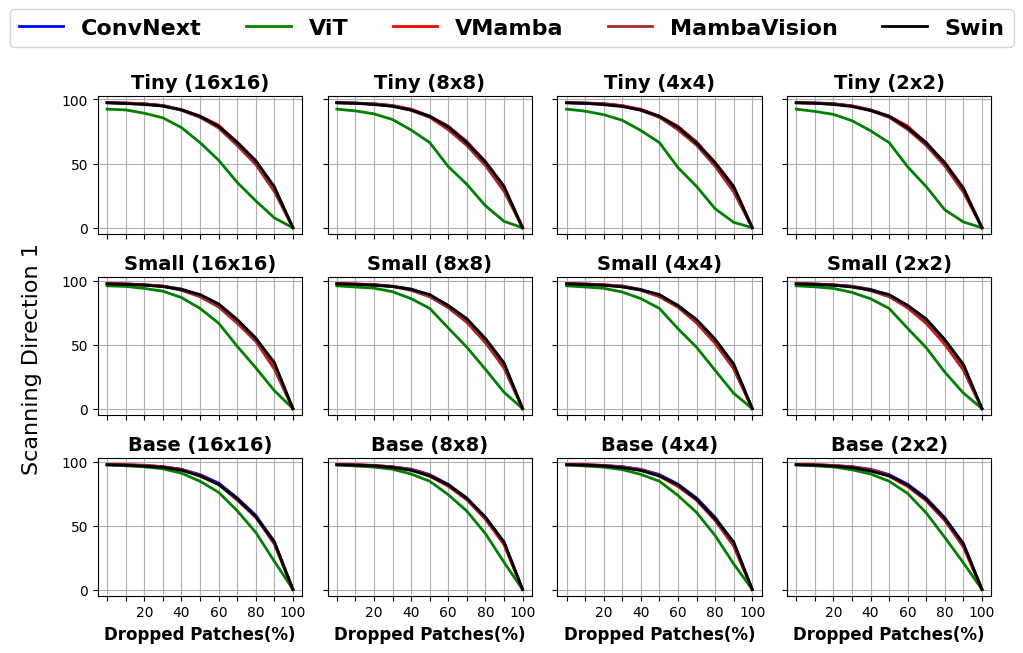}
    \end{minipage}
\end{minipage}
\begin{minipage}{0.49\textwidth}
 \begin{minipage}{\textwidth}
        \centering
        \includegraphics[height=4.5cm, keepaspectratio]{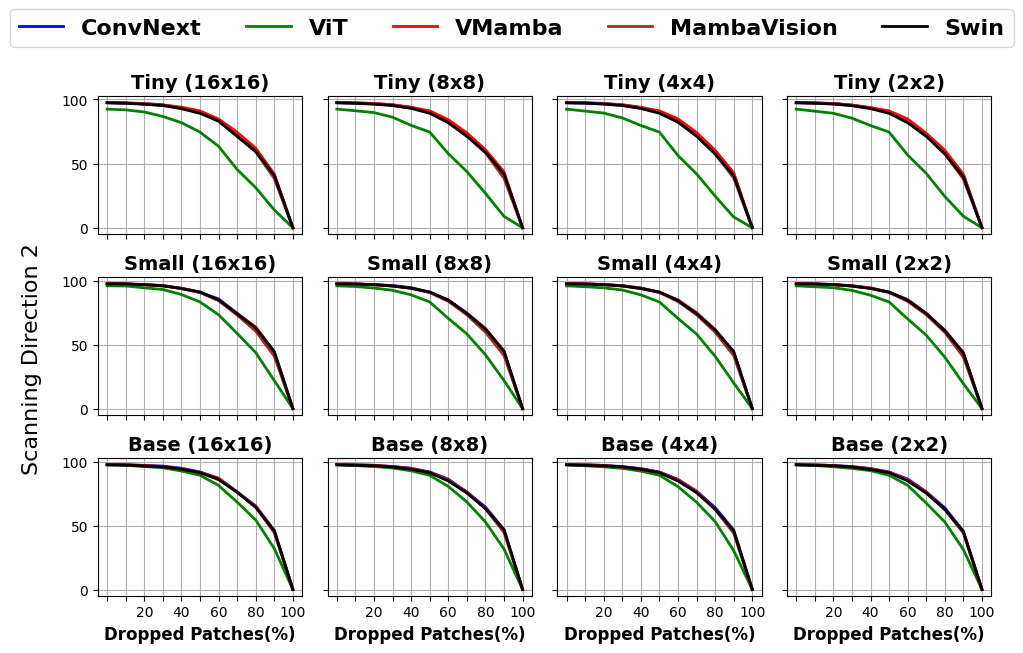}
    \end{minipage}

\end{minipage}
\begin{minipage}{0.49\textwidth}
    \begin{minipage}{\textwidth}
        \centering
        \includegraphics[height=4.5cm, keepaspectratio]{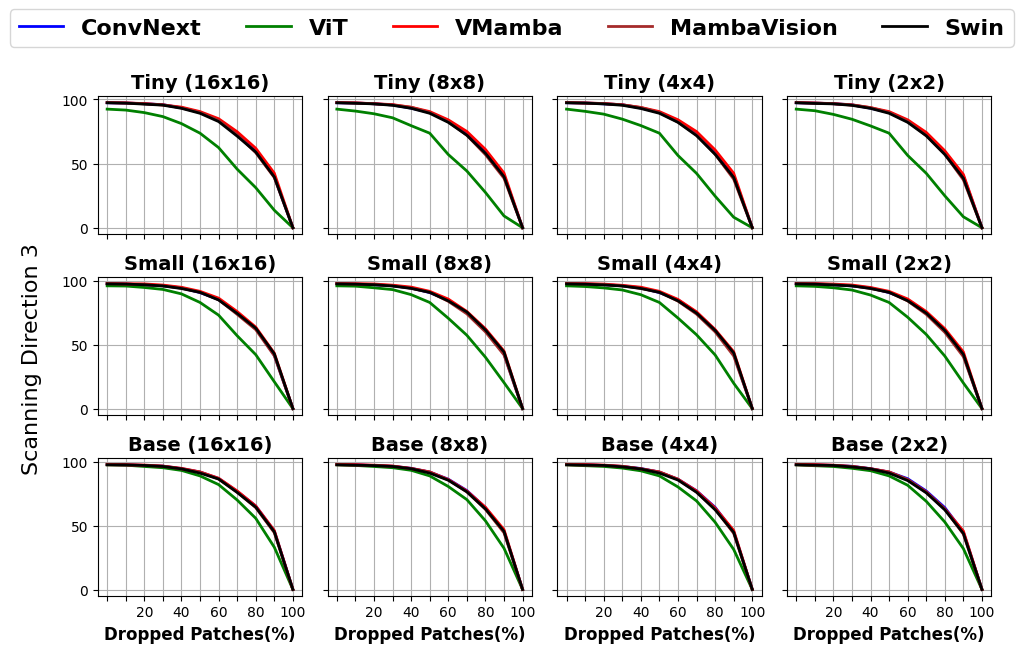}
    \end{minipage}
\end{minipage}
\begin{minipage}{0.49\textwidth}
 \begin{minipage}{\textwidth}
        \centering
        \includegraphics[height=4.5cm, keepaspectratio]{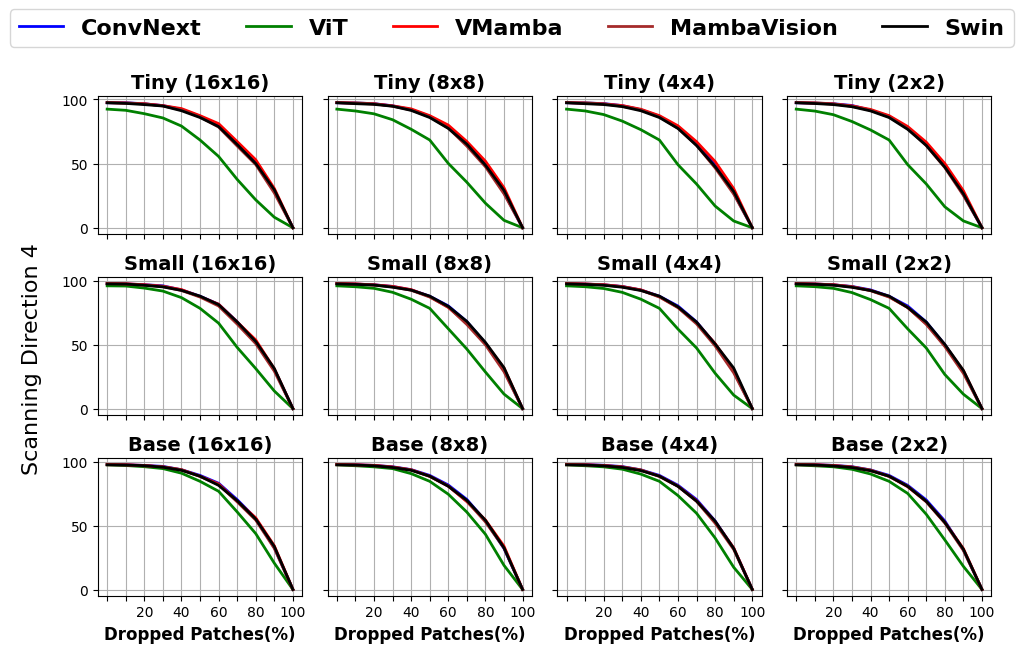}
    \end{minipage}

\end{minipage}
    \caption{\small Information drop of Tiny and Small family of models along the scanning direction: the image is split into a sequence of fixed-size non-overlapping patches of size 16x16, 8x8. 4x4, and 2x2. We report results of sequentially dropping patches along the scanning direction. The top row shows results for top-to-bottom \emph{(Direction 1)} and left-to-right direction\emph{(Direction 2)}.  The bottom row shows results for right-to-left \emph{(Direction 3)} and bottom-to-top direction\emph{(Direction 4)}.}
\label{fig:scan_exp_v3_all}

\end{figure*}

We investigate the models' behavior when information is dropped along the scanning directions. We consider three settings: (1) linearly increasing the amount of information dropped in each patch along the scanning direction, with the most information dropped in patches that are traversed last by the scanning operation, (2) dropping most of the information in the center of the scanning directions while preserving most of the information in patches that are traversed at the beginning and the end, and (3) sequentially dropping patches along the scanning directions. In Figure \ref{fig:scanning_example} we show qualitative samples for showing information drop along scanning directions.  Figure \ref{fig:scan_exp_linear_increase_all} we report results across the first experiment setting across all the scanning directions. In Figure \ref{fig:scan_exp_max_center_all} we report results for the second experiment setting. Furthermore, in Figure \ref{fig:scan_exp_v3_all} we report results across the third experiment setting across all the scanning directions.

\subsection{Random Patch Drop}
\label{app:randomdrop}
\begin{figure*}
\centering
\includegraphics[width=0.90\textwidth]{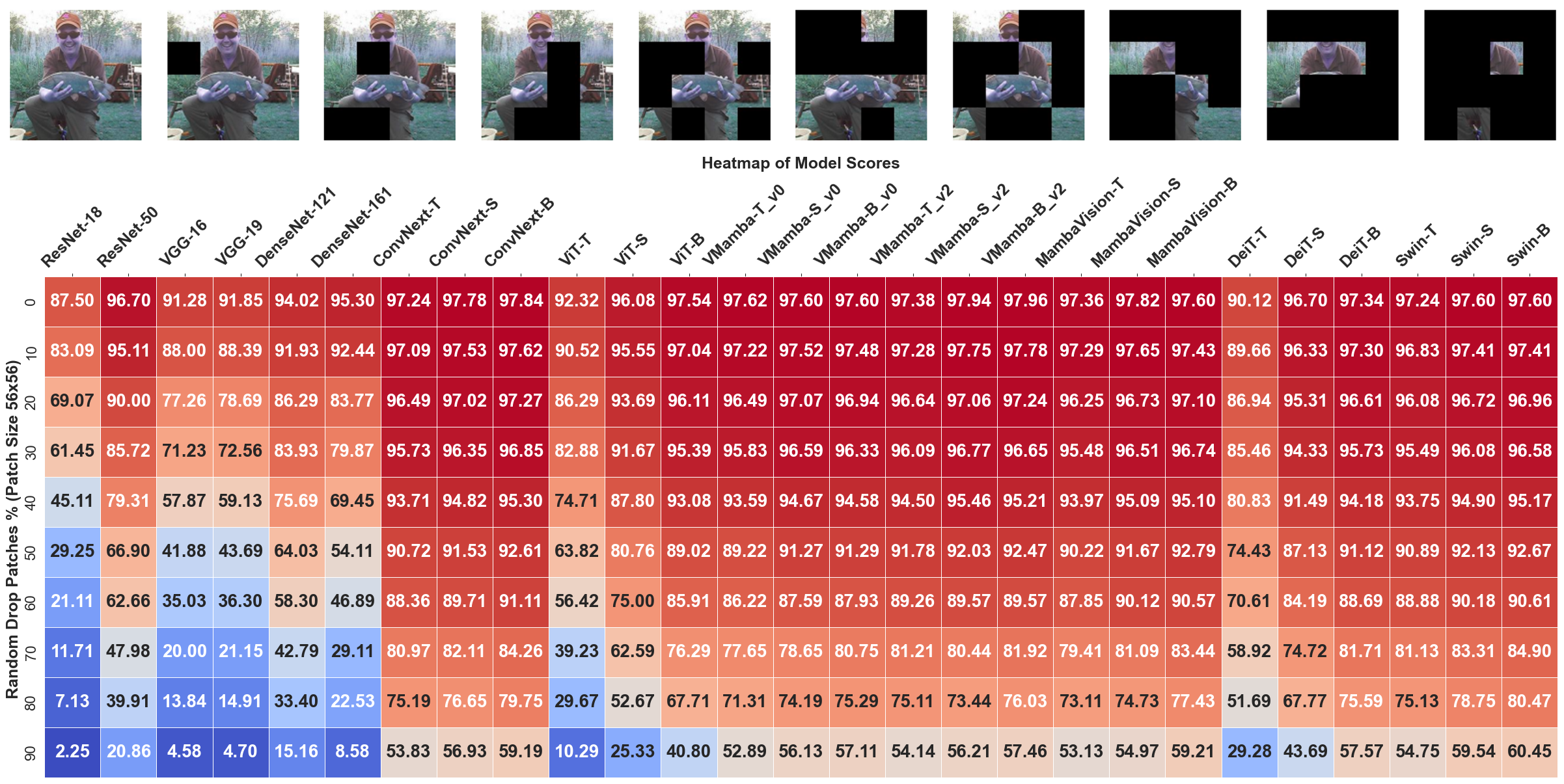}
\caption{Top-1 classification accuracy of various architectures under random patch drop occlusions, using $56\times56$  patch size.}
\label{fig:random_drop_4_4_app}
\end{figure*}

\begin{figure*}
\centering
\includegraphics[width=0.90\textwidth]{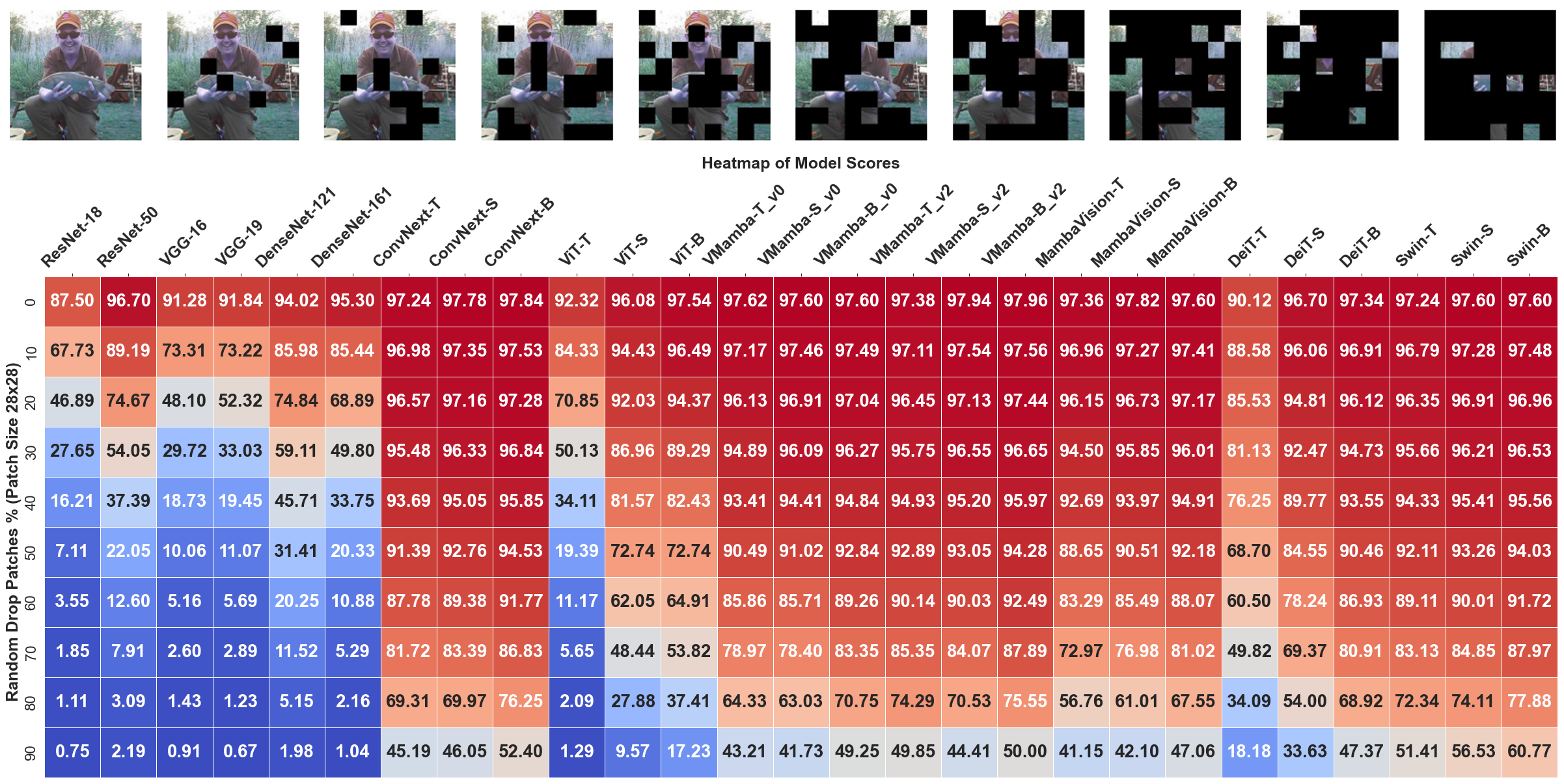}
\caption{Top-1 classification accuracy of various architectures under random patch drop occlusions, using  $28\times28$ patch size.}
\label{fig:random_drop_8_8_app}
\end{figure*}

\begin{figure*}
\centering
\includegraphics[width=0.90\textwidth]{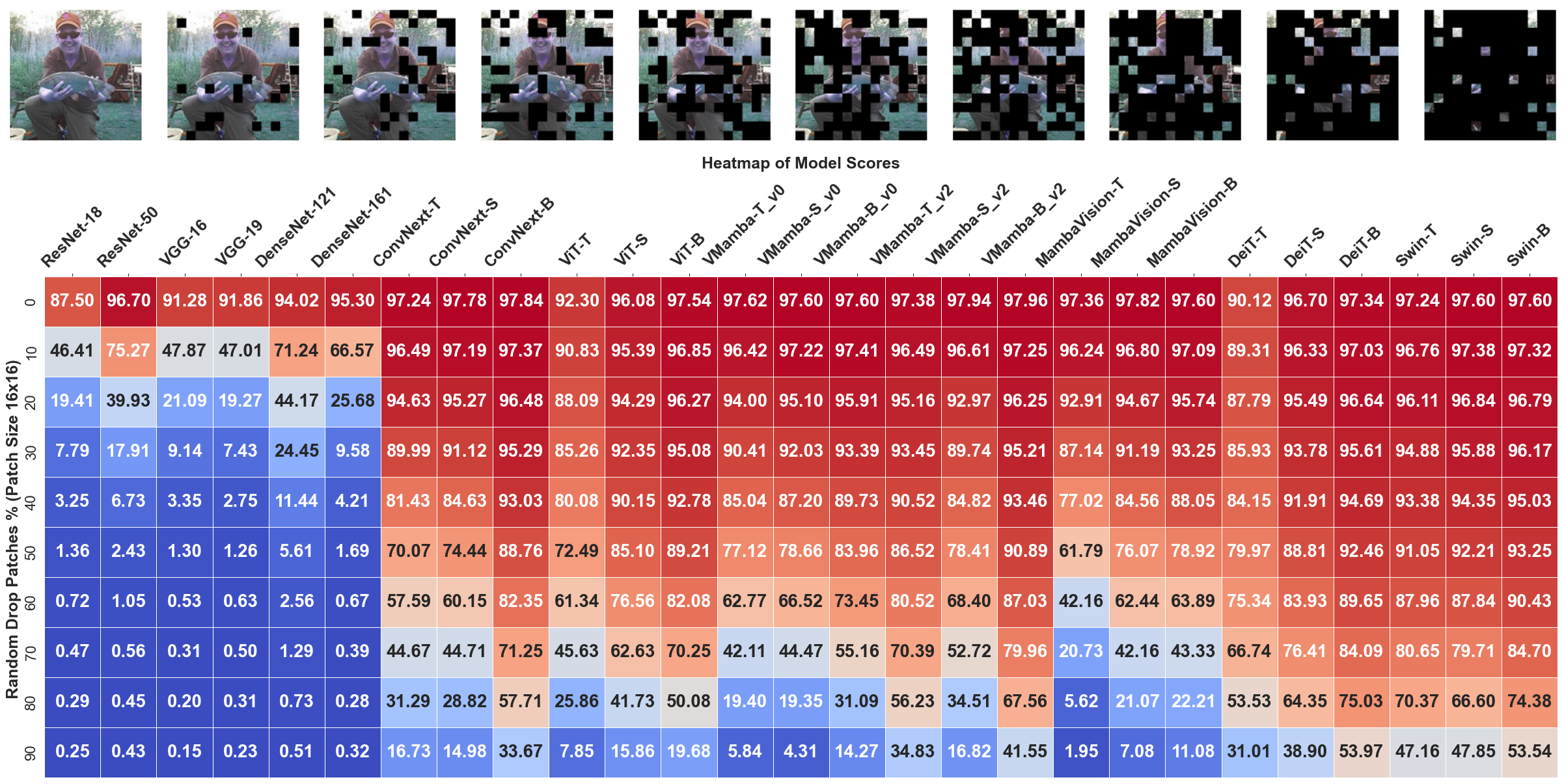}
\caption{Top-1 classification accuracy of various architectures under random patch drop occlusions, using  $16\times16$ patch size.}
\label{fig:random_drop_14_14_app}
\end{figure*}

\begin{figure*}
\centering
\includegraphics[width=0.90\textwidth]{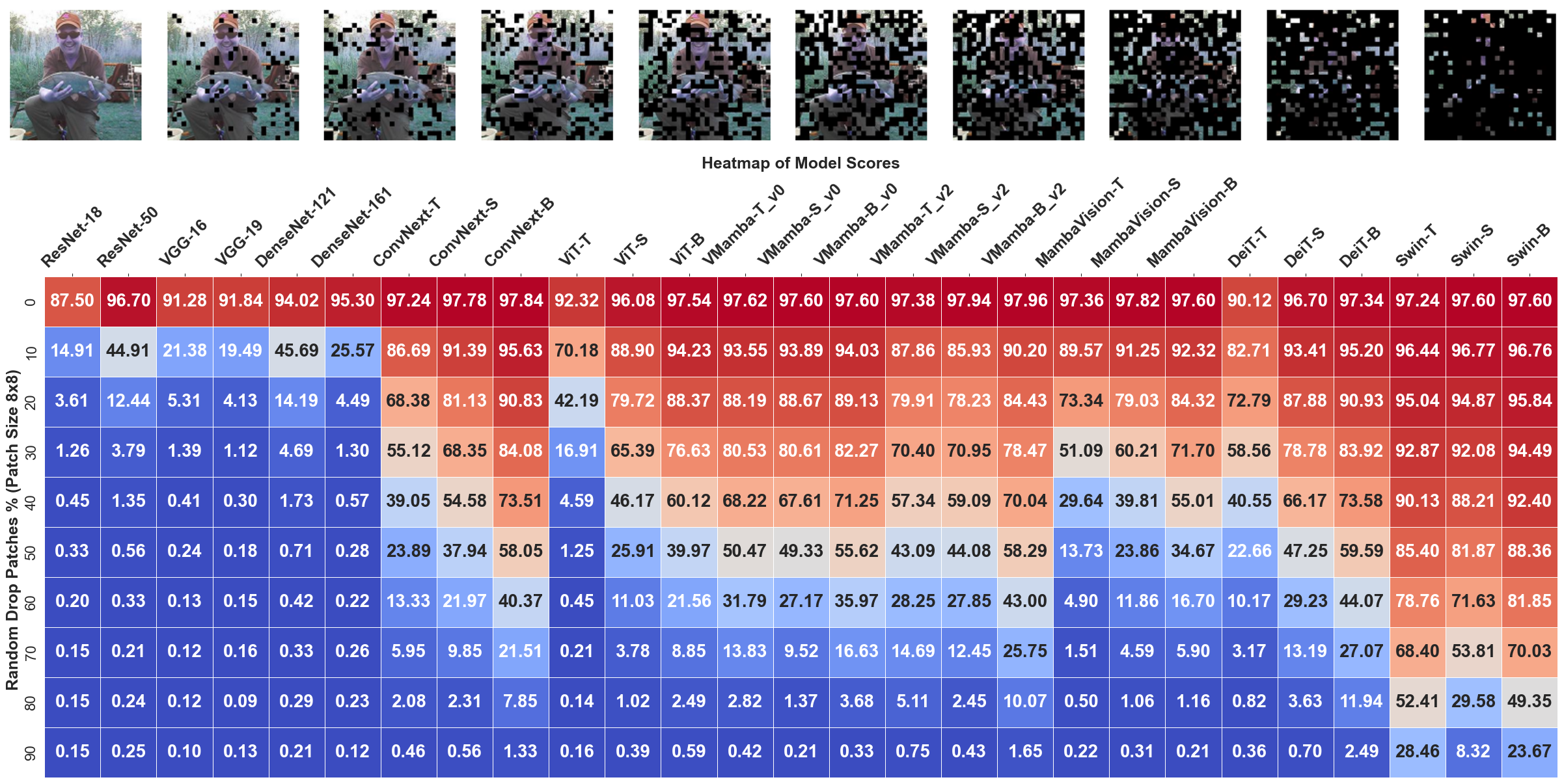}
\caption{Top-1 classification accuracy of various architectures under random patch drop occlusions, using  $8\times8$ patch size.}
\label{fig:random_drop_28_28_app}
\end{figure*}

\begin{figure*}
\centering
\includegraphics[width=0.90\textwidth]{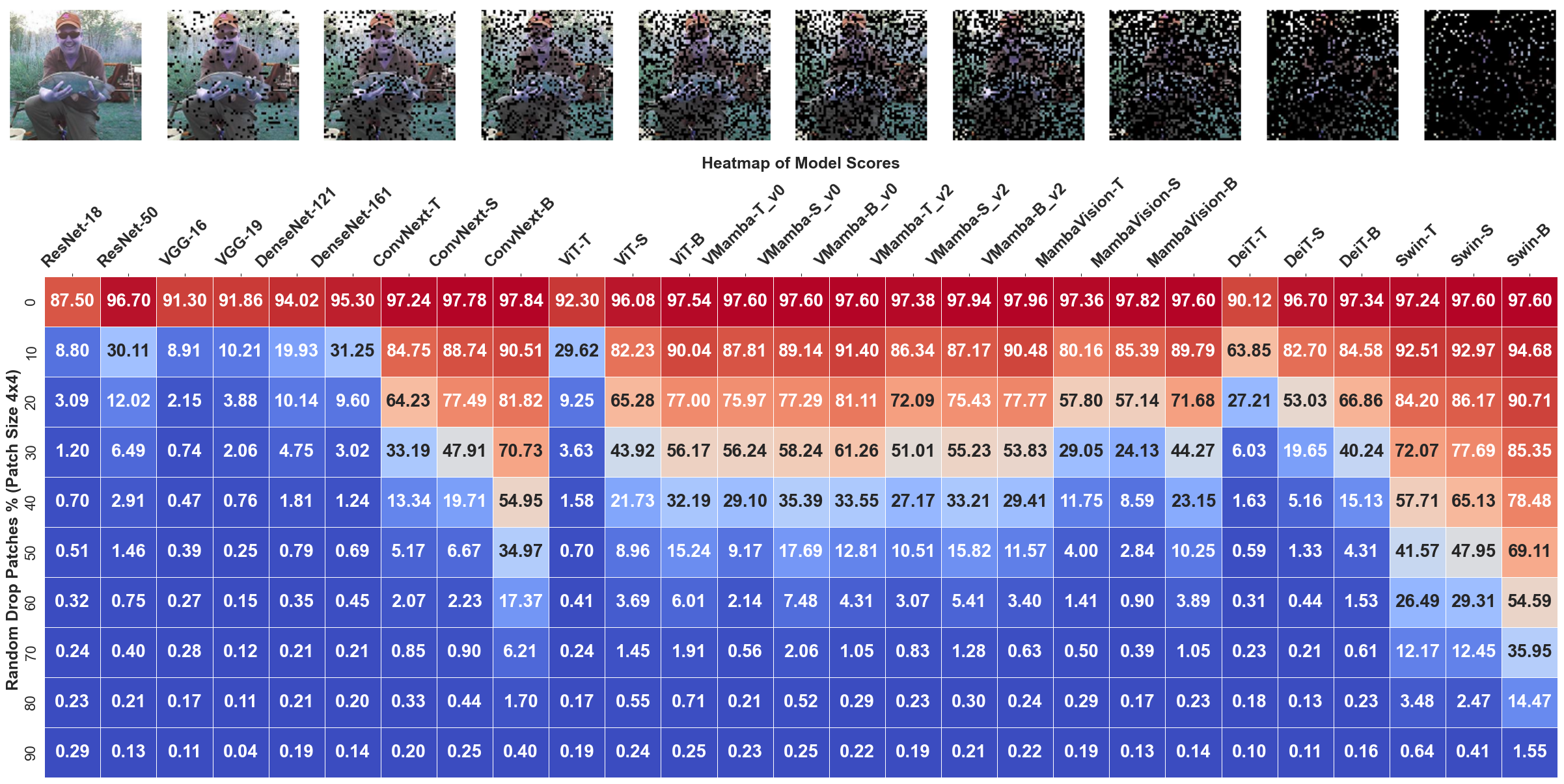}
\caption{Top-1 classification accuracy of various architectures under random patch drop occlusions, using  $4\times4$ patch size.}
\label{fig:random_drop_56_56_app}
\end{figure*}

\begin{figure*}
\centering
\includegraphics[width=0.90\textwidth]{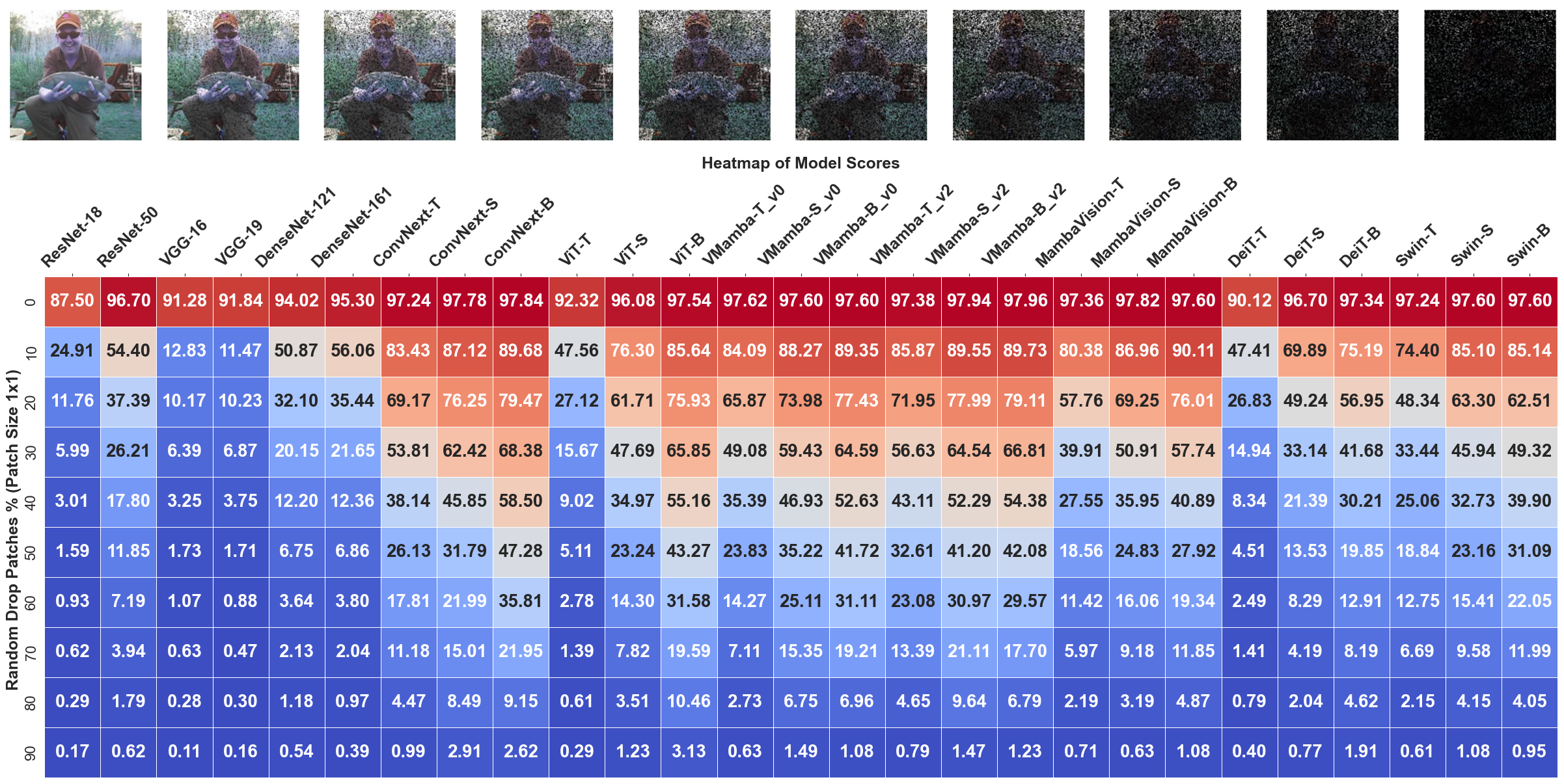}
\caption{Top-1 classification accuracy of various architectures under random patch drop occlusions, using  $1\times1$ patch size.}
\label{fig:random_drop_224_224_app}
\end{figure*}

In Table \ref{tab:randdrop_all_patch_sizes} we expand the random patch drop experiments from the main paper (Table \ref{tab:random_drop_main}) to patch sizes 56$\times$56 and 224$\times$224. Furthermore, in Figure \ref{fig:random_drop_4_4_app}, \ref{fig:random_drop_8_8_app}, \ref{fig:random_drop_14_14_app}, \ref{fig:random_drop_28_28_app}, \ref{fig:random_drop_56_56_app}, and \ref{fig:random_drop_224_224_app} we expand our analysis on random patch drop to several other models.

\subsection{Salient and Non-Salient Patch Drop}
\label{app:dinodrop}
\begin{figure*}
\centering
\includegraphics[width=\textwidth]{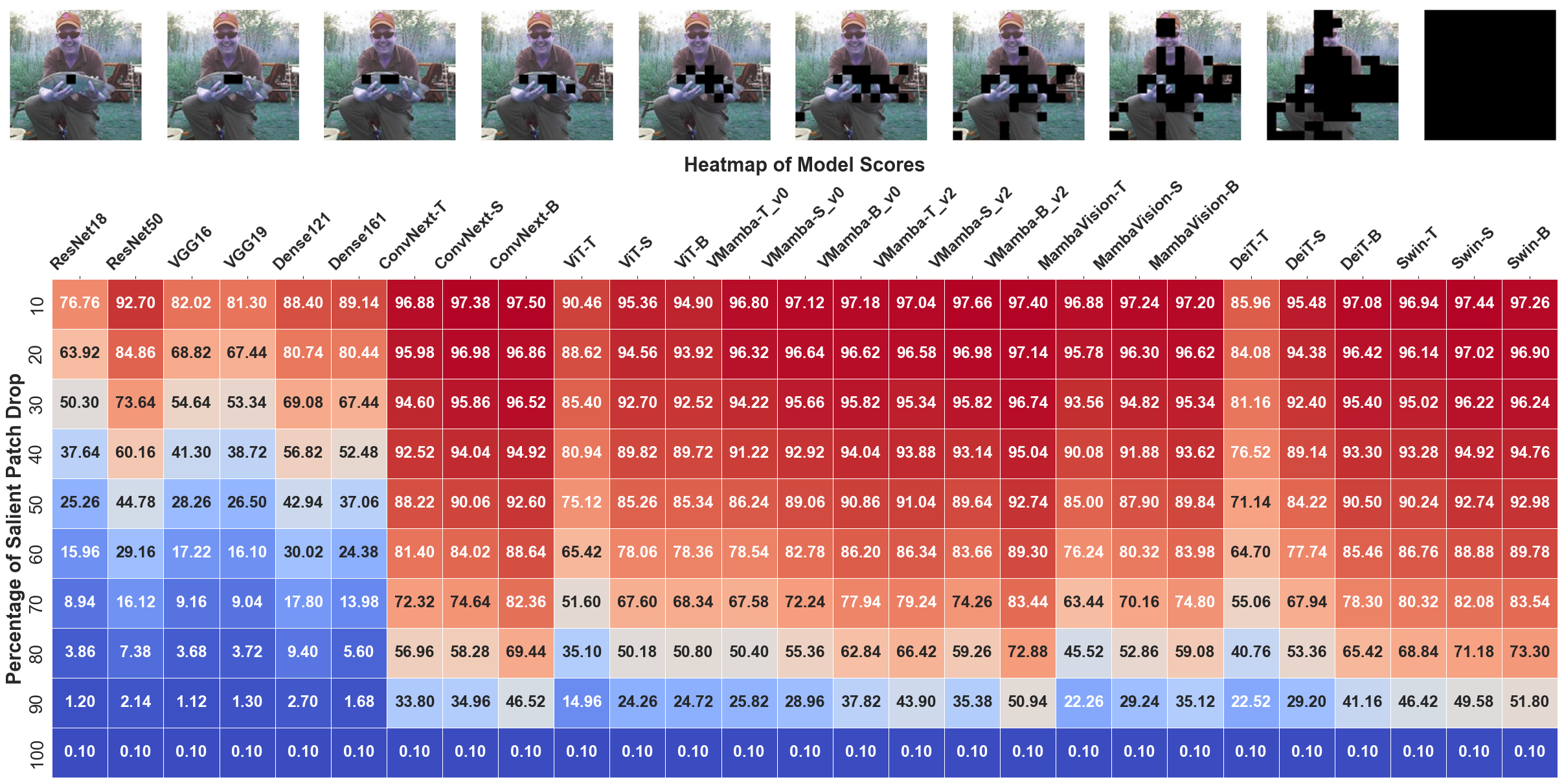}
\caption{ Top-1 classification accuracy reported under salient patch drop occlusion using $16\times16$ patch size.}
\label{fig:dino_drop_best}
\end{figure*}

\begin{figure*}
\centering
\includegraphics[width=\textwidth]{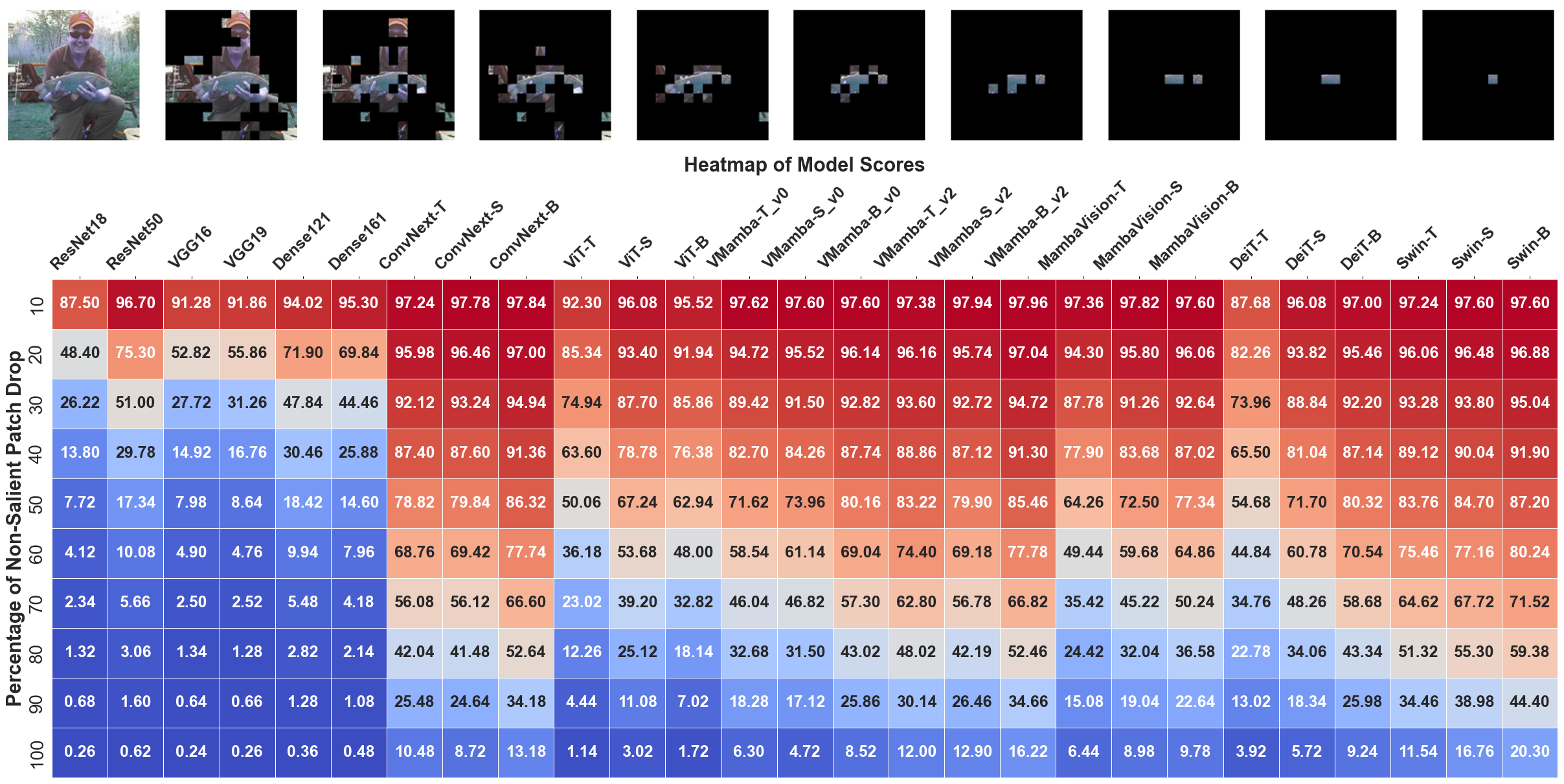}
\caption{ Top-1 classification accuracy reported under non-salient patch drop occlusion using $16\times16$ patch size.}
\label{fig:dino_drop_worst}
\end{figure*}

In Figure \ref{fig:dino_drop_best} and \ref{fig:dino_drop_worst}, we report results on salient and non-salient patch drop of information.

\subsection{Patch Shuffling}
\label{app:shuffle}
\begin{figure*}
\centering
\includegraphics[width=0.90\textwidth]{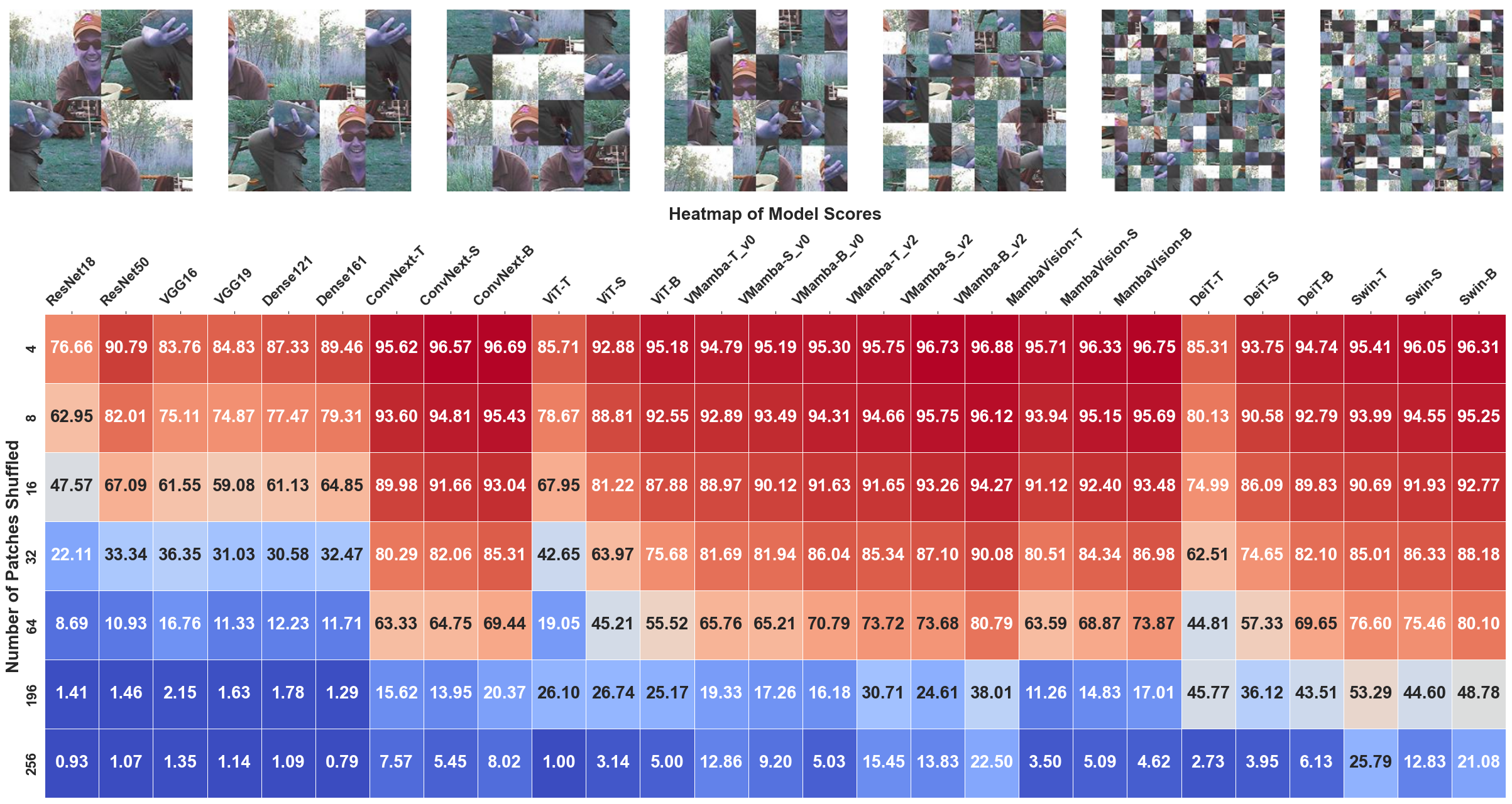}
\caption{ Top-1 classification accuracy among various architectures under increasing patch shuffling.}
\label{fig:shuffle_app}
\end{figure*}

In Figure \ref{fig:shuffle_app}, we expand the patch shuffling experiment from the main paper to several CNN and transformer-based architectures.

\section{Robustness against Image Corruptions}
\label{app:image_corruptions}
\subsection{Robustness against Global Corruptions}
\label{app:imagent_domain}
\begin{figure*}
\centering
\includegraphics[width=\textwidth]{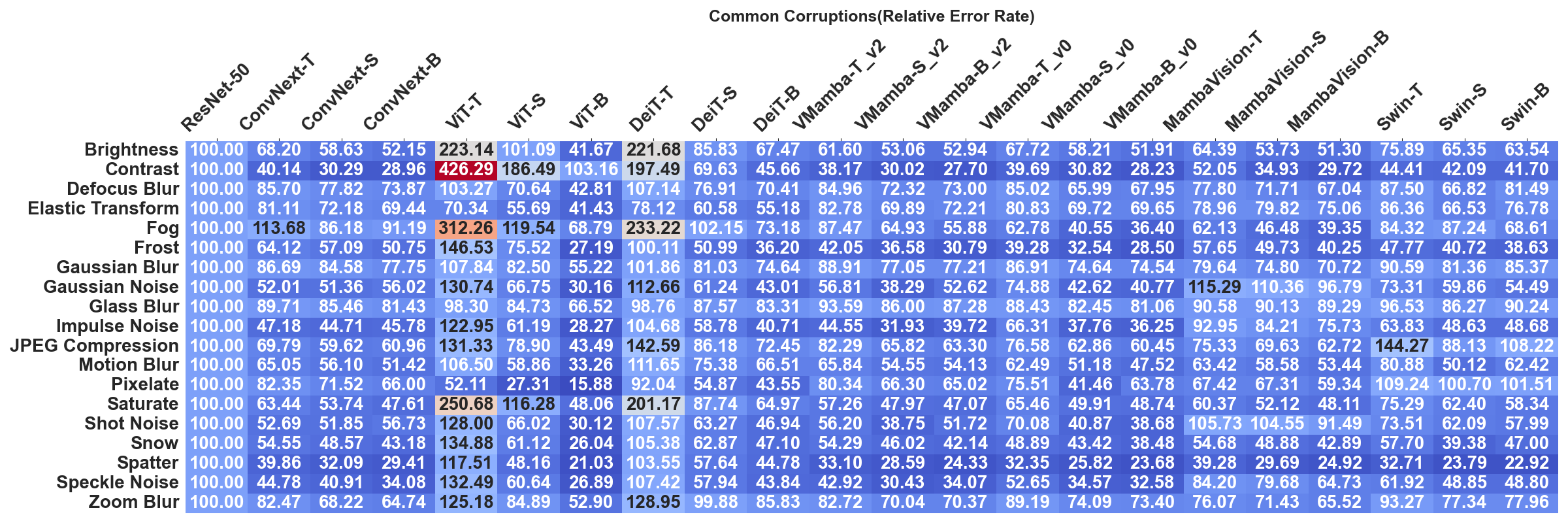}
\caption{Corruption Error(CE) values for different corruptions and architectures on ImageNet-C, with error rates relative to ResNet-50.}
\label{fig:common_c_imagent_app}
\end{figure*}

\begin{figure*}
\centering
\includegraphics[width=\textwidth]{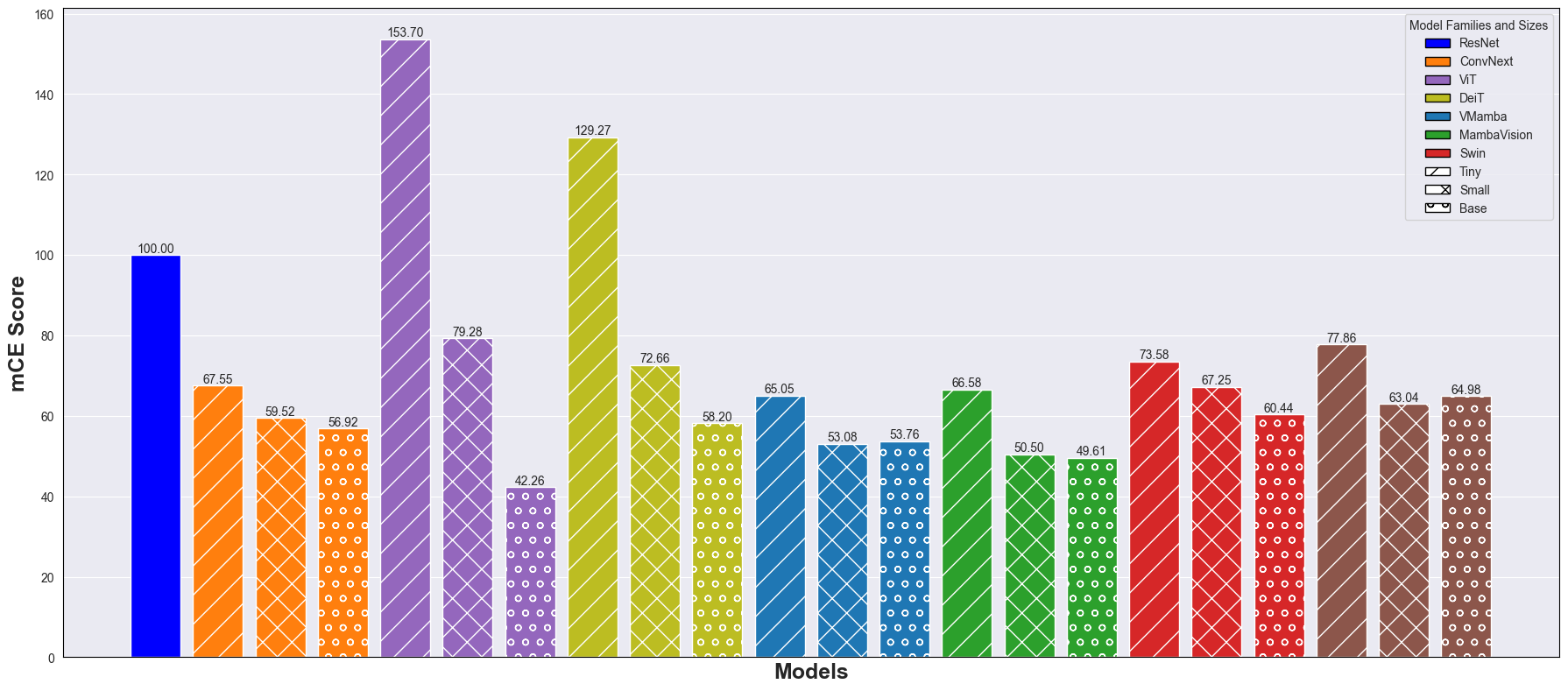}
\caption{mCE for different corruptions and architectures on ImageNet-C, with error rates relative to ResNet-50.}
\label{fig:common_c_mce_imagent_app}
\end{figure*}

In Table \ref{tab:common_corrup_main} relative corruption error is reported across models discussed in the main paper. In Figure \ref{fig:common_c_imagent_app} and \ref{fig:common_c_mce_imagent_app}, we present the relative corruption error and mean corruption error (mCE) of various models subjected to all 19 corruption methods applied to the ImageNet dataset. The common corruption consists of  various types of synthetic corruptions to assess the models' robustness against image distortions like noise, blur, and compression artifacts.  In Table \ref{tab:robustness_} we report results on \textit{\textbf{ImageNetV2}}~\cite{recht2019imagenet}, \textit{\textbf{ImageNet-A}}~\cite{hendrycks2021natural}, \textit{\textbf{ImageNet-R}}~\cite{hendrycks2021many}, and \textit{\textbf{ImageNet-S}}~\cite{gao2022large}. These datasets are designed to evaluate the robustness and generalization capabilities of models trained on the original ImageNet dataset. ImageNetV2~\cite{recht2019imagenet} tests the models' performance on previously unseen images that follow the same distribution as the training data, while ImageNet-A~\cite{hendrycks2021natural} contains naturally occurring adversarial examples that are difficult for models to classify correctly. ImageNet-R~\cite{hendrycks2021many} consists of images with different artistic renditions to test the models' ability to generalize to different visual domains and styles, and ImageNet-S~\cite{gao2022large} consists of sketch-based images.

\subsection{Model Calibration}
\label{app:calibration}

\begin{figure*}[h]
\begin{minipage}{\textwidth}

\centering




\begin{minipage}{0.19\textwidth}
  \centering
  \includegraphics[height=3.8cm, width=\linewidth , keepaspectratio]{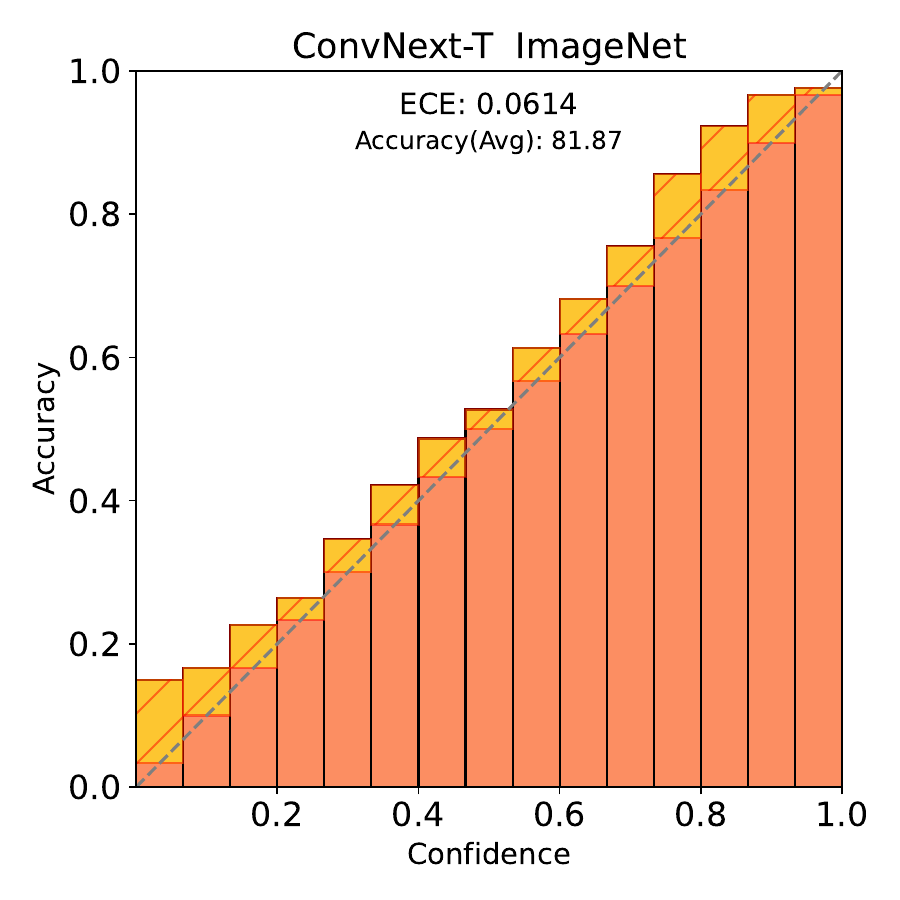}
\end{minipage}
\begin{minipage}{0.19\textwidth}
  \centering
  \includegraphics[height=3.8cm, width=\linewidth, keepaspectratio ]{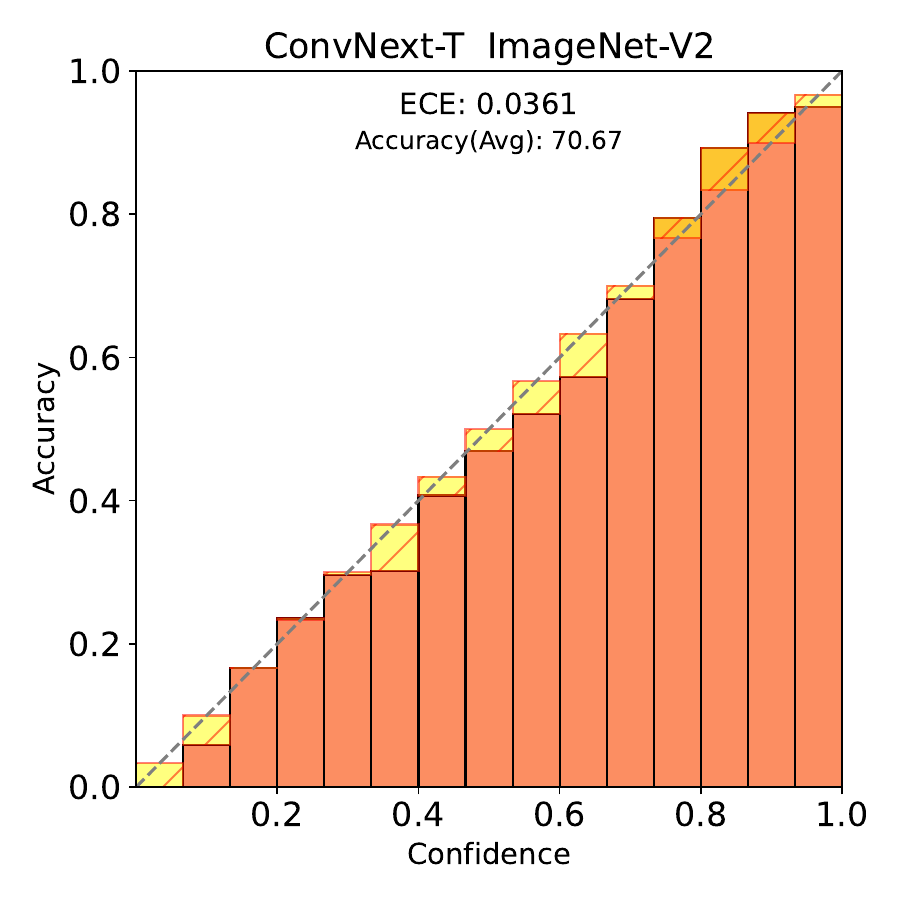}
\end{minipage}
\begin{minipage}{0.19\textwidth}
  \centering
  \includegraphics[height=3.8cm, width=\linewidth, keepaspectratio]{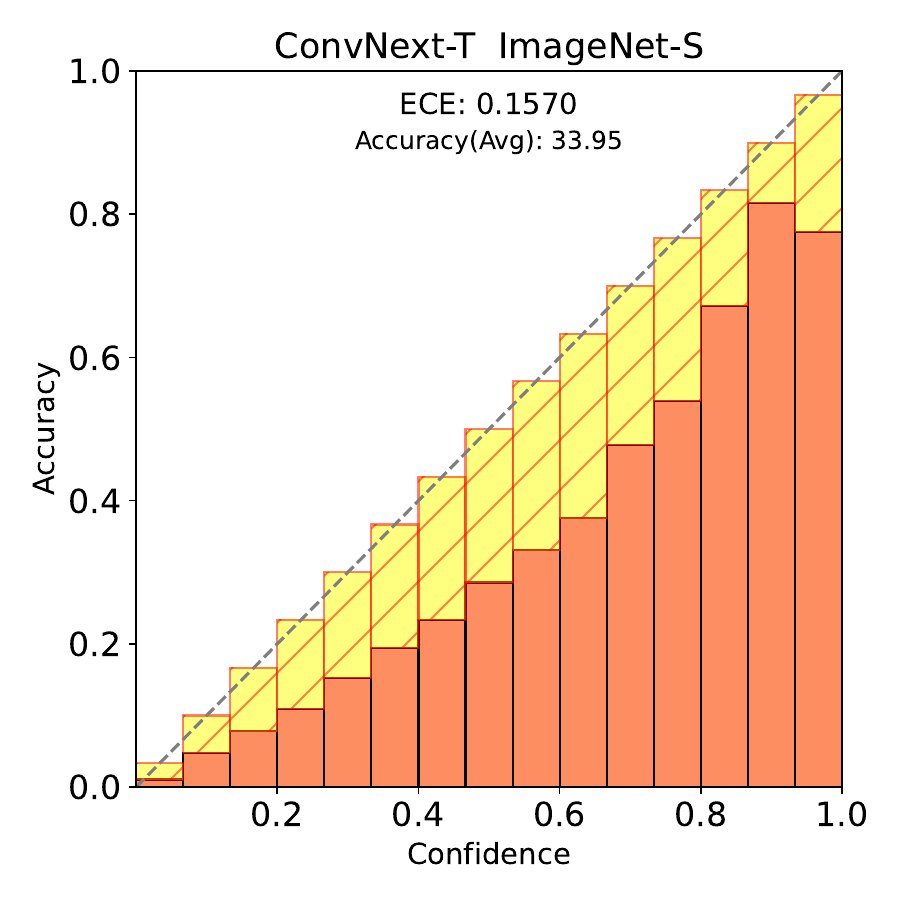}
\end{minipage}
\begin{minipage}{0.19\textwidth}
  \centering
  \includegraphics[height=3.8cm, width=\linewidth, keepaspectratio]{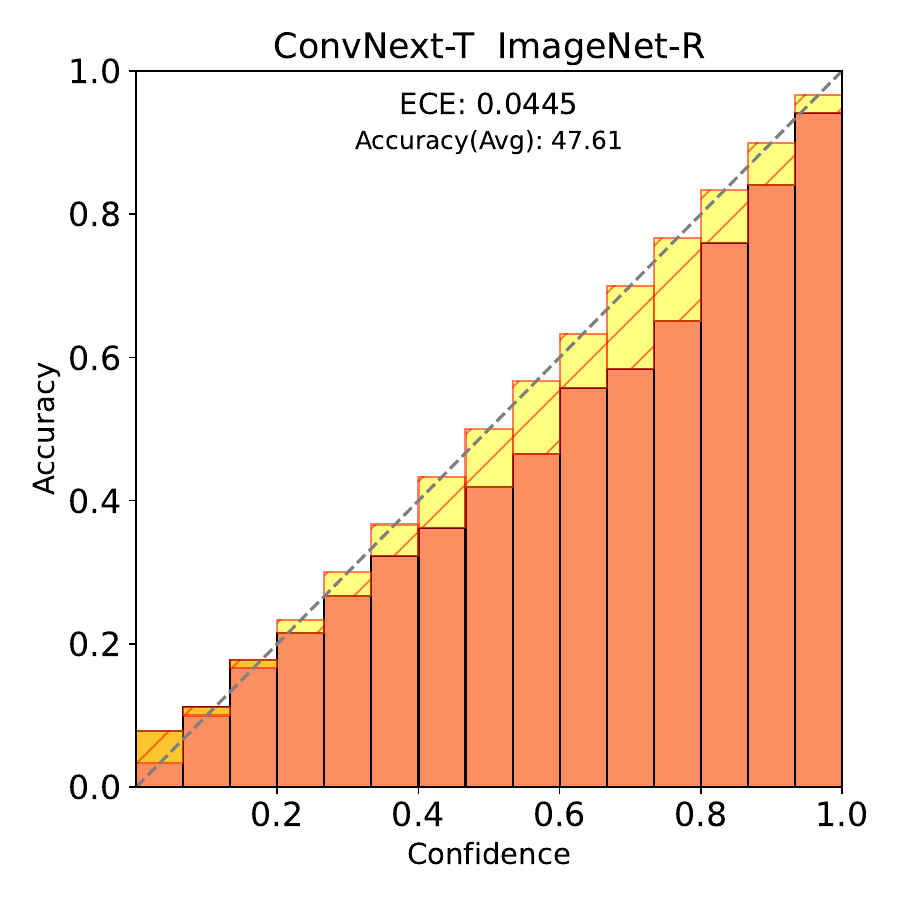}
\end{minipage}
\begin{minipage}{0.19\textwidth}
  \centering
  \includegraphics[height=3.8cm, width=\linewidth, keepaspectratio]{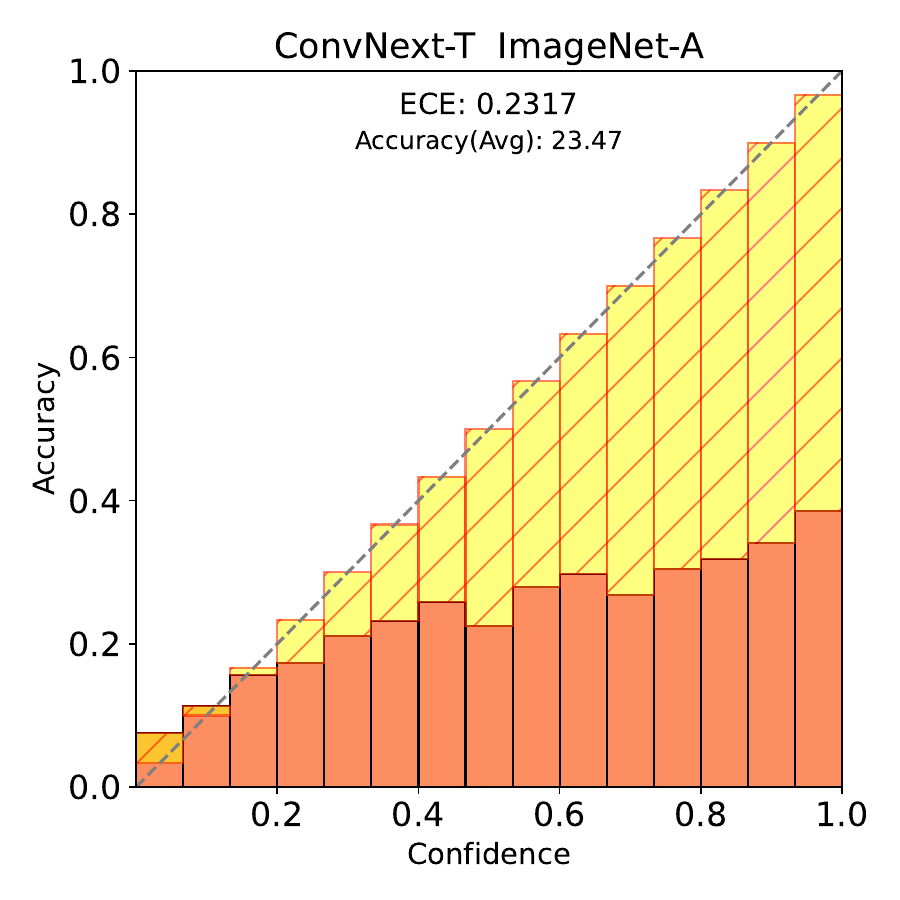}
\end{minipage}

\begin{minipage}{0.19\textwidth}
  \centering
  \includegraphics[height=3.8cm, width=\linewidth , keepaspectratio]{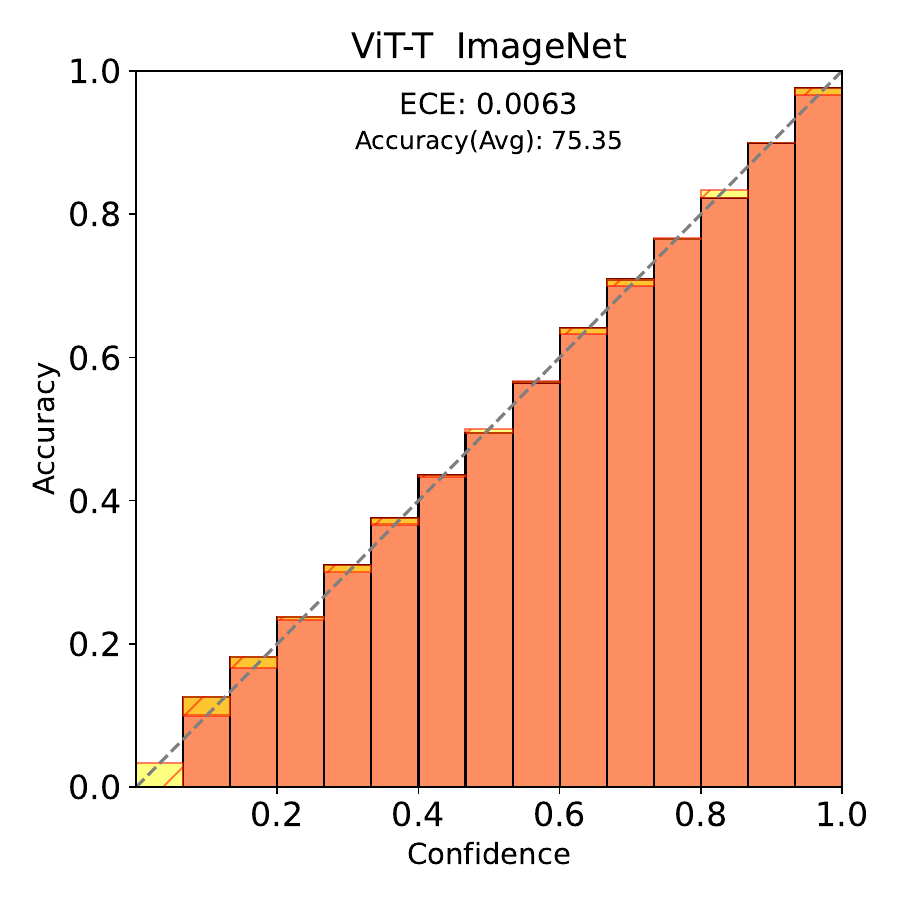}
\end{minipage}
\begin{minipage}{0.19\textwidth}
  \centering
  \includegraphics[height=3.8cm, width=\linewidth, keepaspectratio ]{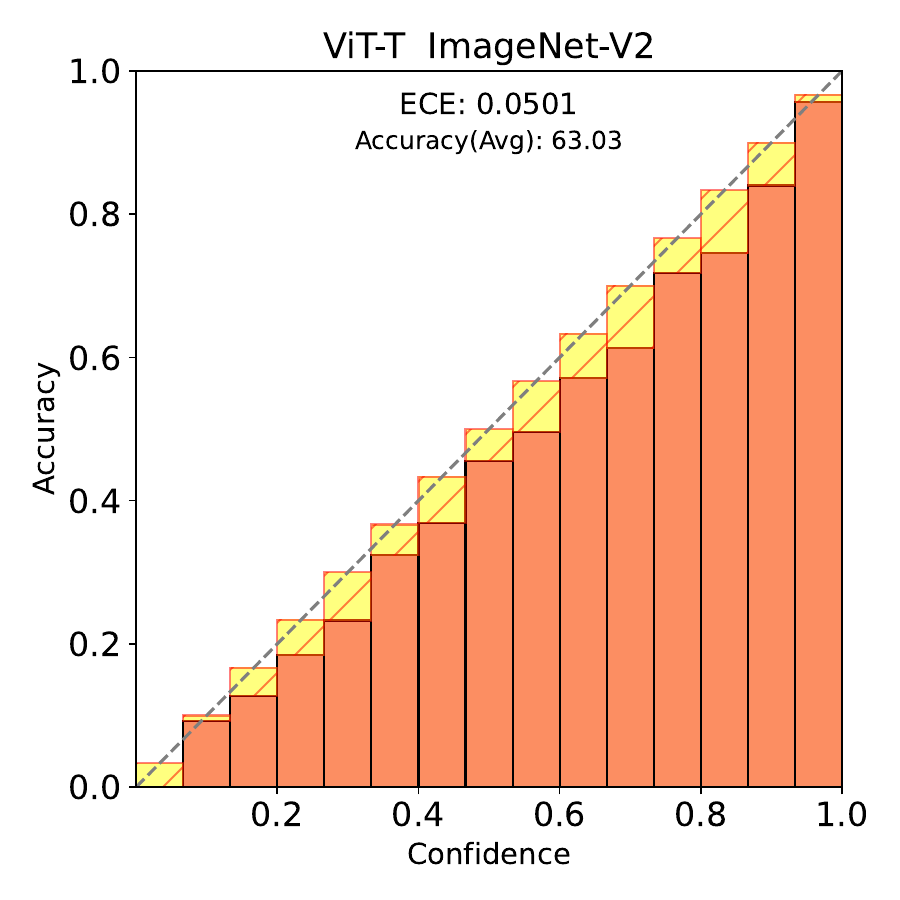}
\end{minipage}
\begin{minipage}{0.19\textwidth}
  \centering
  \includegraphics[height=3.8cm, width=\linewidth, keepaspectratio]{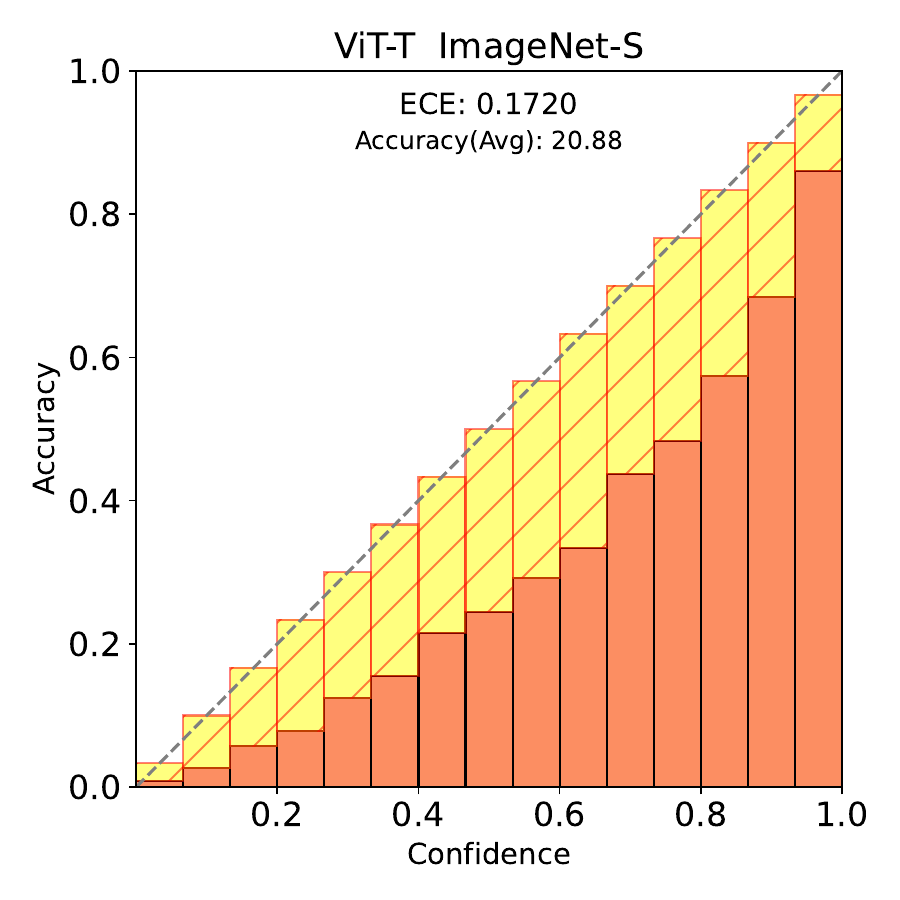}
\end{minipage}
\begin{minipage}{0.19\textwidth}
  \centering
  \includegraphics[height=3.8cm, width=\linewidth, keepaspectratio]{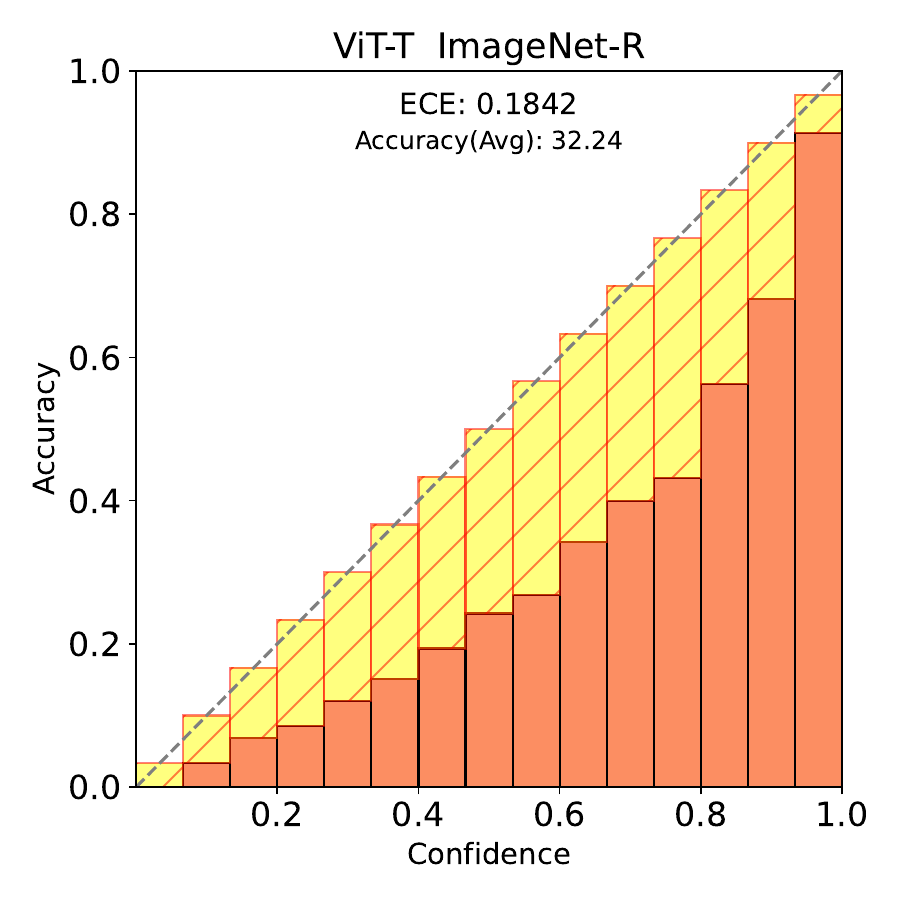}
\end{minipage}
\begin{minipage}{0.19\textwidth}
  \centering
  \includegraphics[height=3.8cm, width=\linewidth, keepaspectratio]{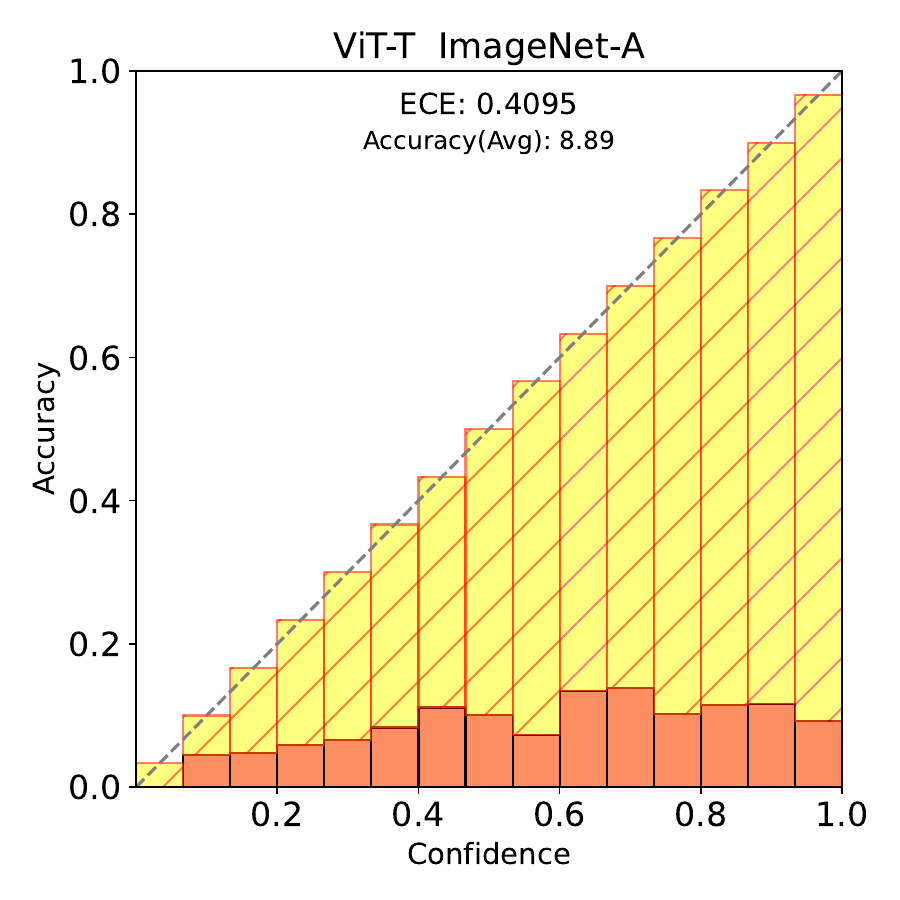}
\end{minipage}


\begin{minipage}{0.19\textwidth}
  \centering
  \includegraphics[height=3.8cm, width=\linewidth , keepaspectratio]{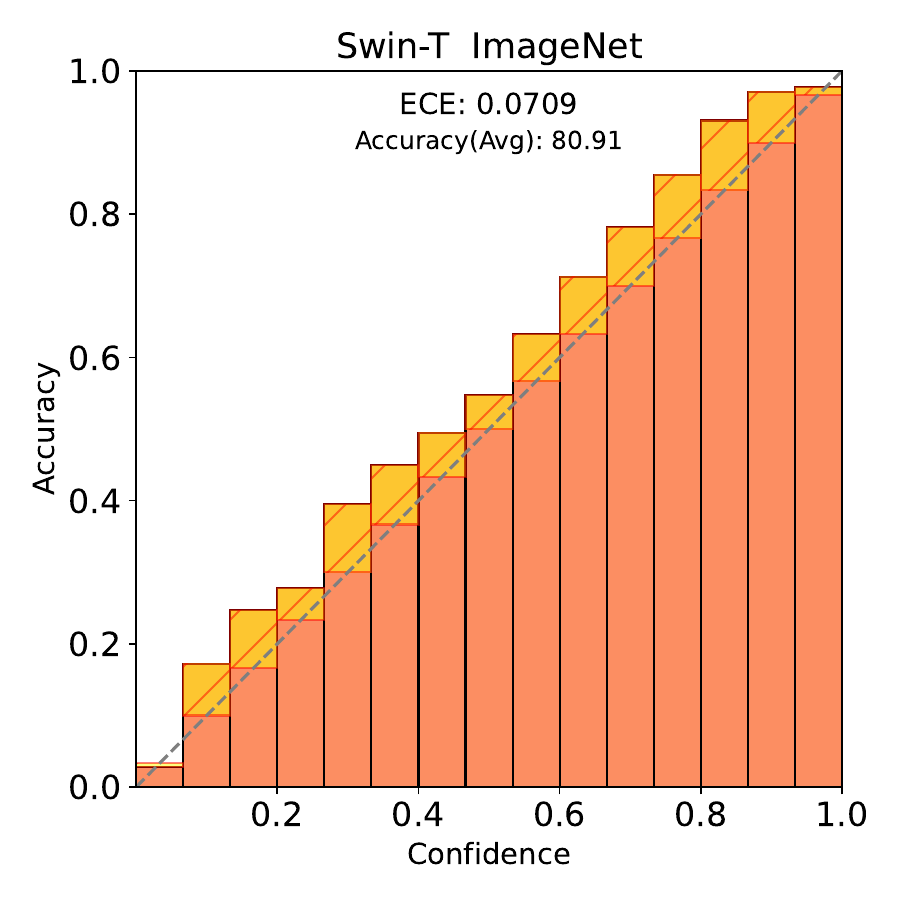}
\end{minipage}
\begin{minipage}{0.19\textwidth}
  \centering
  \includegraphics[height=3.8cm, width=\linewidth, keepaspectratio ]{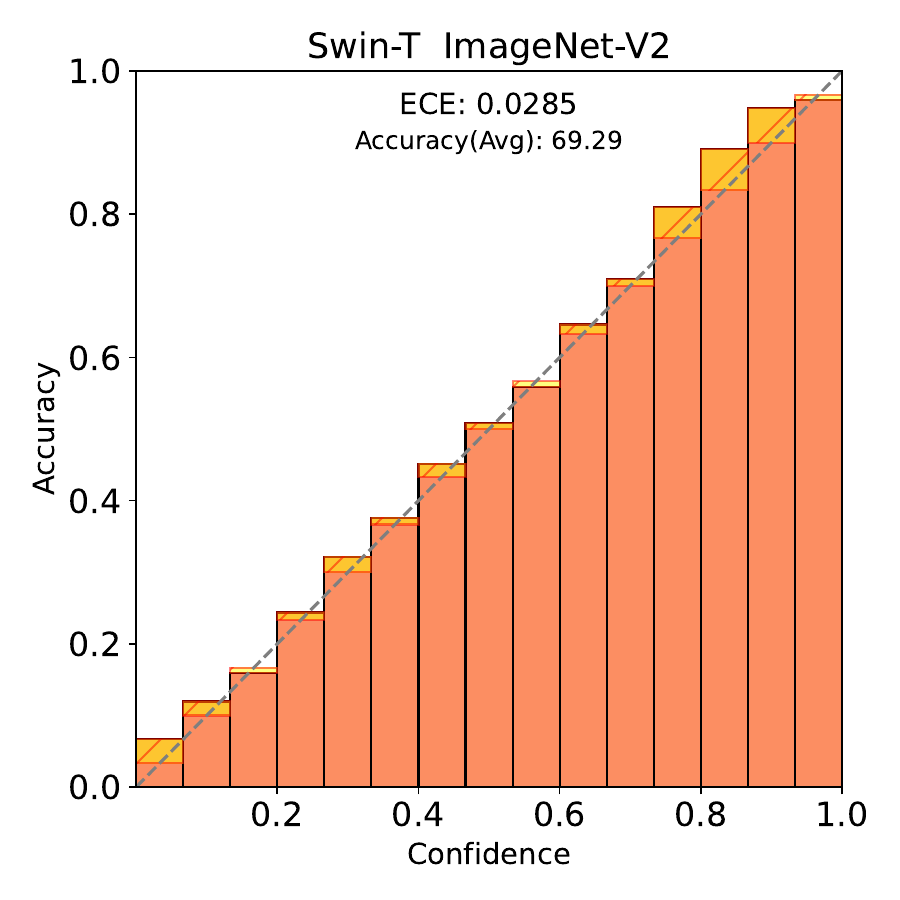}
\end{minipage}
\begin{minipage}{0.19\textwidth}
  \centering
  \includegraphics[height=3.8cm, width=\linewidth, keepaspectratio]{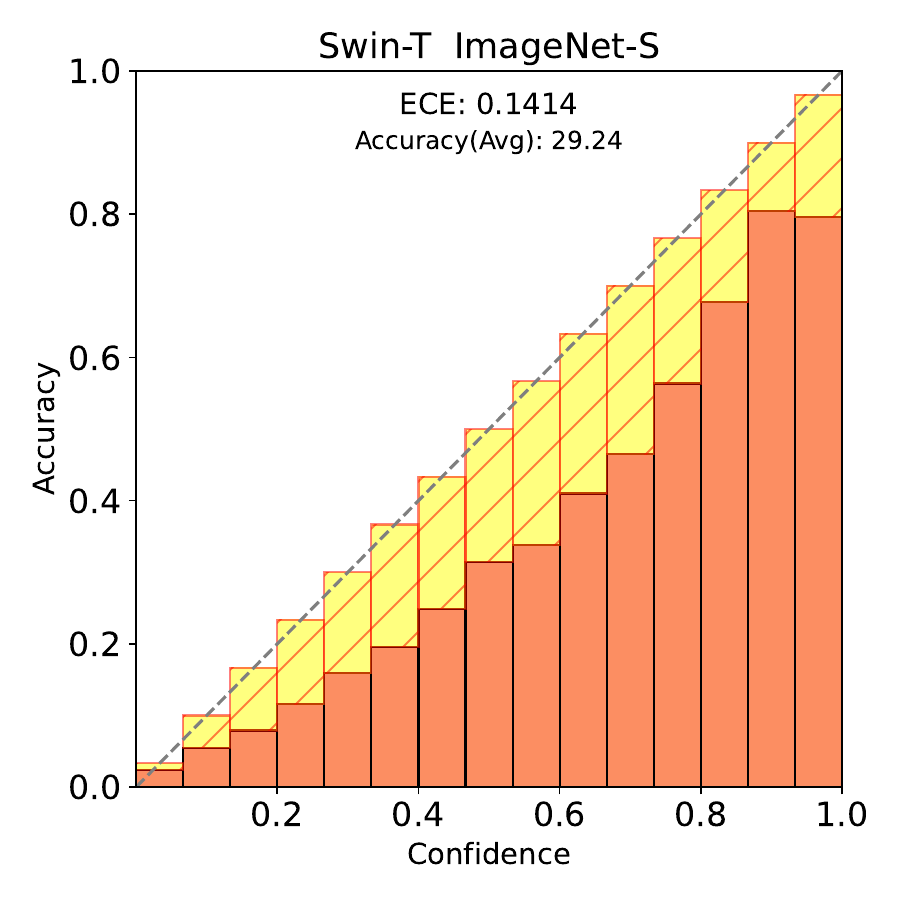}
\end{minipage}
\begin{minipage}{0.19\textwidth}
  \centering
  \includegraphics[height=3.8cm, width=\linewidth, keepaspectratio]{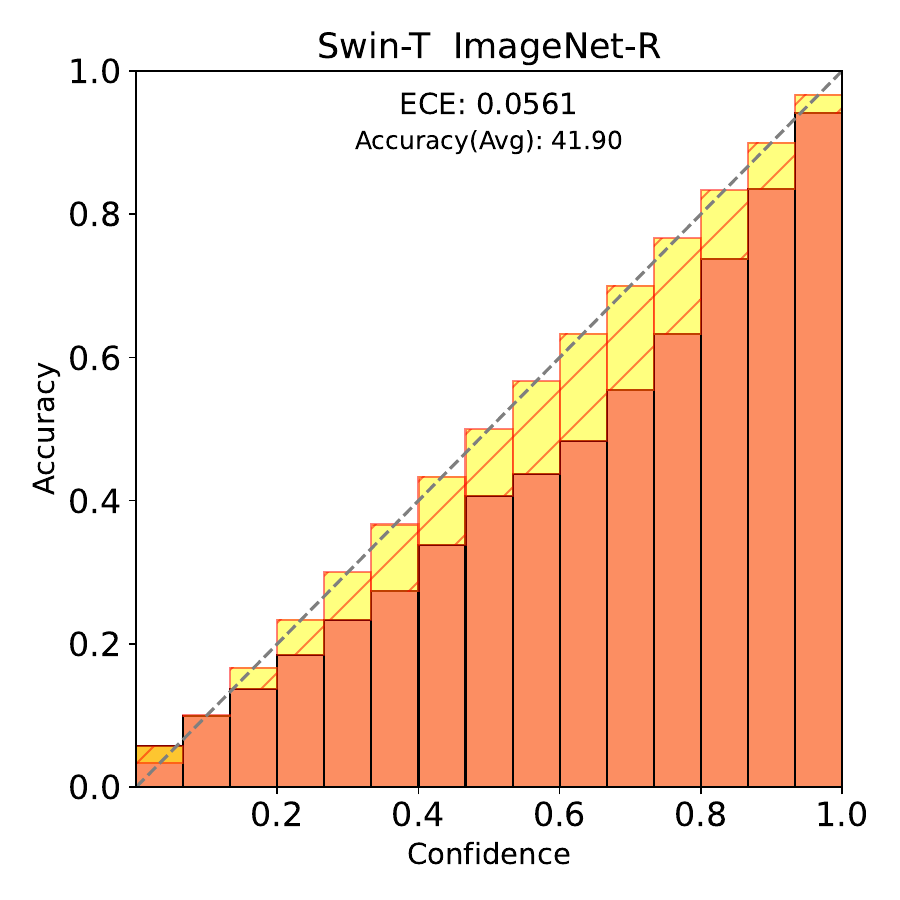}
\end{minipage}
\begin{minipage}{0.19\textwidth}
  \centering
  \includegraphics[height=3.8cm, width=\linewidth, keepaspectratio]{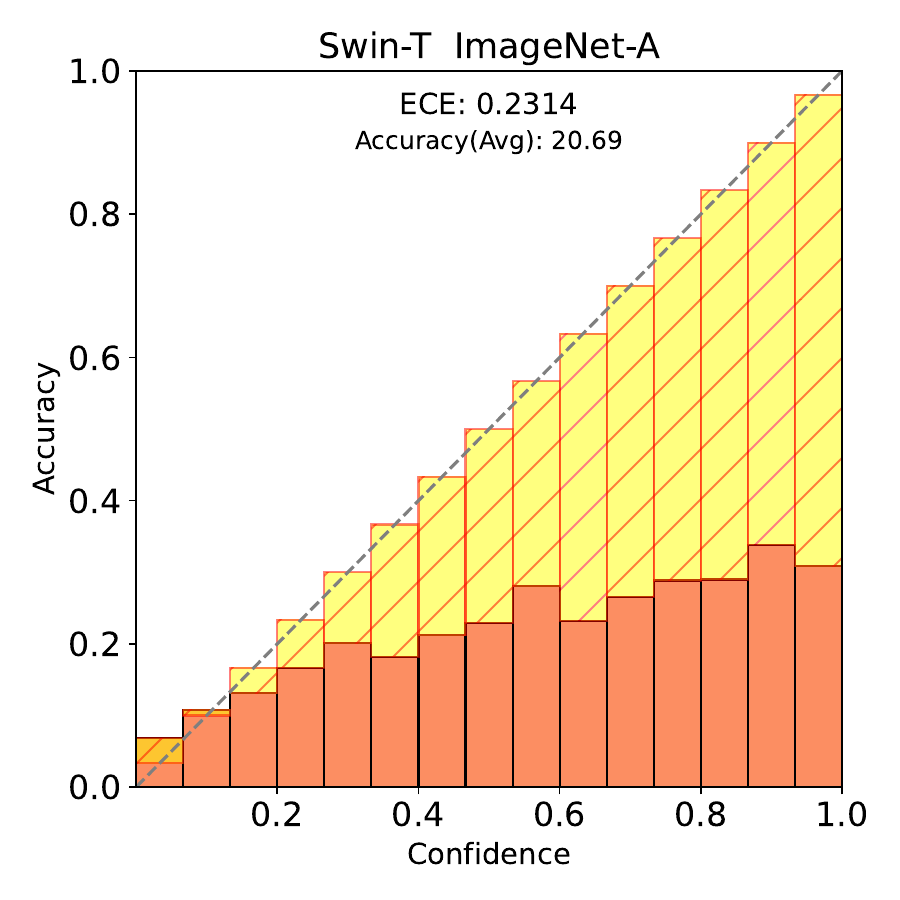}
\end{minipage}


\begin{minipage}{0.19\textwidth}
  \centering
  \includegraphics[height=3.8cm, width=\linewidth , keepaspectratio]{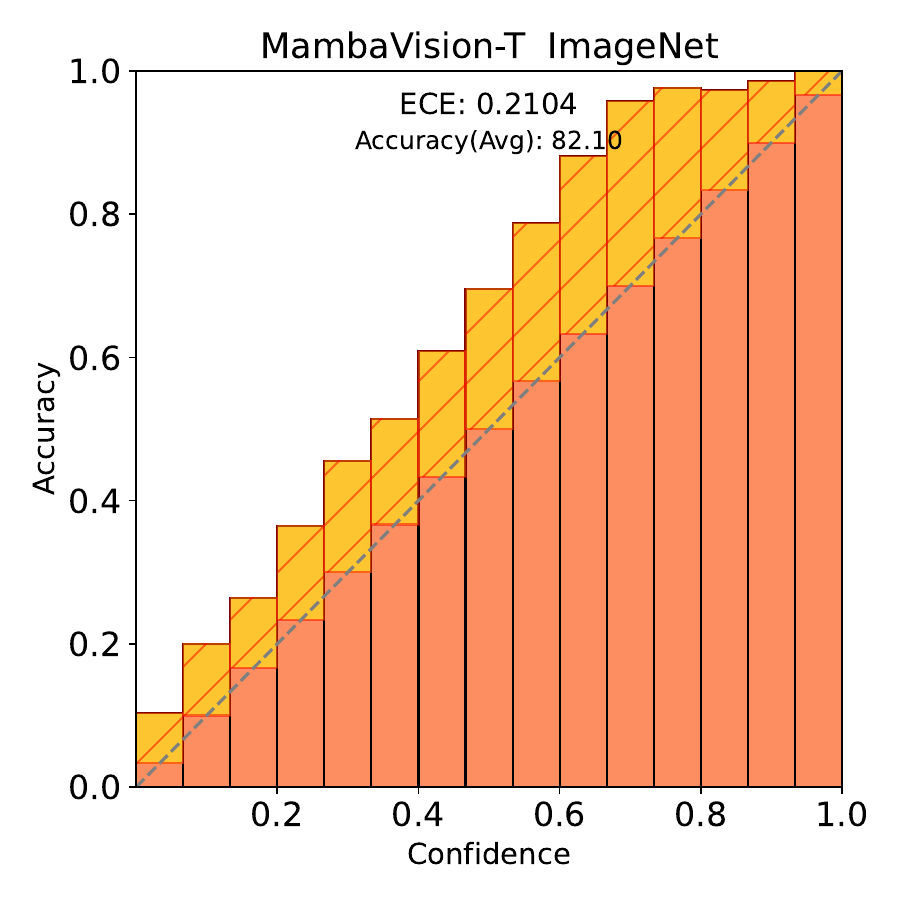}
\end{minipage}
\begin{minipage}{0.19\textwidth}
  \centering
  \includegraphics[height=3.8cm, width=\linewidth, keepaspectratio ]{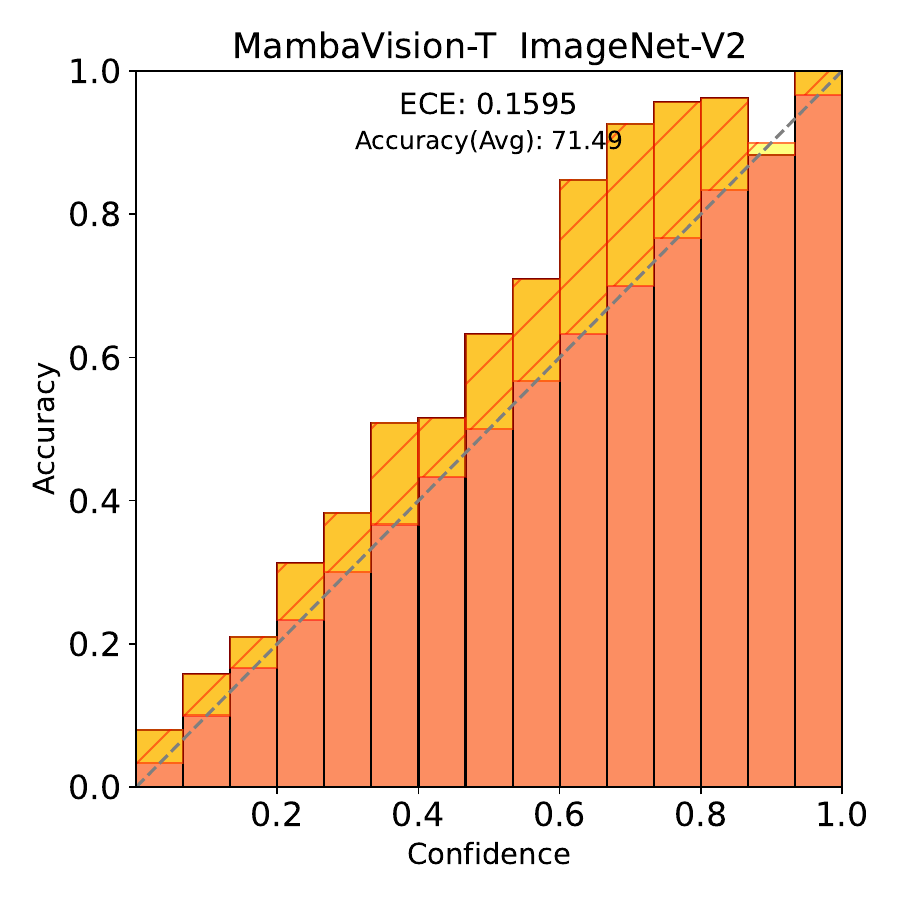}
\end{minipage}
\begin{minipage}{0.19\textwidth}
  \centering
  \includegraphics[height=3.8cm, width=\linewidth, keepaspectratio]{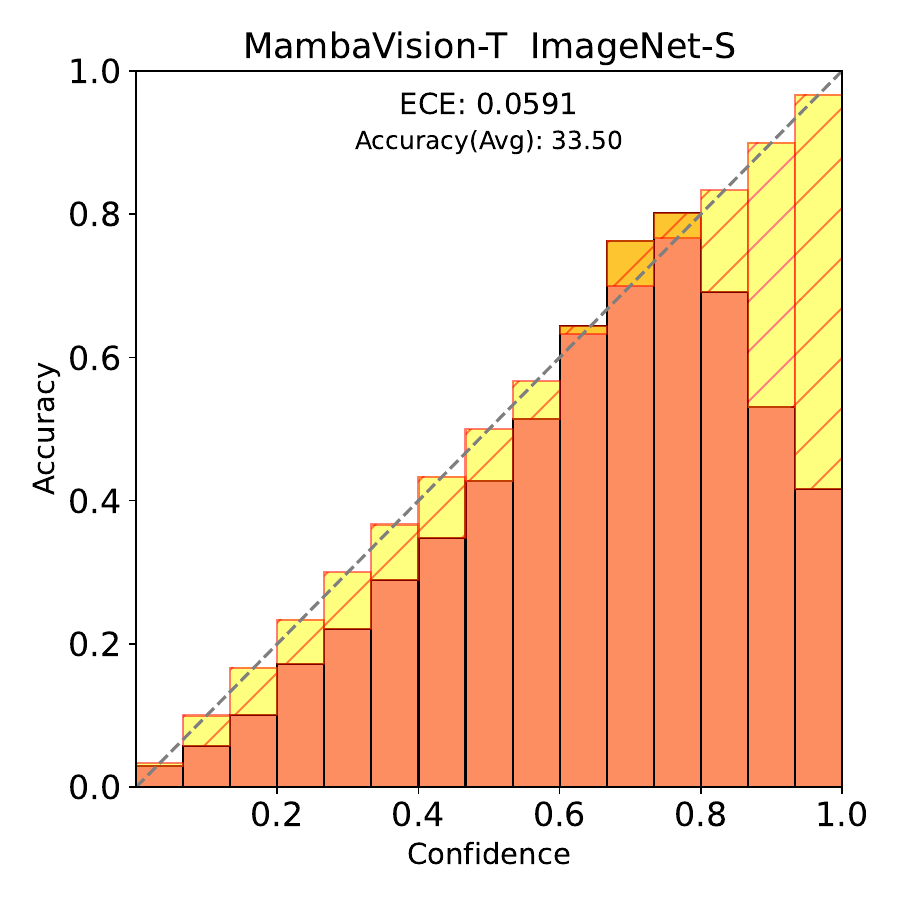}
\end{minipage}
\begin{minipage}{0.19\textwidth}
  \centering
  \includegraphics[height=3.8cm, width=\linewidth, keepaspectratio]{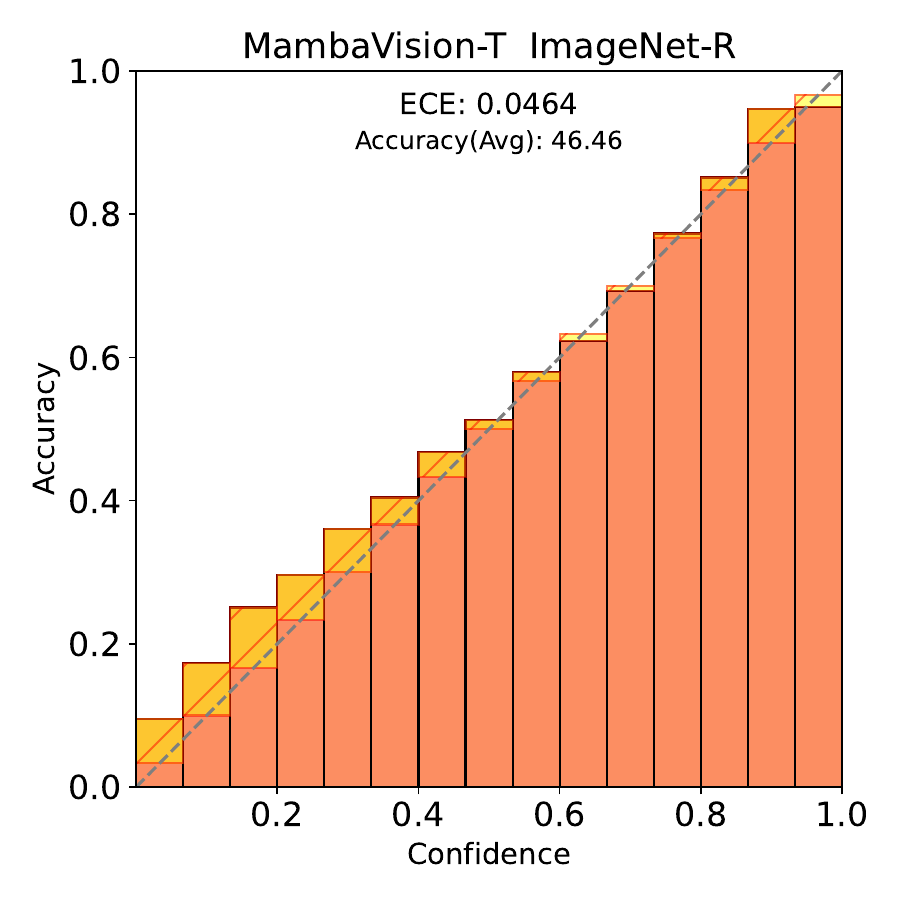}
\end{minipage}
\begin{minipage}{0.19\textwidth}
  \centering
  \includegraphics[height=3.8cm, width=\linewidth, keepaspectratio]{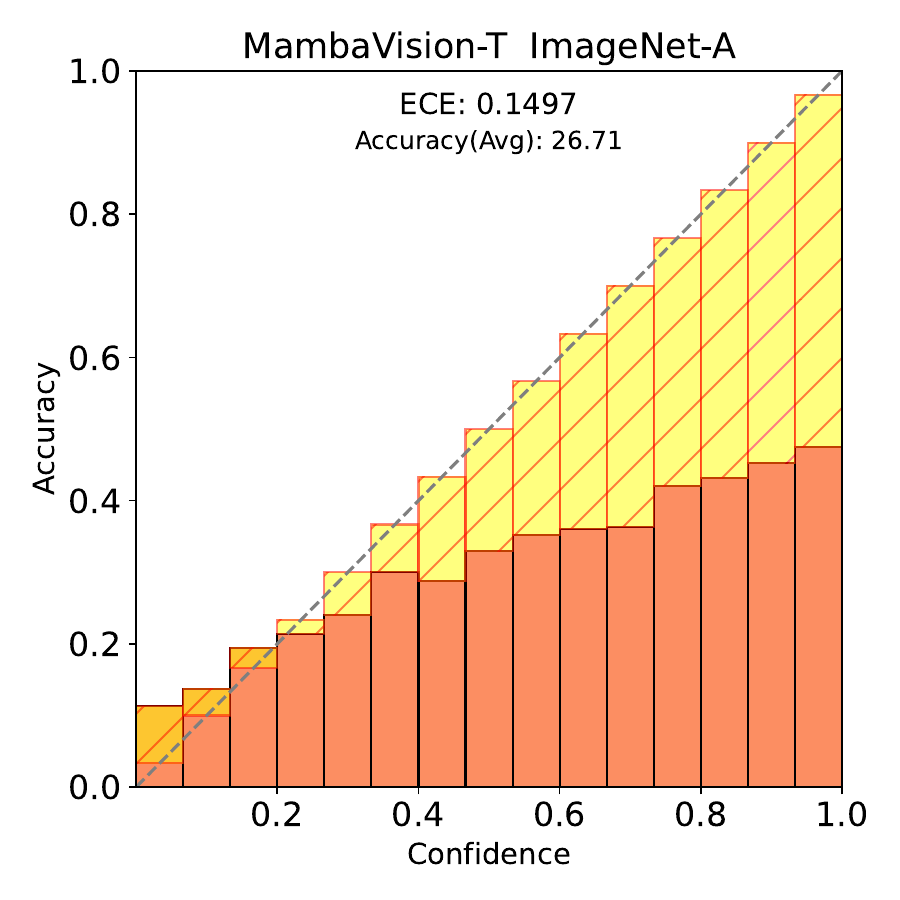}
\end{minipage}


\begin{minipage}{0.19\textwidth}
  \centering
  \includegraphics[height=3.8cm, width=\linewidth , keepaspectratio]{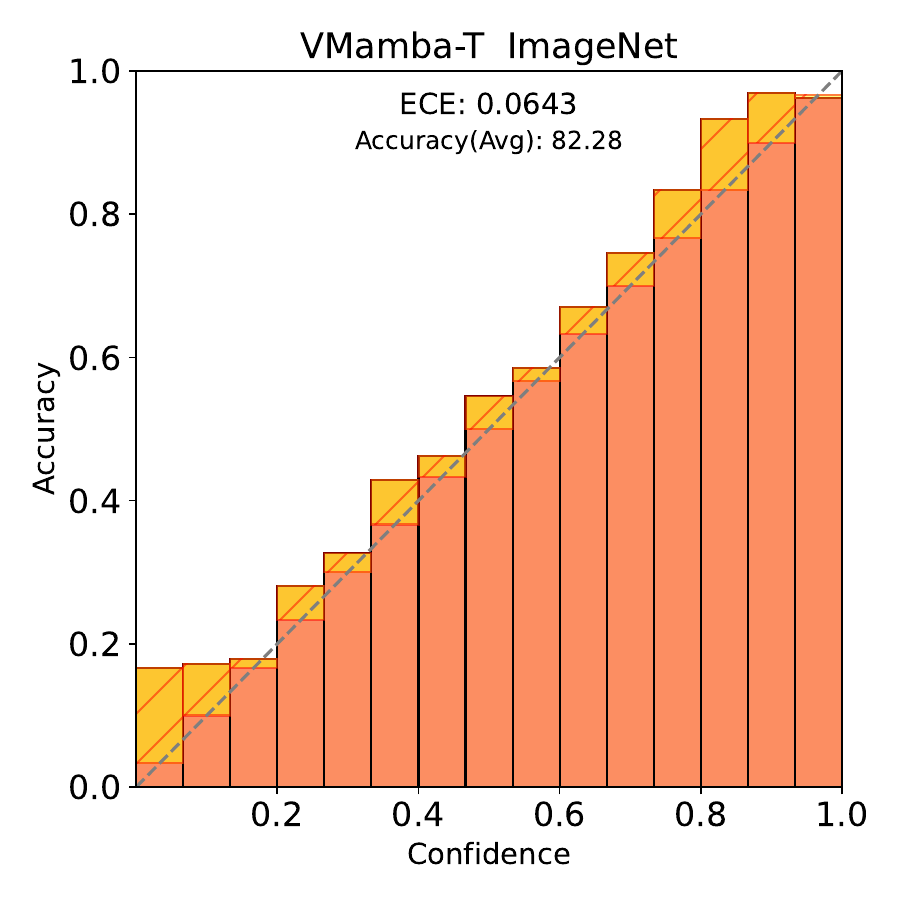}
\end{minipage}
\begin{minipage}{0.19\textwidth}
  \centering
  \includegraphics[height=3.8cm, width=\linewidth, keepaspectratio ]{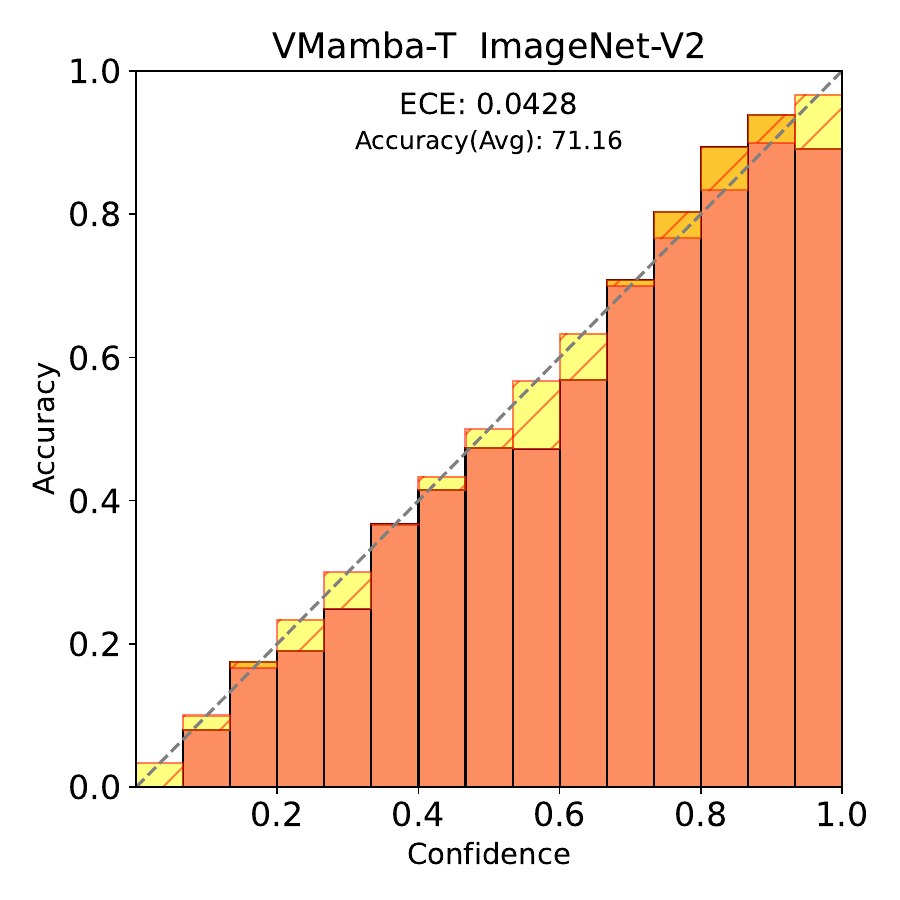}
\end{minipage}
\begin{minipage}{0.19\textwidth}
  \centering
  \includegraphics[height=3.8cm, width=\linewidth, keepaspectratio]{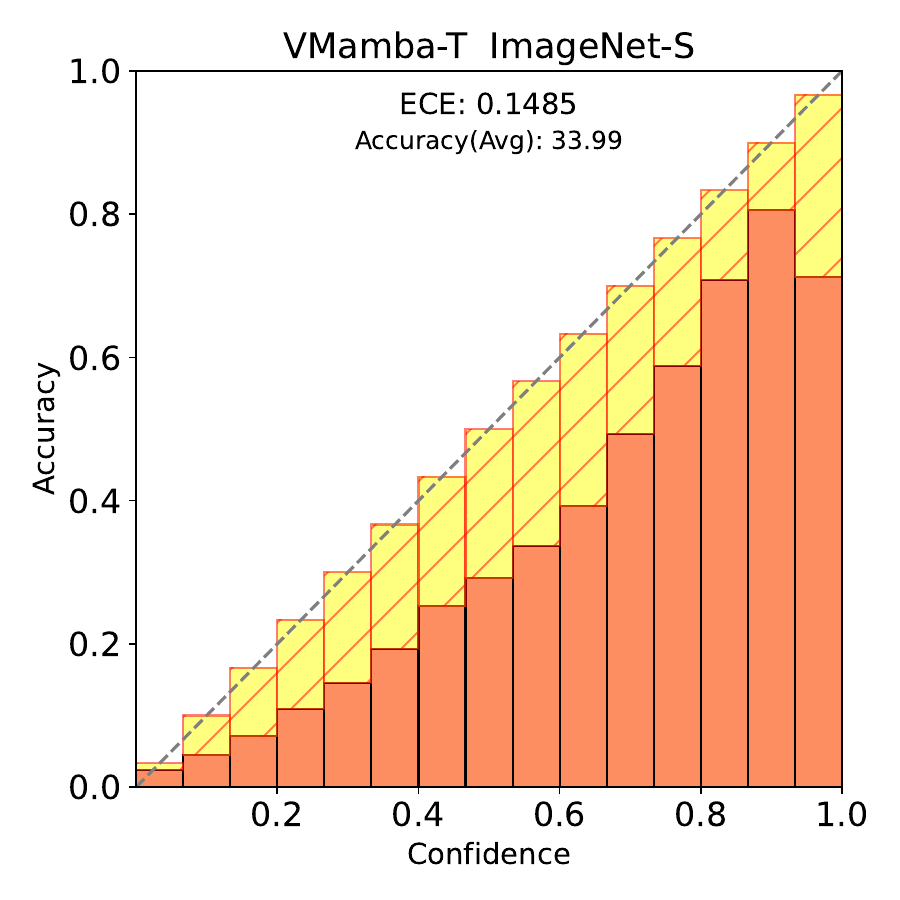}
\end{minipage}
\begin{minipage}{0.19\textwidth}
  \centering
  \includegraphics[height=3.8cm, width=\linewidth, keepaspectratio]{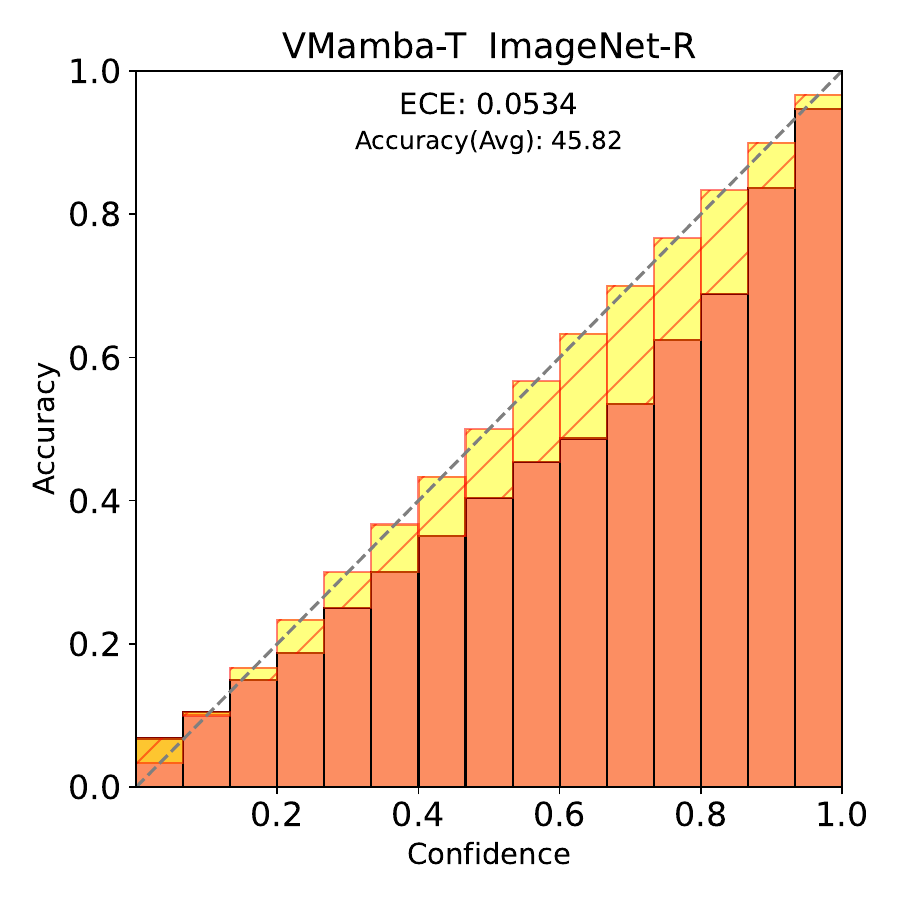}
\end{minipage}
\begin{minipage}{0.19\textwidth}
  \centering
  \includegraphics[height=3.8cm, width=\linewidth, keepaspectratio]{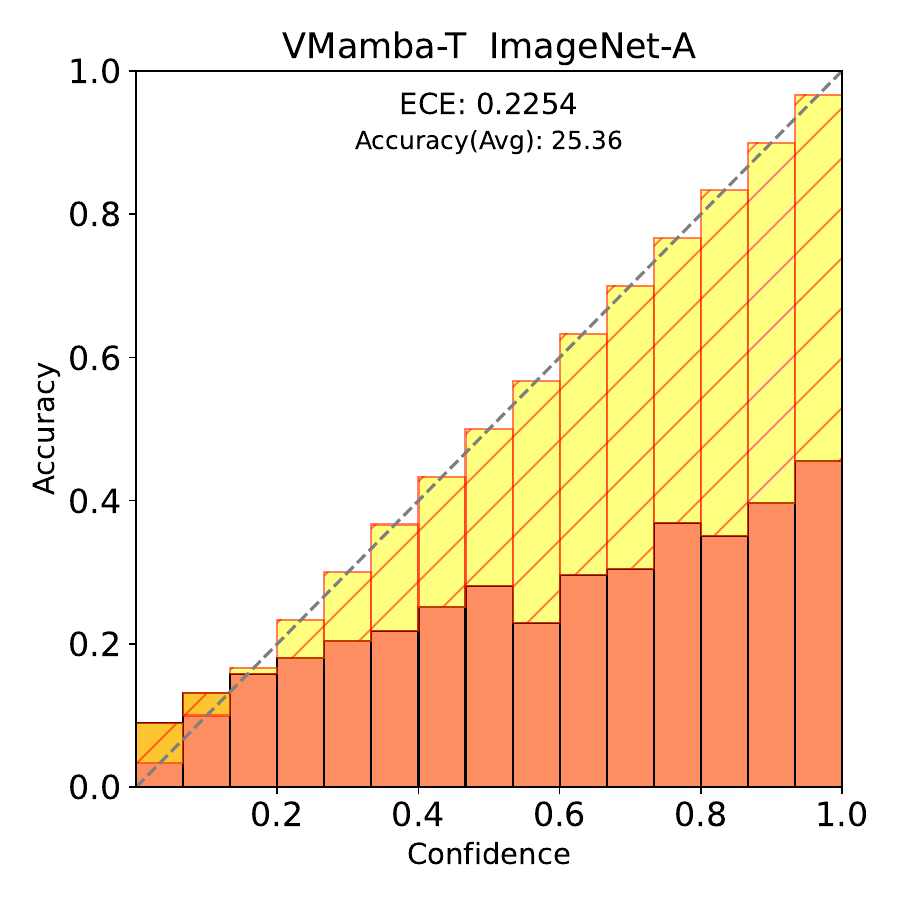}
\end{minipage}

\end{minipage}
\hfill
  \caption{Caliberation Results: Reliability diagrams and ECE on ConvNext-T, ViT-T, Swin-T, VMamba-T, and MambaVision-T across ImageNet, ImageNet-V2, ImageNet-S, ImageNet-R, and ImageNet-A.}
  \label{fig:calib_app_tiny}
\vspace{-1em}
\end{figure*}

\begin{figure*}[h]
\begin{minipage}{\textwidth}

\centering


\begin{minipage}{0.19\textwidth}
  \centering
  \includegraphics[height=3.8cm, width=\linewidth , keepaspectratio]{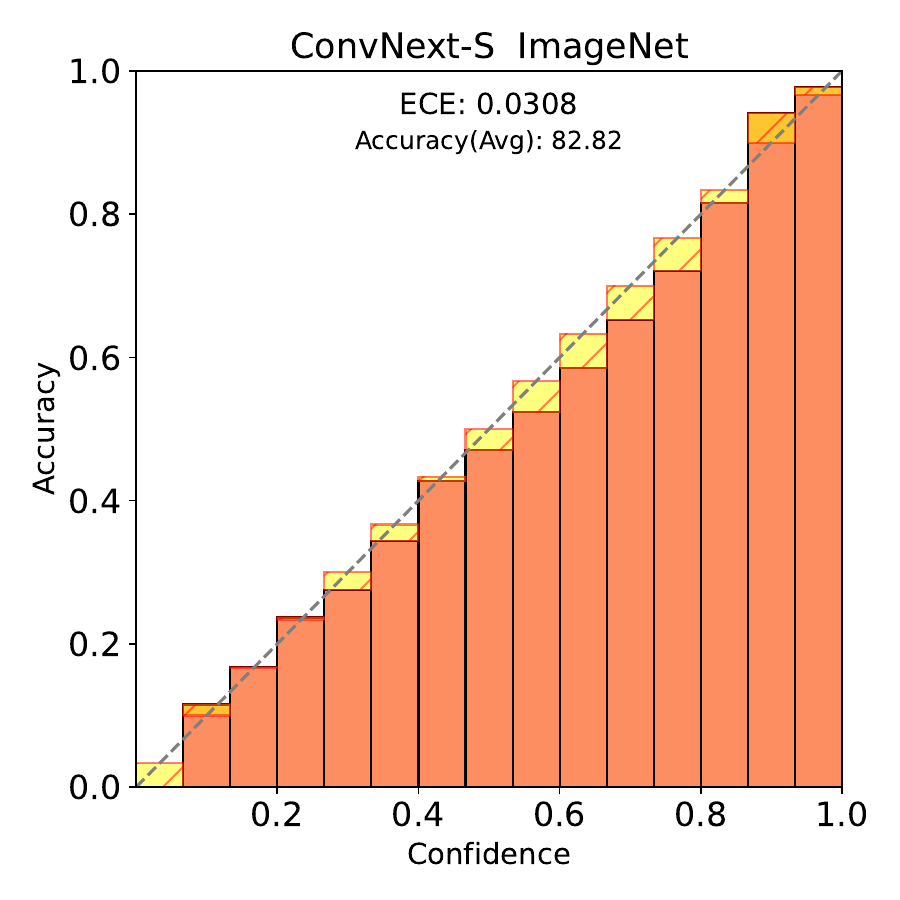}
\end{minipage}
\begin{minipage}{0.19\textwidth}
  \centering
  \includegraphics[height=3.8cm, width=\linewidth, keepaspectratio ]{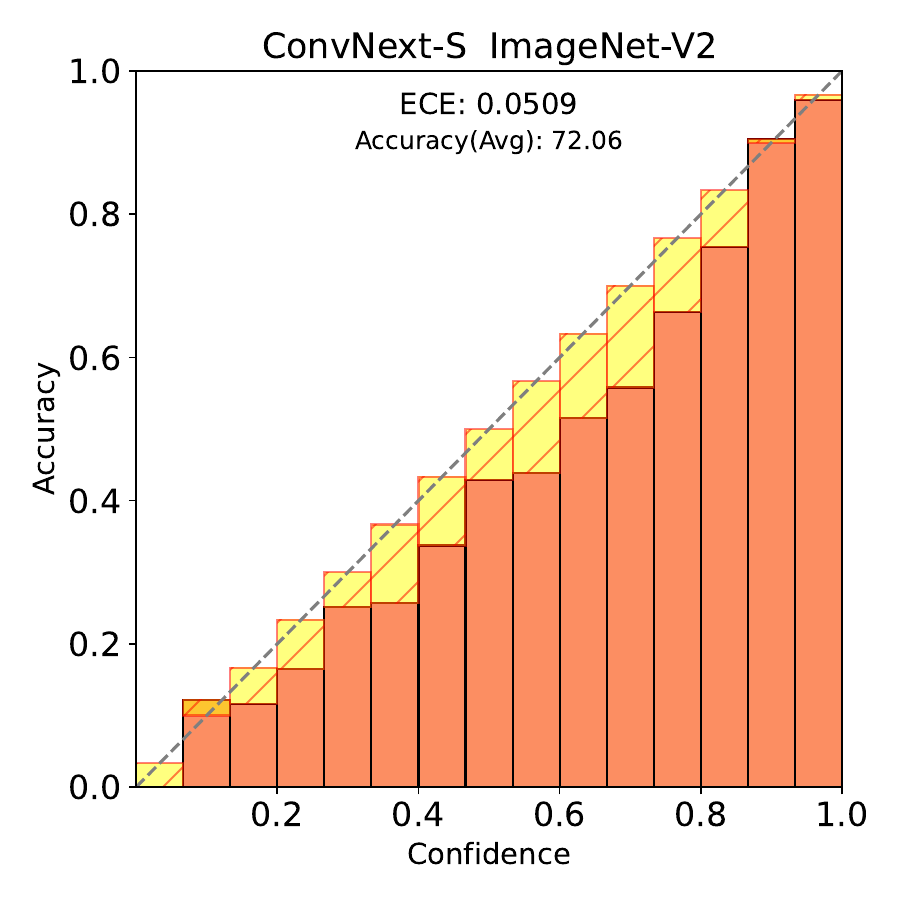}
\end{minipage}
\begin{minipage}{0.19\textwidth}
  \centering
  \includegraphics[height=3.8cm, width=\linewidth, keepaspectratio]{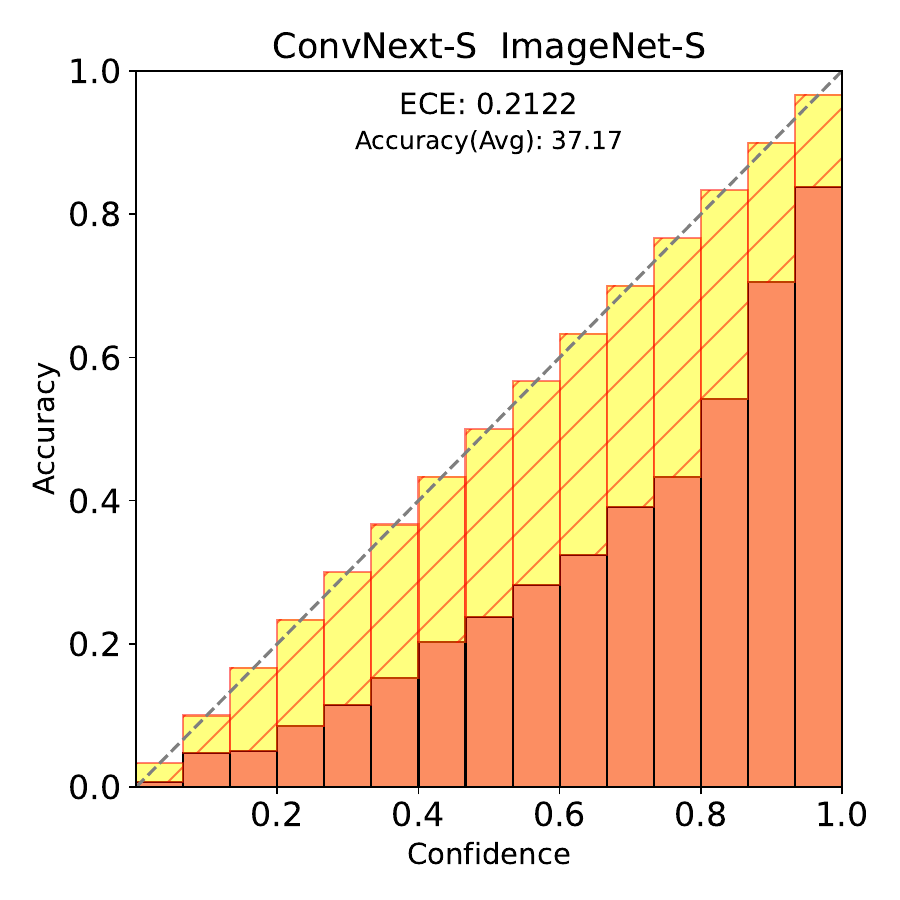}
\end{minipage}
\begin{minipage}{0.19\textwidth}
  \centering
  \includegraphics[height=3.8cm, width=\linewidth, keepaspectratio]{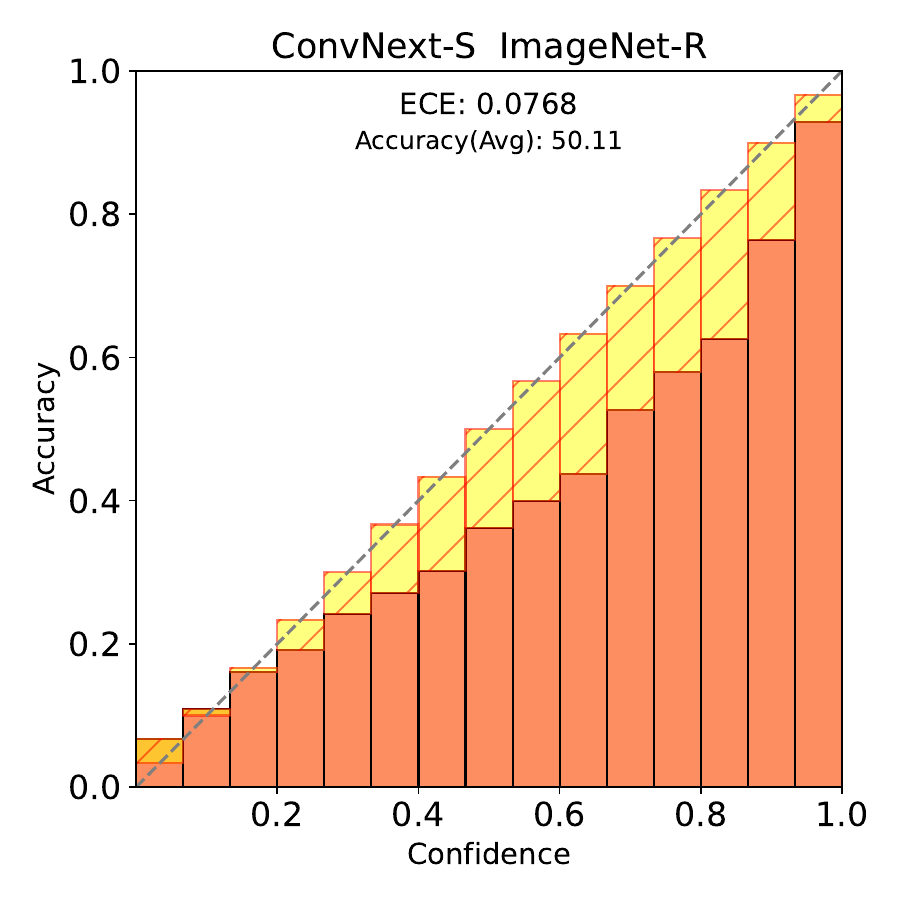}
\end{minipage}
\begin{minipage}{0.19\textwidth}
  \centering
  \includegraphics[height=3.8cm, width=\linewidth, keepaspectratio]{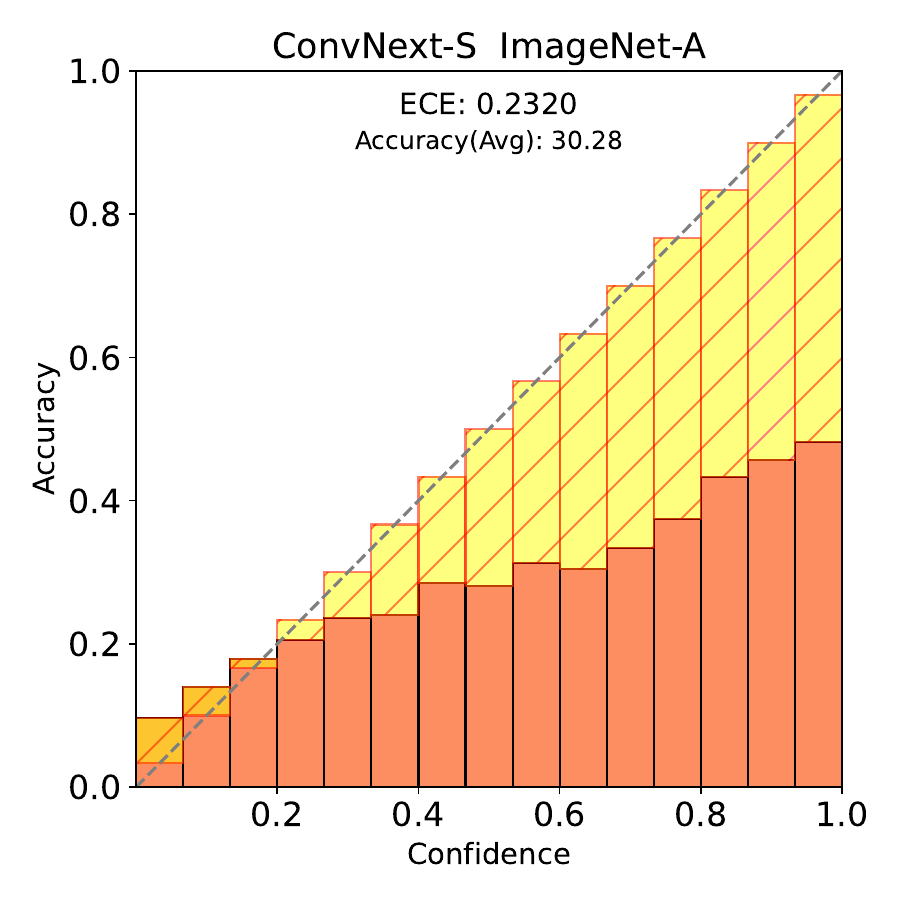}
\end{minipage}

\begin{minipage}{0.19\textwidth}
  \centering
  \includegraphics[height=3.8cm, width=\linewidth , keepaspectratio]{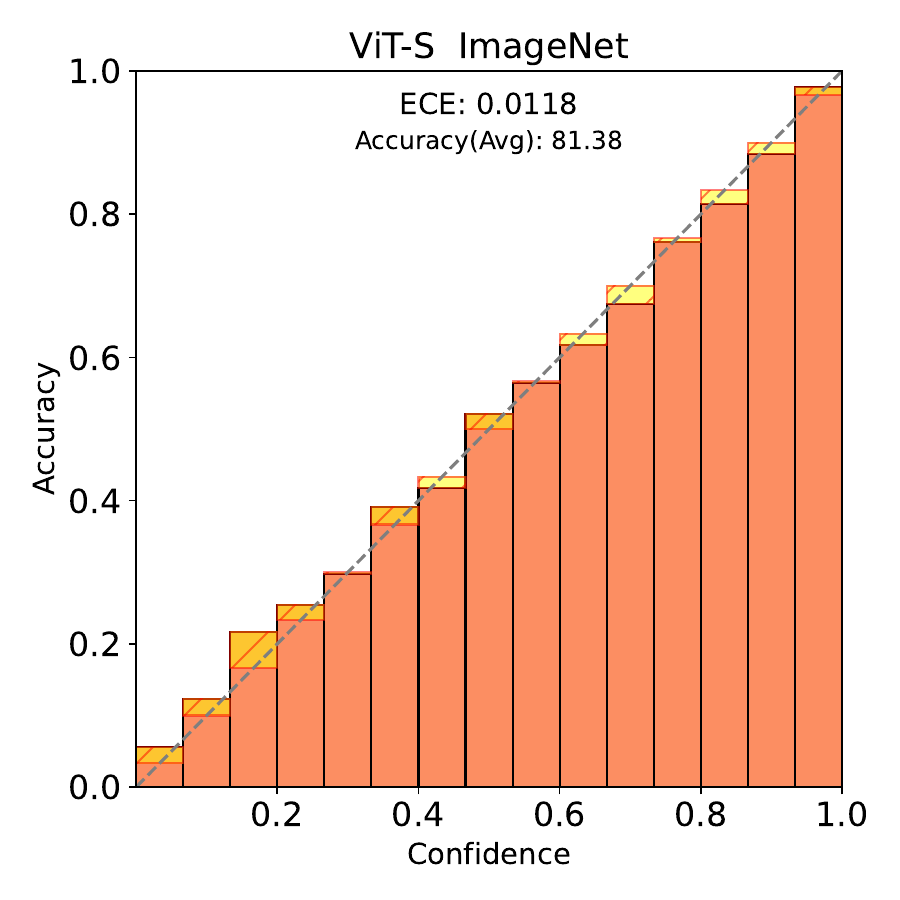}
\end{minipage}
\begin{minipage}{0.19\textwidth}
  \centering
  \includegraphics[height=3.8cm, width=\linewidth, keepaspectratio ]{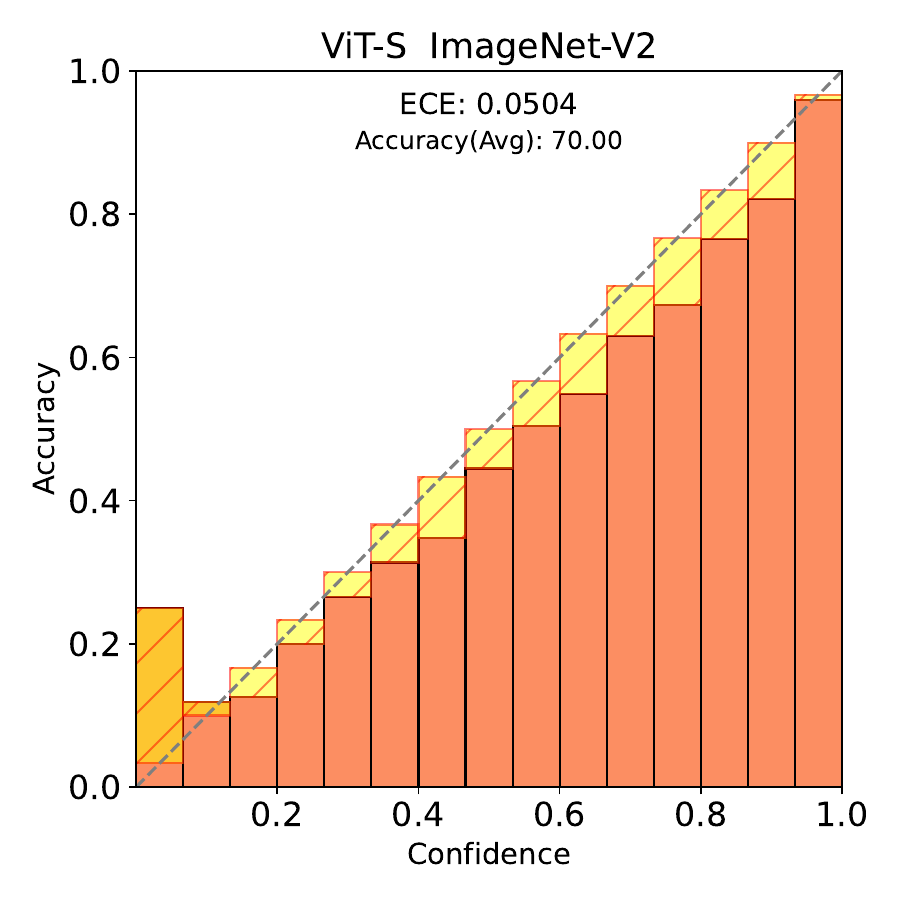}
\end{minipage}
\begin{minipage}{0.19\textwidth}
  \centering
  \includegraphics[height=3.8cm, width=\linewidth, keepaspectratio]{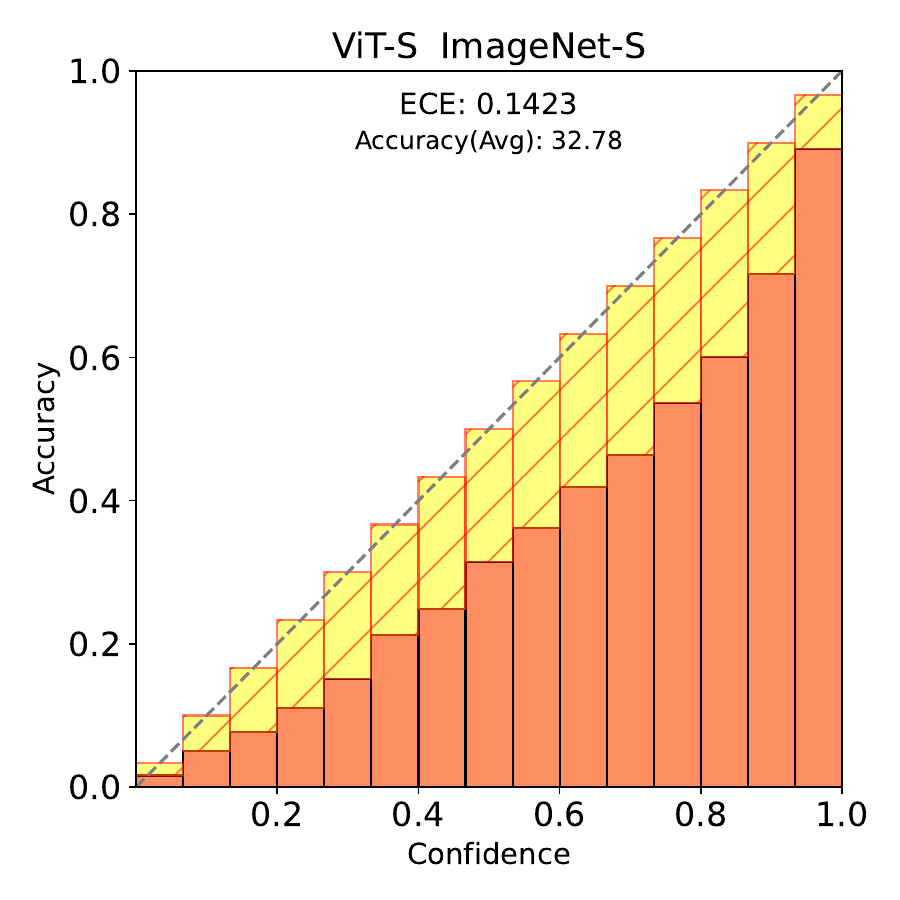}
\end{minipage}
\begin{minipage}{0.19\textwidth}
  \centering
  \includegraphics[height=3.8cm, width=\linewidth, keepaspectratio]{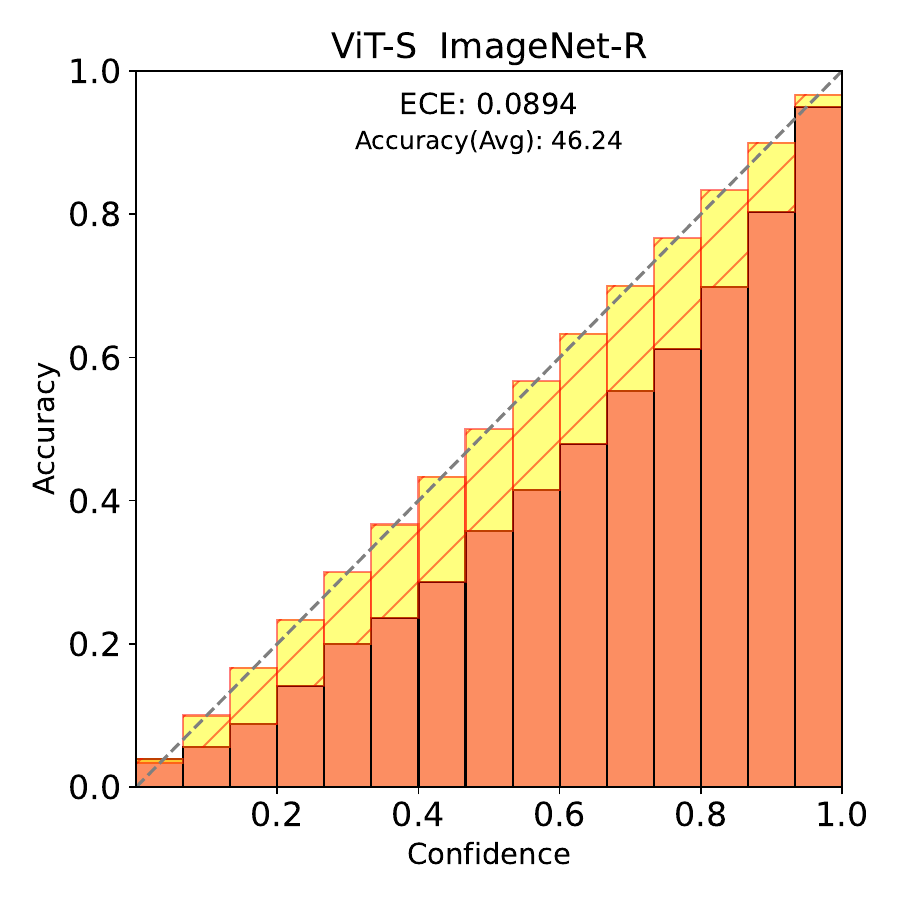}
\end{minipage}
\begin{minipage}{0.19\textwidth}
  \centering
  \includegraphics[height=3.8cm, width=\linewidth, keepaspectratio]{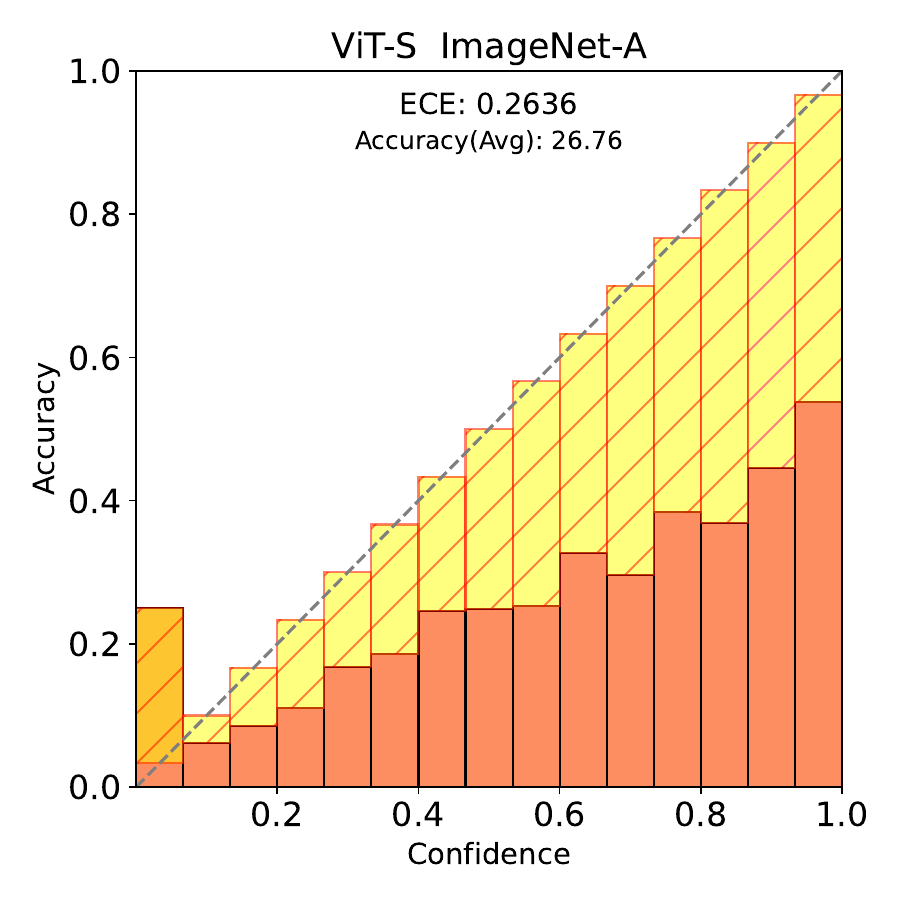}
\end{minipage}


\begin{minipage}{0.19\textwidth}
  \centering
  \includegraphics[height=3.8cm, width=\linewidth , keepaspectratio]{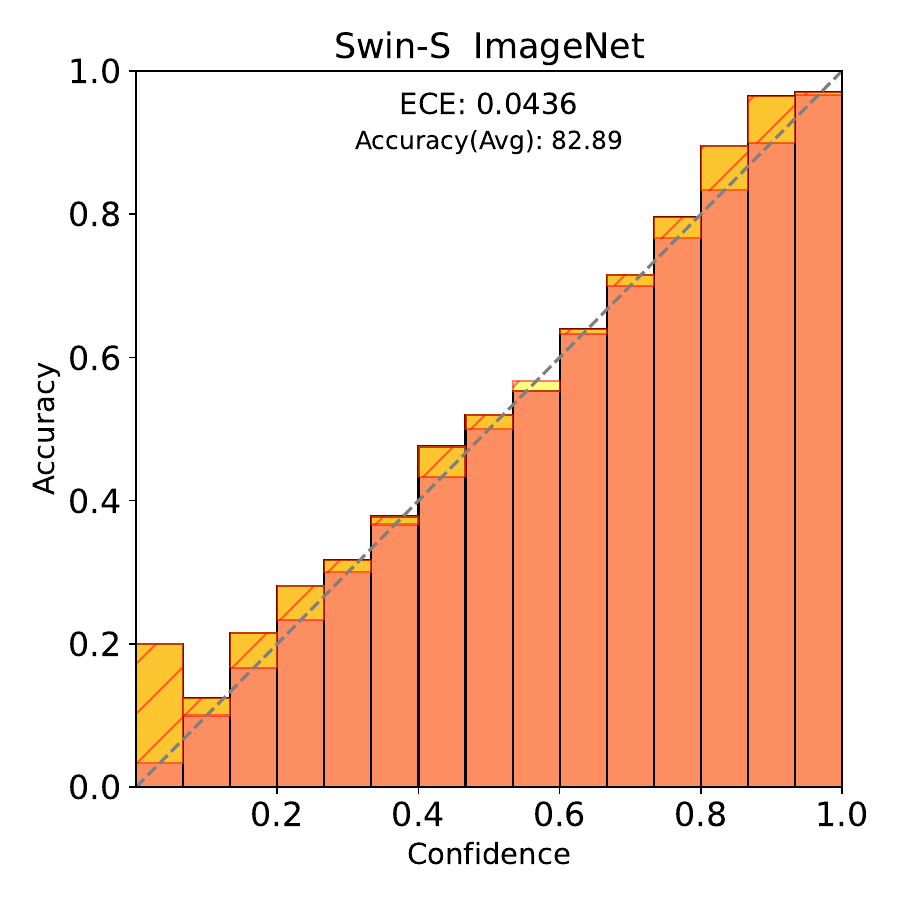}
\end{minipage}
\begin{minipage}{0.19\textwidth}
  \centering
  \includegraphics[height=3.8cm, width=\linewidth, keepaspectratio ]{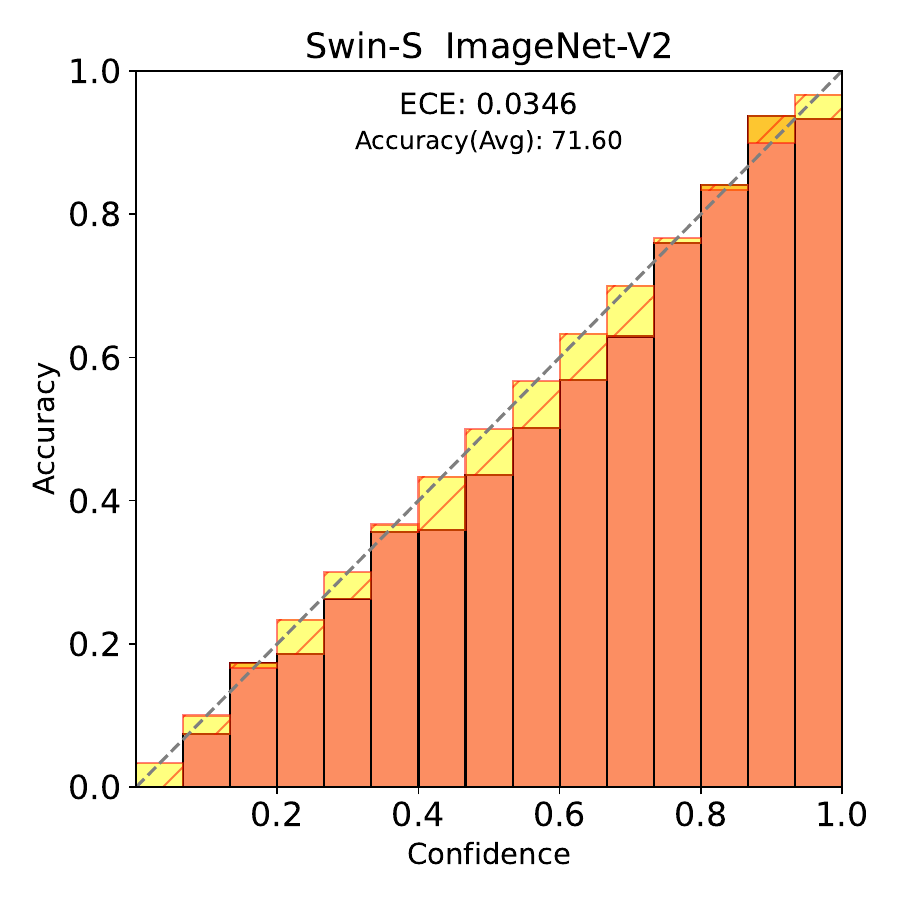}
\end{minipage}
\begin{minipage}{0.19\textwidth}
  \centering
  \includegraphics[height=3.8cm, width=\linewidth, keepaspectratio]{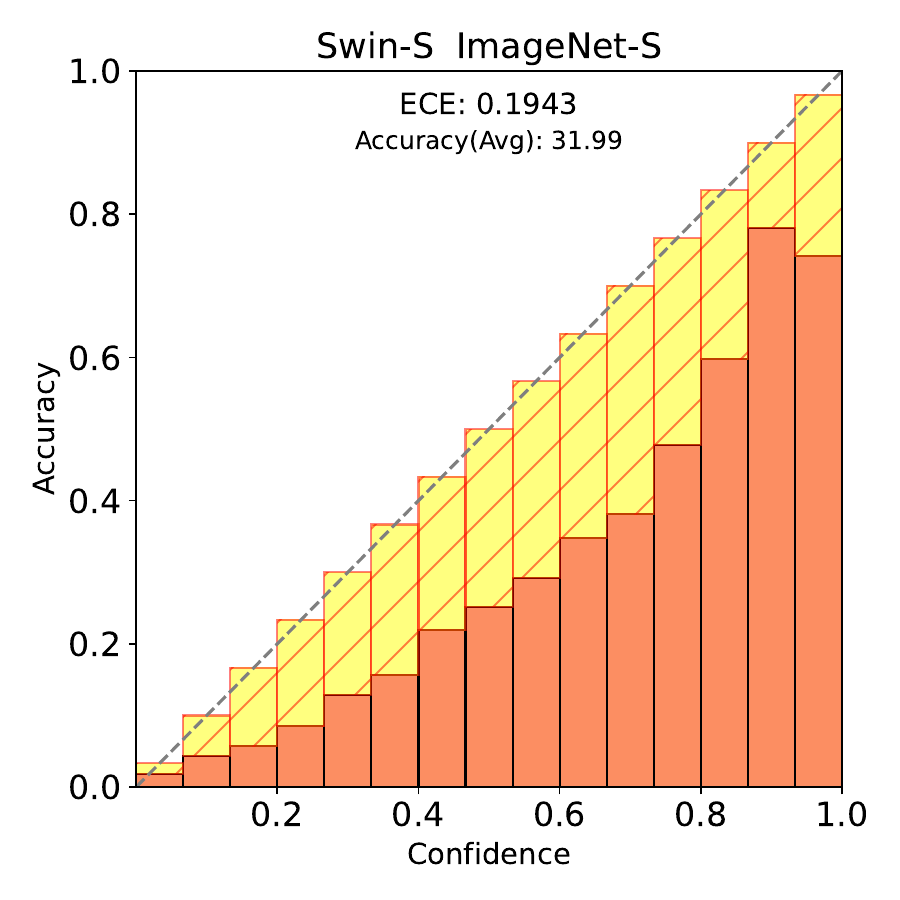}
\end{minipage}
\begin{minipage}{0.19\textwidth}
  \centering
  \includegraphics[height=3.8cm, width=\linewidth, keepaspectratio]{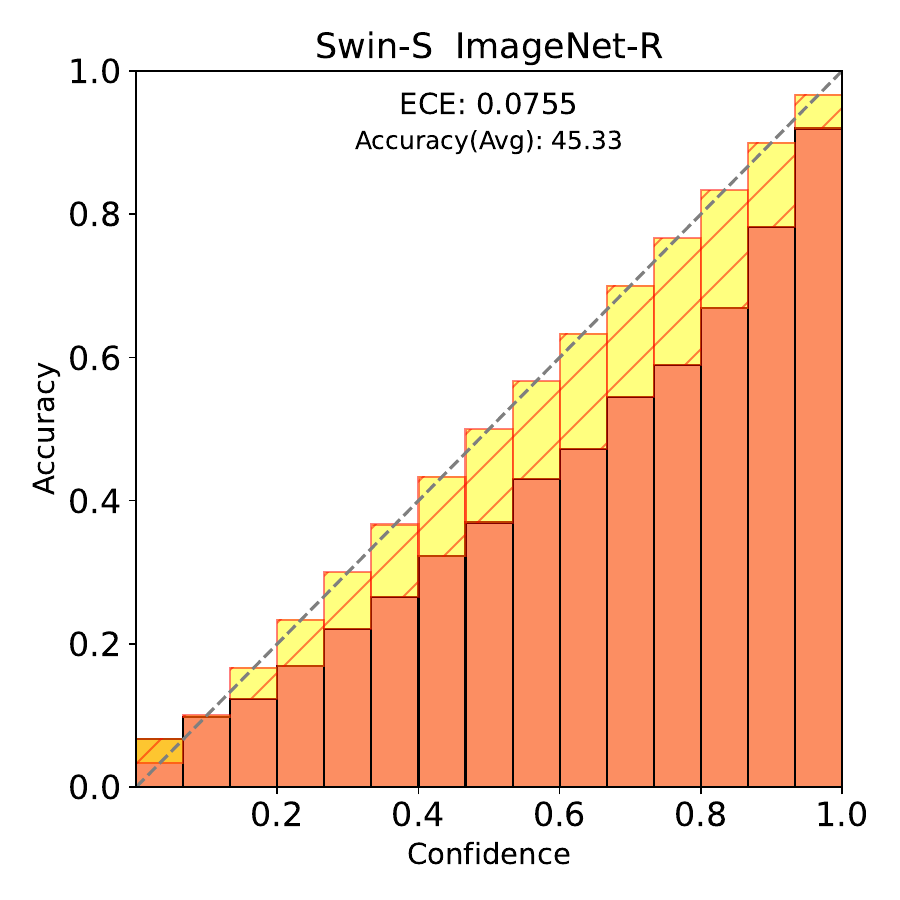}
\end{minipage}
\begin{minipage}{0.19\textwidth}
  \centering
  \includegraphics[height=3.8cm, width=\linewidth, keepaspectratio]{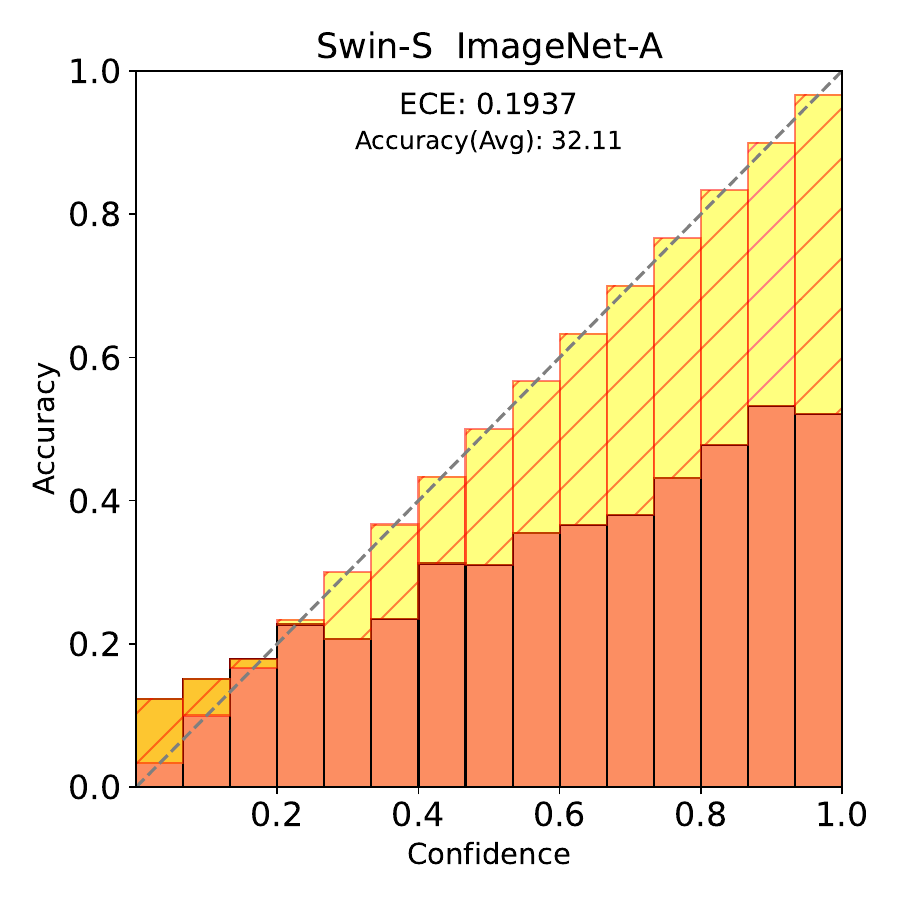}
\end{minipage}


\begin{minipage}{0.19\textwidth}
  \centering
  \includegraphics[height=3.8cm, width=\linewidth , keepaspectratio]{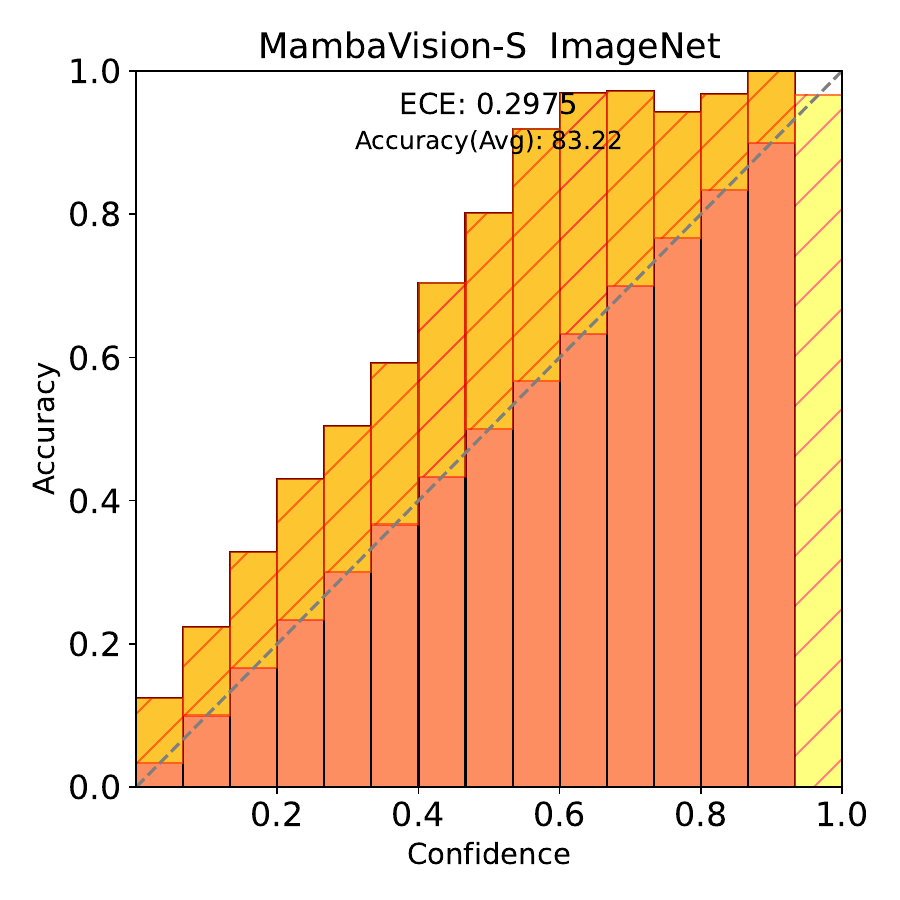}
\end{minipage}
\begin{minipage}{0.19\textwidth}
  \centering
  \includegraphics[height=3.8cm, width=\linewidth, keepaspectratio ]{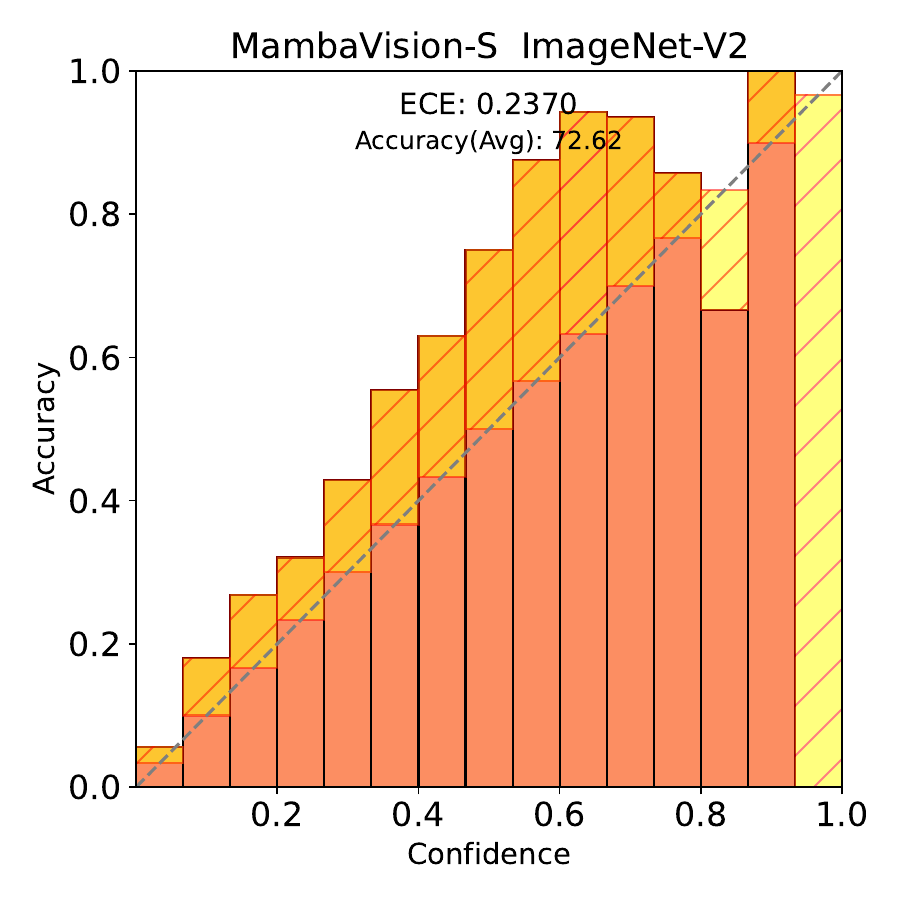}
\end{minipage}
\begin{minipage}{0.19\textwidth}
  \centering
  \includegraphics[height=3.8cm, width=\linewidth, keepaspectratio]{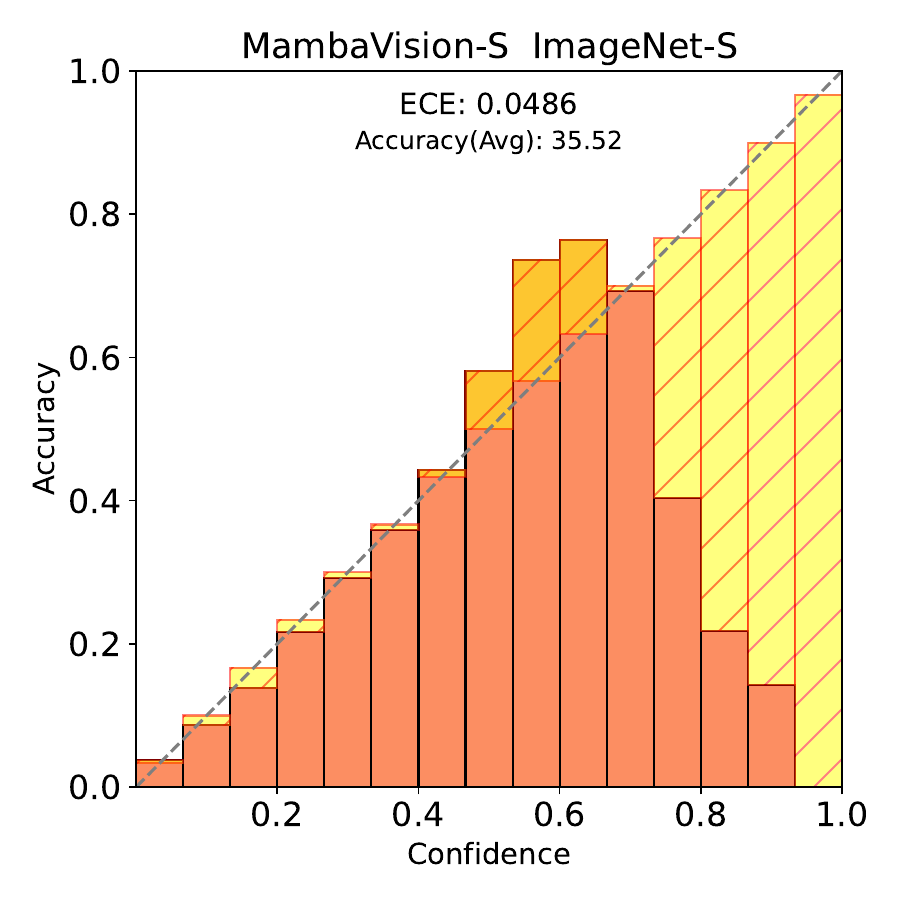}
\end{minipage}
\begin{minipage}{0.19\textwidth}
  \centering
  \includegraphics[height=3.8cm, width=\linewidth, keepaspectratio]{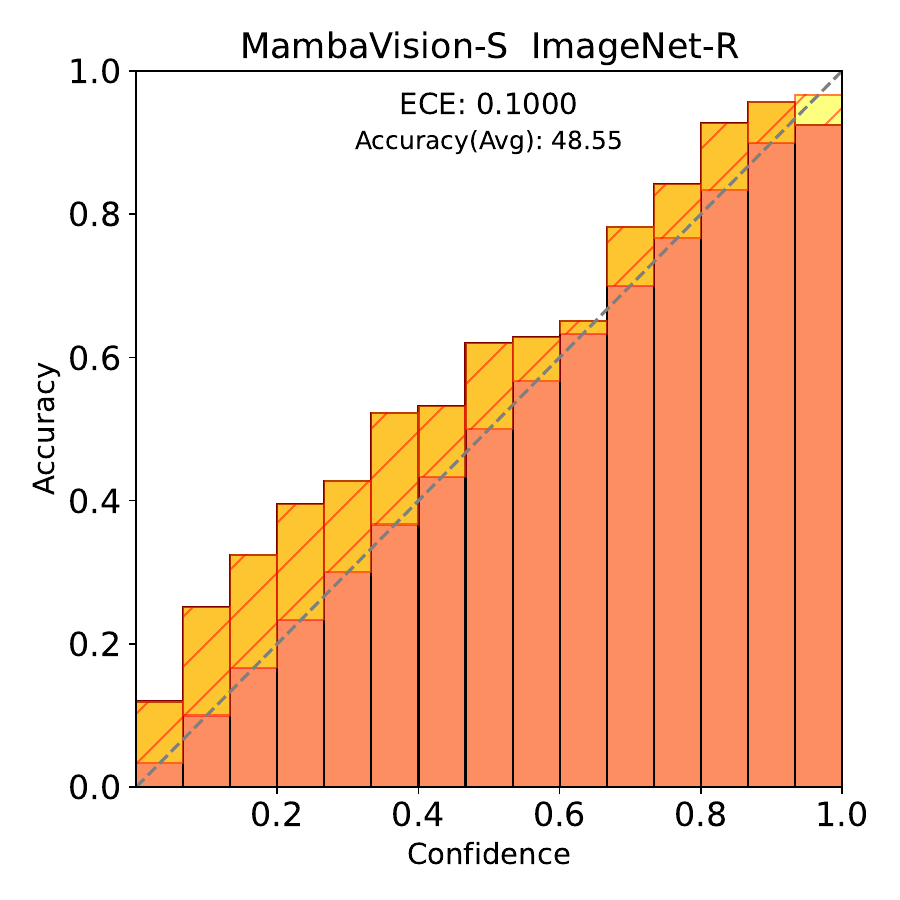}
\end{minipage}
\begin{minipage}{0.19\textwidth}
  \centering
  \includegraphics[height=3.8cm, width=\linewidth, keepaspectratio]{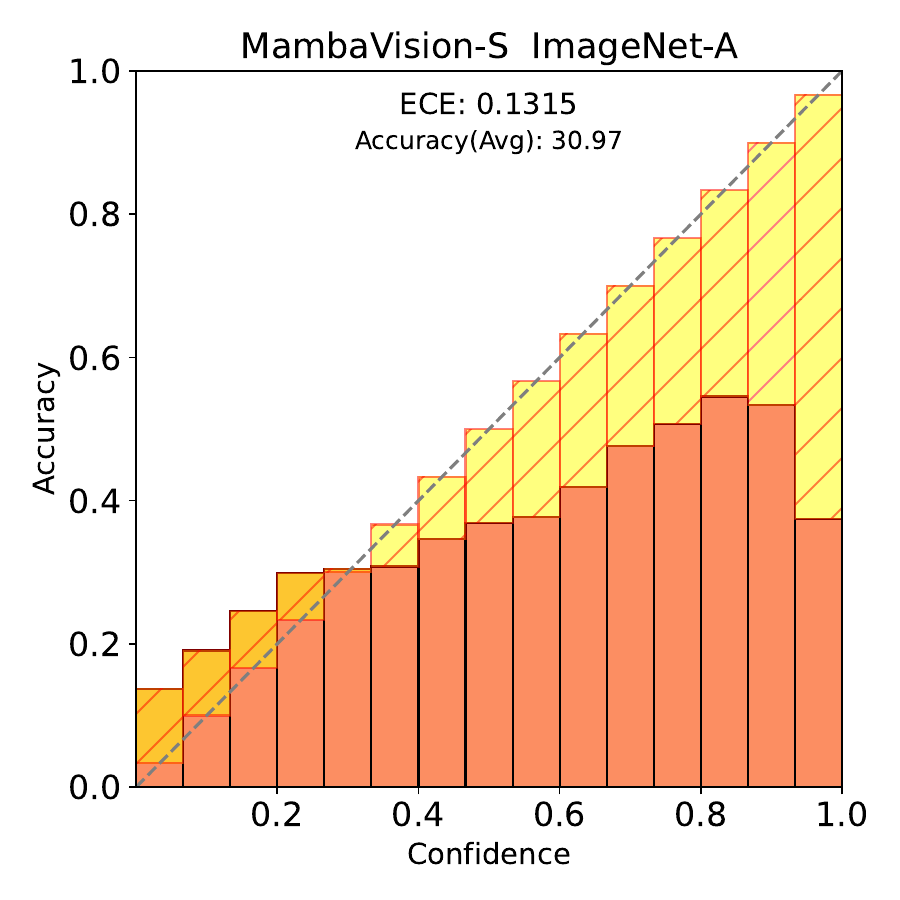}
\end{minipage}


\begin{minipage}{0.19\textwidth}
  \centering
  \includegraphics[height=3.8cm, width=\linewidth , keepaspectratio]{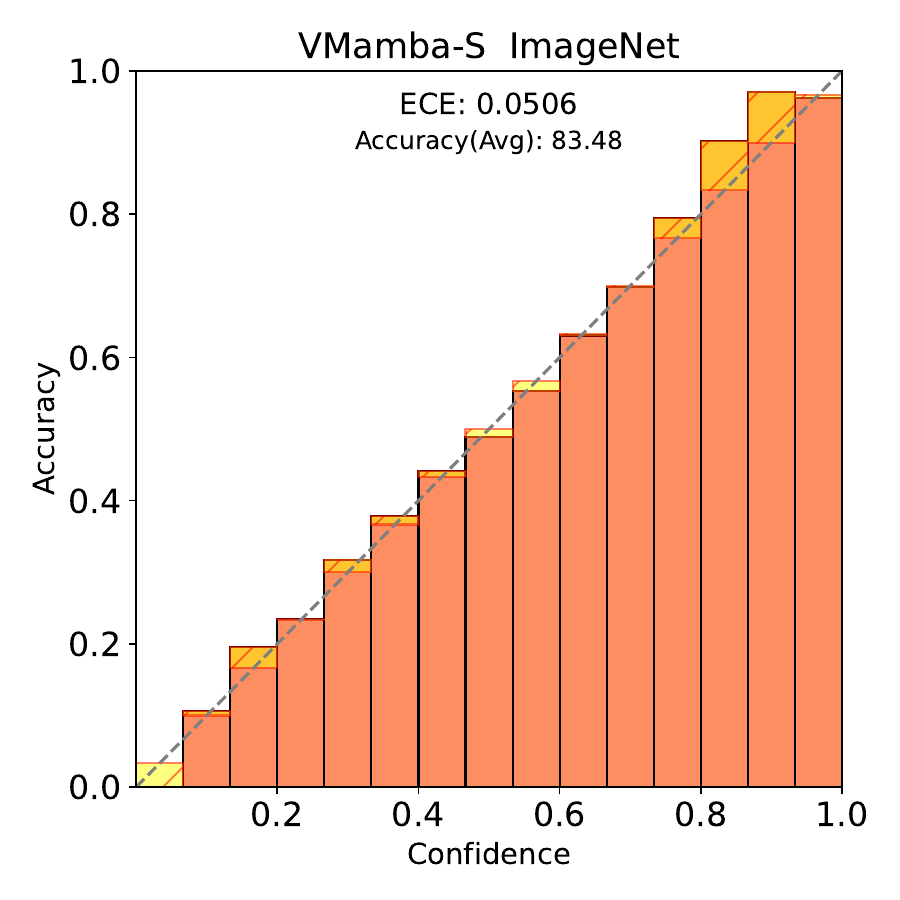}
\end{minipage}
\begin{minipage}{0.19\textwidth}
  \centering
  \includegraphics[height=3.8cm, width=\linewidth, keepaspectratio ]{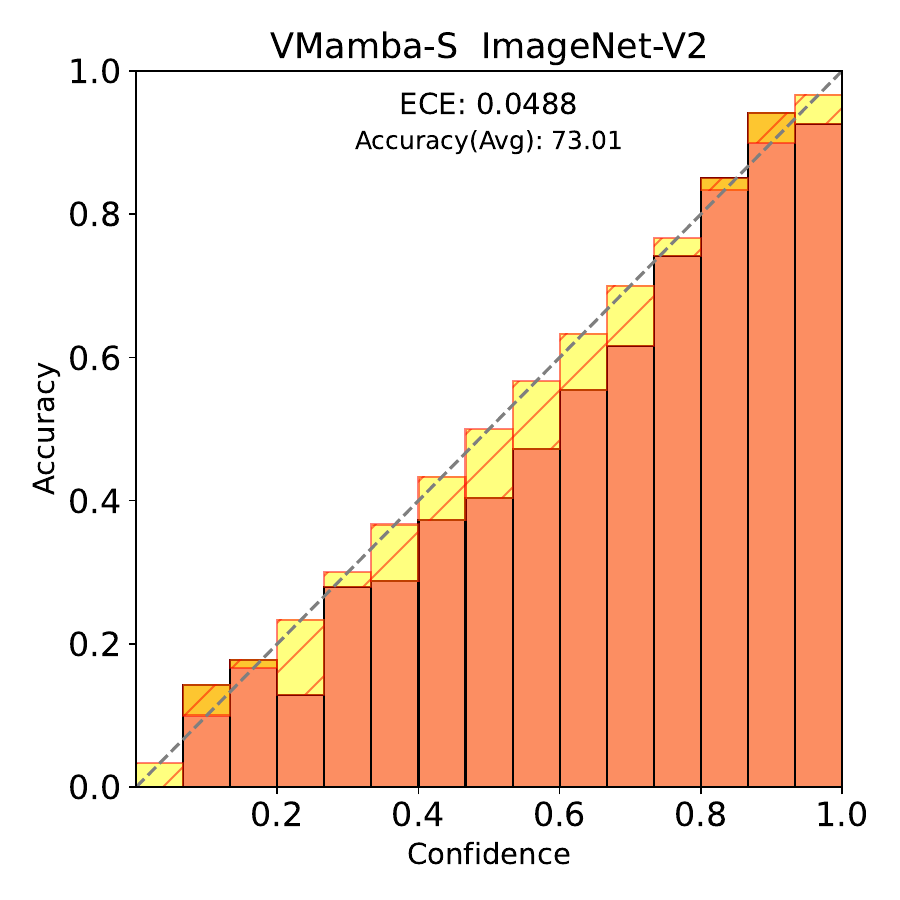}
\end{minipage}
\begin{minipage}{0.19\textwidth}
  \centering
  \includegraphics[height=3.8cm, width=\linewidth, keepaspectratio]{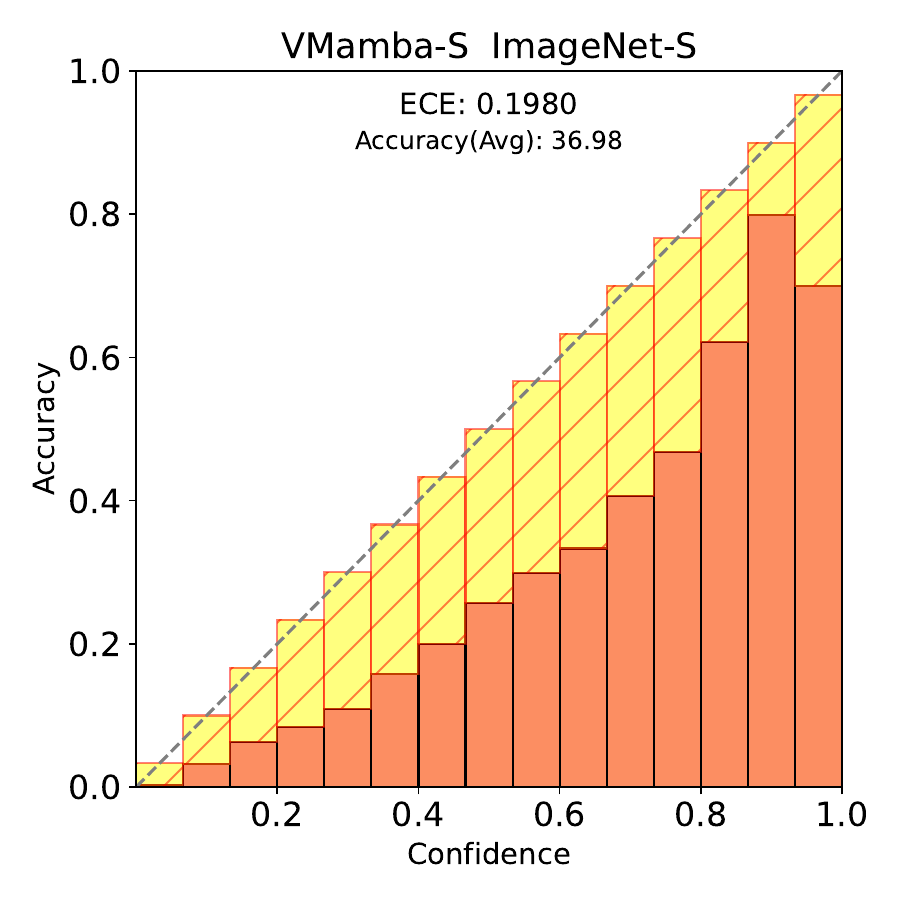}
\end{minipage}
\begin{minipage}{0.19\textwidth}
  \centering
  \includegraphics[height=3.8cm, width=\linewidth, keepaspectratio]{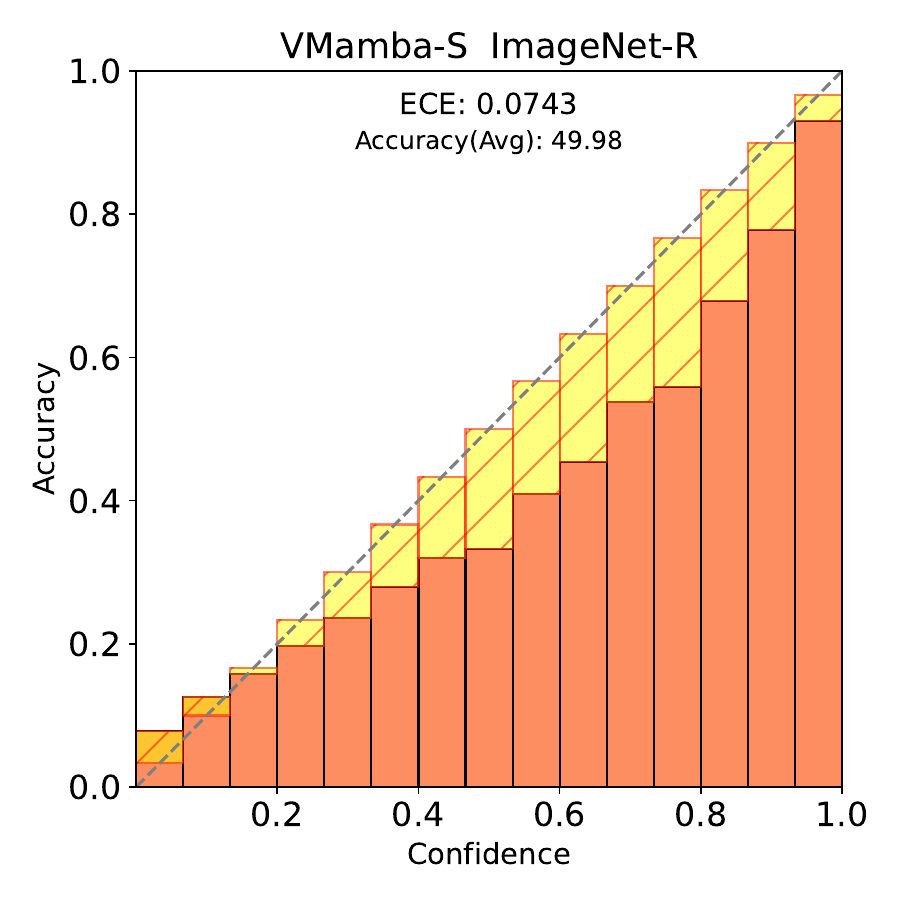}
\end{minipage}
\begin{minipage}{0.19\textwidth}
  \centering
  \includegraphics[height=3.8cm, width=\linewidth, keepaspectratio]{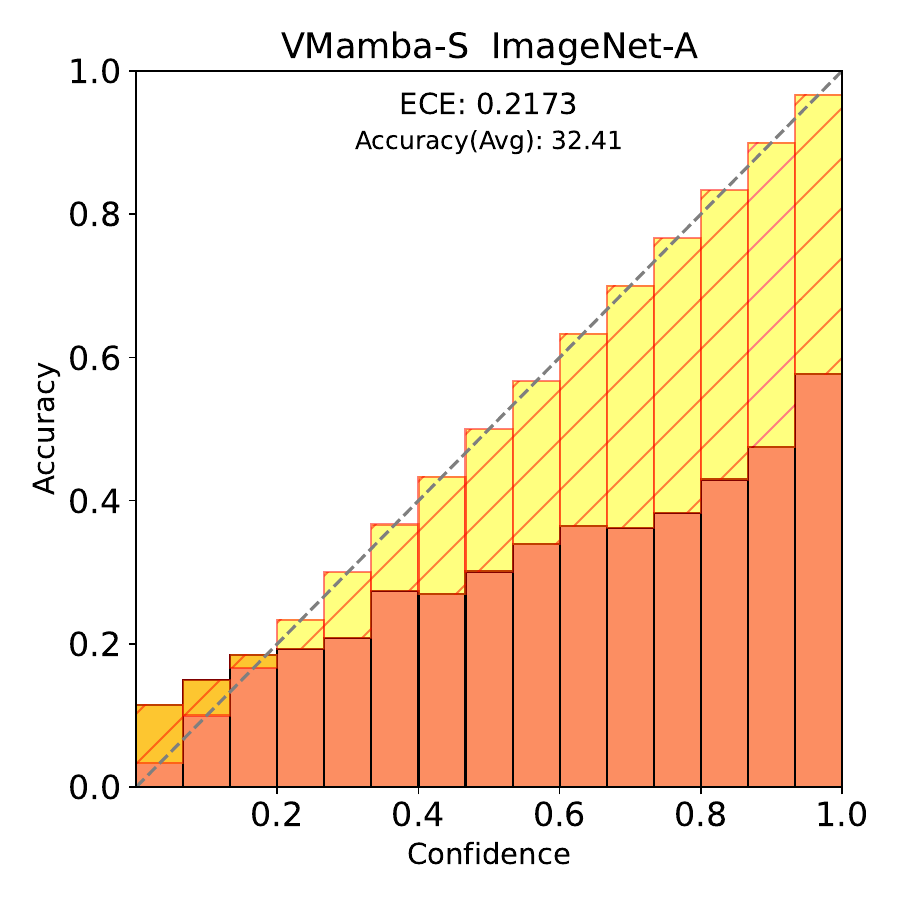}
\end{minipage}

\end{minipage}
\hfill
  \caption{Caliberation Results: Reliability diagrams and ECE on ConvNext-S, ViT-S, Swin-S, VMamba-S, and MambaVision-S across ImageNet, ImageNet-V2, ImageNet-S, ImageNet-R, and ImageNet-A.}
  \label{fig:calib_app_small}
\vspace{-1em}
\end{figure*}

\begin{figure*}[h]
\begin{minipage}{\textwidth}

\centering


\begin{minipage}{0.19\textwidth}
  \centering
  \includegraphics[height=3.8cm, width=\linewidth , keepaspectratio]{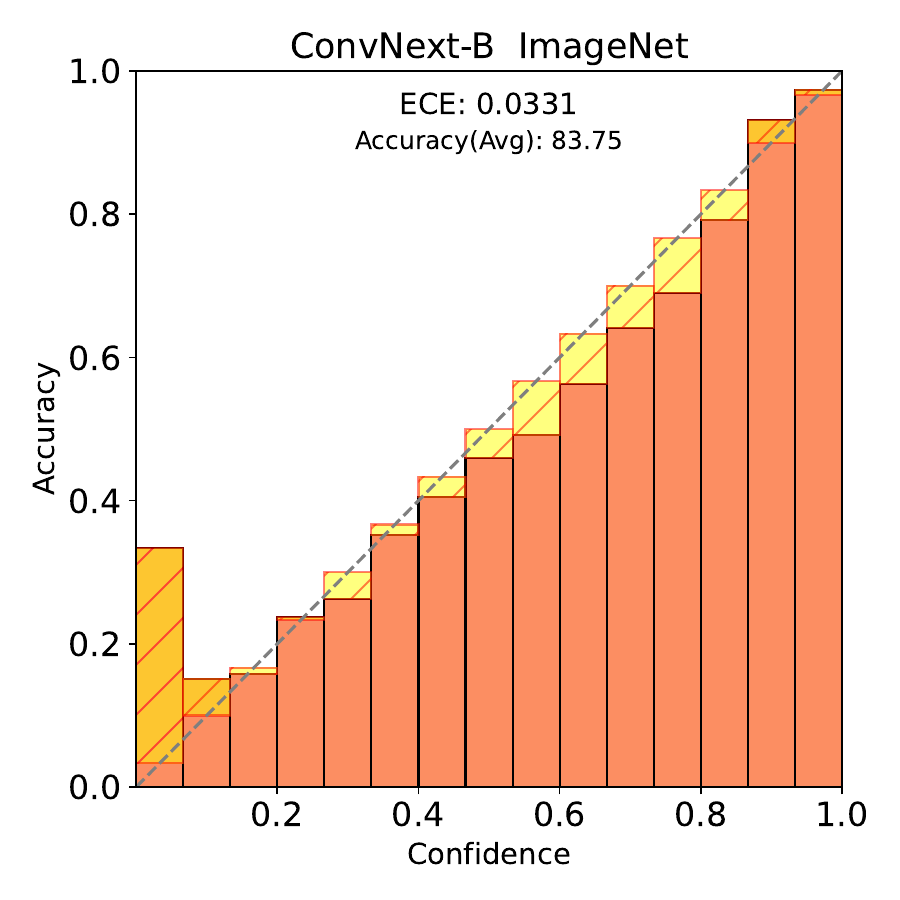}
\end{minipage}
\begin{minipage}{0.19\textwidth}
  \centering
  \includegraphics[height=3.8cm, width=\linewidth, keepaspectratio ]{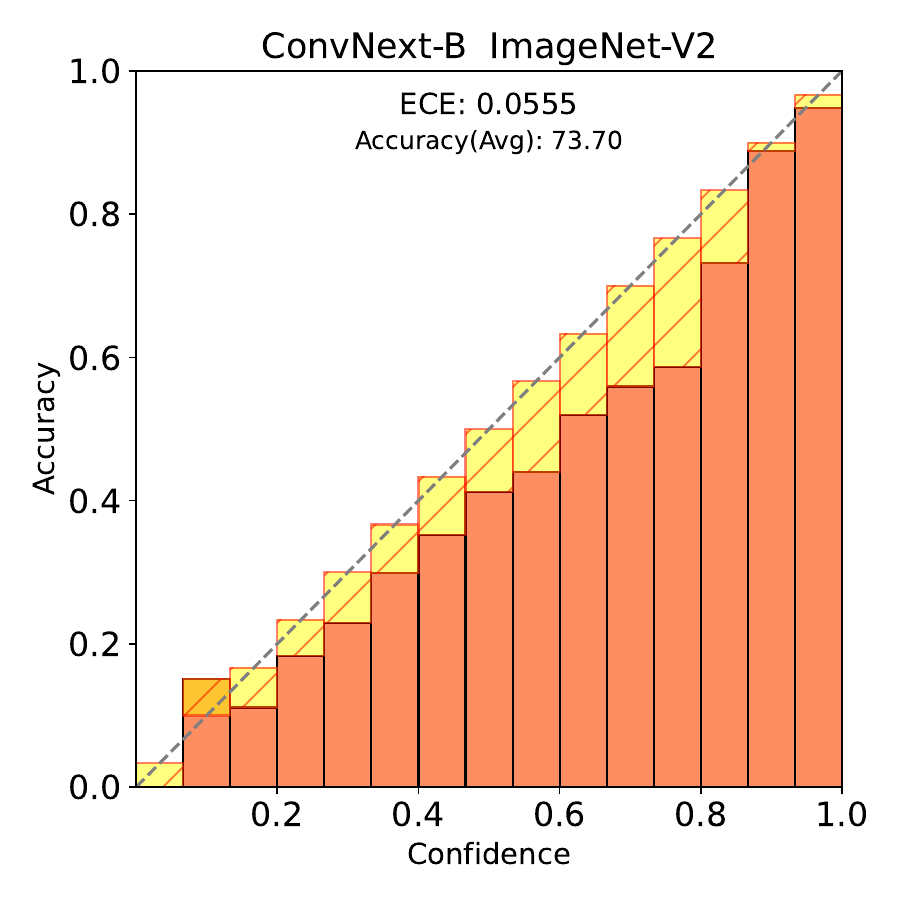}
\end{minipage}
\begin{minipage}{0.19\textwidth}
  \centering
  \includegraphics[height=3.8cm, width=\linewidth, keepaspectratio]{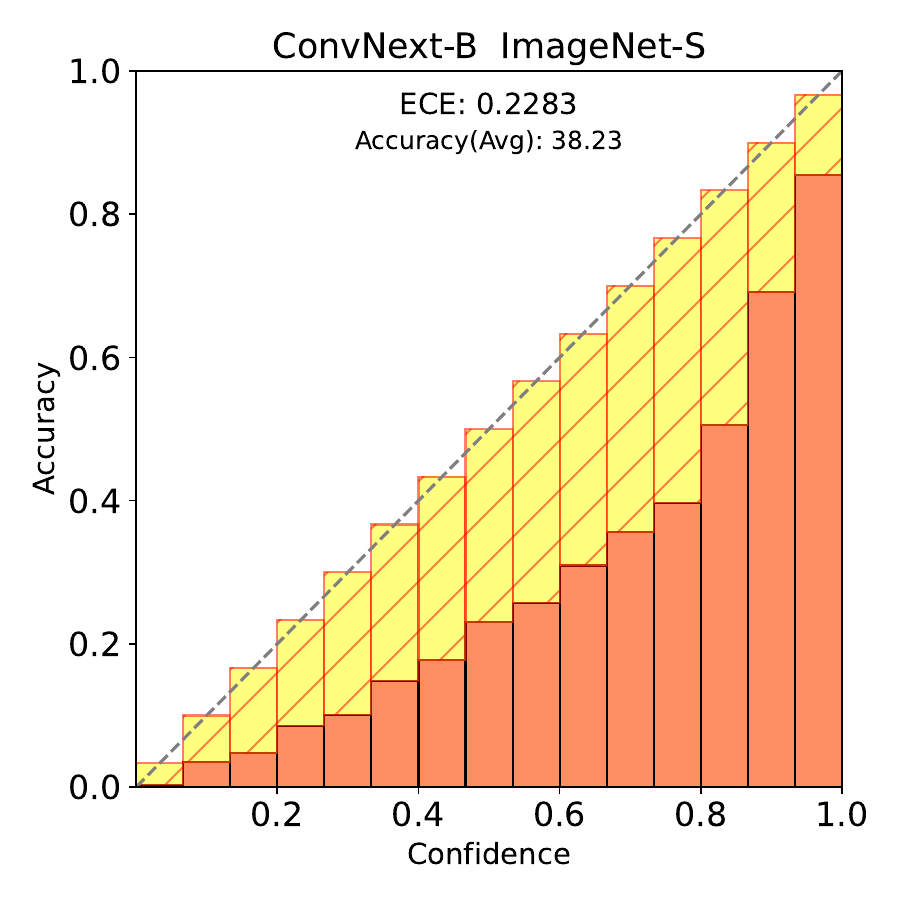}
\end{minipage}
\begin{minipage}{0.19\textwidth}
  \centering
  \includegraphics[height=3.8cm, width=\linewidth, keepaspectratio]{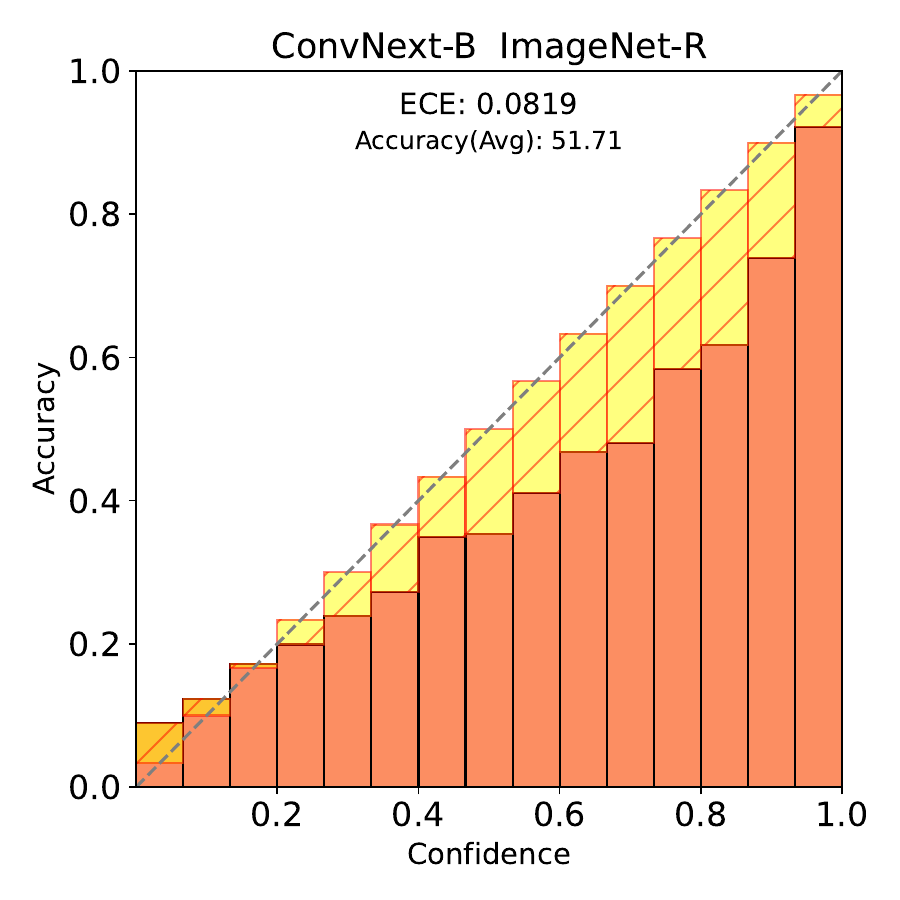}
\end{minipage}
\begin{minipage}{0.19\textwidth}
  \centering
  \includegraphics[height=3.8cm, width=\linewidth, keepaspectratio]{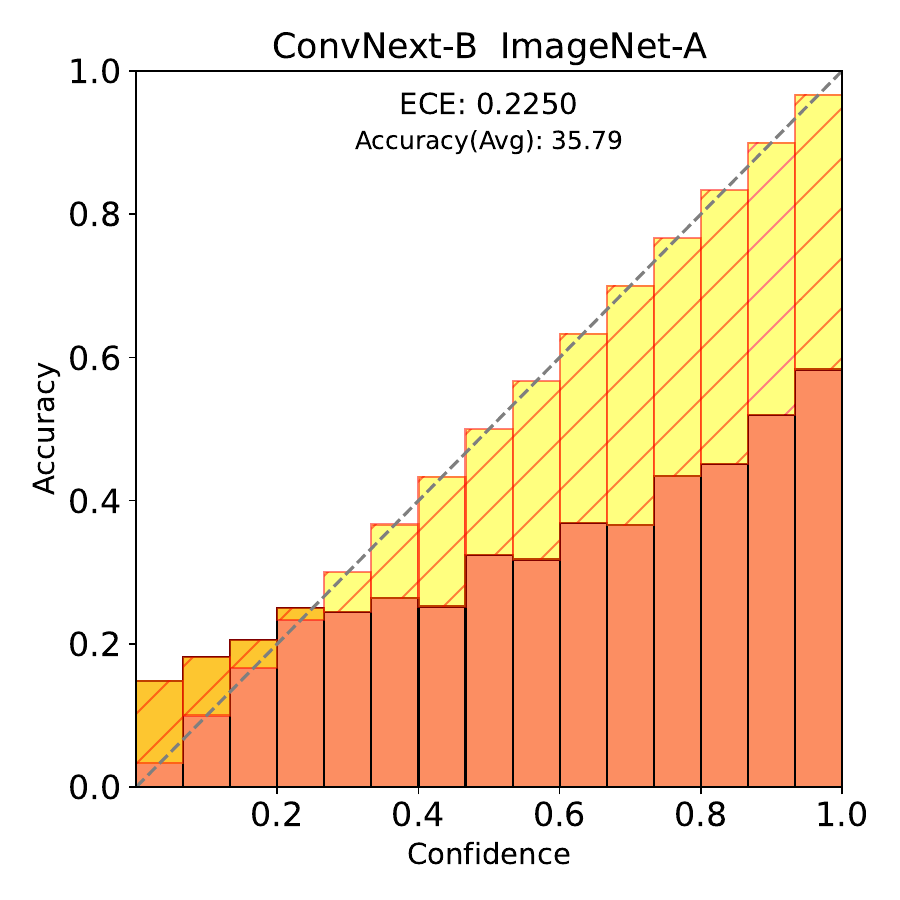}
\end{minipage}

\begin{minipage}{0.19\textwidth}
  \centering
  \includegraphics[height=3.8cm, width=\linewidth , keepaspectratio]{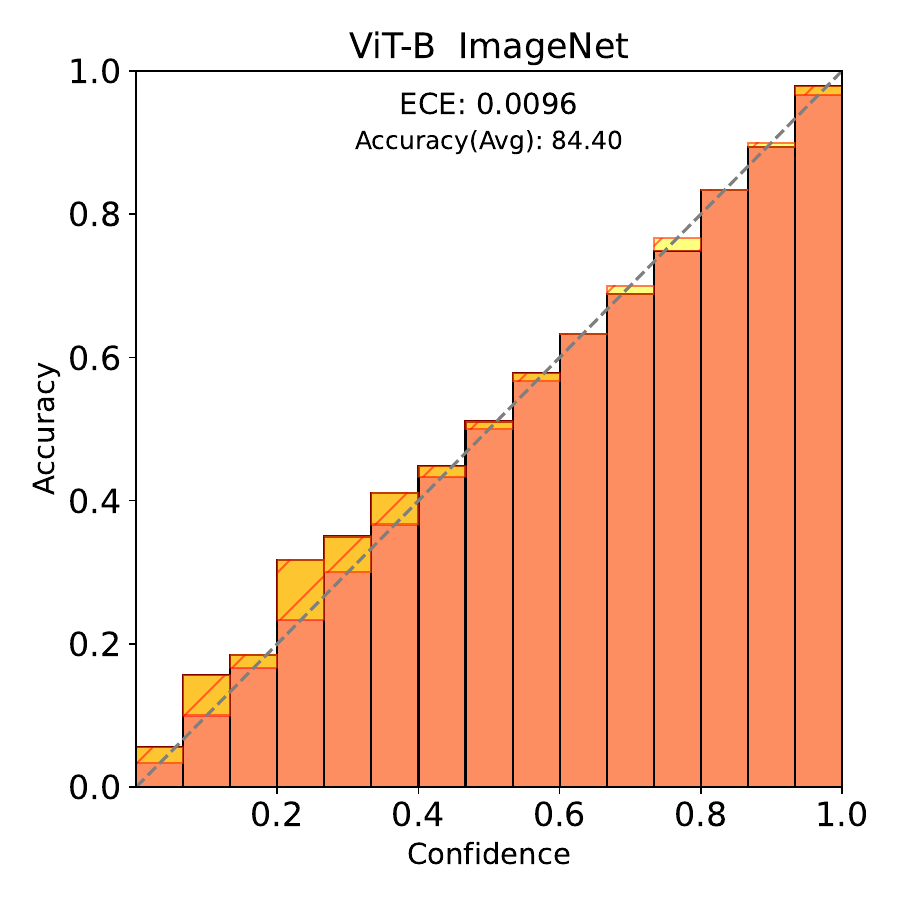}
\end{minipage}
\begin{minipage}{0.19\textwidth}
  \centering
  \includegraphics[height=3.8cm, width=\linewidth, keepaspectratio ]{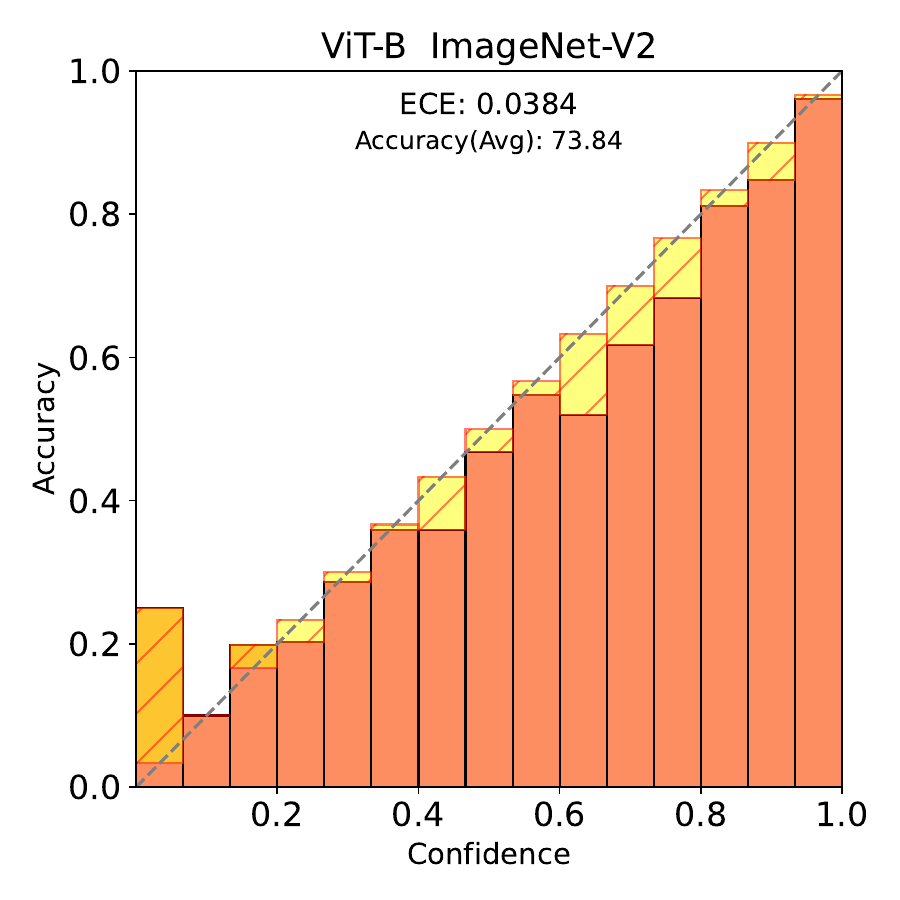}
\end{minipage}
\begin{minipage}{0.19\textwidth}
  \centering
  \includegraphics[height=3.8cm, width=\linewidth, keepaspectratio]{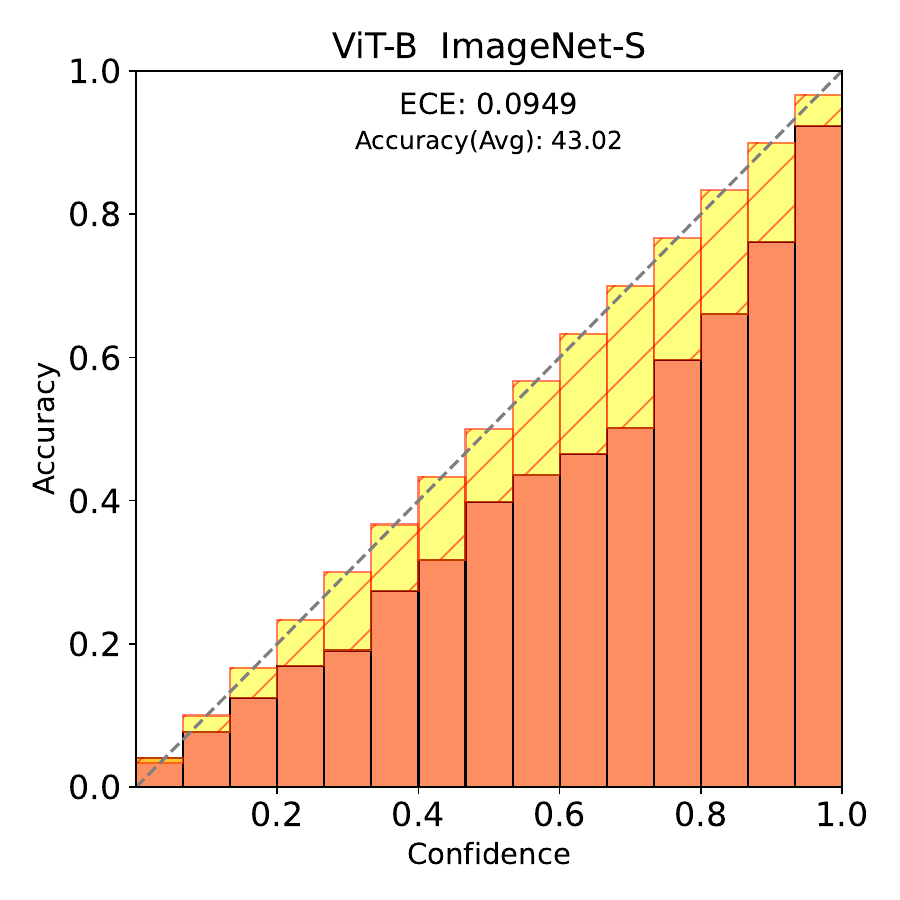}
\end{minipage}
\begin{minipage}{0.19\textwidth}
  \centering
  \includegraphics[height=3.8cm, width=\linewidth, keepaspectratio]{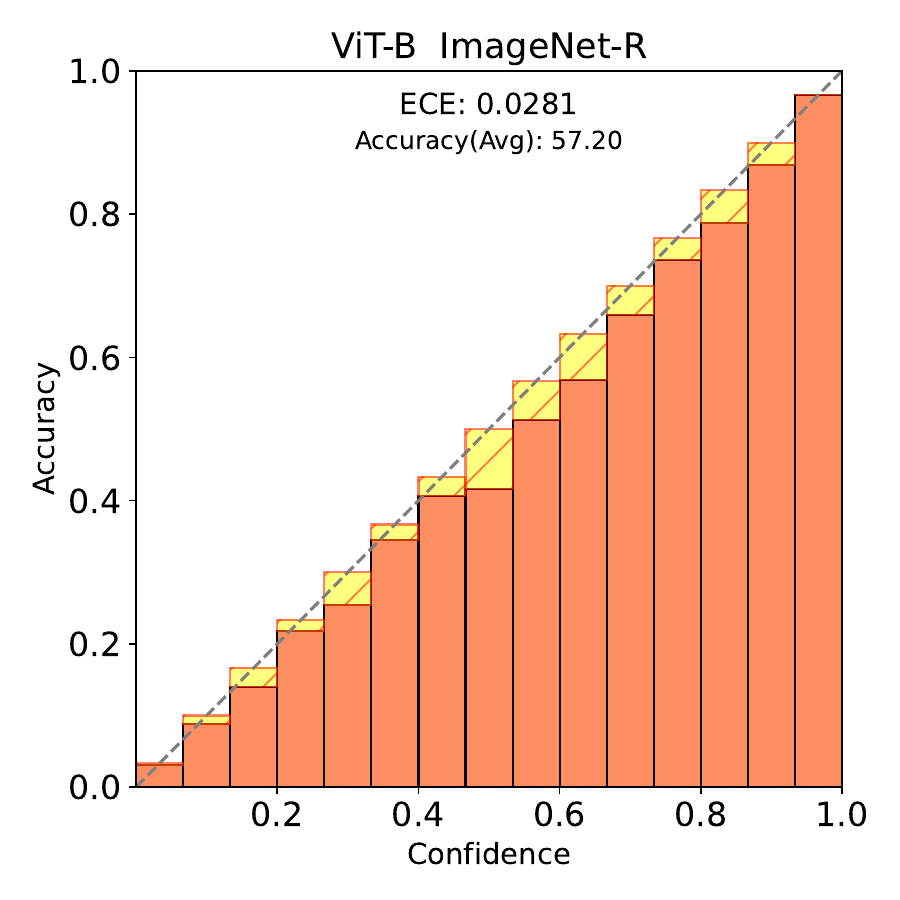}
\end{minipage}
\begin{minipage}{0.19\textwidth}
  \centering
  \includegraphics[height=3.8cm, width=\linewidth, keepaspectratio]{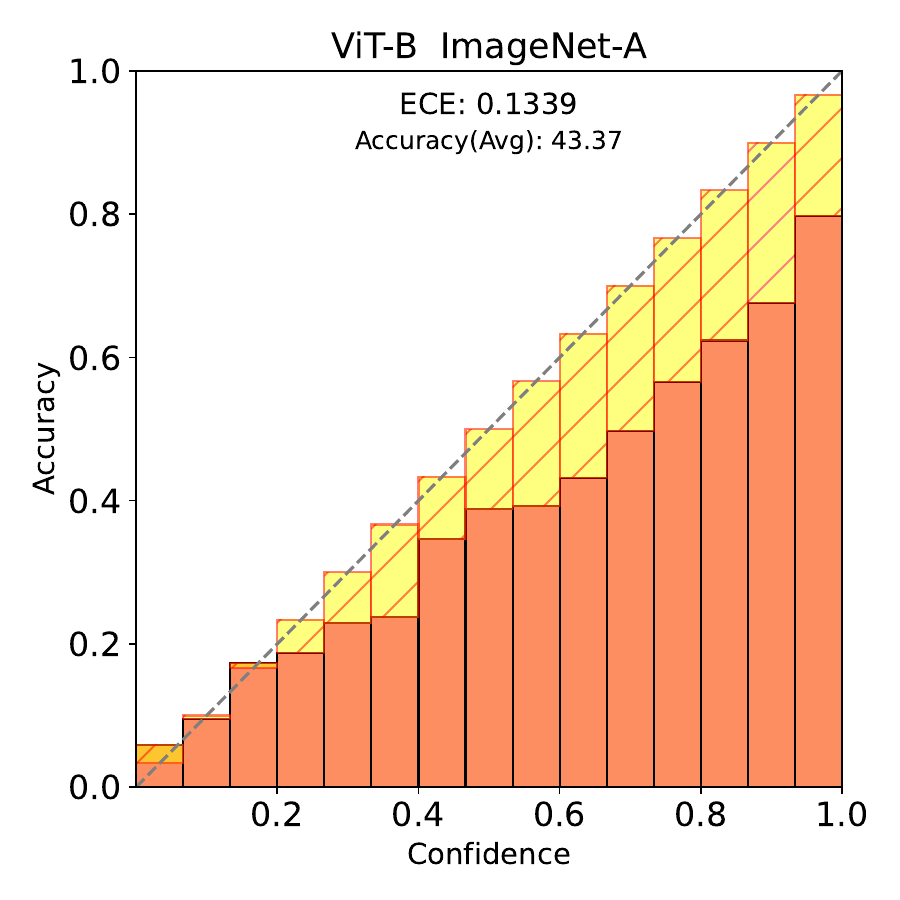}
\end{minipage}


\begin{minipage}{0.19\textwidth}
  \centering
  \includegraphics[height=3.8cm, width=\linewidth , keepaspectratio]{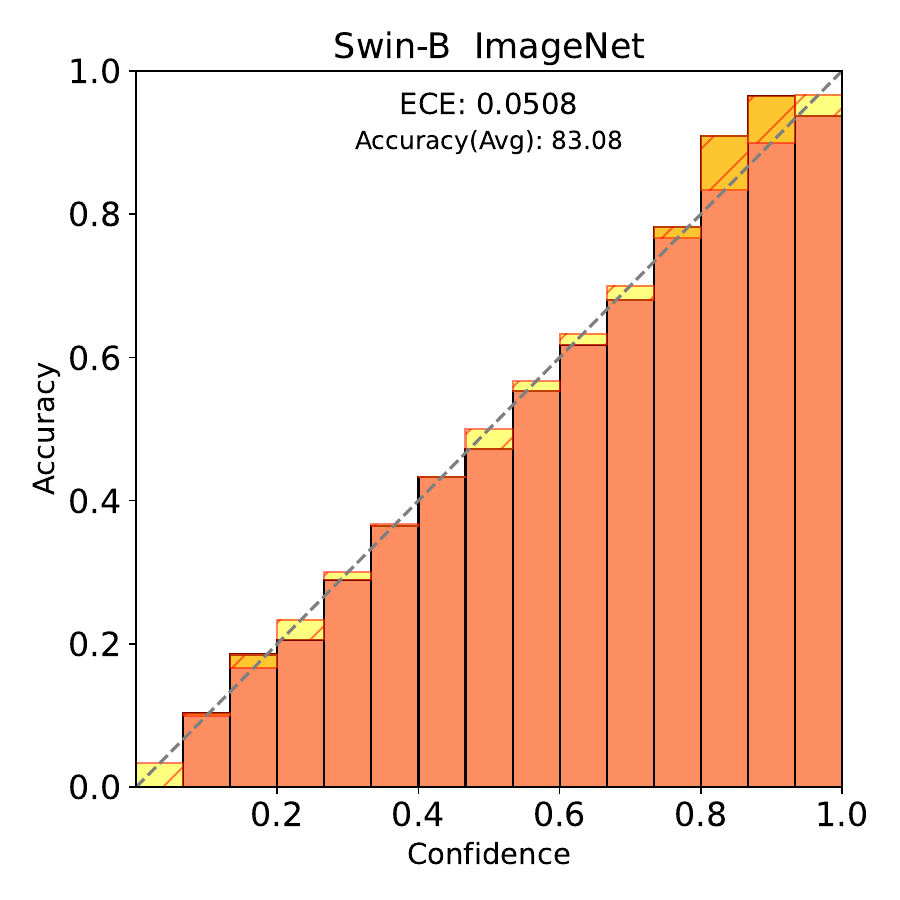}
\end{minipage}
\begin{minipage}{0.19\textwidth}
  \centering
  \includegraphics[height=3.8cm, width=\linewidth, keepaspectratio ]{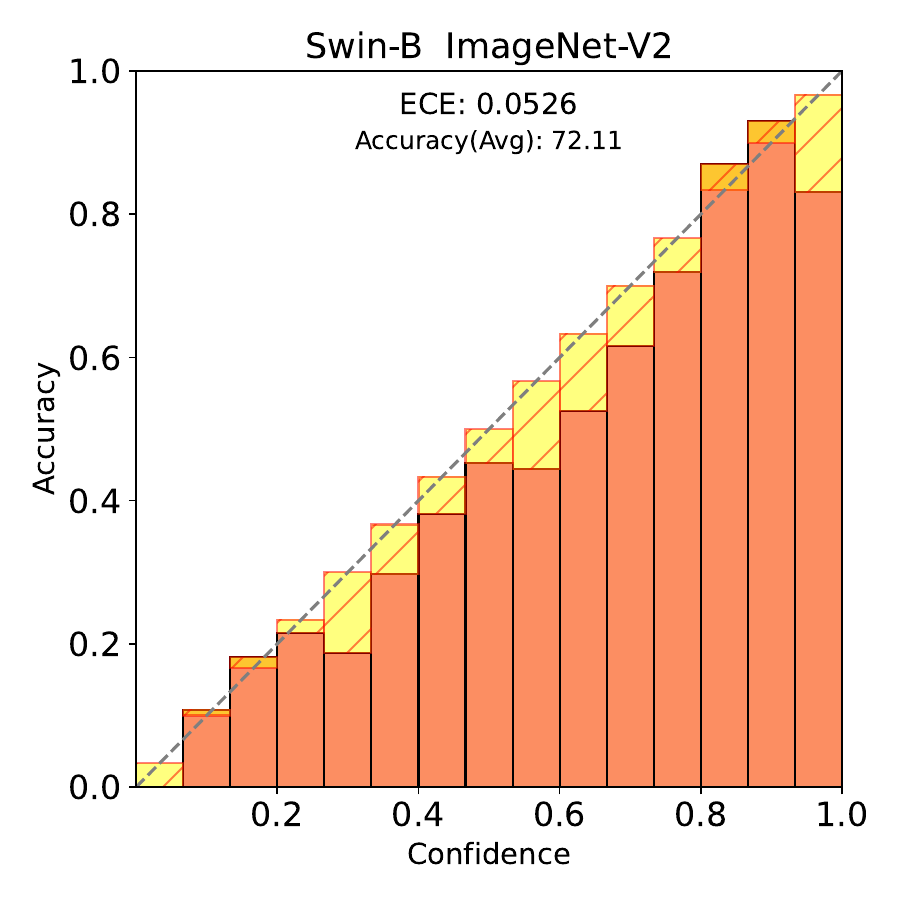}
\end{minipage}
\begin{minipage}{0.19\textwidth}
  \centering
  \includegraphics[height=3.8cm, width=\linewidth, keepaspectratio]{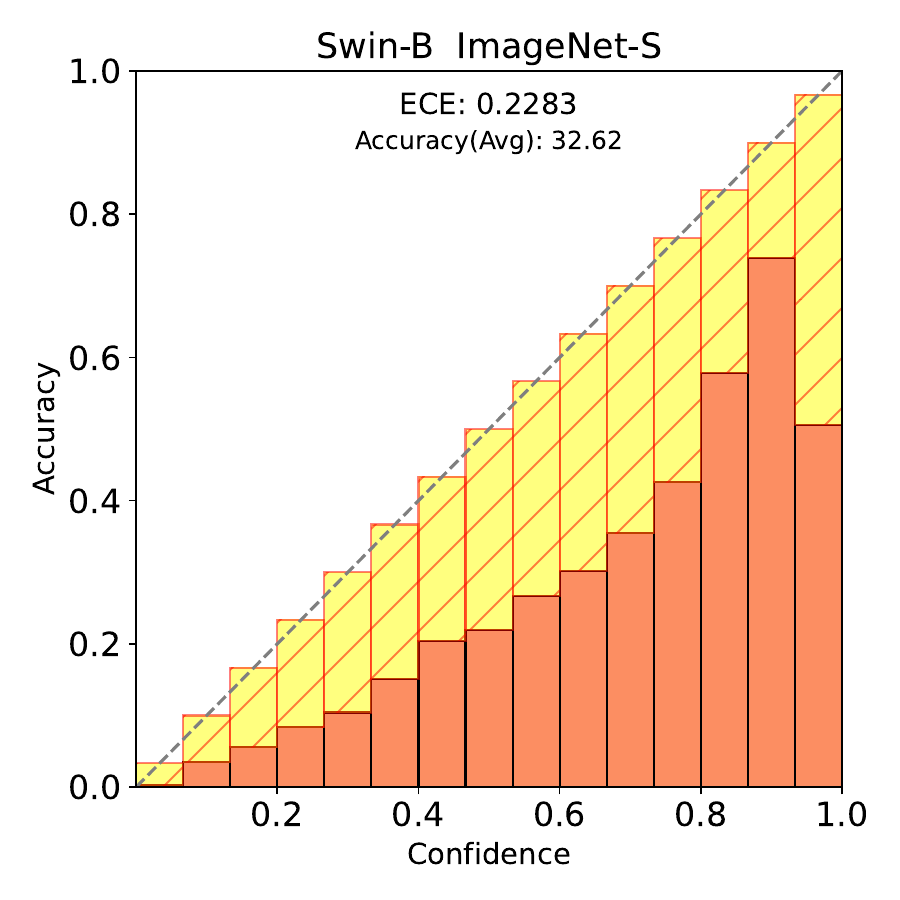}
\end{minipage}
\begin{minipage}{0.19\textwidth}
  \centering
  \includegraphics[height=3.8cm, width=\linewidth, keepaspectratio]{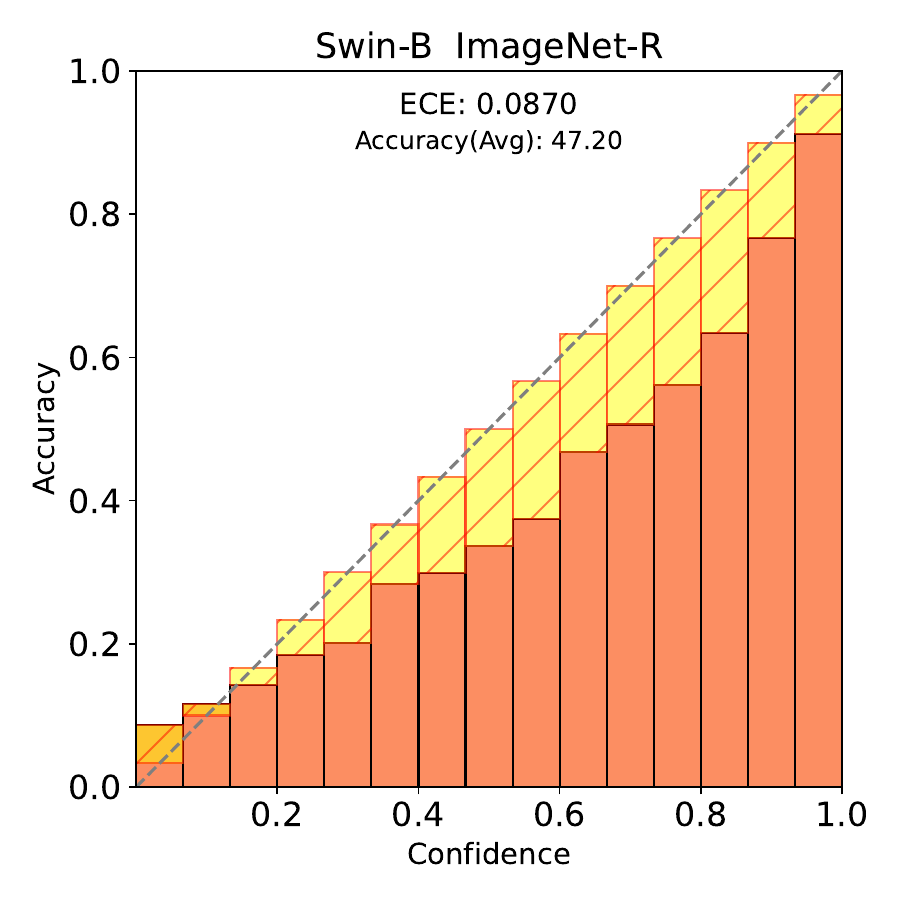}
\end{minipage}
\begin{minipage}{0.19\textwidth}
  \centering
  \includegraphics[height=3.8cm, width=\linewidth, keepaspectratio]{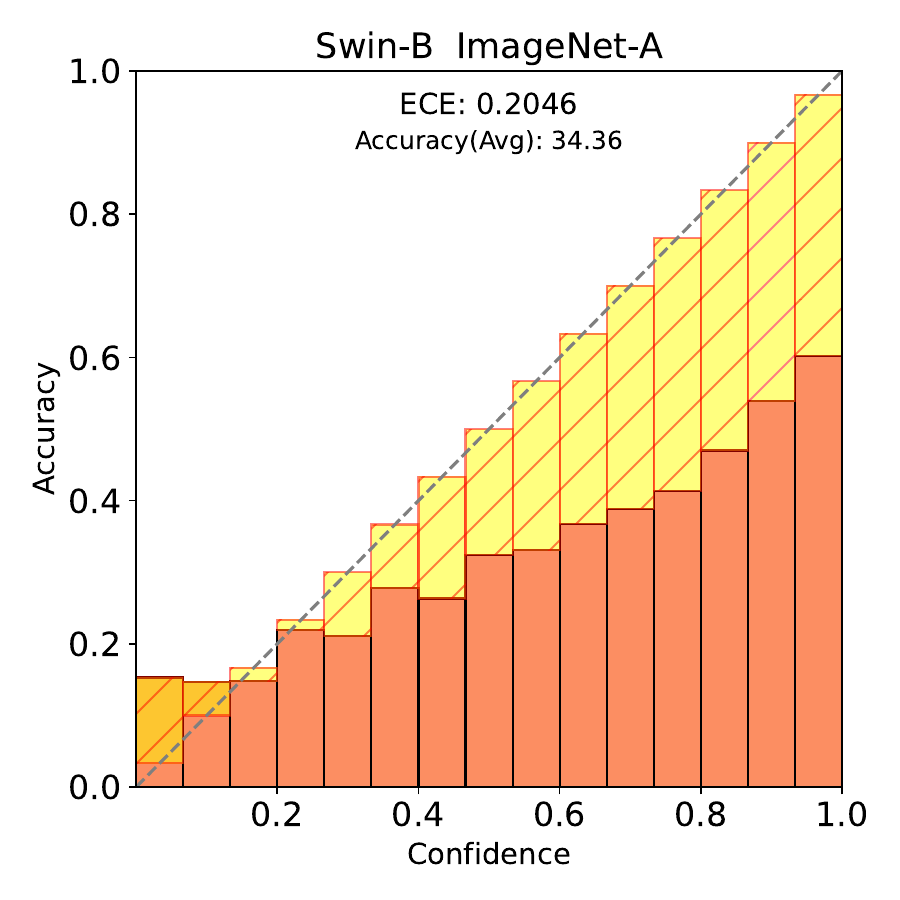}
\end{minipage}


\begin{minipage}{0.19\textwidth}
  \centering
  \includegraphics[height=3.8cm, width=\linewidth , keepaspectratio]{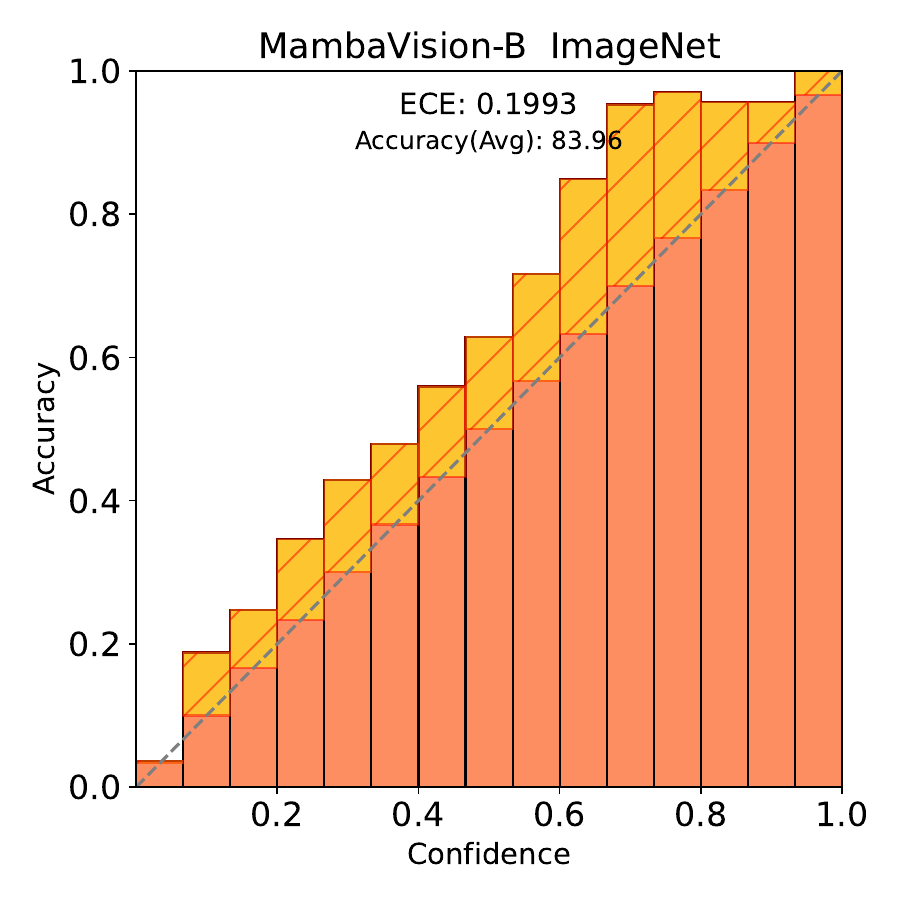}
\end{minipage}
\begin{minipage}{0.19\textwidth}
  \centering
  \includegraphics[height=3.8cm, width=\linewidth, keepaspectratio ]{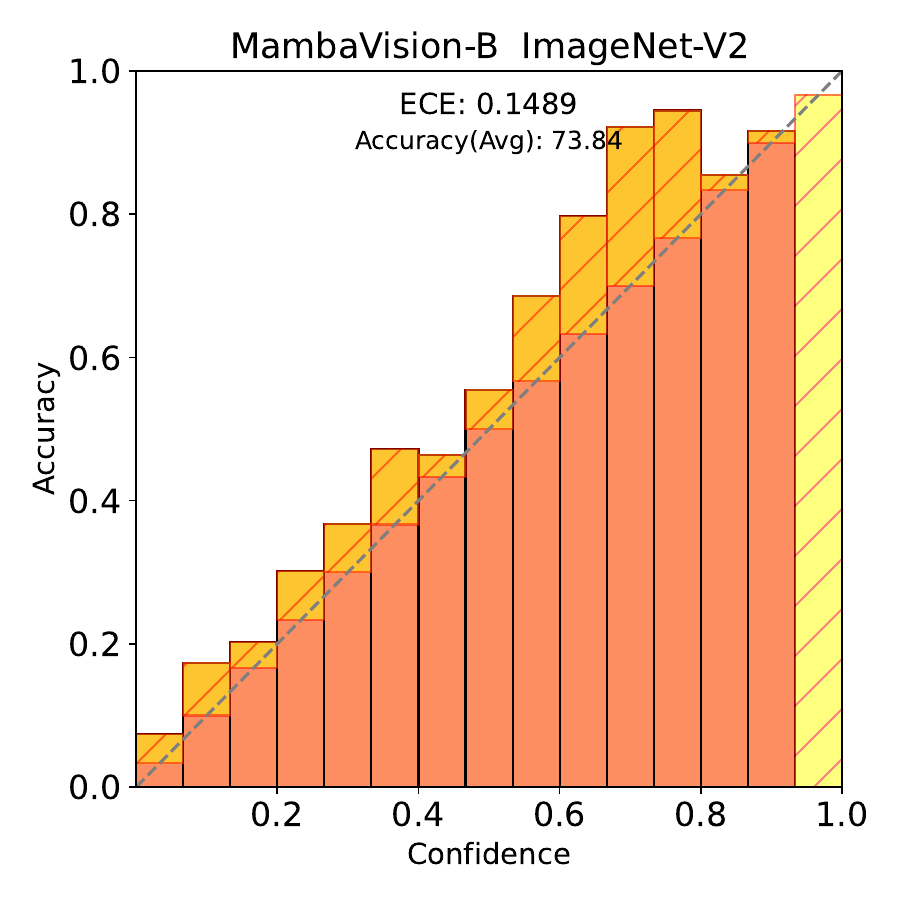}
\end{minipage}
\begin{minipage}{0.19\textwidth}
  \centering
  \includegraphics[height=3.8cm, width=\linewidth, keepaspectratio]{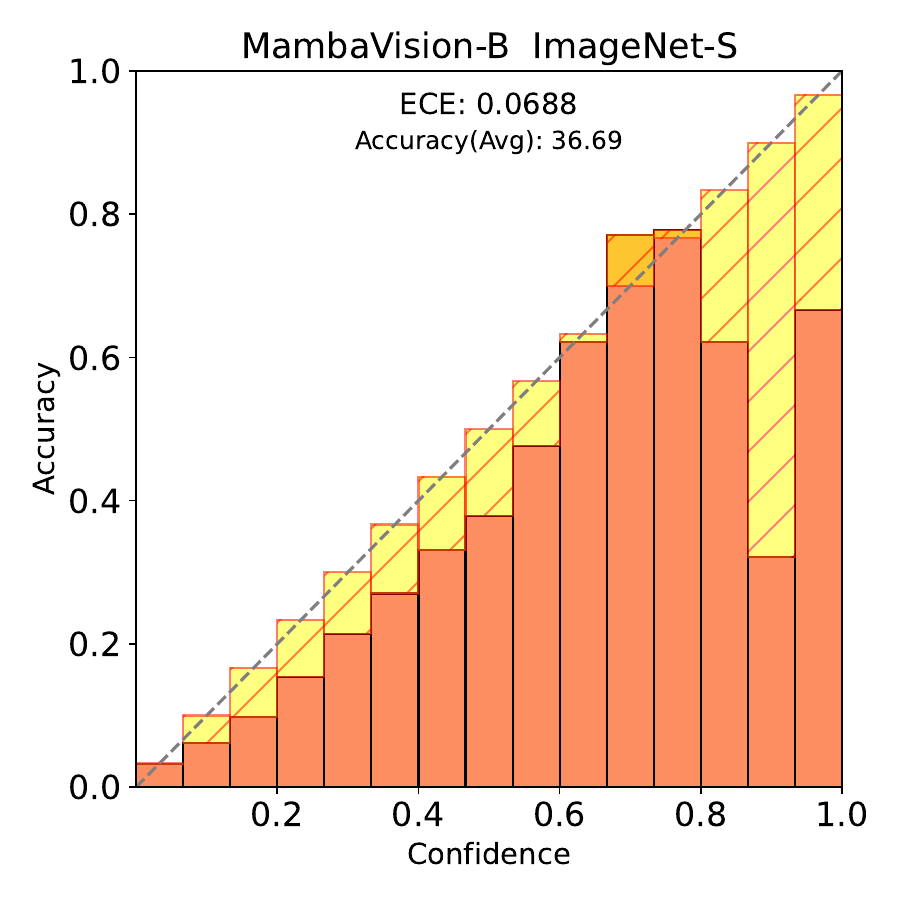}
\end{minipage}
\begin{minipage}{0.19\textwidth}
  \centering
  \includegraphics[height=3.8cm, width=\linewidth, keepaspectratio]{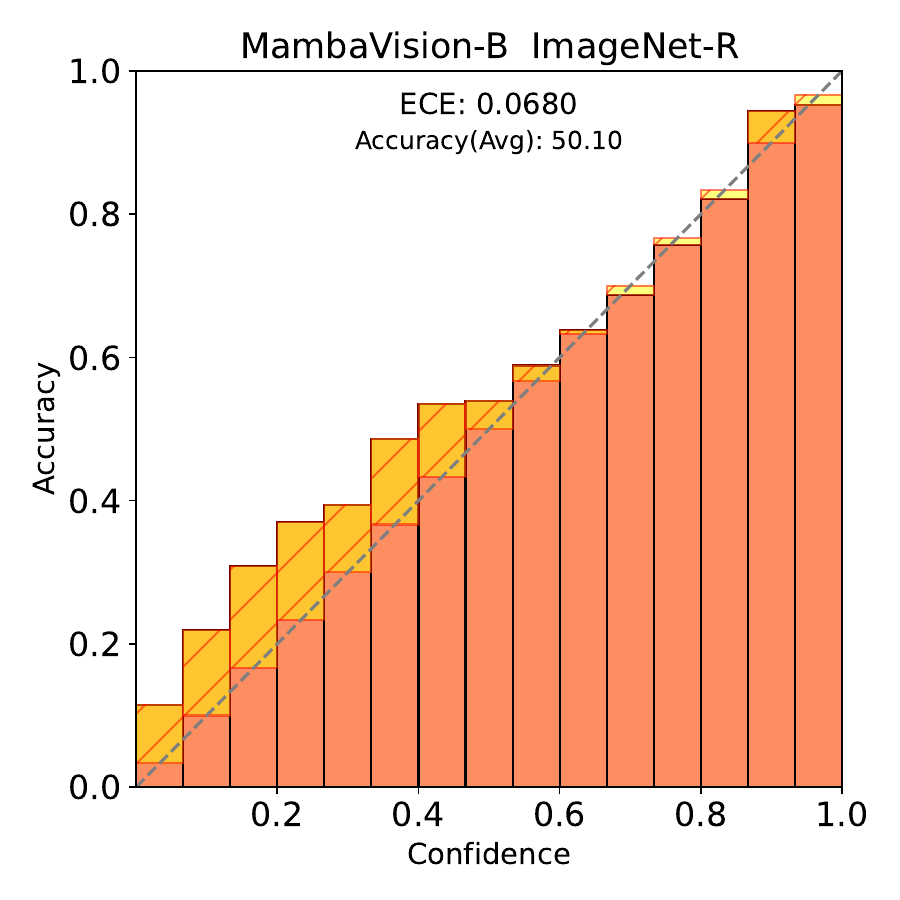}
\end{minipage}
\begin{minipage}{0.19\textwidth}
  \centering
  \includegraphics[height=3.8cm, width=\linewidth, keepaspectratio]{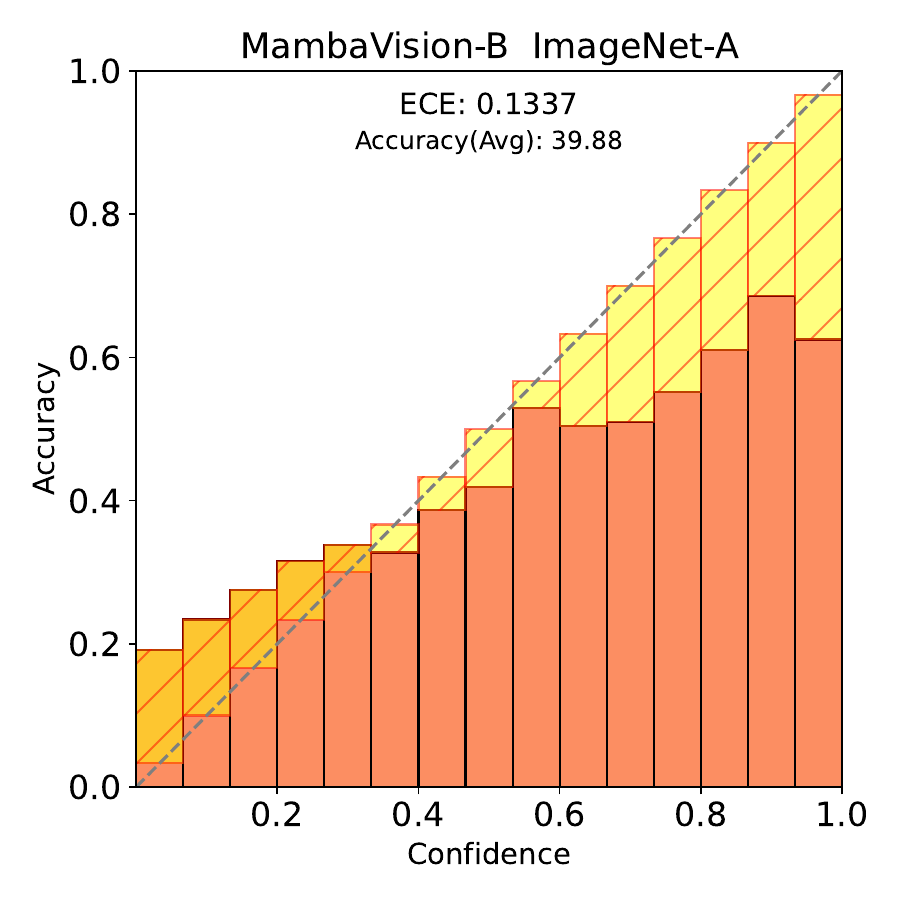}
\end{minipage}


\begin{minipage}{0.19\textwidth}
  \centering
  \includegraphics[height=3.8cm, width=\linewidth , keepaspectratio]{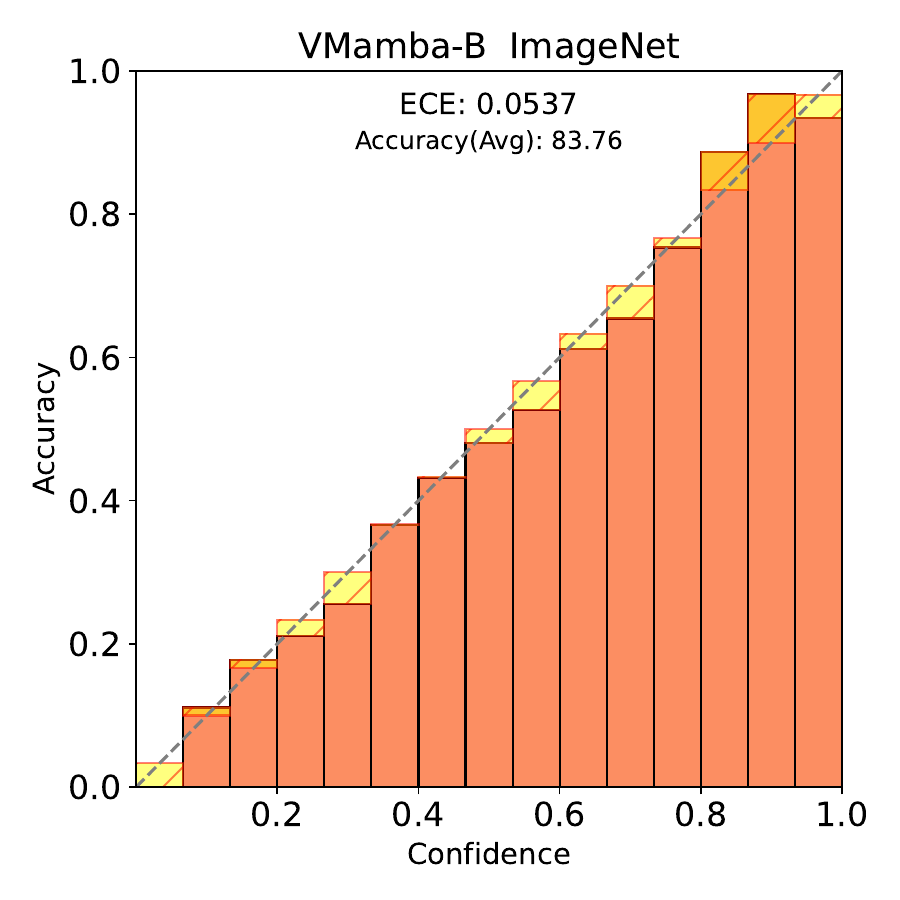}
\end{minipage}
\begin{minipage}{0.19\textwidth}
  \centering
  \includegraphics[height=3.8cm, width=\linewidth, keepaspectratio ]{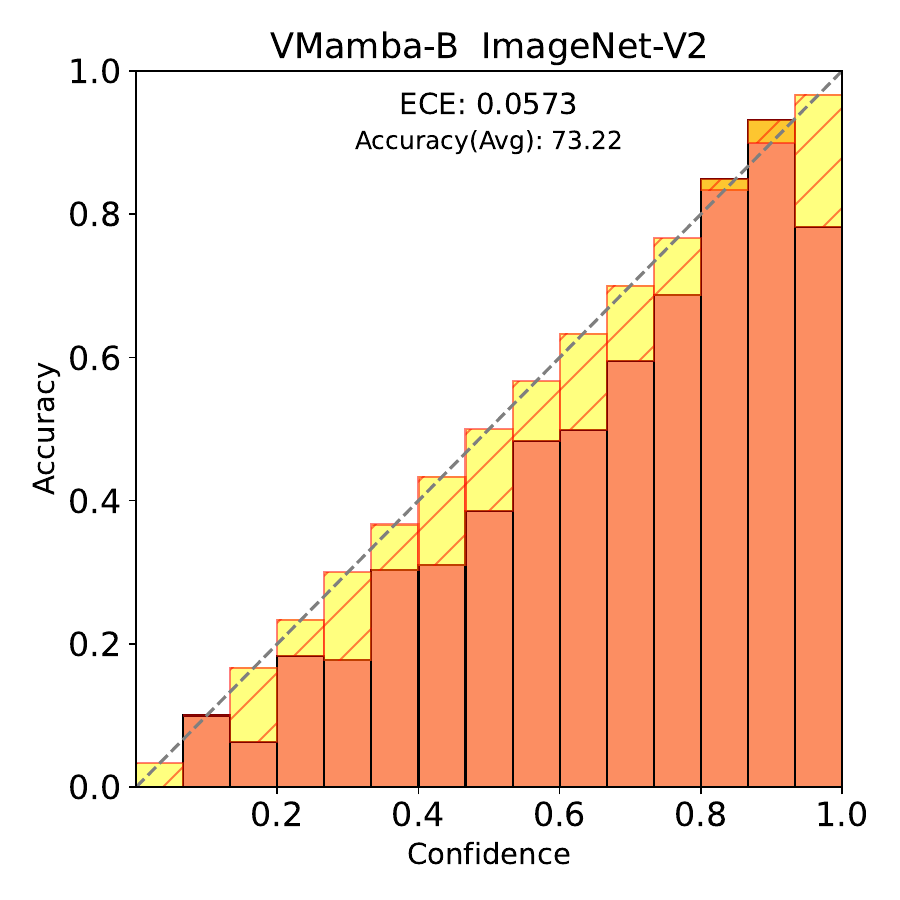}
\end{minipage}
\begin{minipage}{0.19\textwidth}
  \centering
  \includegraphics[height=3.8cm, width=\linewidth, keepaspectratio]{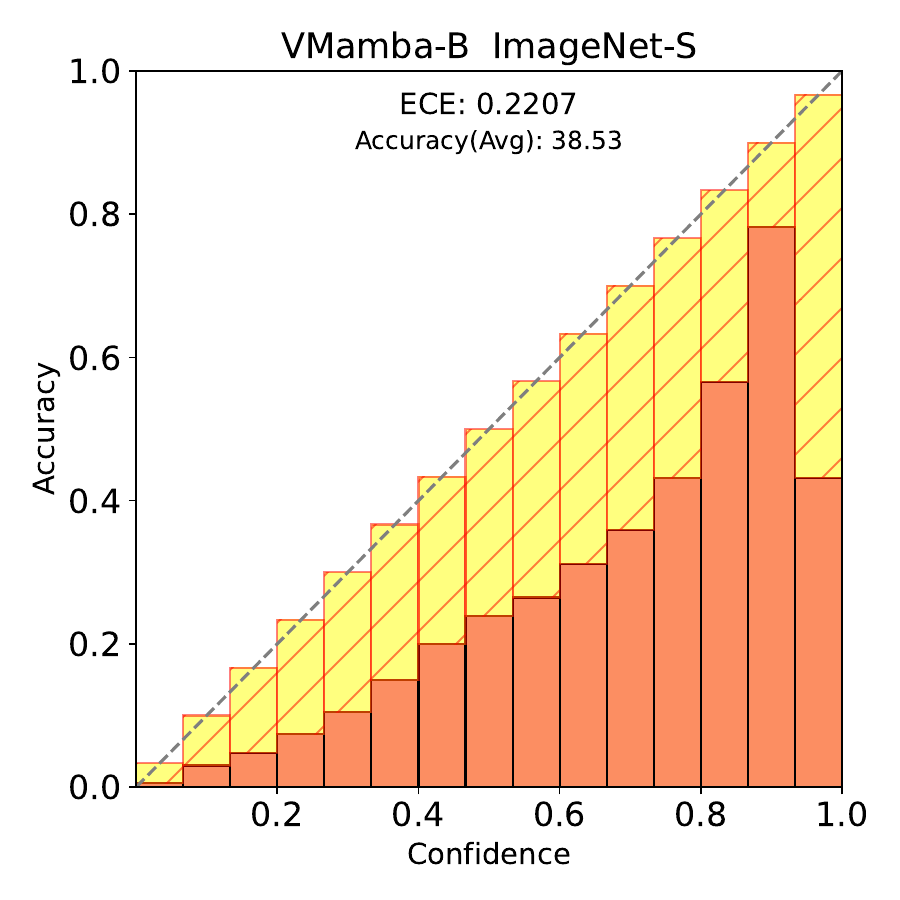}
\end{minipage}
\begin{minipage}{0.19\textwidth}
  \centering
  \includegraphics[height=3.8cm, width=\linewidth, keepaspectratio]{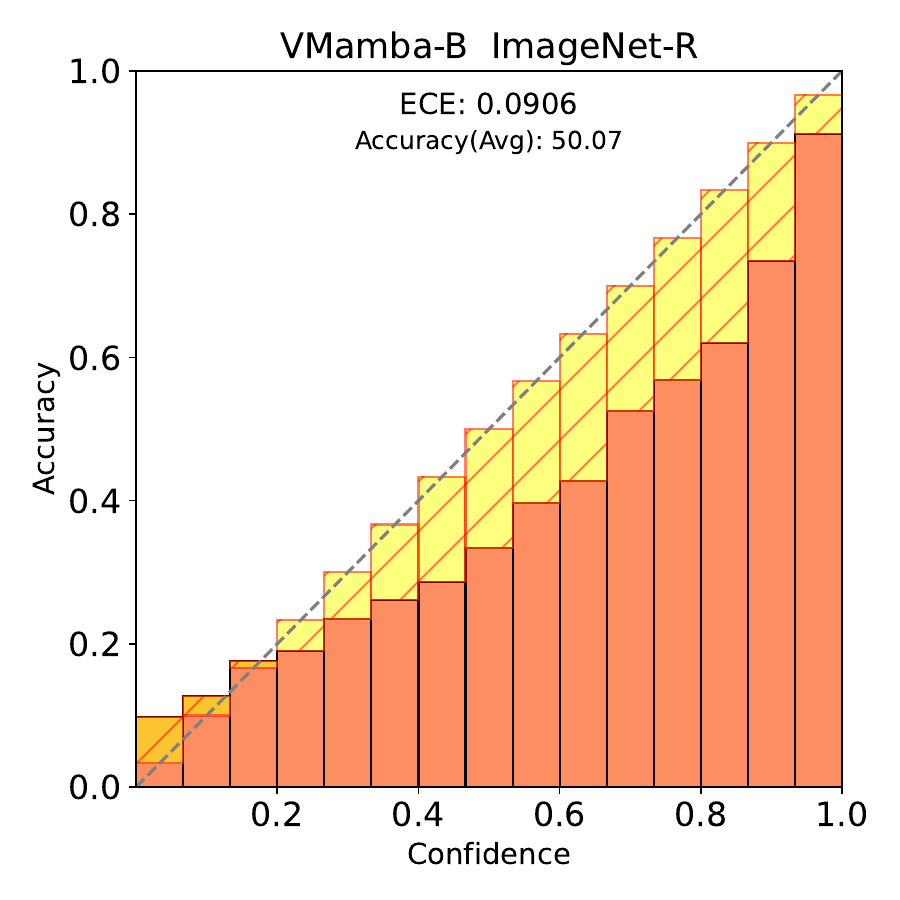}
\end{minipage}
\begin{minipage}{0.19\textwidth}
  \centering
  \includegraphics[height=3.8cm, width=\linewidth, keepaspectratio]{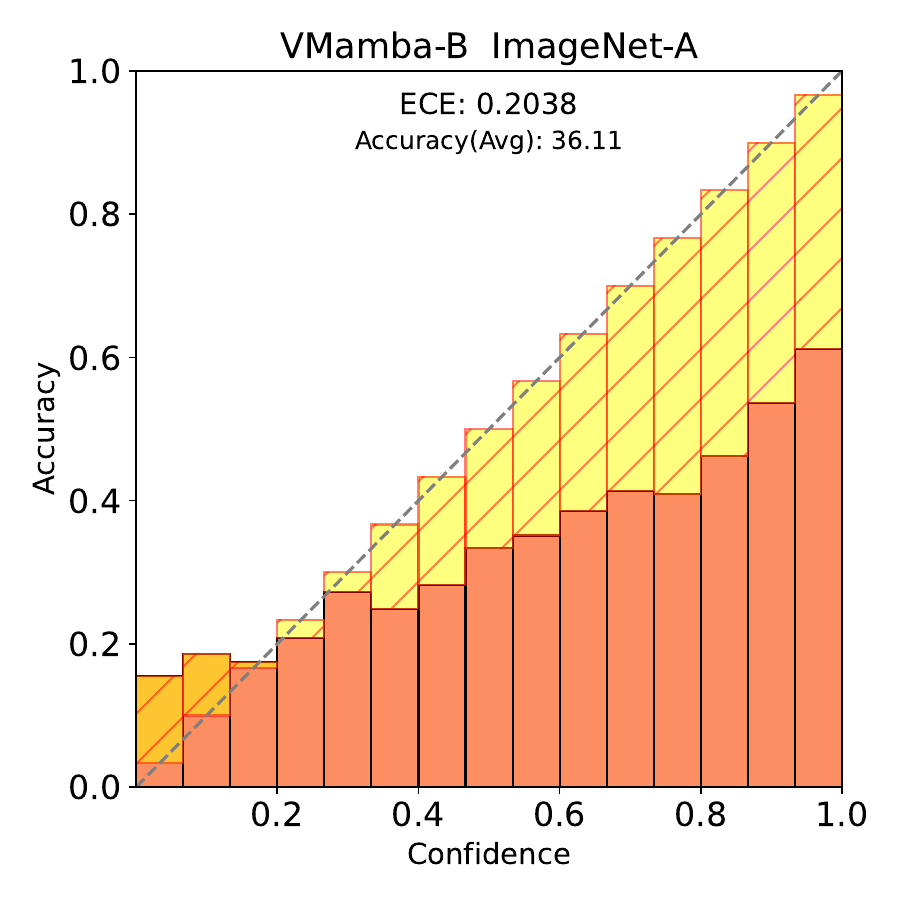}
\end{minipage}

\end{minipage}
\hfill
  \caption{Caliberation Results: Reliability diagrams and ECE on ViT-B, Swin-B, VMamba-B, and MambaVision-B across ImageNet, ImageNet-V2, ImageNet-S, ImageNet-R, and ImageNet-A.}
  \label{fig:calib_app_base}
\vspace{-1em}
\end{figure*}

\begin{figure*}[h]
\includegraphics[width=\textwidth, trim = 0cm 0cm 0cm 0cm, clip]{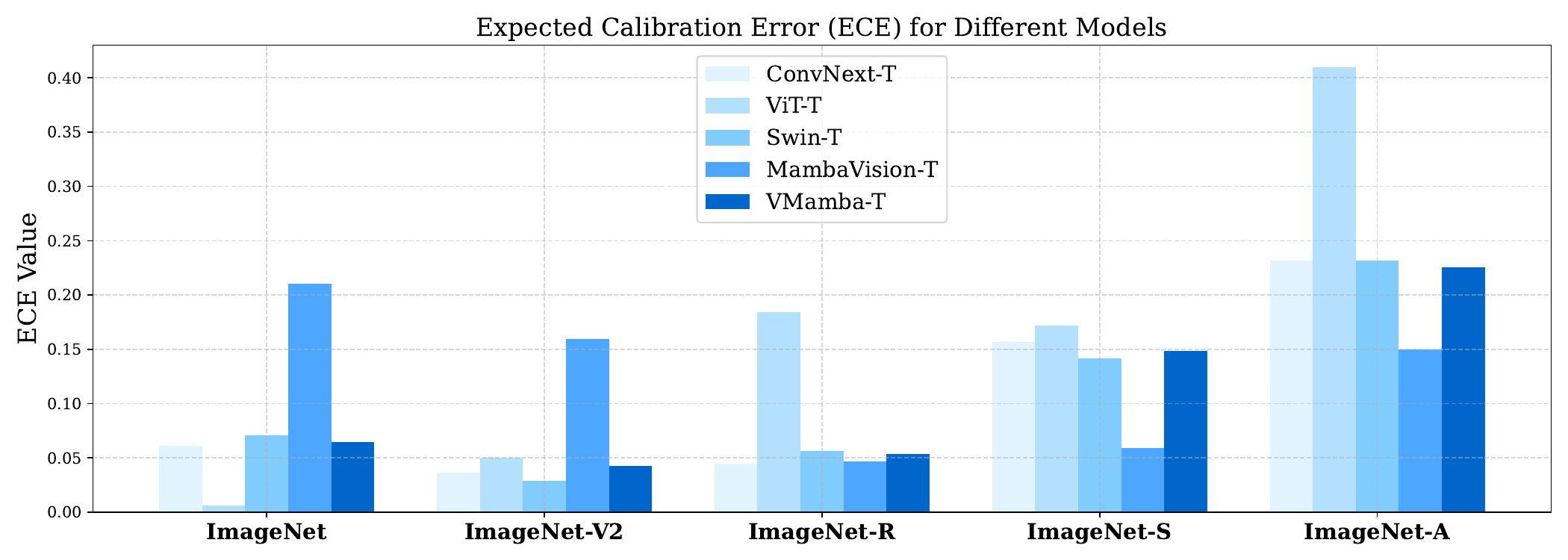}
\includegraphics[width=\textwidth, trim = 0cm 0cm 0cm 0cm, clip]{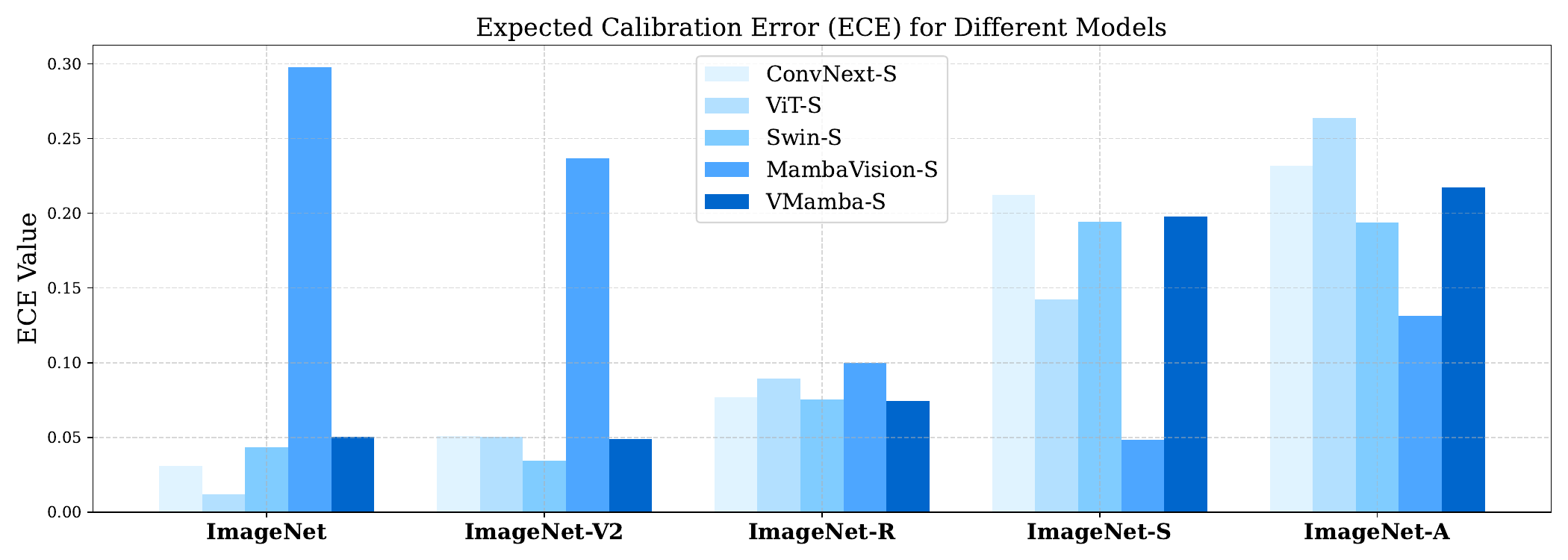}
\includegraphics[width=\textwidth, trim = 0cm 0cm 0cm 0cm, clip]{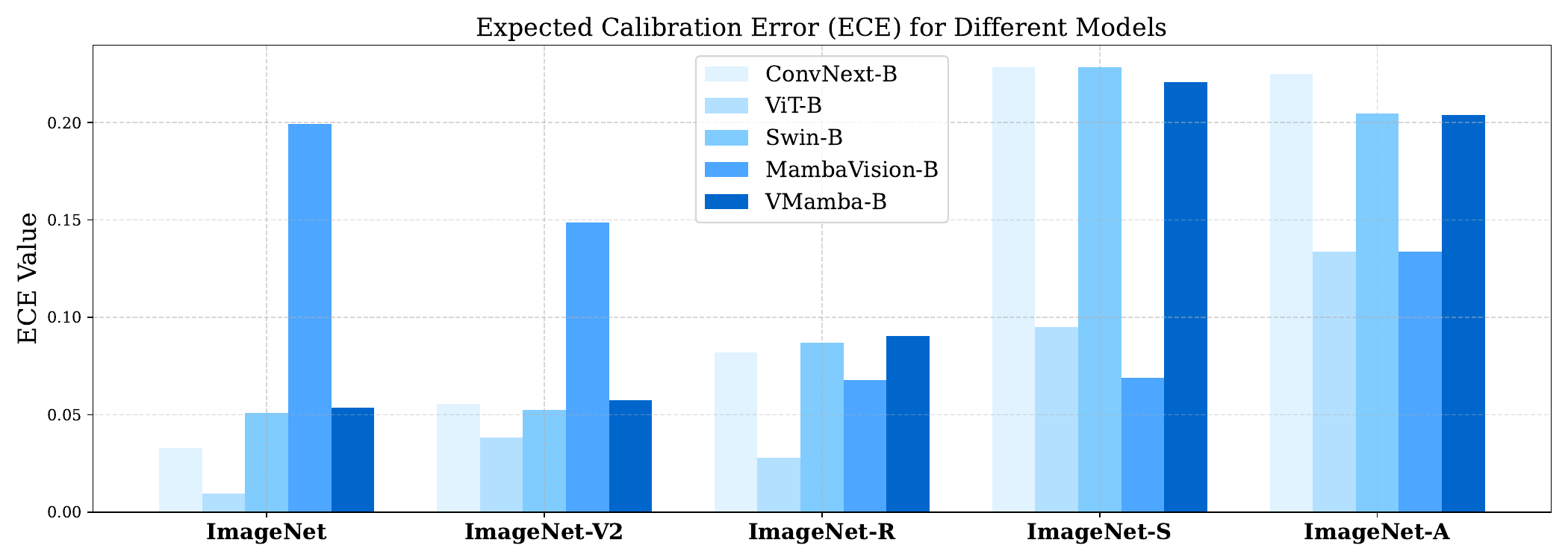}
\vspace{-1em}
\caption{\small ECE error across classification models across ImageNet, ImageNet-V2, ImageNet-S, ImageNet-R, and ImageNet-A. }
\label{fig:ece_error_ap}
\end{figure*}

Model calibration assesses how well a model's predicted confidence aligns with its actual accuracy. For example, if a model predicts a confidence level of 70\% for its predictions, a well-calibrated model should have an actual accuracy close to 70\%. To quantify this alignment, we use the Expected Calibration Error (ECE). ECE involves dividing predictions into \( M \) bins based on their confidence levels (e.g., 60\%-70\%, 70\%-80\%). For each bin, the average accuracy and confidence are computed, and the ECE is the weighted average of the differences between these values. Calibration can also be evaluated visually using reliability diagrams, which plot predicted confidence against actual accuracy; a well-calibrated model should show points near the diagonal. Additionally, confidence histograms reveal the distribution of prediction confidences. Evaluation is performed on both in-distribution data (e.g., ImageNet, ImageNetV2) and out-of-distribution data (e.g., ImageNet-R, ImageNet-S, ImageNet-A), with \( M = 15 \) bins used in all experiments. In Figure \ref{fig:calib_app_tiny}, \ref{fig:calib_app_small}, and \ref{fig:calib_app_base}, we report the ECE plot the reliability diagrams for Tiny, Small, and Base version of different architectures, respectively. Figure \ref{fig:ece_error_ap}, we report the ECE score. We observe that while ViT models report a low ECE error on in-distribution datasets, the error increases significantly for out-of-distribution datasets. An exception is the Base model, which can be attributed to the ViT-B model being pretrained on a larger dataset (ImageNet-21k) comapared to other models. 
For the hybrid-based VSSM model MambaVision we observe a contrasting behaviour with ViT models,  while the ECE score on in-distribution datasets is relatively high, it improves significantly on out-of-distribution datasets. The improvement in ECE for MambaVision on out-of-distribution datasets suggests that the hybrid approach of combining vision transformers with other model components may enhance calibration performance in these scenarios.

\subsection{Robustness against Fine-grained Corruptions}
\label{app:corruptions}

In Table~\ref{tab:imagenet_e_b_app} \emph{(left)} we report results on ImageNet-E dataset across several CNNs, Transformers, and VSSM-based models, and in Table~\ref{tab:imagenet_e_b_app} \emph{(right)}, we report results on ImageNet-B dataset.

\subsection{Robustness Evaluation for Object Detection and Semantic Segmentation}
\label{app:detection_corruptions}

\begin{figure*}[h]
\includegraphics[width=\textwidth, trim = 0cm 0cm 0cm 0cm, clip]{images/non_adversarial/ade/ade_c_results_miou.pdf}
\includegraphics[width=\textwidth, trim = 0cm 0cm 0cm 0cm, clip]{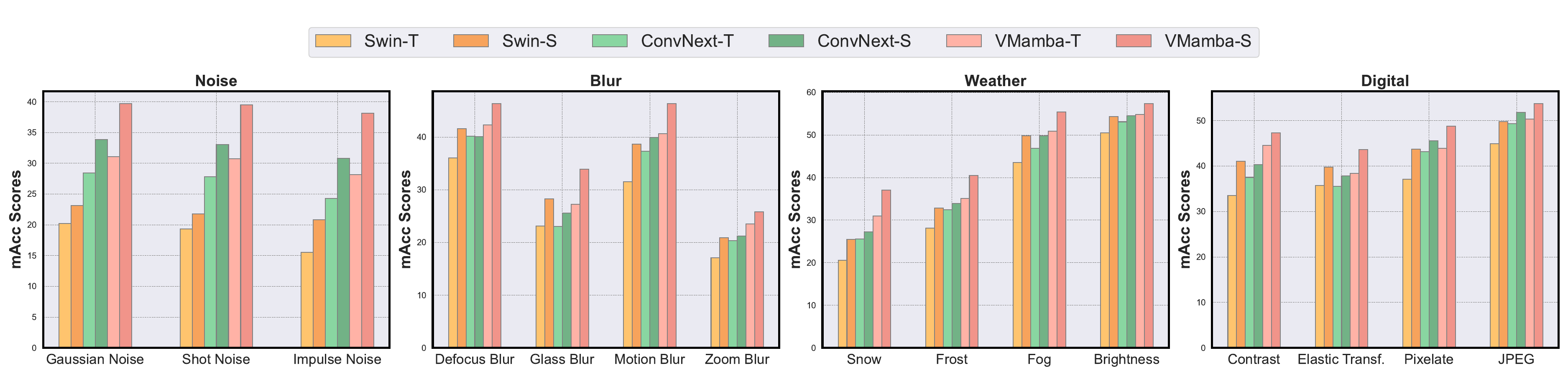}
\includegraphics[width=\textwidth, trim = 0cm 0cm 0cm 0cm, clip]{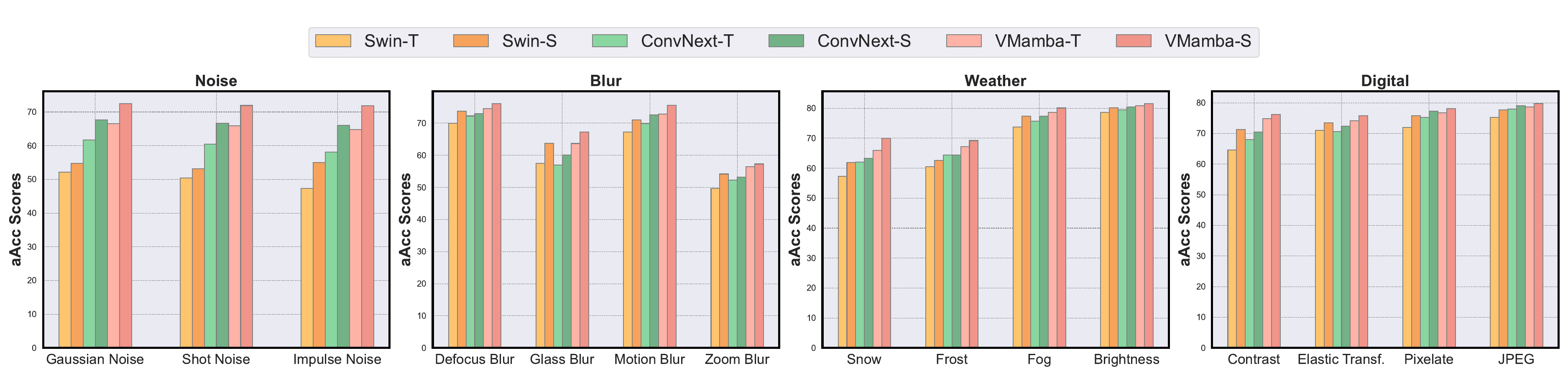}
\vspace{-1em}
\caption{\small mIoU, mAcc, and aAcc score for different architectures on AED20k-C dataset}
\label{fig:aed_c_all_metrics}
\end{figure*}

\begin{figure*}[h]
\includegraphics[width=\textwidth, trim = 0cm 0cm 0cm 0cm, clip]{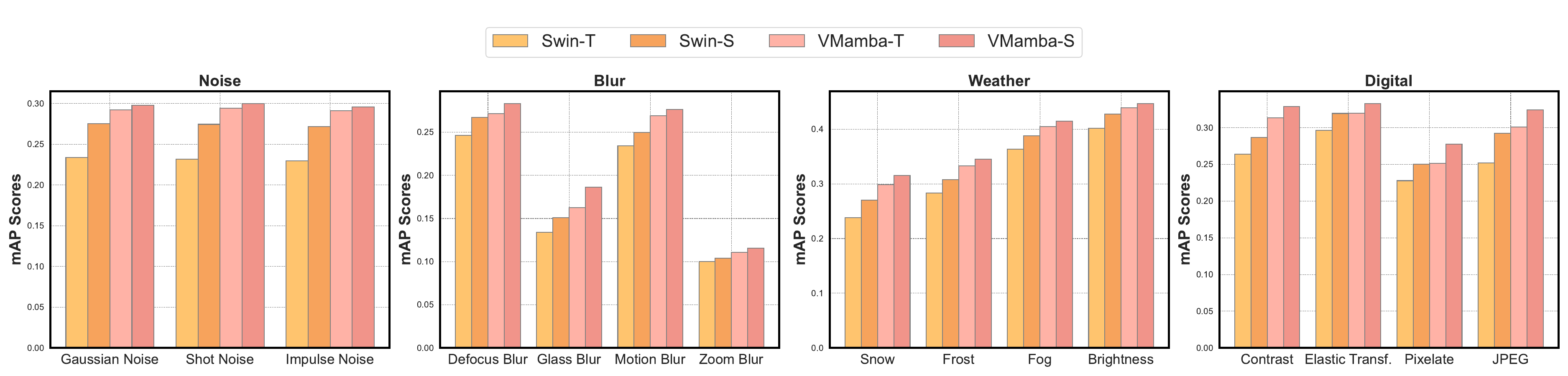}
\includegraphics[width=\textwidth, trim = 0cm 0cm 0cm 0cm, clip]{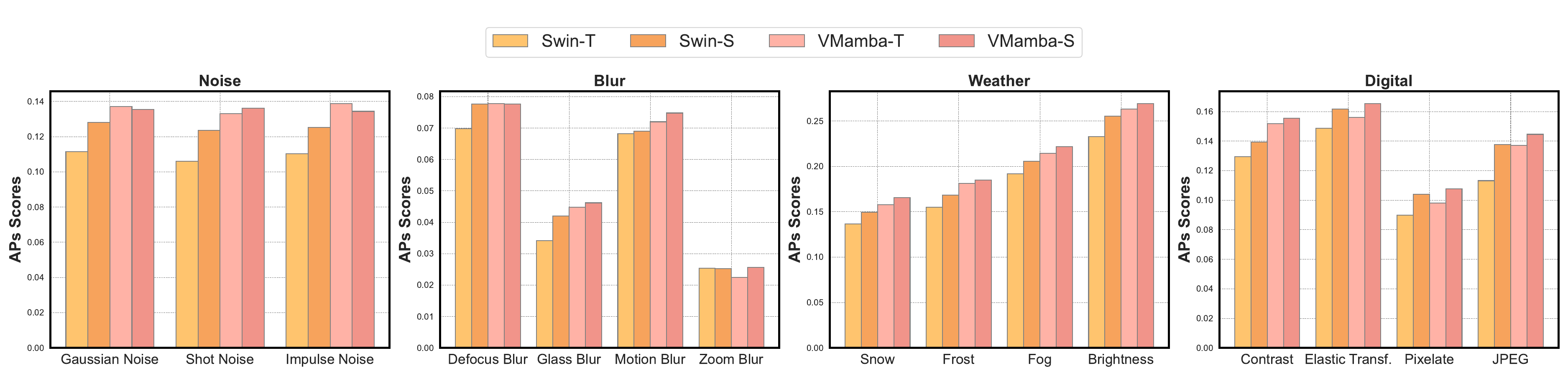}
\includegraphics[width=\textwidth, trim = 0cm 0cm 0cm 0cm, clip]{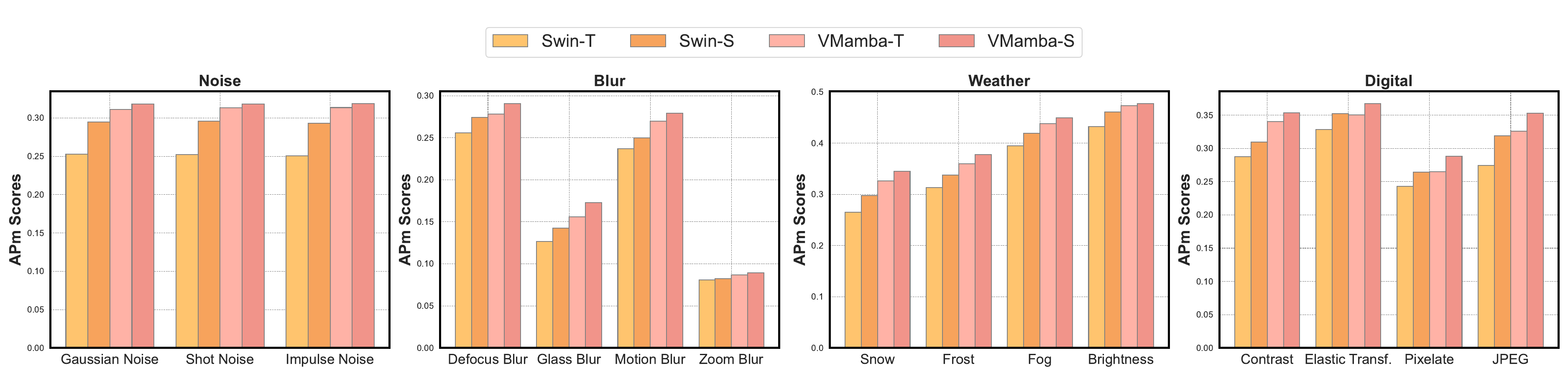}
\includegraphics[width=\textwidth, trim = 0cm 0cm 0cm 0cm, clip]{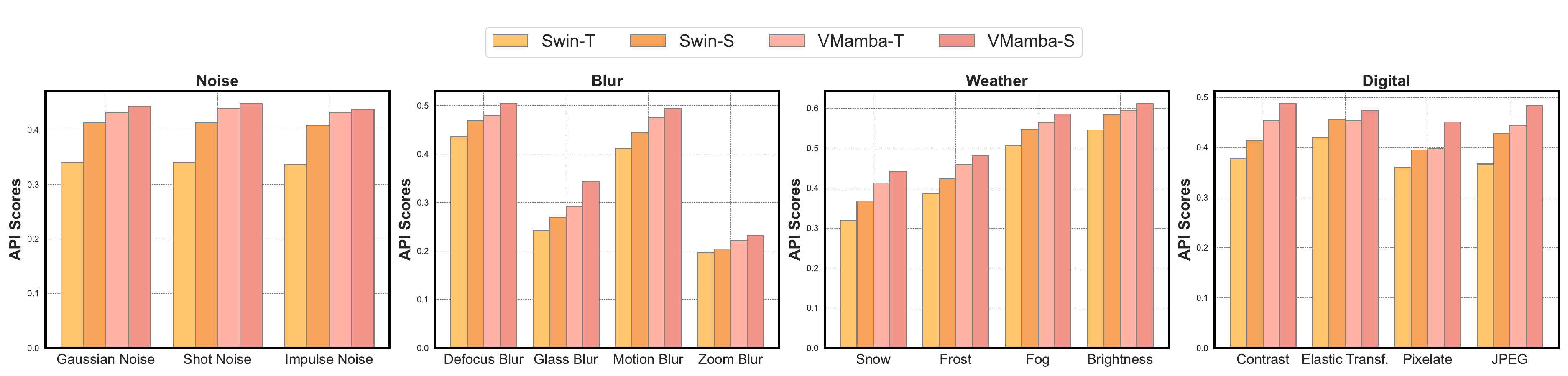}
\vspace{-1em}
\caption{\small $mAP$, $AP_s$, $AP_m$, and $AP_l$ score for different architectures on COCO-C dataset}
\label{fig:coco_c_all_ap}
\end{figure*}

In Table \ref{tab:coco_dc}, we report AP scores of different architectures on COCO-DC dataset. On COCO-DC, we observe that VMamba models' high performance on clean images does not translate to color and texture background variations. Swin-S model achieves the highest average AP score of $56.70$ across the background variations, followed by score of $56.20$ by VMamba-S model.
Figure \ref{fig:aed_c_all_metrics} shows the mIoU, mAcc, and aAcc scores on ADE-20K after applying common corruptions to the dataset. Similarly, Figure \ref{fig:coco_c_all_ap} displays the mAP, APs, APm, and APl scores on the COCO-C dataset. All the scores are averaged across all severity levels of each corruption. 
\section{Robustness against Adversarial Attacks}
\label{app:adv_attack}

\begin{figure*}[h]
\centering
\includegraphics[width=\textwidth]{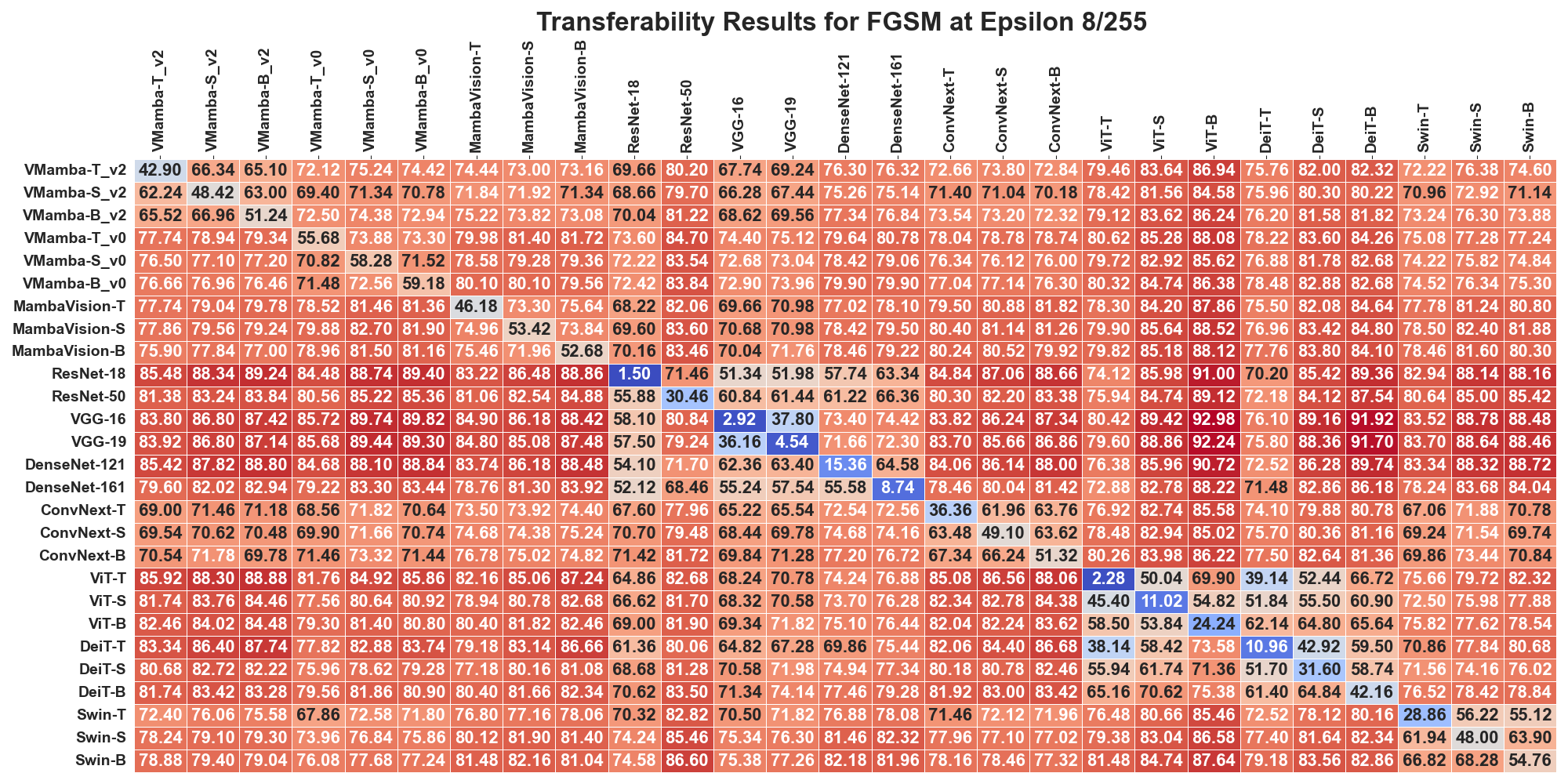}

\caption{Robust accuracy of various architectures under white-box and black-box settings for FGSM  attack. Adversarial examples are crafted at a perturbation budget $\epsilon=\frac{8}{255}$.}
\label{fig:fgsm_app}
\end{figure*}

\begin{figure*}[!t]
\centering
\includegraphics[width=\textwidth]{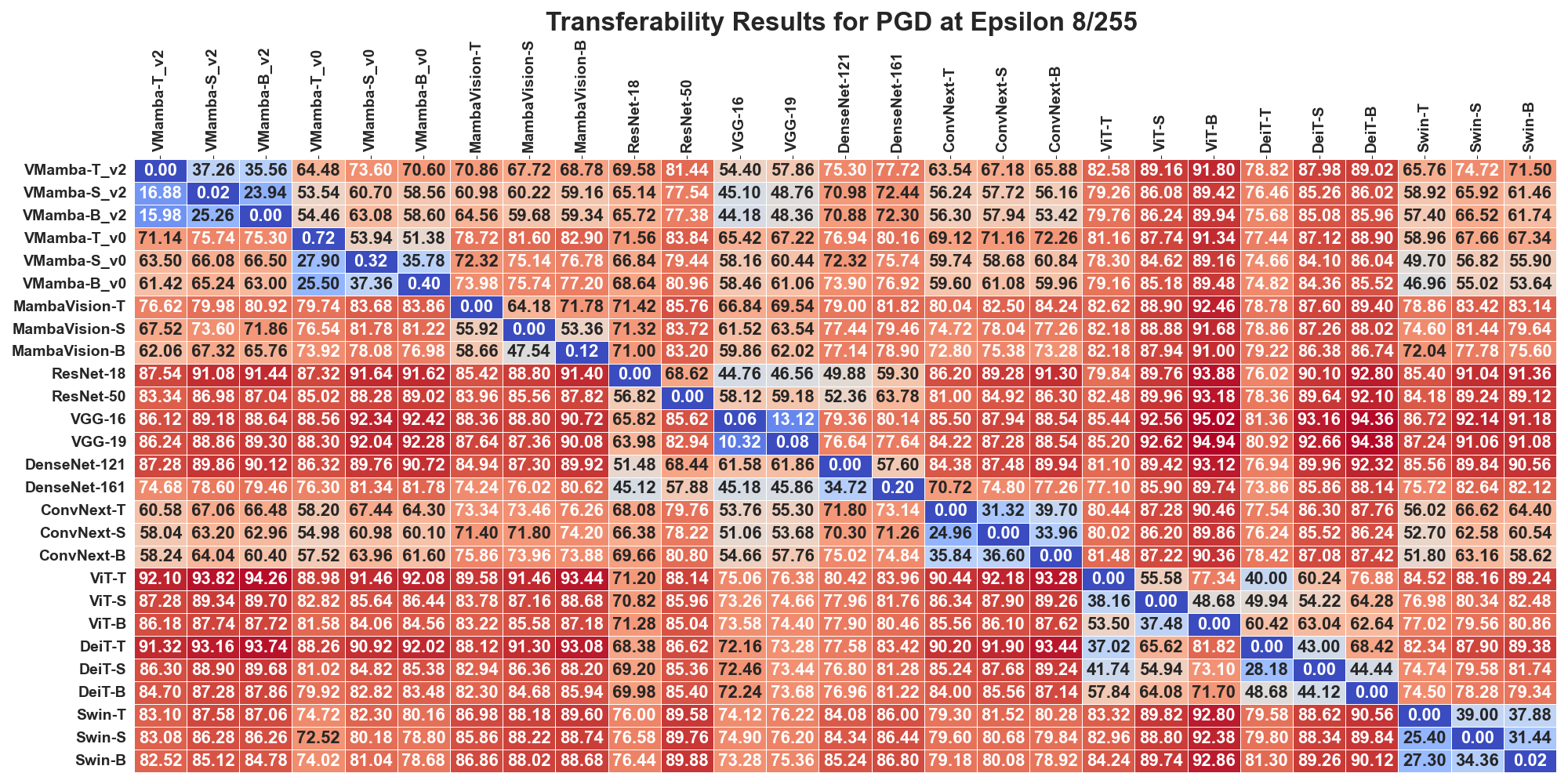}

\caption{Robust accuracy of various architectures under white-box and black-box settings for PGD attack. Adversarial examples are crafted at a perturbation budget $\epsilon=\frac{8}{255}$.}
\label{fig:pgd_app}
\end{figure*}

\begin{figure*}[!t]
\centering
\includegraphics[width=\textwidth]{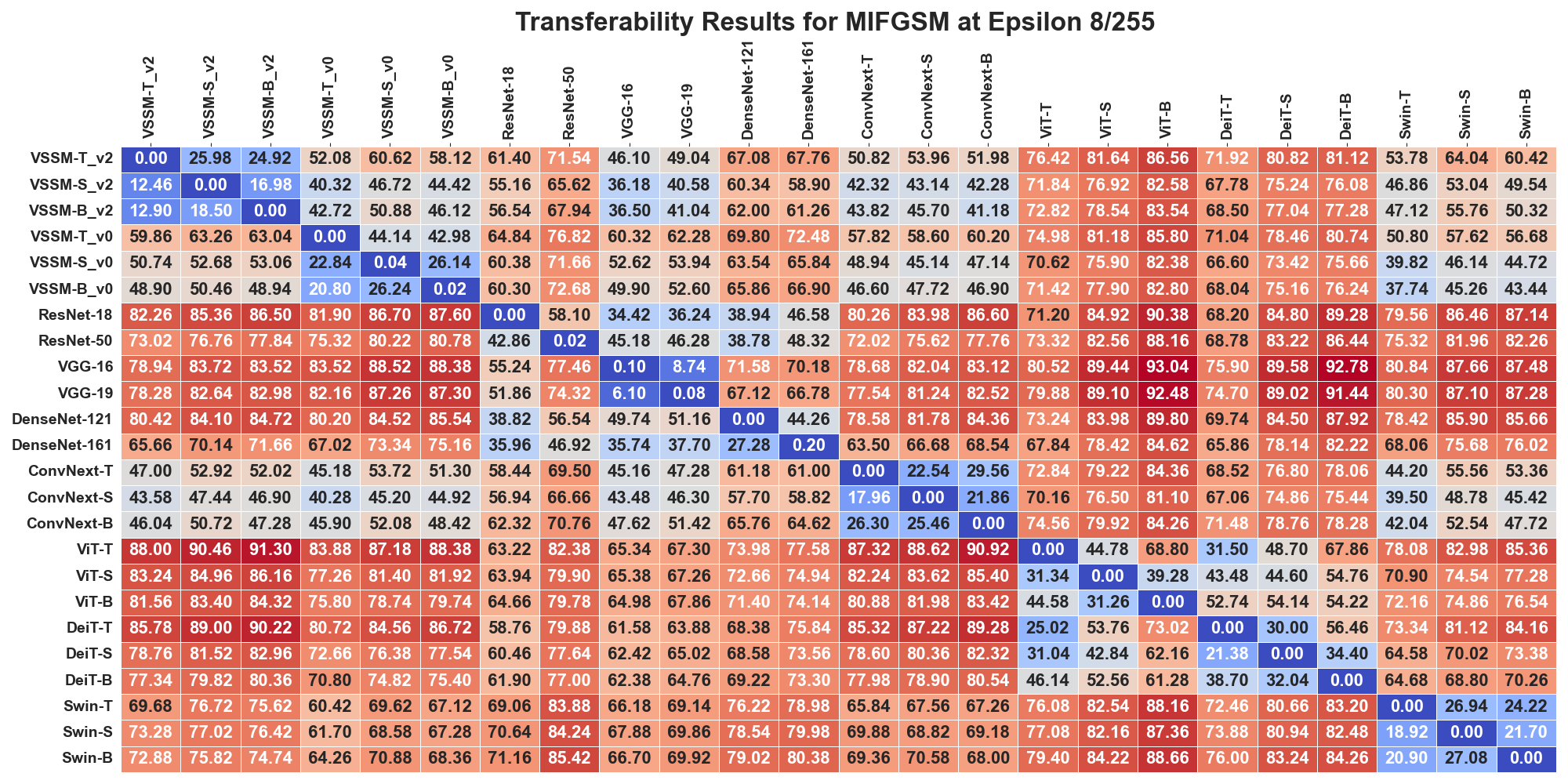}

\caption{Robust accuracy of various architectures under white-box and black-box settings for MIFGSM attack. Adversarial examples are crafted at a perturbation budget $\epsilon=\frac{8}{255}$.}
\label{fig:mifgsm_app}
\end{figure*}

In this section, we expand our analysis on adversarial attacks and their transferability across different models. We include more model families, such as DeiT, DenseNet, and VGG. Figures \ref{fig:fgsm_app}, \ref{fig:pgd_app}, and \ref{fig:mifgsm_app} present the robust accuracy of various models under both white-box and black-box settings for FGSM, PGD, and MIFGSM attacks, respectively. All adversarial examples are crafted with a perturbation budget of $\epsilon=\frac{8}{255}$. For PGD and MIFGSM attacks, we use 20 iterations to craft the adversarial examples. 

In Table \ref{tab:adv_freq_main}, we evaluate the robustness of VMamba and MambaVision against frequency-specific adversarial attacks crafted using Projected Gradient Descent (PGD). These perturbations are constrained to designated frequency bands through a discrete cosine transform (DCT) mask filter~\cite{maiya2021frequency}. Results in Tab.~\ref{tab:adv_freq_main} (\textit{left}) demonstrate that VMamba and MambaVision maintain robustness above 90$\%$ for low-frequency perturbations up to $\epsilon=16$, indicating strong resilience similar to the ConvNext and Swin transformer counterparts. ViT models exhibit the most performance drop under low-frequency adversarial attacks. For high-frequency adversarial attacks (Tab.~\ref{tab:adv_freq_main} (\textit{middle})), the robustness of all models decreases more rapidly with increasing perturbation strength, although ViT-based models show the highest robustness.  Finally, for standard attacks where the complete frequency range is used to generate adversarial examples, VSSM models display higher robustness compared to other models, including ConvNext, ViT, and Swin.

\begin{table*}[h]
\centering
\caption{\small  Top-1 classification accuracy reported random patch drop occlusion using $16\times16$, $8\times8$, $4\times4$ and $1\times1$ patch sizes.}
\label{tab:randdrop_all_patch_sizes}
\setlength{\tabcolsep}{2.0pt}
\resizebox{\linewidth}{!}{
\begin{tabular}{|c c c c c c c c c c c c c c c c|}
\hline
\rowcolor{LightCyan}
\rotatebox{0}{\scriptsize ResNet-50} & \rotatebox{0}{\scriptsize ConvNext-T} & \rotatebox{0}{\scriptsize ConvNext-S} & \rotatebox{0}{\scriptsize ConvNext-B} & \rotatebox{0}{\scriptsize ViT-T} & \rotatebox{0}{\scriptsize ViT-S} & \rotatebox{0}{\scriptsize ViT-B} & \rotatebox{0}{\scriptsize VMamba-T} & \rotatebox{0}{\scriptsize VMamba-S} & \rotatebox{0}{\scriptsize VMamba-B}  & \rotatebox{0}{\scriptsize MambaVision-T}  & \rotatebox{0}{\scriptsize MambaVision-S}  & \rotatebox{0}{\scriptsize MambaVision-B}  & \rotatebox{0}{\scriptsize Swin-T} & \rotatebox{0}{\scriptsize Swin-S} & \rotatebox{0}{\scriptsize Swin-B} \\
\hline
\rowcolor{gray!5}
\multicolumn{16}{|c|}{Patch Size $16\times16$ \emph{(Percentage of patch drop increasing from top to bottom (10\% to 90\%))}} \\
\hline
\heatmapcolor{96.70} & \heatmapcolor{97.24}  & \heatmapcolor{97.78}  & \heatmapcolor{97.84}  & \heatmapcolor{92.30}  & \heatmapcolor{96.08}  & \heatmapcolor{97.54}  & \heatmapcolor{97.38}  & \heatmapcolor{97.94}  & \heatmapcolor{97.96}  & \heatmapcolor{97.36} & \heatmapcolor{97.82} & \heatmapcolor{97.60} & \heatmapcolor{97.24}  & \heatmapcolor{97.60} & \heatmapcolor{97.60}  \\
\hline

\heatmapcolor{75.27}  & \heatmapcolor{96.49}  & \heatmapcolor{97.19}  & \heatmapcolor{97.37}  & \heatmapcolor{90.83}  & \heatmapcolor{95.39}  & \heatmapcolor{96.85}  & \heatmapcolor{96.49}  & \heatmapcolor{96.61}  & \heatmapcolor{97.25}  & \heatmapcolor{96.24} & \heatmapcolor{96.80} & \heatmapcolor{97.09} & \heatmapcolor{96.76}  & \heatmapcolor{97.38}  & \heatmapcolor{97.32}   \\
\hline

\heatmapcolor{39.93}  & \heatmapcolor{94.63}  & \heatmapcolor{95.27}  & \heatmapcolor{96.48}  & \heatmapcolor{88.09}  & \heatmapcolor{94.29}  & \heatmapcolor{96.27}  & \heatmapcolor{95.16}  & \heatmapcolor{92.97}  & \heatmapcolor{96.25}  & \heatmapcolor{92.91} & \heatmapcolor{94.67} & \heatmapcolor{95.74} & \heatmapcolor{96.11}  & \heatmapcolor{96.84}  & \heatmapcolor{96.79}   \\
\hline

\heatmapcolor{17.91}  & \heatmapcolor{89.99}  & \heatmapcolor{91.12}  & \heatmapcolor{95.29}  & \heatmapcolor{85.26}  & \heatmapcolor{92.35}  & \heatmapcolor{95.08}  & \heatmapcolor{93.45}  & \heatmapcolor{89.74}  & \heatmapcolor{95.21}   & \heatmapcolor{87.14} & \heatmapcolor{91.19} & \heatmapcolor{93.25}& \heatmapcolor{94.88}  & \heatmapcolor{95.88}  & \heatmapcolor{96.17}  \\
\hline

\heatmapcolor{6.73}  & \heatmapcolor{81.43} & \heatmapcolor{84.63} & \heatmapcolor{93.03}  & \heatmapcolor{80.08}  & \heatmapcolor{90.15}  & \heatmapcolor{92.78}  & \heatmapcolor{90.52}  & \heatmapcolor{84.82}  & \heatmapcolor{93.46} & \heatmapcolor{77.02} & \heatmapcolor{84.56} & \heatmapcolor{88.05} & \heatmapcolor{93.38}  & \heatmapcolor{94.35} & \heatmapcolor{95.03}    \\
\hline
\heatmapcolor{2.43}  & \heatmapcolor{70.07} & \heatmapcolor{74.44} & \heatmapcolor{88.76}  & \heatmapcolor{72.49}  & \heatmapcolor{85.10}  & \heatmapcolor{89.21}  & \heatmapcolor{86.52}  & \heatmapcolor{78.41}  & \heatmapcolor{90.89}  & \heatmapcolor{61.79} & \heatmapcolor{76.07} & \heatmapcolor{78.92}& \heatmapcolor{91.05}  & \heatmapcolor{92.21} & \heatmapcolor{93.25}    \\
\hline
\heatmapcolor{1.05}  & \heatmapcolor{57.59} & \heatmapcolor{60.15} & \heatmapcolor{82.35}  & \heatmapcolor{61.34}  & \heatmapcolor{76.56}  & \heatmapcolor{82.08}  & \heatmapcolor{80.52}  & \heatmapcolor{68.40}  & \heatmapcolor{87.03}  & \heatmapcolor{42.16} & \heatmapcolor{62.44} & \heatmapcolor{63.89}& \heatmapcolor{87.96}  & \heatmapcolor{87.84} & \heatmapcolor{90.43}    \\
\hline
\heatmapcolor{0.56}  & \heatmapcolor{44.67} & \heatmapcolor{44.71} & \heatmapcolor{71.25}  & \heatmapcolor{45.63}  & \heatmapcolor{62.63}  & \heatmapcolor{70.25}  & \heatmapcolor{70.39}  & \heatmapcolor{52.72}  & \heatmapcolor{79.96}  & \heatmapcolor{20.73} & \heatmapcolor{42.16} & \heatmapcolor{43.33}& \heatmapcolor{80.65}  & \heatmapcolor{79.71} & \heatmapcolor{84.70}    \\
\hline
\heatmapcolor{0.45}  & \heatmapcolor{31.29} & \heatmapcolor{28.82} & \heatmapcolor{57.71}  & \heatmapcolor{25.86}  & \heatmapcolor{41.73}  & \heatmapcolor{50.08}  & \heatmapcolor{56.23}  & \heatmapcolor{34.51}  & \heatmapcolor{67.56}  & \heatmapcolor{5.62} & \heatmapcolor{21.07} & \heatmapcolor{22.21}& \heatmapcolor{70.37}  & \heatmapcolor{66.60} & \heatmapcolor{74.38}    \\
\hline
\heatmapcolor{0.43}  & \heatmapcolor{16.73} & \heatmapcolor{14.98} & \heatmapcolor{33.67}  & \heatmapcolor{7.85}  & \heatmapcolor{15.86}  & \heatmapcolor{19.68}  & \heatmapcolor{34.83}  & \heatmapcolor{16.82}  & \heatmapcolor{41.55} & \heatmapcolor{1.95} & \heatmapcolor{7.08} & \heatmapcolor{11.08} & \heatmapcolor{47.16}  & \heatmapcolor{47.85} & \heatmapcolor{53.54}    \\
\hline

\rowcolor{gray!5}
\multicolumn{16}{|c|}{Patch Size $8\times8$ \emph{(Percentage of patch drop increasing from top to bottom (10\% to 90\%))}} \\
\hline
\heatmapcolor{96.70} & \heatmapcolor{97.24}  & \heatmapcolor{97.78}  & \heatmapcolor{97.84}  & \heatmapcolor{92.32}  & \heatmapcolor{96.08}  & \heatmapcolor{97.54}  & \heatmapcolor{97.38}  & \heatmapcolor{97.94}  & \heatmapcolor{97.96}  & \heatmapcolor{97.36} & \heatmapcolor{97.82} & \heatmapcolor{97.60} & \heatmapcolor{97.24}  & \heatmapcolor{97.60} & \heatmapcolor{97.60}  \\
\hline

\heatmapcolor{44.91}  & \heatmapcolor{86.69}  & \heatmapcolor{91.39}  & \heatmapcolor{95.63}  & \heatmapcolor{70.18}  & \heatmapcolor{88.90}  & \heatmapcolor{94.23}  & \heatmapcolor{87.86}  & \heatmapcolor{85.93}  & \heatmapcolor{90.20}  & \heatmapcolor{89.57} & \heatmapcolor{91.25} & \heatmapcolor{92.32} & \heatmapcolor{96.44}  & \heatmapcolor{96.77}  & \heatmapcolor{96.76}   \\
\hline

\heatmapcolor{12.44}  & \heatmapcolor{68.38}  & \heatmapcolor{81.13}  & \heatmapcolor{90.83}  & \heatmapcolor{42.19}  & \heatmapcolor{79.72}  & \heatmapcolor{88.37}  & \heatmapcolor{79.91}  & \heatmapcolor{78.23}  & \heatmapcolor{84.43}  & \heatmapcolor{73.34} & \heatmapcolor{79.03} & \heatmapcolor{84.32} & \heatmapcolor{95.04}  & \heatmapcolor{94.87}  & \heatmapcolor{95.84}   \\
\hline

\heatmapcolor{3.79}  & \heatmapcolor{55.12}  & \heatmapcolor{68.35}  & \heatmapcolor{84.08}  & \heatmapcolor{16.91}  & \heatmapcolor{65.39}  & \heatmapcolor{76.63}  & \heatmapcolor{70.40}  & \heatmapcolor{70.95}  & \heatmapcolor{78.47}  & \heatmapcolor{51.09} & \heatmapcolor{60.21} & \heatmapcolor{71.70} & \heatmapcolor{92.87}  & \heatmapcolor{92.08}  & \heatmapcolor{94.49}  \\
\hline

\heatmapcolor{1.35}  & \heatmapcolor{39.05} & \heatmapcolor{54.58} & \heatmapcolor{73.51}  & \heatmapcolor{4.59}  & \heatmapcolor{ 46.17}  & \heatmapcolor{60.12}  & \heatmapcolor{57.34}  & \heatmapcolor{59.09}  & \heatmapcolor{70.04} & \heatmapcolor{29.64} & \heatmapcolor{39.81} & \heatmapcolor{55.01} & \heatmapcolor{90.13}  & \heatmapcolor{88.21} & \heatmapcolor{92.40}    \\
\hline
\heatmapcolor{0.56}  & \heatmapcolor{23.89} & \heatmapcolor{37.94} & \heatmapcolor{58.05}  & \heatmapcolor{1.25}  & \heatmapcolor{25.91}  & \heatmapcolor{39.97}  & \heatmapcolor{43.09}  & \heatmapcolor{44.08}  & \heatmapcolor{58.29}  & \heatmapcolor{13.73} & \heatmapcolor{23.86} & \heatmapcolor{34.67}& \heatmapcolor{85.40}  & \heatmapcolor{81.87} & \heatmapcolor{88.36}    \\
\hline
\heatmapcolor{0.33}  & \heatmapcolor{13.33} & \heatmapcolor{21.97} & \heatmapcolor{40.37}  & \heatmapcolor{0.45}  & \heatmapcolor{11.03}  & \heatmapcolor{21.56}  & \heatmapcolor{28.25}  & \heatmapcolor{27.85}  & \heatmapcolor{43.00}  & \heatmapcolor{4.90} & \heatmapcolor{11.86} & \heatmapcolor{16.70}& \heatmapcolor{78.76}  & \heatmapcolor{71.63} & \heatmapcolor{81.85}    \\
\hline
\heatmapcolor{0.21}  & \heatmapcolor{5.95} & \heatmapcolor{9.85} & \heatmapcolor{21.51}  & \heatmapcolor{0.21}  & \heatmapcolor{3.78}  & \heatmapcolor{8.85}  & \heatmapcolor{14.69}  & \heatmapcolor{12.45}  & \heatmapcolor{25.75}  & \heatmapcolor{1.51} & \heatmapcolor{4.59} & \heatmapcolor{5.90} & \heatmapcolor{68.40}  & \heatmapcolor{53.81} & \heatmapcolor{70.03}    \\
\hline
\heatmapcolor{0.24}  & \heatmapcolor{2.08} & \heatmapcolor{2.31} & \heatmapcolor{7.85}  & \heatmapcolor{0.14}  & \heatmapcolor{1.02}  & \heatmapcolor{ 2.49}  & \heatmapcolor{5.11}  & \heatmapcolor{2.45}  & \heatmapcolor{10.07} & \heatmapcolor{0.50} & \heatmapcolor{1.06} & \heatmapcolor{1.16} & \heatmapcolor{52.41}  & \heatmapcolor{29.58} & \heatmapcolor{49.35}    \\
\hline
\heatmapcolor{0.25}  & \heatmapcolor{0.46} & \heatmapcolor{0.56} & \heatmapcolor{1.33}  & \heatmapcolor{0.16}  & \heatmapcolor{0.39}  & \heatmapcolor{0.59}  & \heatmapcolor{0.75}  & \heatmapcolor{0.43}  & \heatmapcolor{1.65} & \heatmapcolor{0.22} & \heatmapcolor{0.31} & \heatmapcolor{0.21} & \heatmapcolor{28.46}  & \heatmapcolor{8.32} & \heatmapcolor{23.67}    \\
\hline

\rowcolor{gray!5}
\multicolumn{16}{|c|}{Patch Size $4\times4$ \emph{(Percentage of patch drop increasing from top to bottom (10\% to 90\%))}} \\
\hline
\heatmapcolor{96.70} & \heatmapcolor{97.24}  & \heatmapcolor{97.78}  & \heatmapcolor{97.84}  & \heatmapcolor{92.30}  & \heatmapcolor{96.08}  & \heatmapcolor{97.54}  & \heatmapcolor{97.38}  & \heatmapcolor{97.94}  & \heatmapcolor{97.96}  & \heatmapcolor{97.36} & \heatmapcolor{97.82} & \heatmapcolor{97.60} & \heatmapcolor{97.24}  & \heatmapcolor{97.60} & \heatmapcolor{97.60}  \\
\hline

\heatmapcolor{30.11}  & \heatmapcolor{84.75}  & \heatmapcolor{88.74}  & \heatmapcolor{90.51}  & \heatmapcolor{29.62}  & \heatmapcolor{82.23}  & \heatmapcolor{90.04}  & \heatmapcolor{86.34}  & \heatmapcolor{87.17}  & \heatmapcolor{90.48}  & \heatmapcolor{80.16} & \heatmapcolor{85.39} & \heatmapcolor{89.79} & \heatmapcolor{92.51}  & \heatmapcolor{92.97}  & \heatmapcolor{94.68}   \\
\hline

\heatmapcolor{12.02}  & \heatmapcolor{64.23}  & \heatmapcolor{77.49}  & \heatmapcolor{81.82}  & \heatmapcolor{9.25}  & \heatmapcolor{65.28}  & \heatmapcolor{77.00}  & \heatmapcolor{72.09}  & \heatmapcolor{75.43}  & \heatmapcolor{77.77}  & \heatmapcolor{57.80} & \heatmapcolor{57.14} & \heatmapcolor{71.68} & \heatmapcolor{84.20}  & \heatmapcolor{86.17}  & \heatmapcolor{90.71}   \\
\hline

\heatmapcolor{6.49}  & \heatmapcolor{33.19}  & \heatmapcolor{47.91}  & \heatmapcolor{70.73}  & \heatmapcolor{3.63}  & \heatmapcolor{43.92}  & \heatmapcolor{56.17}  & \heatmapcolor{51.01}  & \heatmapcolor{55.23}  & \heatmapcolor{53.83} & \heatmapcolor{29.05} & \heatmapcolor{24.13} & \heatmapcolor{44.27}  & \heatmapcolor{72.07}  & \heatmapcolor{77.69}  & \heatmapcolor{85.35}  \\
\hline

\heatmapcolor{2.91}  & \heatmapcolor{13.34} & \heatmapcolor{19.71} & \heatmapcolor{54.95}  & \heatmapcolor{1.58}  & \heatmapcolor{21.73}  & \heatmapcolor{32.19}  & \heatmapcolor{27.17}  & \heatmapcolor{33.21}  & \heatmapcolor{29.41} & \heatmapcolor{11.75} & \heatmapcolor{8.59} & \heatmapcolor{23.15} & \heatmapcolor{57.71}  & \heatmapcolor{65.13} & \heatmapcolor{78.48}    \\
\hline
\heatmapcolor{1.46}  & \heatmapcolor{5.17} & \heatmapcolor{6.67} & \heatmapcolor{34.97}  & \heatmapcolor{0.70}  & \heatmapcolor{8.96}  & \heatmapcolor{15.24}  & \heatmapcolor{10.51}  & \heatmapcolor{15.82}  & \heatmapcolor{11.57}  & \heatmapcolor{4.00} & \heatmapcolor{2.84} & \heatmapcolor{10.25}& \heatmapcolor{41.57}  & \heatmapcolor{47.95} & \heatmapcolor{69.11}    \\
\hline
\heatmapcolor{0.75}  & \heatmapcolor{2.07} & \heatmapcolor{2.23} & \heatmapcolor{17.37}  & \heatmapcolor{0.41}  & \heatmapcolor{3.69}  & \heatmapcolor{6.01}  & \heatmapcolor{3.07}  & \heatmapcolor{5.41}  & \heatmapcolor{3.40} & \heatmapcolor{1.41} & \heatmapcolor{0.90} & \heatmapcolor{3.89} & \heatmapcolor{26.49}  & \heatmapcolor{29.31} & \heatmapcolor{54.59}    \\
\hline
\heatmapcolor{0.40}  & \heatmapcolor{0.85} & \heatmapcolor{0.90} & \heatmapcolor{6.21}  & \heatmapcolor{0.24}  & \heatmapcolor{1.45}  & \heatmapcolor{1.91}  & \heatmapcolor{0.83}  & \heatmapcolor{1.28}  & \heatmapcolor{0.63} & \heatmapcolor{0.50} & \heatmapcolor{0.39} & \heatmapcolor{1.05} & \heatmapcolor{12.17}  & \heatmapcolor{12.45} & \heatmapcolor{35.95}    \\
\hline
\heatmapcolor{0.21}  & \heatmapcolor{0.33} & \heatmapcolor{0.44} & \heatmapcolor{1.70}  & \heatmapcolor{0.17}  & \heatmapcolor{0.55}  & \heatmapcolor{0.71}  & \heatmapcolor{0.23}  & \heatmapcolor{0.30}  & \heatmapcolor{0.24}  & \heatmapcolor{0.29} & \heatmapcolor{0.17} & \heatmapcolor{0.23}& \heatmapcolor{3.48}  & \heatmapcolor{2.47} & \heatmapcolor{14.47}    \\
\hline
\heatmapcolor{0.13}  & \heatmapcolor{0.20} & \heatmapcolor{0.25} & \heatmapcolor{0.40}  & \heatmapcolor{0.19}  & \heatmapcolor{0.24}  & \heatmapcolor{0.25}   & \heatmapcolor{0.19}  & \heatmapcolor{0.21} & \heatmapcolor{0.22}  & \heatmapcolor{0.19} & \heatmapcolor{0.13} & \heatmapcolor{0.14} & \heatmapcolor{0.64} & \heatmapcolor{0.41}  & \heatmapcolor{1.55}   \\
\hline

\rowcolor{gray!5}
\multicolumn{16}{|c|}{Patch Size $1\times1$ \emph{(Percentage of patch drop increasing from top to bottom (10\% to 90\%))}} \\
\hline
\heatmapcolor{96.70} & \heatmapcolor{97.24}  & \heatmapcolor{97.78}  & \heatmapcolor{97.84}  & \heatmapcolor{92.32}  & \heatmapcolor{96.08}  & \heatmapcolor{97.54}  & \heatmapcolor{97.38}  & \heatmapcolor{97.94}  & \heatmapcolor{97.96}  & \heatmapcolor{97.36} & \heatmapcolor{97.82} & \heatmapcolor{97.60}  & \heatmapcolor{97.24}  & \heatmapcolor{97.60} & \heatmapcolor{97.60}  \\
\hline

\heatmapcolor{54.40}  & \heatmapcolor{83.43}  & \heatmapcolor{87.12}  & \heatmapcolor{89.68}  & \heatmapcolor{47.56}  & \heatmapcolor{76.30}  & \heatmapcolor{85.64}  & \heatmapcolor{85.87}  & \heatmapcolor{89.55}  & \heatmapcolor{89.73}  & \heatmapcolor{80.38} & \heatmapcolor{86.96} & \heatmapcolor{90.11} & \heatmapcolor{74.40}  & \heatmapcolor{85.10}  & \heatmapcolor{85.14}   \\
\hline

\heatmapcolor{37.39}  & \heatmapcolor{69.17}  & \heatmapcolor{76.25}  & \heatmapcolor{79.47}  & \heatmapcolor{27.12}  & \heatmapcolor{61.71}  & \heatmapcolor{75.93}  & \heatmapcolor{71.95}  & \heatmapcolor{77.99}  & \heatmapcolor{79.11}  & \heatmapcolor{57.76} & \heatmapcolor{69.25} & \heatmapcolor{76.01} & \heatmapcolor{48.34}  & \heatmapcolor{63.30}  & \heatmapcolor{62.51}   \\
\hline

\heatmapcolor{26.21}  & \heatmapcolor{53.81}  & \heatmapcolor{62.42}  & \heatmapcolor{68.38}  & \heatmapcolor{15.67}  & \heatmapcolor{47.69}  & \heatmapcolor{65.85}  & \heatmapcolor{56.63}  & \heatmapcolor{64.54}  & \heatmapcolor{66.81}  & \heatmapcolor{39.91} & \heatmapcolor{50.91} & \heatmapcolor{57.74} & \heatmapcolor{33.44}  & \heatmapcolor{45.94}  & \heatmapcolor{49.32}  \\
\hline

\heatmapcolor{17.80}  & \heatmapcolor{38.14} & \heatmapcolor{45.85} & \heatmapcolor{58.50}  & \heatmapcolor{9.02}  & \heatmapcolor{ 34.97}  & \heatmapcolor{55.16}  & \heatmapcolor{43.11}  & \heatmapcolor{52.29}  & \heatmapcolor{54.38}  & \heatmapcolor{27.55} & \heatmapcolor{35.95} & \heatmapcolor{40.89}& \heatmapcolor{25.06}  & \heatmapcolor{32.73} & \heatmapcolor{39.90}    \\
\hline
\heatmapcolor{11.85}  & \heatmapcolor{26.13} & \heatmapcolor{31.79} & \heatmapcolor{47.28}  & \heatmapcolor{5.11}  & \heatmapcolor{23.24}  & \heatmapcolor{43.27}  & \heatmapcolor{32.61}  & \heatmapcolor{41.20}  & \heatmapcolor{42.08} & \heatmapcolor{18.56} & \heatmapcolor{24.83} & \heatmapcolor{27.92} & \heatmapcolor{18.84}  & \heatmapcolor{23.16} & \heatmapcolor{31.09}    \\
\hline
\heatmapcolor{7.19}  & \heatmapcolor{17.81} & \heatmapcolor{21.99} & \heatmapcolor{35.81}  & \heatmapcolor{2.78}  & \heatmapcolor{14.30}  & \heatmapcolor{31.58}  & \heatmapcolor{23.08}  & \heatmapcolor{30.97}  & \heatmapcolor{29.57} & \heatmapcolor{11.42} & \heatmapcolor{16.06} & \heatmapcolor{19.34} & \heatmapcolor{12.75}  & \heatmapcolor{15.41} & \heatmapcolor{22.05}    \\
\hline
\heatmapcolor{3.94}  & \heatmapcolor{11.18} & \heatmapcolor{15.01} & \heatmapcolor{21.95}  & \heatmapcolor{1.39}  & \heatmapcolor{7.82}  & \heatmapcolor{19.59}  & \heatmapcolor{13.39}  & \heatmapcolor{21.11}  & \heatmapcolor{17.70}  & \heatmapcolor{5.97} & \heatmapcolor{9.18} & \heatmapcolor{11.85}& \heatmapcolor{6.69}  & \heatmapcolor{9.58} & \heatmapcolor{11.99}    \\
\hline
\heatmapcolor{1.79}  & \heatmapcolor{4.47} & \heatmapcolor{8.49} & \heatmapcolor{9.15}  & \heatmapcolor{0.61}  & \heatmapcolor{3.51}  & \heatmapcolor{10.46}  & \heatmapcolor{4.65}  & \heatmapcolor{9.64}  & \heatmapcolor{6.79} & \heatmapcolor{2.19} & \heatmapcolor{3.19} & \heatmapcolor{4.87} & \heatmapcolor{2.15}  & \heatmapcolor{4.15} & \heatmapcolor{4.05}    \\
\hline
\heatmapcolor{0.62}  & \heatmapcolor{0.99} & \heatmapcolor{2.91} & \heatmapcolor{2.62}  & \heatmapcolor{0.29}  & \heatmapcolor{1.23}  & \heatmapcolor{3.13}  & \heatmapcolor{0.79}  & \heatmapcolor{1.47}  & \heatmapcolor{1.23}  & \heatmapcolor{0.71} & \heatmapcolor{0.63} & \heatmapcolor{1.08}& \heatmapcolor{0.61}  & \heatmapcolor{1.08} & \heatmapcolor{0.95}    \\
\hline

\end{tabular}
}
\end{table*}


\begin{table*}[t]
\centering
\caption{\small  Corruption Error (CE) of various architectures on ImageNet-C corruptions over multiple intensity levels. Results are relative to the CE on ResNet-50 and are evaluated on 5000 ImageNet validation set images.}
\label{tab:common_corrup_main}
\setlength{\tabcolsep}{2.0pt}
\resizebox{\linewidth}{!}{
\begin{tabular}{|l|c c c c c c c c c c c c c c c c|}

\hline
\rowcolor{LightCyan}
\textbf{Corruptions} $\downarrow$ & \rotatebox{0}{\scriptsize \textbf{ResNet-50}} & \rotatebox{0}{\scriptsize \textbf{ConvNext-T}} & \rotatebox{0}{\scriptsize \textbf{ConvNext-S}} & \rotatebox{0}{\scriptsize \textbf{ConvNext-B}} & \rotatebox{0}{\scriptsize \textbf{ViT-T}} & \rotatebox{0}{\scriptsize \textbf{ViT-S}} & \rotatebox{0}{\scriptsize \textbf{ViT-B}} & \rotatebox{0}{\scriptsize \textbf{VMamba-T}} & \rotatebox{0}{\scriptsize \textbf{VMamba-S}} & \rotatebox{0}{\scriptsize \textbf{VMamba-B}} & \rotatebox{0}{\scriptsize \textbf{MambaVision-T}}& \rotatebox{0}{\scriptsize \textbf{MambaVision-S}}& \rotatebox{0}{\scriptsize \textbf{MambaVision-B}}& \rotatebox{0}{\scriptsize \textbf{Swin-T}} & \rotatebox{0}{\scriptsize \textbf{Swin-S}} & \rotatebox{0}{\scriptsize \textbf{Swin-B}} \\
\hline

Brightness & \heatmapcolors{100.00} & \heatmapcolors{68.20}  & \heatmapcolors{58.63}  & \heatmapcolors{52.15}  & \heatmapcolors{223.14}  & \heatmapcolors{101.09}  & \heatmapcolors{41.67}  & \heatmapcolors{61.60}  & \heatmapcolors{53.06}  & \heatmapcolors{52.94} & \heatmapcolors{64.39}& \heatmapcolors{53.73}& \heatmapcolors{51.30} & \heatmapcolors{75.89}  & \heatmapcolors{65.35} & \heatmapcolors{63.54}  \\
\hline

Contrast & \heatmapcolors{100.00}  & \heatmapcolors{40.14}  & \heatmapcolors{30.29}  & \heatmapcolors{28.96}  & \heatmapcolors{426.29}  & \heatmapcolors{186.49}  & \heatmapcolors{103.16}  & \heatmapcolors{38.17}  & \heatmapcolors{30.02}  & \heatmapcolors{27.70} & \heatmapcolors{52.05}& \heatmapcolors{34.93}& \heatmapcolors{29.72} & \heatmapcolors{44.41}  & \heatmapcolors{42.09}  & \heatmapcolors{41.70}   \\
\hline

Defocus Blur & \heatmapcolors{100.00}  & \heatmapcolors{85.70}  & \heatmapcolors{77.82}  & \heatmapcolors{73.87}  & \heatmapcolors{103.27}  & \heatmapcolors{70.64}  & \heatmapcolors{42.81}  & \heatmapcolors{84.96}  & \heatmapcolors{72.32}  & \heatmapcolors{73.00} & \heatmapcolors{77.80}& \heatmapcolors{71.71}& \heatmapcolors{67.04} & \heatmapcolors{87.50}  & \heatmapcolors{66.82}  & \heatmapcolors{81.49}   \\
\hline

Elastic Transform & \heatmapcolors{100.00}  & \heatmapcolors{81.11}  & \heatmapcolors{72.18}  & \heatmapcolors{69.44}  & \heatmapcolors{70.34}  & \heatmapcolors{55.69}  & \heatmapcolors{41.43}  & \heatmapcolors{82.78}  & \heatmapcolors{69.89}  & \heatmapcolors{72.21} & \heatmapcolors{78.96}& \heatmapcolors{79.82}& \heatmapcolors{75.06} & \heatmapcolors{86.36}  & \heatmapcolors{66.53}  & \heatmapcolors{76.78}  \\
\hline

Fog & \heatmapcolors{100.00}  & \heatmapcolors{113.68} & \heatmapcolors{86.18} & \heatmapcolors{91.19}  & \heatmapcolors{312.26}  & \heatmapcolors{119.54}  & \heatmapcolors{68.79}  & \heatmapcolors{87.47}  & \heatmapcolors{64.93}  & \heatmapcolors{55.88}& \heatmapcolors{62.13}& \heatmapcolors{46.48}& \heatmapcolors{39.35} & \heatmapcolors{84.32}  & \heatmapcolors{87.24} & \heatmapcolors{68.61}    \\
\hline
Frost & \heatmapcolors{100.00}  & \heatmapcolors{64.12} & \heatmapcolors{57.09} & \heatmapcolors{50.75}  & \heatmapcolors{146.53}  & \heatmapcolors{75.52}  & \heatmapcolors{27.19}  & \heatmapcolors{42.05}  & \heatmapcolors{36.58}  & \heatmapcolors{30.79}& \heatmapcolors{57.65}& \heatmapcolors{49.73}& \heatmapcolors{40.25} & \heatmapcolors{47.77}  & \heatmapcolors{40.72} & \heatmapcolors{38.63}    \\
\hline
Gaussian Blur & \heatmapcolors{100.00}  & \heatmapcolors{86.69} & \heatmapcolors{84.58} & \heatmapcolors{77.75}  & \heatmapcolors{107.84}  & \heatmapcolors{82.50}  & \heatmapcolors{55.22}  & \heatmapcolors{88.91}  & \heatmapcolors{77.05}  & \heatmapcolors{77.21} & \heatmapcolors{79.64}& \heatmapcolors{74.80}& \heatmapcolors{70.72} & \heatmapcolors{90.59}  & \heatmapcolors{81.36} & \heatmapcolors{85.37}    \\
\hline
Gaussian Noise & \heatmapcolors{100.00}  & \heatmapcolors{52.01} & \heatmapcolors{51.36} & \heatmapcolors{56.02}  & \heatmapcolors{130.74}  & \heatmapcolors{66.75}  & \heatmapcolors{30.16}  & \heatmapcolors{56.81}  & \heatmapcolors{38.29}  & \heatmapcolors{52.62}& \heatmapcolors{115.29}& \heatmapcolors{110.36}& \heatmapcolors{96.79} & \heatmapcolors{73.31}  & \heatmapcolors{59.86} & \heatmapcolors{54.49}    \\
\hline
Glass Blur & \heatmapcolors{100.00}  & \heatmapcolors{89.71} & \heatmapcolors{85.46} & \heatmapcolors{81.43}  & \heatmapcolors{98.30}  & \heatmapcolors{84.73}  & \heatmapcolors{66.52}  & \heatmapcolors{93.59}  & \heatmapcolors{86.00}  & \heatmapcolors{87.28}& \heatmapcolors{90.58}& \heatmapcolors{90.13}& \heatmapcolors{89.29} & \heatmapcolors{96.53}  & \heatmapcolors{86.27} & \heatmapcolors{90.24}    \\
\hline
Impulse Noise & \heatmapcolors{100.00}  & \heatmapcolors{47.18} & \heatmapcolors{44.71} & \heatmapcolors{45.78}  & \heatmapcolors{122.95}  & \heatmapcolors{61.19}  & \heatmapcolors{28.27}  & \heatmapcolors{44.55}  & \heatmapcolors{31.93}  & \heatmapcolors{39.72} & \heatmapcolors{92.95}& \heatmapcolors{84.21}& \heatmapcolors{75.73}& \heatmapcolors{63.83}  & \heatmapcolors{48.63} & \heatmapcolors{48.68}    \\
\hline
JPEG Compression & \heatmapcolors{100.00}  & \heatmapcolors{69.79} & \heatmapcolors{59.62} & \heatmapcolors{60.96}  & \heatmapcolors{131.33}  & \heatmapcolors{78.90}  & \heatmapcolors{43.49}  & \heatmapcolors{82.29}  & \heatmapcolors{65.82}  & \heatmapcolors{63.30} & \heatmapcolors{74.33}& \heatmapcolors{69.63}& \heatmapcolors{62.72}& \heatmapcolors{144.27}  & \heatmapcolors{88.13} & \heatmapcolors{108.22}    \\
\hline
Motion Blur & \heatmapcolors{100.00}  & \heatmapcolors{65.05} & \heatmapcolors{56.10} & \heatmapcolors{51.42}  & \heatmapcolors{106.50}  & \heatmapcolors{58.86}  & \heatmapcolors{33.26}  & \heatmapcolors{65.84}  & \heatmapcolors{54.55}  & \heatmapcolors{54.13}& \heatmapcolors{63.42}& \heatmapcolors{58.58}& \heatmapcolors{53.44} & \heatmapcolors{80.88}  & \heatmapcolors{50.12} & \heatmapcolors{62.42}    \\
\hline
Pixelate & \heatmapcolors{100.00}  & \heatmapcolors{82.35} & \heatmapcolors{71.52} & \heatmapcolors{66.00}  & \heatmapcolors{52.11}  & \heatmapcolors{27.31}  & \heatmapcolors{15.88}  & \heatmapcolors{80.34}  & \heatmapcolors{66.30}  & \heatmapcolors{65.02} & \heatmapcolors{67.42}& \heatmapcolors{67.31}& \heatmapcolors{59.34}& \heatmapcolors{109.24}  & \heatmapcolors{100.70} & \heatmapcolors{101.51}    \\
\hline
Saturate & \heatmapcolors{100.00}  & \heatmapcolors{63.44} & \heatmapcolors{53.74} & \heatmapcolors{47.61}  & \heatmapcolors{250.68}  & \heatmapcolors{116.28}  & \heatmapcolors{48.06}  & \heatmapcolors{57.26}  & \heatmapcolors{47.97}  & \heatmapcolors{47.07} & \heatmapcolors{60.37}& \heatmapcolors{52.12}& \heatmapcolors{48.11}& \heatmapcolors{75.29}  & \heatmapcolors{62.40} & \heatmapcolors{58.34}    \\
\hline
Shot Noise & \heatmapcolors{100.00}  & \heatmapcolors{52.69} & \heatmapcolors{51.85} & \heatmapcolors{56.73}  & \heatmapcolors{128.00}  & \heatmapcolors{66.02}  & \heatmapcolors{30.12}  & \heatmapcolors{56.20}  & \heatmapcolors{38.75}  & \heatmapcolors{51.72} & \heatmapcolors{105.73}& \heatmapcolors{104.55}& \heatmapcolors{91.49}& \heatmapcolors{73.51}  & \heatmapcolors{62.09} & \heatmapcolors{57.99}    \\
\hline
Snow & \heatmapcolors{100.00}  & \heatmapcolors{54.55} & \heatmapcolors{48.57} & \heatmapcolors{43.18}  & \heatmapcolors{134.88}  & \heatmapcolors{61.12}  & \heatmapcolors{26.04}  & \heatmapcolors{54.29}  & \heatmapcolors{46.02}  & \heatmapcolors{42.14} & \heatmapcolors{54.68}& \heatmapcolors{48.88}& \heatmapcolors{42.89}& \heatmapcolors{57.70}  & \heatmapcolors{39.38} & \heatmapcolors{47.00}    \\
\hline
Spatter & \heatmapcolors{100.00}  & \heatmapcolors{39.86} & \heatmapcolors{32.09} & \heatmapcolors{29.41}  & \heatmapcolors{117.51}  & \heatmapcolors{48.16}  & \heatmapcolors{21.03}  & \heatmapcolors{33.10}  & \heatmapcolors{28.59}  & \heatmapcolors{24.33} & \heatmapcolors{39.28}& \heatmapcolors{29.69}& \heatmapcolors{24.92}& \heatmapcolors{32.71}  & \heatmapcolors{23.79} & \heatmapcolors{22.92}    \\
\hline
Speckle Noise & \heatmapcolors{100.00}  & \heatmapcolors{44.78} & \heatmapcolors{40.91} & \heatmapcolors{34.08}  & \heatmapcolors{132.49}  & \heatmapcolors{60.64}  & \heatmapcolors{26.89}  & \heatmapcolors{42.92}  & \heatmapcolors{30.43}  & \heatmapcolors{34.07}& \heatmapcolors{84.20}& \heatmapcolors{79.68}& \heatmapcolors{64.73} & \heatmapcolors{61.92}  & \heatmapcolors{48.85} & \heatmapcolors{48.80}    \\
\hline
Zoom Blur & \heatmapcolors{100.00}  & \heatmapcolors{82.47} & \heatmapcolors{68.22} & \heatmapcolors{64.74}  & \heatmapcolors{125.18}  & \heatmapcolors{84.89}  & \heatmapcolors{52.90}  & \heatmapcolors{82.72}  & \heatmapcolors{70.04}  & \heatmapcolors{70.37} & \heatmapcolors{76.07}& \heatmapcolors{71.43}& \heatmapcolors{65.52}& \heatmapcolors{93.27}  & \heatmapcolors{77.34} & \heatmapcolors{77.96}    \\
\hline
\hline

\rowcolor{gray!15}\textbf{mCE} & \heatmapcolors{100.00} & \heatmapcolors{67.55} & \heatmapcolors{59.52} & \heatmapcolors{56.92} & \heatmapcolors{153.7} & \heatmapcolors{79.28} & \heatmapcolors{42.26} & \heatmapcolors{65.05} & \heatmapcolors{53.08} & \heatmapcolors{53.76} & \heatmapcolors{66.58}& \heatmapcolors{50.50}& \heatmapcolors{49.61} & \heatmapcolors{77.86} & \heatmapcolors{63.04} & \heatmapcolors{64.98} \\
\hline

\end{tabular}
}
\end{table*}

\begin{table*}[t]
\caption{\small Comparison in domain generalization setting. Models trained on ImageNet are evaluated on datasets with domain shifts.} 
\label{tab:robustness_}
\setlength{\tabcolsep}{10.0pt}
\resizebox{\linewidth}{!}{
\begin{tabular}{|l|c|cccc|l|}
\toprule
\rowcolor{LightCyan}
    \small{Model} $\downarrow$ & ImageNet & ImageNetV2 & ImageNet-S & ImageNet-A &  ImageNet-R  & Average \\
    \midrule
     \cellcolor{gray!15}ConvNext-T  &  \cellcolor{gray!5}81.87 & 70.67\dec{11.20} & 33.96\dec{47.91} & 10.48\dec{71.39} & 32.53\dec{49.34} & 36.90\dec{44.97} \\
    \cellcolor{gray!15}ViT-T& \cellcolor{gray!5}75.35 & 63.05\dec{12.30} & 20.88\dec{54.47} & 3.31\dec{72.04} & 20.29\dec{55.06} & 26.88\dec{48.47} \\
    \cellcolor{gray!15}Swin-T  & \cellcolor{gray!5}80.91 & 69.31\dec{11.60} & 29.24\dec{51.67} & 8.93\dec{71.98} & 28.52\dec{52.39} & 34.00\dec{46.91} \\
    \cellcolor{gray!15}VMamba-T\emph{(v0)} & \cellcolor{gray!5}81.92 & 70.82\dec{11.10} & 33.01\dec{48.91} & 13.56\dec{68.36} & 32.11\dec{49.81} & 37.37\dec{44.55} \\
    \cellcolor{gray!15}VMamba-T\emph{(v2)} & \cellcolor{gray!5}82.28 & 71.16\dec{11.12} & 33.99\dec{48.29} & 12.08\dec{70.20} & 32.05\dec{50.23} & 37.32\dec{44.96} \\
        \cellcolor{gray!15}MambaVision-T & \cellcolor{gray!5}82.10 & 71.50\dec{10.60} & 33.49\dec{48.61} & 13.39\dec{68.71} & 32.28\dec{49.82} & 37.66\dec{44.44} \\

    \midrule
     \cellcolor{gray!15}ConvNext-S & \cellcolor{gray!5}82.82 & 72.07\dec{10.75} & 37.16\dec{45.66} & 14.43\dec{68.39} & 35.57\dec{47.25} & 39.81\dec{43.01} \\
    \cellcolor{gray!15}ViT-S & \cellcolor{gray!5}81.40 & 69.98\dec{11.42} & 32.77\dec{48.63} & 13.09\dec{68.31} & 31.14\dec{50.26} & 36.74\dec{44.66} \\
    \cellcolor{gray!15}Swin-S  & \cellcolor{gray!5}82.90 & 71.62\dec{11.28} & 31.97\dec{50.93} & 15.72\dec{67.18} & 31.93\dec{50.97} & 37.81\dec{45.09} \\
    \cellcolor{gray!15}VMamba-S\emph{(v0)} & \cellcolor{gray!5}83.15 & 72.89\dec{10.26} & 38.05\dec{45.10} & 17.28\dec{65.87} & 37.05\dec{46.10} & 41.32\dec{41.83} \\
    \cellcolor{gray!15}VMamba-S\emph{(v2)}  & \cellcolor{gray!5}83.48 & 73.01\dec{10.47} & 36.98\dec{46.50} & 16.45\dec{67.03} & 35.75\dec{47.73} & 40.55\dec{42.93} \\
        \cellcolor{gray!15}MambaVision-S & \cellcolor{gray!5}83.22 & 72.62\dec{10.60} & 35.53\dec{47.69} & 15.97\dec{67.25} & 33.96\dec{49.26} & 39.52\dec{43.70} \\

    \midrule
    \cellcolor{gray!15}ConvNext-B & \cellcolor{gray!5}83.75 & 73.68\dec{10.07} & 38.23\dec{45.52} & 18.16\dec{65.59} & 36.66\dec{47.09} & 41.68\dec{42.07} \\
    \cellcolor{gray!15}ViT-B & \cellcolor{gray!5}84.40 & 73.84\dec{10.56} & 43.01\dec{41.39} & 24.09\dec{60.31} & 41.03\dec{43.37} & 45.49\dec{38.91} \\
          \cellcolor{gray!15}Swin-B  & \cellcolor{gray!5}83.08 & 72.09\dec{10.99} & 32.62\dec{50.46} & 17.95\dec{65.13} & 33.23\dec{49.85} & 38.97\dec{44.11} \\
    \cellcolor{gray!15}VMamba-B\emph{(v0)}  & \cellcolor{gray!5}83.48 & 72.97\dec{10.51} & 38.24\dec{45.24} & 18.88\dec{64.60} & 37.33\dec{46.15} & 41.86\dec{41.62} \\
    \cellcolor{gray!15}VMamba-B\emph{(v2)}  & \cellcolor{gray!5}83.76 & 73.22\dec{10.54} & 38.53\dec{45.23} & 18.35\dec{65.41} & 35.99\dec{47.77} & 41.52\dec{42.24} \\
        \cellcolor{gray!15}MambaVision-B & \cellcolor{gray!5}83.96 & 73.84\dec{10.12} & 36.69\dec{47.27} & 21.69\dec{62.27} & 35.72\dec{48.24} & 41.98\dec{41.98} \\

    \bottomrule
    \end{tabular}
    }
\end{table*}





\begin{table*}[h]
\centering
\caption{\small Top-1 classification accuracy of various architectures on the ImageNet-E dataset~\cite{li2023imagenet} (\textit{left}) and ImageNet-B dataset~\cite{malik2024objectcompose} (\textit{right}). }

\label{tab:imagenet_e_b_app}
\setlength{\tabcolsep}{1.0pt}
\resizebox{\linewidth}{!}{
\begin{tabular}{|c| c c c c c c c c c | c c c c c|}
\hline
\rowcolor{LightCyan}
\multicolumn{1}{|c|}{Dataset $\rightarrow$} & \multicolumn{9}{|c|}{\textbf{ImageNet-E}} & \multicolumn{5}{|c|}{\textbf{ImageNet-B}} \\
\hline

\rowcolor{gray!5}
\rotatebox{0}{Model $\downarrow$} & \rotatebox{0}{\scriptsize $\lambda=-20$} & \rotatebox{0}{\scriptsize $\lambda=20$} & \rotatebox{0}{\scriptsize $\lambda=20$(adv)} & \rotatebox{0}{\scriptsize Random-BG} & \rotatebox{0}{\scriptsize $0.1$} & \rotatebox{0}{\scriptsize $0.08$} & \rotatebox{0}{\scriptsize $0.05$} & \rotatebox{0}{\scriptsize Random Pos.} & \rotatebox{0}{\scriptsize Original}  & \rotatebox{0}{\scriptsize Original}  & \rotatebox{0}{\scriptsize Caption}  & \rotatebox{0}{\scriptsize Class}  & \rotatebox{0}{\scriptsize Color} & \rotatebox{0}{\scriptsize Texture}  \\
\hline

ResNet-50 & \heatmapcolor{88.74} & \heatmapcolor{86.76}  & \heatmapcolor{73.02}  & \heatmapcolor{84.05}  & \heatmapcolor{89.19}  & \heatmapcolor{86.60}  & \heatmapcolor{77.34}  & \heatmapcolor{73.30}  & \heatmapcolor{94.55} & \heatmapcolor{98.60} & \heatmapcolor{94.00}  & \heatmapcolor{96.60}  & \heatmapcolor{88.20}  & \heatmapcolor{85.70}   \\
\hline

VGG16 & \heatmapcolor{84.14} & \heatmapcolor{79.62}  & \heatmapcolor{62.59}  & \heatmapcolor{77.27}  & \heatmapcolor{83.63}  & \heatmapcolor{80.16}  & \heatmapcolor{70.93}  & \heatmapcolor{64.15}  & \heatmapcolor{91.06} & \heatmapcolor{94.10} & \heatmapcolor{88.20}  & \heatmapcolor{93.70}  & \heatmapcolor{74.80}  & \heatmapcolor{75.00}   \\
\hline
VGG19 & \heatmapcolor{83.89} & \heatmapcolor{80.15}  & \heatmapcolor{63.21}  & \heatmapcolor{77.80}  & \heatmapcolor{81.16}  & \heatmapcolor{81.01}  & \heatmapcolor{70.54}  & \heatmapcolor{64.91}  & \heatmapcolor{92.05} & \heatmapcolor{94.50} & \heatmapcolor{87.80}  & \heatmapcolor{93.30}  & \heatmapcolor{77.10}  & \heatmapcolor{77.80}   \\
\hline

DenseNet-121 & \heatmapcolor{84.93} & \heatmapcolor{81.78}  & \heatmapcolor{62.36}  & \heatmapcolor{79.55}  & \heatmapcolor{85.15}  & \heatmapcolor{81.17}  & \heatmapcolor{70.08}  & \heatmapcolor{64.50}  & \heatmapcolor{92.78} & \heatmapcolor{96.20} & \heatmapcolor{90.20}  & \heatmapcolor{95.10}  & \heatmapcolor{81.10}  & \heatmapcolor{80.00}   \\
\hline
DenseNet-161 & \heatmapcolor{87.48} & \heatmapcolor{85.48}  & \heatmapcolor{67.72}  & \heatmapcolor{82.79}  & \heatmapcolor{87.22}  & \heatmapcolor{84.41}  & \heatmapcolor{75.00}  & \heatmapcolor{70.08}  & \heatmapcolor{93.04} & \heatmapcolor{97.50} & \heatmapcolor{90.70}  & \heatmapcolor{94.70}  & \heatmapcolor{81.10}  & \heatmapcolor{80.10}   \\
\hline

ConvNext-T & \heatmapcolor{90.95}  & \heatmapcolor{90.03}  & \heatmapcolor{76.88}  & \heatmapcolor{88.09}  & \heatmapcolor{93.01}  & \heatmapcolor{90.87}  & \heatmapcolor{83.09}  & \heatmapcolor{80.19}  & \heatmapcolor{96.09} & \heatmapcolor{98.20}  & \heatmapcolor{93.20}  & \heatmapcolor{95.10}  & \heatmapcolor{88.80}  & \heatmapcolor{87.40}    \\
\hline


ConvNext-S & \heatmapcolor{91.96}  & \heatmapcolor{90.76}  & \heatmapcolor{78.52}  & \heatmapcolor{88.99}  & \heatmapcolor{93.61}  & \heatmapcolor{91.66}  & \heatmapcolor{85.34}  & \heatmapcolor{82.19}  & \heatmapcolor{96.07}  & \heatmapcolor{98.80}  & \heatmapcolor{94.00}  & \heatmapcolor{96.70}  & \heatmapcolor{90.70}  & \heatmapcolor{89.60}     \\
\hline


ConvNext-B & \heatmapcolor{92.30}  & \heatmapcolor{91.52}  & \heatmapcolor{80.44}  & \heatmapcolor{90.00}  & \heatmapcolor{93.91}  & \heatmapcolor{93.01}  & \heatmapcolor{86.65}  & \heatmapcolor{83.75}  & \heatmapcolor{96.41}  & \heatmapcolor{99.20}  & \heatmapcolor{93.60}  & \heatmapcolor{96.40}  & \heatmapcolor{90.60}  & \heatmapcolor{91.40}    \\
\hline


ViT-T & \heatmapcolor{80.81}  & \heatmapcolor{77.07}  & \heatmapcolor{46.78}  & \heatmapcolor{69.07}  & \heatmapcolor{81.06}  & \heatmapcolor{76.55}  & \heatmapcolor{64.13}  & \heatmapcolor{57.86}  & \heatmapcolor{91.08} & \heatmapcolor{95.20}  & \heatmapcolor{85.50}  & \heatmapcolor{90.40}  & \heatmapcolor{67.30}  & \heatmapcolor{64.50}    \\
\hline

ViT-S & \heatmapcolor{86.77}  & \heatmapcolor{83.46}  & \heatmapcolor{63.19}  & \heatmapcolor{80.58}  & \heatmapcolor{87.98}  & \heatmapcolor{84.05}  & \heatmapcolor{74.29}  & \heatmapcolor{69.94}  & \heatmapcolor{94.74} & \heatmapcolor{97.70}  & \heatmapcolor{89.20}  & \heatmapcolor{94.30}  & \heatmapcolor{84.20}  & \heatmapcolor{80.60}     \\
\hline

ViT-B & \heatmapcolor{90.07}  & \heatmapcolor{87.48}  & \heatmapcolor{71.28}  & \heatmapcolor{84.88}  & \heatmapcolor{91.01}  & \heatmapcolor{88.64}  & \heatmapcolor{79.99}  & \heatmapcolor{76.42}  & \heatmapcolor{95.66} & \heatmapcolor{98.00}  & \heatmapcolor{90.40}  & \heatmapcolor{93.80}  & \heatmapcolor{86.20}  & \heatmapcolor{84.80}     \\
\hline

DeiT-T & \heatmapcolor{80.68}  & \heatmapcolor{77.21}  & \heatmapcolor{51.24}  & \heatmapcolor{71.83}  & \heatmapcolor{80.30}  & \heatmapcolor{76.83}  & \heatmapcolor{65.03}  & \heatmapcolor{59.54}  & \heatmapcolor{89.94} & \heatmapcolor{91.60}  & \heatmapcolor{86.50}  & \heatmapcolor{90.50}  & \heatmapcolor{73.90}  & \heatmapcolor{73.20}    \\
\hline

DeiT-S & \heatmapcolor{87.52}  & \heatmapcolor{84.88}  & \heatmapcolor{63.03}  & \heatmapcolor{80.70}  & \heatmapcolor{89.13}  & \heatmapcolor{85.70}  & \heatmapcolor{76.75}  & \heatmapcolor{71.37}  & \heatmapcolor{94.14} & \heatmapcolor{98.30}  & \heatmapcolor{91.40}  & \heatmapcolor{95.20}  & \heatmapcolor{85.70}  & \heatmapcolor{84.10}     \\
\hline

DeiT-B & \heatmapcolor{89.66}  & \heatmapcolor{86.79}  & \heatmapcolor{68.77}  & \heatmapcolor{84.26}  & \heatmapcolor{91.10}  & \heatmapcolor{89.19}  & \heatmapcolor{80.31}  & \heatmapcolor{77.25}  & \heatmapcolor{95.38} & \heatmapcolor{98.80}  & \heatmapcolor{92.30}  & \heatmapcolor{96.00}  & \heatmapcolor{86.70}  & \heatmapcolor{84.30}     \\
\hline

VMamba-T(v0) & \heatmapcolor{90.03}  & \heatmapcolor{89.31}  & \heatmapcolor{70.38}  & \heatmapcolor{85.59}  & \heatmapcolor{91.49}  & \heatmapcolor{89.59}  & \heatmapcolor{82.01}  & \heatmapcolor{78.63}  & \heatmapcolor{95.73}   & \heatmapcolor{98.00}  & \heatmapcolor{91.60}  & \heatmapcolor{94.90}  & \heatmapcolor{86.10}  & \heatmapcolor{86.70}   \\
\hline

VMamba-S(v0) & \heatmapcolor{90.53}  & \heatmapcolor{90.76}  & \heatmapcolor{73.39}  & \heatmapcolor{87.78}  & \heatmapcolor{93.13}  & \heatmapcolor{90.71}  & \heatmapcolor{83.92}  & \heatmapcolor{80.42}  & \heatmapcolor{96.16}  & \heatmapcolor{99.10}  & \heatmapcolor{92.80}  & \heatmapcolor{95.40}  & \heatmapcolor{89.40}  & \heatmapcolor{88.50}    \\
\hline

VMamba-B(v0) & \heatmapcolor{91.75}  & \heatmapcolor{90.62}  & \heatmapcolor{73.90}  & \heatmapcolor{88.33}  & \heatmapcolor{93.29}  & \heatmapcolor{91.19}  & \heatmapcolor{83.96}  & \heatmapcolor{81.66}  & \heatmapcolor{96.00}  & \heatmapcolor{98.80}  & \heatmapcolor{93.50}  & \heatmapcolor{96.20}  & \heatmapcolor{89.70}  & \heatmapcolor{88.40}     \\
\hline

VMamba-T(v2) & \heatmapcolor{91.15}  & \heatmapcolor{89.87}  & \heatmapcolor{75.18}  & \heatmapcolor{87.41}  & \heatmapcolor{92.09}  & \heatmapcolor{91.06}  & \heatmapcolor{83.66}  & \heatmapcolor{79.71}  & \heatmapcolor{95.84}   & \heatmapcolor{98.50}  & \heatmapcolor{92.20}  & \heatmapcolor{96.30}  & \heatmapcolor{87.20}  & \heatmapcolor{86.80}   \\
\hline

VMamba-S(v2) & \heatmapcolor{92.03}  & \heatmapcolor{90.79}  & \heatmapcolor{76.15}  & \heatmapcolor{88.81}  & \heatmapcolor{93.22}  & \heatmapcolor{92.25}  & \heatmapcolor{85.57}  & \heatmapcolor{81.89}  & \heatmapcolor{96.37}  & \heatmapcolor{99.20}  & \heatmapcolor{94.10}  & \heatmapcolor{97.40}  & \heatmapcolor{90.90}  & \heatmapcolor{89.50}    \\
\hline

VMamba-B(v2) & \heatmapcolor{92.37}  & \heatmapcolor{91.27}  & \heatmapcolor{77.30}  & \heatmapcolor{89.11}  & \heatmapcolor{93.70}  & \heatmapcolor{92.64}  & \heatmapcolor{86.03}  & \heatmapcolor{83.62}  & \heatmapcolor{96.37}  & \heatmapcolor{99.10}  & \heatmapcolor{94.00}  & \heatmapcolor{96.50}  & \heatmapcolor{90.80}  & \heatmapcolor{89.80}     \\
\hline

MambaVision-T & \heatmapcolor{90.67}  & \heatmapcolor{88.83}  & \heatmapcolor{73.07}  & \heatmapcolor{86.40}  & \heatmapcolor{91.93}  & \heatmapcolor{90.07}  & \heatmapcolor{81.87}  & \heatmapcolor{78.42}  & \heatmapcolor{95.73}  & \heatmapcolor{98.60}  & \heatmapcolor{93.70}  & \heatmapcolor{96.60}  & \heatmapcolor{89.10}  & \heatmapcolor{87.70}     \\
\hline

MambaVision-S & \heatmapcolor{91.22}  & \heatmapcolor{90.19}  & \heatmapcolor{75.18}  & \heatmapcolor{88.19}  & \heatmapcolor{92.78}  & \heatmapcolor{91.19}  & \heatmapcolor{84.01}  & \heatmapcolor{81.18}  & \heatmapcolor{96.03}  & \heatmapcolor{99.40}  & \heatmapcolor{94.40}  & \heatmapcolor{97.80}  & \heatmapcolor{91.40}  & \heatmapcolor{90.10}     \\
\hline

MambaVision-B & \heatmapcolor{91.77}  & \heatmapcolor{90.65}  & \heatmapcolor{78.42}  & \heatmapcolor{89.18}  & \heatmapcolor{93.70}  & \heatmapcolor{92.81}  & \heatmapcolor{86.35}  & \heatmapcolor{83.78}  & \heatmapcolor{96.30}  & \heatmapcolor{99.10}  & \heatmapcolor{94.60}  & \heatmapcolor{97.20}  & \heatmapcolor{91.40}  & \heatmapcolor{90.60}     \\
\hline

Swin-T & \heatmapcolor{90.05}  & \heatmapcolor{88.83}  & \heatmapcolor{71.51}  & \heatmapcolor{86.19}  & \heatmapcolor{91.08}  & \heatmapcolor{88.94}  & \heatmapcolor{79.39}  & \heatmapcolor{76.49}  & \heatmapcolor{95.27}  & \heatmapcolor{97.90}  & \heatmapcolor{91.70}  & \heatmapcolor{95.30}  & \heatmapcolor{85.50}  & \heatmapcolor{84.00}     \\
\hline
Swin-S & \heatmapcolor{90.67}  & \heatmapcolor{88.86}  & \heatmapcolor{73.35}  & \heatmapcolor{87.25}  & \heatmapcolor{91.91}  & \heatmapcolor{89.68}  & \heatmapcolor{81.55}  & \heatmapcolor{78.81}  & \heatmapcolor{96.25}   & \heatmapcolor{98.30}  & \heatmapcolor{91.80}  & \heatmapcolor{95.50}  & \heatmapcolor{86.10}  & \heatmapcolor{85.40}   \\
\hline
Swin-B & \heatmapcolor{91.08}  & \heatmapcolor{89.96}  & \heatmapcolor{75.09}  & \heatmapcolor{87.87}  & \heatmapcolor{92.62}  & \heatmapcolor{91.22}  & \heatmapcolor{83.43}  & \heatmapcolor{80.65}  & \heatmapcolor{95.95} & \heatmapcolor{98.60}  & \heatmapcolor{92.30}  & \heatmapcolor{95.60}  & \heatmapcolor{89.20}  & \heatmapcolor{87.40}      \\
\hline

\end{tabular}
}
\end{table*}

\begin{table*}[h]
\caption{\small Average Precision (AP) scores for different architectures on the COCO-DC dataset~\cite{malik2024objectcompose}, detailing results for small (APs), medium (APm), and large objects (APl).}
\label{tab:coco_dc}
\setlength{\tabcolsep}{4.0pt}
\definecolor{lightblue}{HTML}{E6F5FF}
\definecolor{lightgreen}{HTML}{E6FFE6}
\definecolor{lightyellow}{HTML}{FFFFE6}
\resizebox{\linewidth}{!}{
\begin{tabular}{l|cccc|cccc|cccc|cccc}
\toprule
\rowcolor{LightCyan}
\textbf{Model} & \textbf{AP} & \textbf{APs} & \textbf{APm} & \textbf{APl} & \textbf{AP} & \textbf{APs} & \textbf{APm} & \textbf{APl} & \textbf{AP} & \textbf{APs} & \textbf{APm} & \textbf{APl} & \textbf{AP} & \textbf{APs} & \textbf{APm} & \textbf{APl} \\ \midrule
\rowcolor{gray!5}
\multicolumn{5}{c}{\hspace{5em}\textbf{Original}} & \multicolumn{4}{c}{\textbf{Color}} & \multicolumn{4}{c}{\textbf{Texture}} & \multicolumn{4}{c}{\textbf{Average} }\\ \midrule
\cellcolor{gray!15}\textbf{ConvNext-T} & \cellcolor{gray!5}66.2 & \cellcolor{gray!5}41.0 & \cellcolor{gray!5}61.3 & \cellcolor{gray!5}71.4 & 55.0\dec{11.20} & 26.3\dec{14.70}& 49.0\dec{12.30}& 60.6 \dec{10.80} & 53.5\dec{12.70} & 26.5\dec{14.50}& 47.5\dec{13.80}& 60.5\dec{10.90}& 54.25 \dec{11.95}& 26.40\dec{14.60} & 48.25\dec{13.05} & 60.55\dec{10.85} \\
\cellcolor{gray!15}\textbf{Swin-T} & \cellcolor{gray!5}66.3& \cellcolor{gray!5}45.1& \cellcolor{gray!5}61.6& \cellcolor{gray!5}71.6 &55.1\dec{11.20}& 29.2\dec{15.90}& 48.0\dec{13.60}& 62.0\dec{09.60} & 53.6\dec{12.70}& 31.4\dec{13.70}& 45.2\dec{16.40}& 61.4\dec{10.20}& 54.35\dec{11.95} & 30.30\dec{14.80} & 46.60 \dec{15.00}& 61.70\dec{09.90} \\
\cellcolor{gray!15}\textbf{Swin-S} & \cellcolor{gray!5}69.1& \cellcolor{gray!5}43.7& \cellcolor{gray!5}62.3& \cellcolor{gray!5}75.9& 57.4\dec{11.70}& 30.1\dec{13.60}& 49.1\dec{13.20}& 64.3\dec{11.60}& 56.0\dec{13.10}& 26.8\dec{16.90}& 45.4\dec{16.90}& 64.3\dec{11.60}& 56.70\dec{12.40} & 28.45\dec{15.25} & 47.25\dec{15.05} & 64.30\dec{11.60} \\
\cellcolor{gray!15}\textbf{VSSM-T} & \cellcolor{gray!5}69.3& \cellcolor{gray!5}47.1& \cellcolor{gray!5}64.4& \cellcolor{gray!5}75.5&56.2\dec{13.10}& 30.9\dec{16.20}& 48.4\dec{16.00}& 63.7\dec{11.80}& 53.2\dec{16.10}& 29.2\dec{17.90}& 44.9\dec{19.50}& 61.0\dec{14.50}& 54.57\dec{14.73}& 30.05\dec{17.05}& 46.65\dec{17.75}& 62.35\dec{13.15}  \\
\cellcolor{gray!15}\textbf{VSSM-S} & \cellcolor{gray!5}70.9& \cellcolor{gray!5}51.7& \cellcolor{gray!5}64.6& \cellcolor{gray!5}77.1& 56.7\dec{14.20}& 31.3\dec{20.40}& 48.9\dec{15.70}& 64.3\dec{12.80}& 55.7\dec{15.20}& 26.8\dec{24.90}& 47.4\dec{17.20}& 64.4\dec{12.70}& 56.20\dec{14.70}& 29.05\dec{22.65}& 48.15\dec{16.45}& 64.35\dec{12.75}\\ \bottomrule
\end{tabular}
}
\end{table*}

\begin{table*}[h]
\centering
\caption{\small Robust accuracy of various architectures against low-frequency and high-frequency-based perturbation using PGD attack at varying level of  perturbation budget $\epsilon \in \{1/255,2/255,4/255,8/255,12/255,16/255\}$. }

\label{tab:adv_freq_main}
\setlength{\tabcolsep}{1.0pt}
\resizebox{\linewidth}{!}{
\begin{tabular}{|l| c c c c c c| c c c  c c c |c c c|}
\hline
\rowcolor{LightCyan}
\multicolumn{1}{|c|}{Filter $\rightarrow$} & \multicolumn{6}{|c|}{\textbf{Low-Pass}}  & \multicolumn{6}{|c|}{\textbf{High-Pass}} & \multicolumn{3}{|c|}{\textbf{All-Pass}} \\
\hline
\rowcolor{gray!5}
\hspace{2em}{$\epsilon \rightarrow$} & 1/255 & 2/255 & 4/255 & 8/255 & 12/255 & 16/255 & 1/255 & 2/255 & 4/255 & 8/255 &12/255 &16/255  &  1/255 & 2/255 & 4/255 \\
\hline

ResNet-50 & \heatmapcolor{94.94} & \heatmapcolor{93.28}  & \heatmapcolor{91.38}  & \heatmapcolor{88.42}  & \heatmapcolor{87.74}  & \heatmapcolor{87.86}  & \heatmapcolor{72.68}  & \heatmapcolor{43.58}  & \heatmapcolor{28.70} & \heatmapcolor{16.38} & \heatmapcolor{11.62}  & \heatmapcolor{10.76}  & \heatmapcolor{29.68}  & \heatmapcolor{2.04} &\heatmapcolor{0.08}\\
\hline

ConvNext-T & \heatmapcolor{95.98}  & \heatmapcolor{95.46}  & \heatmapcolor{94.60}  & \heatmapcolor{91.76}  & \heatmapcolor{89.70}  & \heatmapcolor{89.50}  & \heatmapcolor{58.26}  & \heatmapcolor{26.46}  & \heatmapcolor{13.62} & \heatmapcolor{4.40}  & \heatmapcolor{2.38}  & \heatmapcolor{2.06}  & \heatmapcolor{20.18}  & \heatmapcolor{1.60}  &\heatmapcolor{0.02}  \\
\hline

ConvNext-S & \heatmapcolor{96.52}  & \heatmapcolor{96.14}  & \heatmapcolor{95.80}  & \heatmapcolor{93.48}  & \heatmapcolor{91.70}  & \heatmapcolor{91.36}  & \heatmapcolor{66.58}  & \heatmapcolor{34.66}  & \heatmapcolor{20.50}  & \heatmapcolor{7.38}  & \heatmapcolor{3.66}  & \heatmapcolor{3.12}  & \heatmapcolor{30.10}  & \heatmapcolor{3.98}  &\heatmapcolor{0.08}   \\
\hline

ConvNext-B & \heatmapcolor{96.84}  & \heatmapcolor{96.76}  & \heatmapcolor{96.14}  & \heatmapcolor{93.62}  & \heatmapcolor{92.40}  & \heatmapcolor{91.72}  & \heatmapcolor{64.96}  & \heatmapcolor{39.26}  & \heatmapcolor{22.98}  & \heatmapcolor{8.76}  & \heatmapcolor{5.46}  & \heatmapcolor{4.56}  & \heatmapcolor{31.02}  & \heatmapcolor{6.24}   &\heatmapcolor{0.18} \\
\hline




ViT-T & \heatmapcolor{85.52}  & \heatmapcolor{77.56}  & \heatmapcolor{68.50}  & \heatmapcolor{53.02}  & \heatmapcolor{47.28}  & \heatmapcolor{46.32}  & \heatmapcolor{69.32}  & \heatmapcolor{38.48}  & \heatmapcolor{19.24} & \heatmapcolor{5.52}  & \heatmapcolor{2.56}  & \heatmapcolor{1.88}  & \heatmapcolor{11.14}  & \heatmapcolor{0.16} &\heatmapcolor{0.00}   \\
\hline

ViT-S & \heatmapcolor{93.12}  & \heatmapcolor{88.60}  & \heatmapcolor{83.50}  & \heatmapcolor{72.84}  & \heatmapcolor{69.22}  & \heatmapcolor{69.06}  & \heatmapcolor{77.14}  & \heatmapcolor{46.48}  & \heatmapcolor{27.94} & \heatmapcolor{11.08}  & \heatmapcolor{6.76}  & \heatmapcolor{5.28}  & \heatmapcolor{18.74}  & \heatmapcolor{0.42}  &\heatmapcolor{0.00}   \\
\hline

ViT-B & \heatmapcolor{94.66}  & \heatmapcolor{90.28}  & \heatmapcolor{87.40}  & \heatmapcolor{80.68}  & \heatmapcolor{77.50}  & \heatmapcolor{77.26}  & \heatmapcolor{85.74}  & \heatmapcolor{61.84}  & \heatmapcolor{46.52} & \heatmapcolor{26.18}  & \heatmapcolor{18.28}  & \heatmapcolor{16.28}  & \heatmapcolor{30.94}  & \heatmapcolor{1.30} &\heatmapcolor{0.08}     \\
\hline

VMamba-T & \heatmapcolor{96.46}  & \heatmapcolor{96.04}  & \heatmapcolor{95.22}  & \heatmapcolor{93.02}  & \heatmapcolor{91.70}  & \heatmapcolor{91.18}  & \heatmapcolor{58.12}  & \heatmapcolor{32.50}  & \heatmapcolor{16.46}   & \heatmapcolor{5.74}  & \heatmapcolor{2.92}  & \heatmapcolor{2.82}  & \heatmapcolor{24.00}  & \heatmapcolor{5.16}  &\heatmapcolor{0.28}  \\
\hline

VMamba-S & \heatmapcolor{97.02}  & \heatmapcolor{96.90}  & \heatmapcolor{96.48}  & \heatmapcolor{94.10}  & \heatmapcolor{92.62}  & \heatmapcolor{92.24}  & \heatmapcolor{67.64}  & \heatmapcolor{44.00}  & \heatmapcolor{28.18}  & \heatmapcolor{11.36}  & \heatmapcolor{6.46}  & \heatmapcolor{5.48}  & \heatmapcolor{32.54}  & \heatmapcolor{10.90}  &\heatmapcolor{0.84}   \\
\hline

VMamba-B & \heatmapcolor{97.28}  & \heatmapcolor{97.00}  & \heatmapcolor{96.56}  & \heatmapcolor{94.28}  & \heatmapcolor{92.90}  & \heatmapcolor{92.24}  & \heatmapcolor{66.60}  & \heatmapcolor{45.44}  & \heatmapcolor{27.68}  & \heatmapcolor{10.36}  & \heatmapcolor{6.08}  & \heatmapcolor{4.78}  & \heatmapcolor{33.10}  & \heatmapcolor{9.88}  &\heatmapcolor{0.40}    \\
\hline

MambaVision-T & \heatmapcolor{96.44}  & \heatmapcolor{95.98}  & \heatmapcolor{95.18}  & \heatmapcolor{93.02}  & \heatmapcolor{92.40}  & \heatmapcolor{92.54}  & \heatmapcolor{60.52}  & \heatmapcolor{35.32}  & \heatmapcolor{22.90}  & \heatmapcolor{9.00}  & \heatmapcolor{4.88}  & \heatmapcolor{3.86}  & \heatmapcolor{18.38}  & \heatmapcolor{3.04}  &\heatmapcolor{0.16}    \\
\hline

MambaVision-S & \heatmapcolor{96.74}  & \heatmapcolor{96.26}  & \heatmapcolor{95.44}  & \heatmapcolor{93.56}  & \heatmapcolor{92.84}  & \heatmapcolor{93.10}  & \heatmapcolor{67.60}  & \heatmapcolor{44.56}  & \heatmapcolor{33.24}  & \heatmapcolor{17.48}  & \heatmapcolor{11.14}  & \heatmapcolor{9.24}  & \heatmapcolor{23.80}  & \heatmapcolor{4.46}  &\heatmapcolor{0.30}    \\
\hline

MambaVision-B & \heatmapcolor{96.80}  & \heatmapcolor{96.24}  & \heatmapcolor{95.82}  & \heatmapcolor{93.86}  & \heatmapcolor{93.04}  & \heatmapcolor{93.22}  & \heatmapcolor{70.28}  & \heatmapcolor{45.68}  & \heatmapcolor{33.20}  & \heatmapcolor{18.36}  & \heatmapcolor{12.86}  & \heatmapcolor{11.24}  & \heatmapcolor{28.80}  & \heatmapcolor{6.40}  &\heatmapcolor{0.72}    \\
\hline

Swin-T & \heatmapcolor{96.24}  & \heatmapcolor{95.86}  & \heatmapcolor{95.24}  & \heatmapcolor{93.28}  & \heatmapcolor{91.94}  & \heatmapcolor{91.30}  & \heatmapcolor{49.12}  & \heatmapcolor{23.42}  & \heatmapcolor{9.78}  & \heatmapcolor{2.06}  & \heatmapcolor{1.10}  & \heatmapcolor{0.88}  & \heatmapcolor{12.60}  & \heatmapcolor{1.86}  &\heatmapcolor{0.00}    \\
\hline
Swin-S & \heatmapcolor{96.92}  & \heatmapcolor{96.60}  & \heatmapcolor{96.16}  & \heatmapcolor{94.26}  & \heatmapcolor{93.30}  & \heatmapcolor{92.70}  & \heatmapcolor{60.70}  & \heatmapcolor{36.50}  & \heatmapcolor{20.18}   & \heatmapcolor{7.48}  & \heatmapcolor{3.90}  & \heatmapcolor{3.30}  & \heatmapcolor{25.92}  & \heatmapcolor{6.74}  &\heatmapcolor{0.42}  \\
\hline
Swin-B & \heatmapcolor{96.82}  & \heatmapcolor{96.94}  & \heatmapcolor{96.24}  & \heatmapcolor{94.72}  & \heatmapcolor{94.10}  & \heatmapcolor{93.34}  & \heatmapcolor{61.04}  & \heatmapcolor{40.50}  & \heatmapcolor{24.68} & \heatmapcolor{9.30}  & \heatmapcolor{5.82}  & \heatmapcolor{4.34}  & \heatmapcolor{30.14}  & \heatmapcolor{9.76}   &\heatmapcolor{0.70}    \\
\hline


\end{tabular}
}
\end{table*}

\end{document}